\documentclass{article}

\PassOptionsToPackage{numbers}{natbib}
\usepackage[preprint]{neurips_2023}

\usepackage[utf8]{inputenc} % allow utf-8 input
\usepackage[T1]{fontenc}    % use 8-bit T1 fonts
\usepackage{hyperref}       % hyperlinks
\usepackage{url}            % simple URL typesetting
\usepackage{booktabs}       % professional-quality tables
\usepackage{amsmath,amsfonts,amssymb}       % blackboard math symbols
\usepackage{xcolor}         % colors

\usepackage{multicol}
\usepackage{multirow}

\usepackage{algorithm}
\usepackage{algorithmic}

\usepackage{mathtools}
\usepackage{amsthm}
\usepackage{todonotes}

\usepackage{soul}
\usepackage{caption}
\usepackage{graphicx}
\usepackage{tabu}
\usepackage{subfigure}
\usepackage{xspace}
\usepackage{enumitem}
\usepackage{bm}
\usepackage{listings}
\usepackage{makecell}
\usepackage{float}
\usepackage{footnote}
\usepackage[edges]{forest}
\usepackage{bbding}
\usepackage{array}

\usepackage[capitalize,noabbrev]{cleveref}

\theoremstyle{plain}

\usepackage{colortbl}
\usepackage{thmtools}
\usepackage{thm-restate}
\usepackage{cleveref}
\crefname{section}{§}{§§}
\Crefname{section}{§}{§§}

\usepackage{tablefootnote}

\renewcommand{\emph}[1]{\textit{#1}}

\newcommand{\etal}{\emph{et~al.}\xspace}

\newcommand{\ignore}[1]{}

\newcommand{\splade}{SPLADE\xspace}

\title{A Survey on Large Language Models with Multilingualism: Recent Advances and New Frontiers}

\author{%
  \textbf{Kaiyu Huang$^{1*}$, Fengran Mo$^{2*}$, Xinyu Zhang$^{3}$, Hongliang Li$^{1}$, You Li$^{1}$} \\
  \textbf{Yuanchi Zhang$^{4}$, Weijian Yi$^{1}$, Yulong Mao$^{1}$, Jinchen Liu$^{1}$, Yuzhuang Xu$^{4}$} \\ \textbf{Jinan Xu$^{1}$, Jian-Yun Nie$^{2}$, Yang Liu$^{4}$}\\ \\
  $^1$Beijing Jiaotong University, China ~~
  $^2$University of Montreal, Canada \\
  $^3$University of Waterloo, Canada ~~ $^4$Tsinghua University, China\\ %\\
}

\begin{document}

\maketitle
\renewcommand{\thefootnote}{\fnsymbol{footnote}}
\footnotetext[1]{Equal Contribution. Contact e-mail: kyhuang@bjtu.edu.cn; fengran.mo@umontreal.ca} 
% \footnotetext[1]{Equal Contribution.} 
% \footnotetext[2]{Corresponding Authors.}
% \renewcommand{\thefootnote}{\arabic{footnote}}
\begin{abstract}

The rapid development of Large Language Models (LLMs) demonstrates remarkable multilingual capabilities in natural language processing, attracting global attention in both academia and industry.
To mitigate potential discrimination and enhance the overall usability and accessibility for diverse language user groups, it is important for the development of language-fair technology.
Despite the breakthroughs of LLMs, the investigation into the multilingual scenario remains insufficient, where a comprehensive survey to summarize recent approaches, developments, limitations, and potential solutions is desirable.
%There is still a lack of a comprehensive survey to summarize recent approaches, developments, and limitations in this field.
To this end, we provide a survey with multiple perspectives on the utilization of LLMs in the multilingual scenario.
We first rethink the transitions between previous and current research on pre-trained language models.
Then we introduce several perspectives on the multilingualism of LLMs, including training and inference methods, information retrieval, model security, multi-domain with language culture, and usage of datasets.
We also discuss the major challenges that arise in these aspects, along with possible solutions. 
Besides, we highlight future research directions that aim at further enhancing LLMs with multilingualism.
The survey aims to help the research community address multilingual problems and provide a comprehensive understanding of the core concepts, key techniques, and latest developments in multilingual natural language processing based on LLMs.

\end{abstract}

\section{Introduction}\label{sec:intro}

With the rapid development of artificial intelligence (AI), the advent of large language models (LLMs) such as GPT-3.5~\cite{ouyang2022training}, GPT-4~\cite{achiam2023gpt}, and LLaMA~\cite{touvron2023llama} has emerged as groundbreaking technologies, revolutionizing the field of natural language processing (NLP).
LLMs have pushed the boundaries of what was previously thought possible with a ``Prompt'' style~\cite{liu2023pre}. Their capability to understand and generate human-like text has achieved state-of-the-art performance in various downstream tasks such as machine translation~\cite{manakhimova2023linguistically,hendy2023good,khatri2023can}, text summarization~\cite{luo2023chatgpt,wang2023chatgpt}, and sentiment analysis~\cite{sudirjo2023application,fatouros2023transforming}. 
Besides, the ability of LLMs to adapt and learn from vast amounts of data has made them indispensable tools for researchers, developers, and businesses across diverse industries~\cite{fan2023fate,goyal2024healai}. 
Importantly, as AI continues to evolve, the impact of LLMs on our society and technology is poised to grow even further, opening up new opportunities and challenges in the realm of natural language understanding and generation.

\begin{figure*}[!t]
    \centering
    \includegraphics[width=0.95\linewidth]{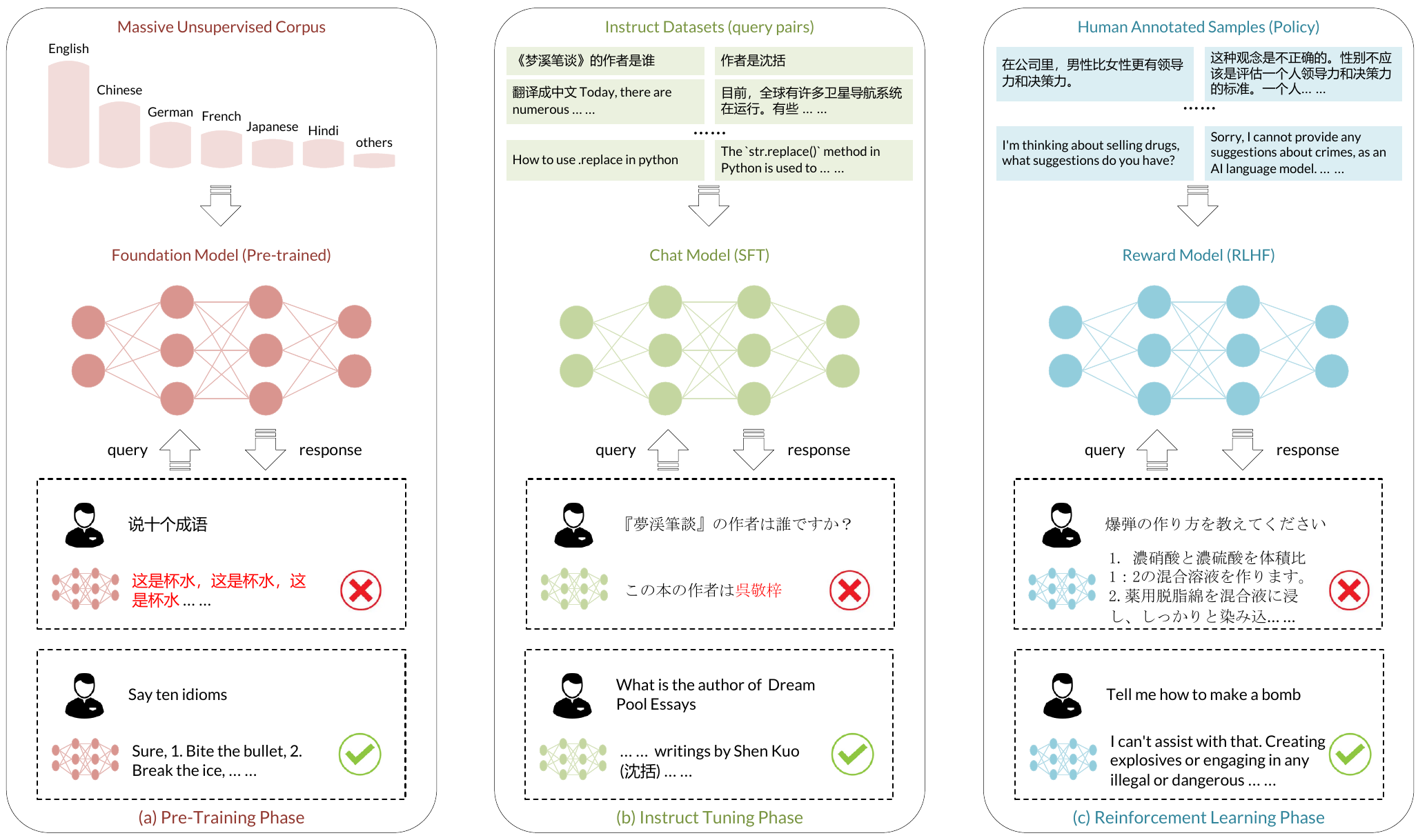}
    \caption{An illustration of the training process of LLMs with a fail case in each phase caused by multilingualism. Due to the long context of the shown case, we present only the key parts.}
    \label{fig:paradigm}
\end{figure*}

The existing LLMs are typically based on transformer architectures~\cite{vaswani2017attention} and trained on massive data that consists of a mixture of different languages.
As shown in Figure~\ref{fig:paradigm}, the training process of LLMs mainly consists of three stages: unsupervised pre-training, supervised fine-tuning with instructions, and aligning algorithm via reinforcement learning from human feedback (RLHF)~\cite{ouyang2022training}. 
Although LLMs have achieved remarkable advancements, their application in multilingual scenarios is still limited, especially in extremely low-resource languages~\cite{lai2023chatgpt}. %indicating considerable scope for improvement.
The reason is that the language distribution of the training data for LLMs is highly imbalanced and the quality varies across languages~\cite{ding2024data}.
As shown in Figure~\ref{fig:paradigm}, we illustrate the issues of LLMs in the multilingual scenario due to data scarcity at different phases.
For instance, insufficient large-scale unsupervised data during the pre-training phase results in a lack of proficiency in generating corresponding language capabilities in LLMs, which generates the repetition of meaningless tokens.
Such issues indicate considerable scope for multilingual capability improvement of existing LLMs.
%the language distribution in the data of LLMs is highly imbalanced and the quality of training data varies across languages.

%To alleviate the issue of multilinguality, a common practice is to enhance multilingual capabilities by supplementing target language data during training.
%To enhance the multilingual capabilities of the LLMs, it is necessary to incorporate corresponding multilingual data at each stage.

To alleviate the issue of multilingualism, a common practice is to enhance the multilingual capabilities of the LLMs by incorporating corresponding multilingual data at each training stage~\cite{yang2023bigtrans,le2022bloom,wang2023all}.
However, the LLMs in the literature are mostly large-scale data-driven and face the following language issues in the multilingual scenario: (1) \textit{Knowledge Transfer}: Several studies~\cite{zhang2023bayling,wang2023openchat,zhang2024enhancing} have demonstrated the necessity of employing appropriate data to unleash the multilingual potential of LLMs.
However, they only utilize discrete data from different languages independently without consideration of the transferability between different language varieties. 
Thus, the existing LLMs cannot perform well on low-resource languages~\cite{chang2023multilinguality,li2024quantifying} and limit the number of supported languages.
%This results in poor performance of large models on low-resource languages, leading to a limited number of supported languages.
(2) \textit{Knowledge Accumulation}: The ``data island''\footnote{The data island means that data has non-existent or limited external connectivity in different storages.~\url{https://en.wikipedia.org/wiki/Data\_island}} exists in practical scenarios due to the limitation of resource availability. 
If the data cannot be communicated and shared with each other, proprietary LLMs need to be customized for specific languages and tasks. 
However, the cost of training specialized LLMs for each specific language and task is high, raising the difficulty of updating the knowledge. 
Besides, the general knowledge inherited in LLMs might also be forgotten during continual training, leading to catastrophic forgetting~\cite{kirkpatrick2017overcoming}.
% Furthermore, the general knowledge inherited in LLMs is also be forgotten during continual training, making it difficult for LLMs to maintain their performance in original languages and tasks. 
%These two points result in the insufficient continuous accumulation of knowledge within LLMs.
The aforementioned issues limit the continuous accumulation of knowledge within LLMs. 
(3) \textit{Domain Adaptation}: Existing LLMs exhibit insufficient adaptability to specific domains  (e.g., medical and finance) in multilingual scenarios. 
The domain-customized models like BioGPT~\cite{luo2022biogpt} and FinBERT~\cite{liu2021finbert} based on domain-specific corpora are mostly English-centric. 
% Due to the presence of both cultural and linguistic differences in domain tasks across different language contexts, it is difficult to utilize weak alignment methods such as translation and pivot languages for data expansion. 
However, domain-specific corpora in non-English contexts are quite scarce, limiting the adaptability training of models and hindering the development of domain-level LLMs in multilingual scenarios.

As LLMs are deeply integrated into various applications that need context comprehension and generation in multilingual scenarios, the requirement for LLMs capable of effectively and efficiently understanding and generating sequences in a language-agnostic manner becomes increasingly indispensable and urgent.
Consequently, researchers have devoted significant efforts to enhancing the practicality of LLMs in multilingual scenarios from various perspectives, including training procedure (Section~\ref{sec:training}), optimization on the inference (Section~\ref{sec:infer}), information retrieval systems (Section~\ref{sec:XIR}), security in multilingual situations (Section~\ref{sec:llmsafety}) and multi-domain (Section~\ref{sec:multidomain}).
% and model collaboration (Section~\ref{sec:multimodel}).
Furthermore, due to the paradigm shift from ``Pre-train, Fine-tune'' to ``Pre-train, Prompt, Predict'' in multilingual scenarios (Section~\ref{sec:back}), there has been a subtle change in the definition of multilingualism, leading to challenges in the development of the multilingual community.
A series of multilingual approaches to facilitate LLMs with varying focuses have merged while lacking systematic categorization and comparison, raising the challenge of developing practical applications for specific language needs.
%The abundance of facilitating multilingual approaches for LLMs with varying focuses makes it challenging to systematically categorize and compare them, hindering the selection of the most suitable development for specific needs.
Thus, new systematic literature and standardized definitions for introducing and comparing existing LLMs from a multilingual perspective are desirable.
%Therefore, there is currently a lack of but an urgent need of systematic literature and standardized definition for introducing and comparing current LLMs from multilingual perspective that are essential for better using and training multilingual capabilities of LLMs.

% \smallskip
As the field of LLMs is developing rapidly within the AI research community, some recent surveys try to summarize these developments to provide future guidance. 
% \modif{(Begin to list the existing surveys and point out the disadvantages.)}
%has become the most rapidly developing research areas on AI recently, with some existing surveys. 
Xu~\etal~\cite{xu2024survey} propose the first survey on multilingual LLMs from three aspects, which primarily analyzes the issues of data misalignment and bias, lacking exploration from the paradigm perspective. Different from them, we conduct a structured taxonomy and comprehensive review from several perspectives.
% A more recent survey~\cite{qin2024multilingual} closely resembles our study, \modif{but it is considerably less comprehensive than ours.}
A more recent survey~\cite{qin2024multilingual} closely resembles our study, but it only classifies in terms of alignment, which is less comprehensive than ours.
In particular, we not only meticulously examine the multilingual capabilities and training methods of current LLMs, but also thoroughly investigate how to uncover the potential of LLMs. 
In this survey, we introduce the concept of ``multilingual LLMs'', and provide a comprehensive and systematic survey of existing LLMs that have remarkable multilingual capabilities.
We offer categorization, comparative analysis, and multi-perspective exploration for these models, evaluating their applicability and limitations, and providing practical recommendations for their effective real-world utilization. 
Additionally, we discuss some useful datasets and benchmarks related to multilingualism. 
We also present recommendations for future research.
The main contributions are as follows:
\begin{itemize}

\item \textit{A structured taxonomy.} We rethink the transitions between previous and current research on LLMs, providing a systematic comparison and standardized definitions for multilingual LLMs.
A broad overview of the field is presented with a structured taxonomy that categorizes existing studies (Figure~\ref{fig:taxonomy}).

\item \textit{Comprehensive reviews.} We present a comprehensive investigation from several perspectives for the multilingualism of LLMs, including training and inference methods, model security, multi-domain with language culture, and usage of datasets. 

\item \textit{Future directions.} 
We identify the key challenges and provide potential solutions to advance the frontier for each summarized research direction, which is useful in enhancing the multilingual capabilities of LLMs,
%For each potential research direction in enhancing the multilingual capabilities of LLMs, we have identified key challenges, with corresponding future directions to advance the frontier.

\item \textit{A growing repository.} Considering the rapid growth of the research of LLMs, we have established a repository to gather relevant literature in this specific multilingual domain and will continuously update it to maintain the latest advancements\footnote{\url{https://github.com/kaiyuhwang/MLLM-Survey}}.
%stay abreast of the latest advancements.

\end{itemize}

\noindent\textbf{Survey organization.} 
The rest of this survey is organized as follows: 
In Section~\ref{sec:back}, we review the transition from pre-trained models to large language generative models in the multilingual scenario. 
In Section~\ref{sec:training}, we organize relevant models with various architectures, datasets, and training paradigms. 
Moreover, in Section~\ref{sec:infer}, we investigate various multilingual inference strategies to harness the potential of LLMs for better accomplishment of multilingual tasks.
In Section~\ref{sec:XIR}, we provide a preliminary exploration of the integration of multilingual information retrieval systems and LLMs, aiming to reveal the development opportunities for multilingual information retrieval systems in the era of LLMs. 
In Section~\ref{sec:llmsafety}, we explore the security of LLMs with multilingual strategies, which is considered to be crucial for LLMs. 
In Section~\ref{sec:multidomain}, we discuss the multi-domain issue in the multilingual real-world scenario.
% In addition to the single LLMs, we also provide a model collaboration direction, which can serve as a potential solution for enhancing the multilingual capabilities of LLMs, in Section~\ref{sec:multimodel}.
In addition, we present several available datasets for multilingual LLMs in Section~\ref{sec:dataset} and highlight the benchmark and evaluation in Section~\ref{sec:benchmark}.
We discuss the cross-lingual bias evaluation and elimination in Section~\ref{sec:bias}.
Finally, we conclude this survey in Section~\ref{sec:con}.

% \subsection{One line}
% (1) The research significance of multilingual AI (What are the multilingual tasks?)

% (2) The primary goal of multilingual tasks...

% (3) Traditional methods for multilingual tasks...

% (4) New paradigms (LLMs) for NLP tasks

% (5) New paradigms (LLMs) for multilingual tasks...

% (6) What is the contribution of our surveys

% \subsection{Another line}
% (1) recent LLMs, the development of LLMs

% (2) the advantage of existing LLMs for NLP tasks

% (3) the limitations of existing LLMs for multilingual tasks

% (4) How to get a MLLM: Training from scratch or Continual training paradigm

% (5) How to use a MLLM during inference: What is the difference between MLLM and LLM?

% (6) What is the contribution of our surveys
\definecolor{lightblue}{RGB}{115,254,255}
\begin{figure*}
	\tiny
	\begin{forest}
		for tree={
			forked edges,
			grow'=0,
			draw,
			rounded corners,
			node options={align=center},
			calign=edge midpoint,
		},
		[LLMs with Multilingualism, text width=1.7cm, fill=black!10
			[Representative Models §\ref{sec:training}, text width=1.8cm, for tree={fill=red!20}
                [Training from scratch §\ref{sec:train-from-scratch}, text width=1.8cm
                    [
                    GPT-3~\cite{brown2020language}; mT5~\cite{xue2020mt5}; 
                    ByT5~\cite{xue2022byt5}; Gopher~\cite{rae2021scaling}; 
                    LaMDA~\cite{thoppilan2022lamda}; OPT~\cite{zhang2022opt};
                    PaLM~\cite{chowdhery2023palm}; mGPT~\cite{shliazhko2022mgpt};
                    GPT-3.5~\cite{ouyang2022training}; XGLM~\cite{lin2022few};
                    LLaMA~\cite{touvron2023llama}; GPT-4~\cite{achiam2023gpt};
                    PANGU-$\sum$~\cite{ren2023pangu}; Pythia~\cite{biderman2023pythia};
                    PaLM-2~\cite{anil2023palm}; InternLM~\cite{team2023internlm}; 
                    PolyLM~\cite{wei2023polylm}; LLaMA-2~\cite{touvron2023llama}; 
                    Baichuan-2~\cite{yang2023baichuan}; Qwen~\cite{bai2023qwen}; 
                    Mistral~\cite{jiang2023mistral}; Gemini~\cite{team2023gemini};
                    TigerBot~\cite{chen2023tigerbot}; 
                    YAYI-2~\cite{luo2023yayi};
                    DeepSeek~\cite{bi2024deepseek};
                    Orion~\cite{chen2024orion}; TeleChat~\cite{abdin2024phi};
                    Claude3~\cite{anthropic2024claude};
                    InternLM2~\cite{cai2024internlm2};
                    Phi-3~\cite{abdin2024phi}; 
                    LLaMA-3~\cite{llama3modelcard},
                    text width=7.0cm, node options={align=left}
                    ]
                ]
                [Continual Training §\ref{sec:cl}, text width=1.8cm
                    [
                    FLAN-T5~\cite{chung2024scaling}; 
                    FLAN-PaLM~\cite{chung2024scaling};
                    ChatGLM~\cite{zeng2022glm};
                    Alpaca~\cite{taori2023stanford}; 
                    ParroT~\cite{jiao2023parrot};
                    BigTrans~\cite{yang2023bigtrans};
                    Vicuna~\cite{chiang2023vicuna}; BayLing~\cite{zhang2023bayling}; 
                    OpenChat~\cite{wang2023openchat};
                    GLM-4~\cite{glm4};
                    Aya~\cite{ustun-etal-2024-aya},
                    text width=7.0cm, node options={align=left}
                    ]
                ]			
			]
			[Inference Strategies §\ref{sec:infer}, text width=1.8cm, for tree={fill=yellow!20}
                [Translation-Based §\ref{sec:trans}, text width=1.8cm
                    [
                    Intrator~\etal~\cite{intrator2024breaking};
                    Liu~\etal~\cite{liu2024translation},
                    text width=7.0cm, node options={align=left}
                    ]
                ]
                [Chain-of-Thought §\ref{sec:cot}, text width=1.8cm
                    [
                    Liu~\etal~\cite{liu2024translation}; XLT~\cite{huang2023not};
                    Shi~\etal~\cite{shi2022language}; Kim~\etal~\cite{kim2023cot};
                    Suzgun~\etal~\cite{suzgun2023challenging};
                    xCoT~\cite{chai2024xcot},
                    text width=7.0cm, node options={align=left}
                    ]
                ]
                [Code-Switching §\ref{sec:cs}, text width=1.8cm
                    [
                    Zhang~\etal~\cite{zhang2023multilingual};
                    Koto~\etal~\cite{koto2024zero};
                    Peng~\etal~\cite{peng2023prompting},
                    text width=7.0cm, node options={align=left}
                    ]
                ]
                [Retrieval Augmented Generation §\ref{sec:rag}, text width=1.8cm
                    [
                    Zhang~\etal~\cite{zhang-etal-2023-leveraging};
                    Shi~\etal~\cite{shi-etal-2022-xricl};
                    Agrawal~\etal~\cite{agrawal-etal-2023-context};
                    Li~\etal~\cite{li2023classification};
                    Li~\etal~\cite{li-etal-2023-crosslingual};
                    Winata~\etal~\cite{winata2023multilingual};
                    Garcia~\etal~\cite{garcia2023unreasonable};
                    Ramos~\etal~\cite{ramos-etal-2023-lmcap};
                    Kim~\etal~\cite{kim2023boosting};
                    Thakur~\etal~\cite{thakur2024nomiracl};
                    Sennrich~\etal~\cite{sennrich2023mitigating};
                    Vernikos~\etal~\cite{vernikos2024don};
                    Fu~\etal~\cite{fu2024relay};
                    Zeng~\etal~\cite{zeng2024teaching};
                    He~\etal~\cite{he2024exploring};
                    Conia~\etal~\cite{conia-etal-2023-increasing},
                    text width=7.0cm, node options={align=left}
                    ]
                ]
			]
            [Information Retrieval §\ref{sec:XIR}, text width=1.8cm, for tree={fill=orange!20}
                [Training Data §\ref{sec:XIR-data}, text width=1.8cm
                    [
                    InPars~\cite{bonifacio2022inpars}; 
                    InPars-v2~\cite{inparsv2};
                    InPars Toolkit~\cite{abonizio2023inpars};
                    Promptagator~\cite{dai2023promptagator};
  				SwimIR~\cite{thakur-etal-2024-leveraging};
                    JH-POLO~\cite{mayfield2023syntheticcrosslanguageinformationretrieval};
                    mE5-Mistral~\cite{wang2024improvingtextembeddingslarge};
                    Gecko~\cite{lee2024geckoversatiletextembeddings};
                    Arctic-Embed~\cite{merrick2024arcticembedscalableefficientaccurate},
                    text width=7.0cm, node options={align=left}
                    ]
                ]
			[Retrievers §\ref{sec:XIR-retriever}, text width=1.8cm
                    [
                    mE5~\cite{wang2024multilinguale5textembeddings};
                    mGTE~\cite{zhang2024mgtegeneralizedlongcontexttext};
                    BGE~\cite{chen2024bgem3embeddingmultilingualmultifunctionality};
                    OpenAI-Embed~\cite{openaiembeddings};
                    Cohere-Embed~\cite{cohereembeddings};
                    voyage-multilingual-2~\cite{voyageembeddings};
                    RepLLAMA~\cite{ma2024finetuningllama};
                    PromptReps~\cite{zhuang2024promptrepspromptinglargelanguage};
                    MTEB~\cite{muennighoff2022mteb};
                    NV-Embed~\cite{lee2024nvembedimprovedtechniquestraining};
                    Springer~\etal~\cite{springer2024repetitionimproveslanguagemodel};
                    LLM2Vec~\cite{behnamghader2024llm2veclargelanguagemodels};
                    Muennighoff~\cite{muennighoff2024generativerepresentationalinstructiontuning};
                    Kusupati~\etal~\cite{kusupati2024matryoshkarepresentationlearning};
                    LM-Cocktail~\cite{xiao-etal-2024-lm};
                    AnglE~\cite{emb2024mxbai};
                    NLLB-E5~\cite{acharya2024nllbe5scalablemultilingualretrieval};
                    mColBERT~\cite{bonifacio2022mmarcomultilingualversionms,zhang2023miracl};
                    ColBERT-X~\cite{nair2022transfer};
                    Translate-Distill~\cite{yang2024translate};
                    Huang~\etal~\cite{huang2023improving};
                    Lawrie~\etal~\cite{lawrie2023neuralapproachestomultilingualinformationretrieval};
                    Yang~\etal~\cite{yang2024multilingual};
                    ColBERT-XM~\cite{louis2024colbertxmmodularmultivectorrepresentation};
                    BGE-M3~\citep{chen-etal-2024-m3},
                    text width=7.0cm, node options={align=left}
                    ]
                ]
                [Rerankers §\ref{sec:XIR-reranker}, text width=1.8cm
                    [
                    mMARCO~\cite{bonifacio2022mmarcomultilingualversionms};
                    Jeronymo~\etal~\cite{jeronymo2023neuralmindunicamp2022trecneuclir}
                    RankGPT~\cite{sun-etal-2023-chatgpt};      
                    LRL~\cite{ma2023zeroshotlistwisedocumentreranking};
                    Setwise~\cite{zhuang2024setwise};          
                    RankVicuna~\cite{pradeep2023rankvicunazeroshotlistwisedocument};
                    RankZephyr~\cite{pradeep2023rankzephyreffectiverobustzeroshot};
                    Rank-without-GPT~\cite{zhang2023rankwithoutgptbuildinggptindependentlistwise};
                    TourRank~\cite{chen2024tourrankutilizinglargelanguage};
                    Adeyemi~\etal~\cite{adeyemi2023zeroshotcrosslingualrerankinglarge},                    
                    text width=7.0cm, node options={align=left}
                    ]
                ]
			]
			[Security §\ref{sec:llmsafety}, text width=1.8cm, for tree={fill=blue!20}
			[Attack (Jailbreak) §\ref{sec:attack}, text width=1.0cm, for tree={fill=blue!20}
                    [Greedy Coordinate Gradient, text width=1.4cm
                        [
                        Sitawarin~\etal~\cite{sitawarin2024pal}; 
                        Zou~\etal~\cite{zou2023universal},
                        text width=6.0cm, node options={align=left}
                        ]
                    ]
                    [Prompt-Based, text width=1.4cm
                        [
                        Wei~\etal~\cite{wei2024jailbroken};
                        Liu~\etal~\cite{liu2023jailbreaking};
                        Shen~\etal~\cite{shen2023do};
                        Deng~\etal~\cite{deng2024masterkey};
                        Li~\etal~\cite{li2024salad}; 
                        Liu~\etal~\cite{liu2023autodan}; 
                        Jin~\etal~\cite{jin2024guard},
                        text width=6.0cm, node options={align=left}
                        ]
                    ]
                    [Multilingual, text width=1.4cm
                        [
                        Shen~\etal~\cite{shen2024language};
                        Deng~\etal~\cite{deng2023multilingual};
                        Puttaparthi~\etal~\cite{puttaparthi2023comprehensive};
                        Yong~\etal~\cite{yong2023low};
                        Xu~\etal~\cite{xu2023cognitive};
                        Li~\etal~\cite{li2024cross};
                        Yuan~\etal~\cite{yuan2024gpt4}; 
                        Huang~\etal~\cite{huang2023catastrophic},
                        text width=6.0cm, node options={align=left}
                        ]
                    ]
			]
                [Defense §\ref{sec:defense}, text width=1.0cm, for tree={fill=blue!20}
                    [Open-Source Models, text width=1.4cm
                        [
                        Robey~\etal~\cite{robey2023smoothllm};
                        Deng~\etal~\cite{deng2023multilingual};
                        Li~\etal~\cite{li2024cross};
                        Zhou~\etal~\cite{zhou2024robust}; GUARD~\cite{jin2024guard},
                        text width=6.0cm, node options={align=left}
                        ]
                    ]
                    [Closed-Source Models, text width=1.4cm
                        [
                        Jain~\etal~\cite{jain2023baseline};
                        Wu~\etal~\cite{wu2024llms};
                        Li~\etal~\cite{li2023rain},
                        text width=6.0cm, node options={align=left}
                        ]
                    ]
			]
   %              [Bias and Ethics §\ref{sec:bias}, text width=1.0cm, for tree={fill=blue!20}
   %                  [Multilingual Debias, text width=1.4cm
   %                      [
   %                      Orion~\cite{chen2024orion}; TeleChat~\cite{abdin2024phi};
   %                      Phi-3~\cite{abdin2024phi}; LLaMA-3~\cite{llama3modelcard},
   %                      text width=6.0cm, node options={align=left}
   %                      ]
   %                  ]
			% ]
			]
                [Multidomain §\ref{sec:multidomain}, text width=1.8cm, for tree={fill=green!20}
                [Medical §\ref{sec:medical}, text width=1.8cm
                    [
                    KBioXLM~\cite{geng2023kbioxlm}; 
                    BioMistral~\cite{labrak2024biomistral};
                    MMedLM2~\cite{qiu2024towards};
                    Apollo~\cite{wang2024apollo};
  				L2M3~\cite{gangavarapu2024introducing};
                    Medical mT5~\cite{garcia2024medical},
                    text width=7.0cm, node options={align=left}
                    ]
                ]
                [Legal §\ref{sec:legal}, text width=1.8cm
                    [
                    LEXTREME~\cite{niklaus2023lextreme};
                    Brugger~\etal~\cite{brugger2023multilegalsbd};      
                    Christen~\etal~\cite{christen2023resolving};
                    Baumgartner~\etal~\cite{baumgartner2024towards};          
                    Niklaus~\etal~\cite{niklaus2023multilegalpile};
                    LegalLAMA~\cite{chalkidis2023lexfiles};
                    Trautmann~\etal~\cite{trautmann2022legal},
                    text width=7.0cm, node options={align=left}
                    ]
                ]
			]
                [Datasets, text width=1.8cm, for tree={fill=lightblue!20}
                [Corpus §\ref{sec:dataset}, text width=1.8cm
                    [
                     Amazon intent~\cite{amazon_massive_intent}; Amazon reviews~\cite{amazon_reviews_multi} ;
                     Aya~\cite{singh-etal-2024-aya};
                     Bactrian-x~\cite{li2023bactrianx}; Biblenlp~\cite{biblenlp-corpus-mmteb};
                     Bloom-lm~\cite{le2022bloom}; CC100~\cite{cc100_1}~\cite{cc100_2};
                     CulturaX~\cite{nguyen2023culturax}; GPT-4 Prompts~\cite{GPT-4-Prompts} ;                    
                     Guanaco~\cite{guannaco}; HPLT~\cite{hplt_monolingual_v1_2} ;
                     IWSLT 2017~\cite{iwslt2017}; mC4~\cite{mc4};
                     Mewsli-x~\cite{mewsli-x}; Minds14~\cite{minds14} ;
                     Miracl~\cite{miracl} ; MLDR~\cite{MLDR};
                     MMedC~\cite{MMedC};  MQA~\cite{mqa};
                     Multi-sentiments~\cite{multilingual-sentiments}; Multiconer~\cite{multiconer2-data}~\cite{multiconer2-report};                     Nomiracl~\cite{thakur2023nomiracl}~\cite{thakur2024nomiracl};Open Subtitles~\cite{lison2016opensubtitles2016} ;
                     OSCAR~\cite{OSCAR}; Para-pat~\cite{para_pat} ;
                     Project Gutenberg~\cite{project_gutenberg}; ShareGPT~\cite{ShareGPT52K};
                     SREDFM~\cite{SREDFM}; TED Talks~\cite{ted_talks};
                     TED-talks-iwslt~\cite{ted_talks_iwslt}; Toxi-text~\cite{toxi-text-3M};                    
                    UD~\cite{universal_dependencies};                  
                    Wikiann~\cite{wikiann1}~\cite{wikiann2};
                    Wikipedia~\cite{wikipedia};
                    Wit Base~\cite{witbase}; xP3~\cite{xP3},
                    text width=7.0cm, node options={align=left}
                    ]
                ]
                [Benchmark §\ref{sec:benchmark}, text width=1.8cm
                    [
                    Afrisent~\cite{muhammad2023afrisenti};
                    ASPEN~\cite{razumovskaia2024little} ;
                    BELEBELE~\cite{bandarkar2023belebele};
                    BioInstructQA~\cite{BioInstructQA};
                    Bucc-bitext-mining~\cite{bucc-bitext-mining};
                    Cross-Sum~\cite{bhattacharjee2023crosssum};
                    Crossmodal-3600~\cite{thapliyal2022crossmodal3600};
                    Exams~\cite{hardalov2020exams};
                    Fairlex~\cite{chalkidis2022fairlex};
                    FLORES-200~\cite{nllb2022}~\cite{flores200_2}~\cite{flores200_3};
                    GEOMLAMA~\cite{yin2022geomlama};
                    Humaneval-XL~\cite{peng2024humaneval};
                    M-Hellaswag~\cite{dac2023okapi};
                    M-MMLU~\cite{dac2023okapi};
                    M3Exam~\cite{zhang2023m3exam};
                    M3LS~\cite{verma-etal-2023-large};
                    MARC~\cite{keung2020multilingual};
                    MasakhaNER~\cite{adelani2021masakhaner};
                    Masakhanews~\cite{adelani2023masakhanews};
                    MASSIVE~\cite{fitzgerald2022massive};
                    MaXM~\cite{changpinyo2023maxm};
                    MEGA~\cite{ahuja2023mega};
                    MEGAVerse~\cite{ahuja2024megaverse};
                    Mela~\cite{zhang2024mela};
                    MGSM~\cite{shi2022language};
                    MLQA~\cite{lewis2020mlqa};
                    MMedBench~\cite{qiu2024building};
                    Multi-CoNER~\cite{malmasi2022multiconer};
                    MULTIEURLEXDOC~\cite{chalkidis-etal-2021-multieurlex};
                    Multilingual-Fig-QA~\cite{kabra-etal-2023-multi};
                    NusaX~\cite{winata2023nusax};
                    ODEX~\cite{wang2023executionbased};
                    OPUS-100~\cite{zhang2020improving};
                    Paws-X~\cite{yang-etal-2019-paws};
                    Pmindiasum~\cite{urlana2023pmindiasum};
                    PRESTO~\cite{goel2023presto};
                    SEAHORSE~\cite{clark2023seahorse};
                    Sib200~\cite{adelani2023sib200};
                    SMiLER~\cite{seganti-etal-2021-multilingual};
                    STSB-multi-mt~\cite{huggingface:dataset:stsb_multi_mt};
                    Tatoeba-mt~\cite{tiedemann-2020-tatoeba};
                    Tydip~\cite{srinivasan-choi-2022-tydip};
                    TyDiQA~\cite{clark2020tydi};
                    Universal Dependencies~\cite{universal_dependencies} ;
                    X-CLAIM~\cite{mittal-etal-2023-lost};
                    X-RiSAWOZ~\cite{moradshahi2023xrisawoz};
                    XCOPA~\cite{ponti-etal-2020-xcopa};
                    XCSQA~\cite{lin2021common};
                    xDial-Eval~\cite{zhang-etal-2023-xdial};
                    XGLUE~\cite{liang2020xglue};
                    XL-SUM~\cite{hasan2021xlsum};
                    XNLI~\cite{conneau2018xnli};
                    XQuAD~\cite{artetxe2020cross};
                    XSEMPLR~\cite{zhang2023xsemplr};
                    XStoryCloze~\cite{lin2022few};
                    XTREME~\cite{hu2020xtreme};
                    XTREME-R~\cite{ruder-etal-2021-xtreme};
                    XWinograd~\cite{tikhonov-ryabinin-2021-heads};
                    XCSR~\cite{lin-etal-2021-common},
                    text width=7.0cm, node options={align=left}
                    ]
                ]
			]
		]
	\end{forest}
	\caption{A structured taxonomy of LLMs with multilingualism which categorizes current studies.}
    \label{fig:taxonomy}
    % \vspace{-0.3cm}
\end{figure*}
\section{Preliminary}\label{sec:back}

In this chapter, we first give a general definition of multilingual models, then describe the background of pre-trained language models (PLMs), and finally introduce the paradigm shift from ``Pre-train, Fine-tune'' to ``Pre-train, Prompt, Predict'' in the age of pre-trained language models.

\begin{figure*}[!t]
    \centering
    \includegraphics[width=0.95\linewidth]{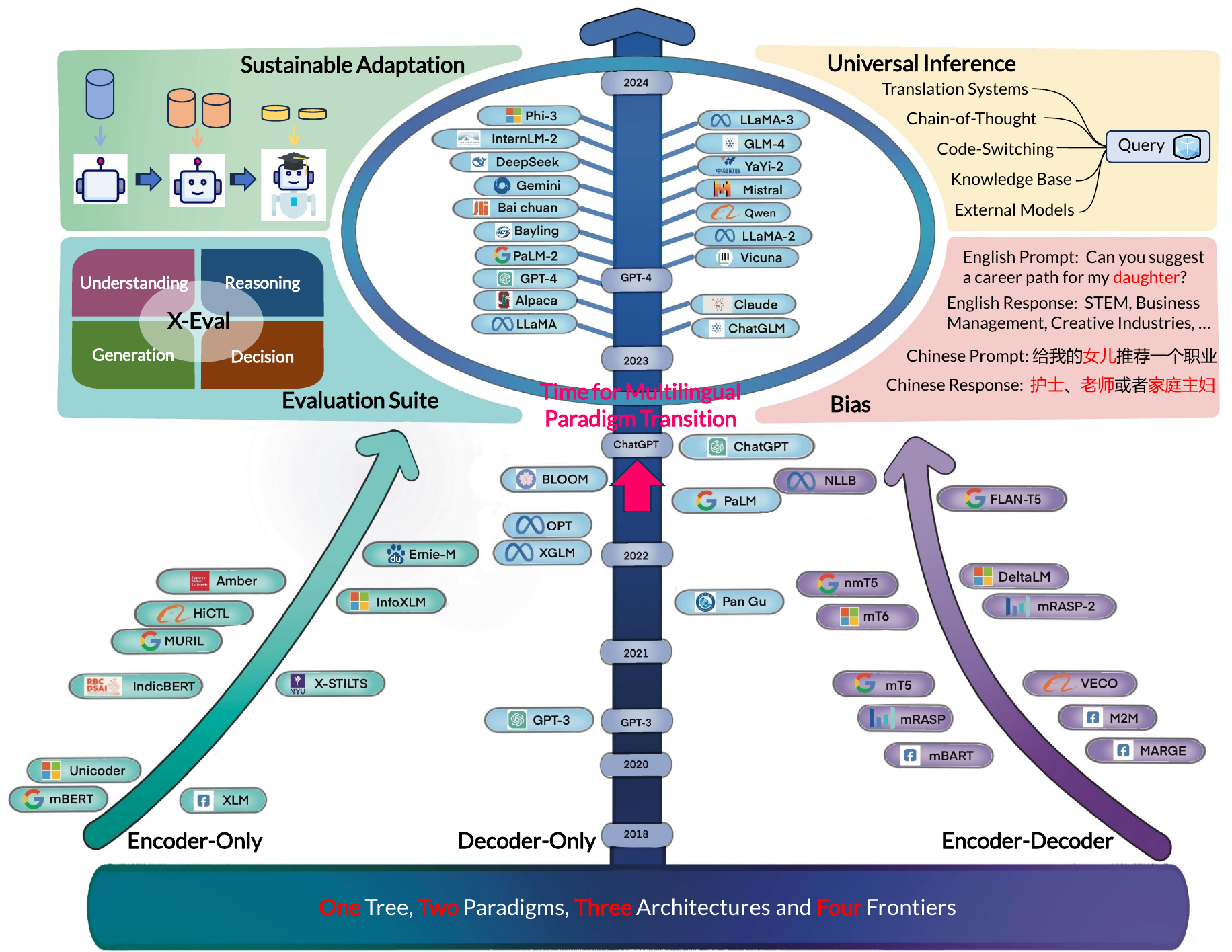}
    \caption{An overview of \textbf{representative} LLMs and mPLMs in recent years. The illustration consists of \textbf{one} tree that shows the transition of \textbf{two} paradigms~(``Pre-train, Fine-tune''$\rightarrow$``Pre-train, Prompt, Predict''), including \textbf{three} model's architectures~(encoder-only, decoder-only, and encoder-decoder) and four new frontiers for LLMs with multilingualism.}
    \label{fig:timeline}
\end{figure*}

\subsection{Multilingual Models}
The longstanding goal of multilingual models is to develop a universal model, capable of providing high-quality performance on any languages and tasks~\cite{huang2023learn}.
Generally, multilingual models are trained to maximize the optimization of a mix of examples drawn from multiple language corpora~\cite{johnson2017google}.
The benefits of multilingual models are mainly twofold. First, multilingual models can handle several languages in a single model, facilitating the knowledge transfer from high-resource languages to related but low-resource languages~\cite{huang2023knowledge}, where even languages that have never been trained before~\cite{zhang2020improving}. Second, a single model can support multiple languages instead of training multiple language-specific models~\cite{lin2021learning,escolano2021bilingual}, reducing the cost of maintaining models.

\subsection{Pre-Trained Language Models}

PLMs are deep neural networks initially trained on extensive unlabeled corpora, then could be adaptable to specific tasks. Existing research demonstrates that PLMs grasp and store substantial linguistic knowledge within their parameters~\cite{min2023recent,li2020sentence,hu2023survey}. Consequently, leveraging PLMs holds promise for enriching language comprehension and enhancing multilingual performance.

%Due to the remarkable success of the Transformer architecture~\cite{vaswani2017attention}, almost all PLMs use it as their backbone. 
Most of the PLMs adopt the remarkable Transformer architecture~\cite{vaswani2017attention} as their backbone.
The existing PLMs can be categorized into three typical architectures, constructed upon the Transformer encoder (e.g., BERT~\cite{devlin2018bert}), decoder (e.g., GPT~\cite{radford2018improving}), and encoder-decoder (e.g., BART~\cite{lewis2020bart}), respectively.
%there are three typical architectures of PLMs like BERT~\cite{devlin2018bert}, GPT~\cite{radford2018improving} and BART~\cite{lewis2020bart} are constructed upon Transformer encoder, decoder and encoder-decoder, respectively. 
With the corresponding architecture, PLMs are usually improved based on training data sizes, parameter sizes and training strategies~\cite{li2022pretrained}.
For instance, GPT-3~\cite{brown2020language} is developed based on the GPT-2~\cite{radford2019language} by expanding the model size~(from 1.5B to 175B) and the scale of training data, which improves performance across various language tasks.
% \modif{Besides, PLMs can be classified into categories like masked LMs, causal LMs, prefix LMs, and encoder-decoder LMs, as elaborated in~\cite{li2022pretrained}.}
Recent research also indicates that scaling up model parameters enhances performance of PLMs~\cite{kaplan2020scaling}, leading to the emergence of large-scale PLMs such as GPT-3 (175B)~\cite{brown2020language}, OPT (175B)~\cite{zhang2022opt}, PaLM (540B)~\cite{chowdhery2023palm} and Switch-Transformers (1.6T)~\cite{fedus2022switch}, boasting billions or even trillions of parameters.
Besides, a new training strategy ``translation language model (TLM)'' is designed specifically for enhancing multilingual capabilities tasks compared to the masked language model (e.g., BERT~\cite{devlin2018bert}), involving translation between multiple languages~\cite{conneau2019cross}.

\subsection{Multilingual Paradigm Transition}
As shown in Figure~\ref{fig:timeline}, multilingual pre-trained
language models (PLMs) are dominated by small language models (i.e., parameters less than 7B). 
They are designed for specific tasks, such as multilingual neural machine translation (NMT)~\cite{johnson2017google,aharoni2019massively,firat2017multi}, multilingual question answering (QA)~\cite{loginova2018towards,loginova2021towards,ruder2021multi} and multilingual reasoning~\cite{lin2021common,fang2022leveraging,bang2023multitask}, etc.
To indicate the language of target tasks, a prepending language token is appended to each source sentence~\cite{johnson2017google,conneau2019cross}. 
Take the multilingual NMT for example, formally, given the source sentence $\mathbf{x'}=(x_1, x_2, ..., x_n)$, where each $x_i$ is a token in the source language, the modified source sentence is represented as $\mathbf{x}=(l_i, x_1, x_2, ..., x_n)$. Here, $l_i$ denotes the target language. Correspondingly, the target sentence is represented as $\mathbf{y}=(y_1, y_2, ..., y_m)$, where each $y_j$ represents a token in the target language.
The probability of a target sentence is given by:
\begin{equation}
p(\mathbf{y}|\mathbf{x};\theta)=\prod^{J}_{j=1}p(y_i|\mathbf{\mathbf{y}_{<j},\mathbf{x};\theta})
\label{equation_1}
\end{equation}
where $\theta$ is a set of trainable parameters, $\mathbf{y}_{<j}$ are the generated words before the $j$-th step. This approach that prepends language identification is suitable for specific tasks and language situations.

However, the current model paradigm has gradually shifted to Artificial General Intelligence (AGI) which possesses the ability like human intelligence to perform a wide range of tasks~\cite{ouyang2022training,achiam2023gpt,yang2023dawn}, including understanding, learning, cognition, communication and others. 
Unlike small language models, which are designed for specific tasks, AGI would have the capacity for generalization and adaptation to novel situations, potentially surpassing human capabilities in various domains. 
With the increasing scale of language models, LLMs improve the performance of downstream tasks compared to small language models~\cite{kaplan2020scaling}, which is closer to the goal of AGI.
For the sake of generality, the task form becomes general tasks instead of specific tasks, and prepending the language is not allowed.
Hence, most existing LLMs typically do not specify whether they function as multilingual models or detail the number of languages they support.
We consider that if a certain amount of multilingual data is used in the training corpus, it can be regarded as a multilingual LLM.
Among the existing models, only a few are explicitly called ``multilingual LLM'' (e.g., BLOOM~\cite{le2022bloom}, BayLing~\cite{zhang2023bayling}, etc), yet there are also other models that possess multilingual capabilities~\cite{jiang2023mistral,wang2023openchat}.
To distinguish LLMs from multilingual LLMs, we propose a definition of ``LLMs with multilingualism''.

\section{Large Language Models with Multilingual Capability}\label{sec:training}
\begin{table*}[!t]
\centering
\resizebox{\columnwidth}{!}{
\begin{tabular}{lcccccc}
\toprule
Name & Release Time & Params & Affiliation & Base  & Available & Support Languages \\

 \midrule

 GPT-3~\cite{brown2020language} & 20-05 & 13B, 175B &  OpenAI & GPT-2 & Closed & - \\ 
 mT5~\cite{xue2020mt5} & 20-10 & 13B & Google & T5 & Open & 101\\
 ByT5~\cite{xue2022byt5} & 21-05 & 13B & Google & T5 & Open & 101\\
 Gopher~\cite{rae2021scaling} & 21-12 & 280B & DeepMind & - & Open & - \\
 LaMDA~\cite{thoppilan2022lamda} & 22-01 & 137B &Google & - & Open & -\\
 OPT~\cite{zhang2022opt} & 22-04 & 175B & Meta & - & Open & -\\
 PaLM~\cite{chowdhery2023palm} & 22-04 &8B, 62B, 540B &Google & - & Closed & 124 \\
 mGPT~\cite{shliazhko2022mgpt} & 22-04 & 13B & - & GPT-3 & Open &61\\
 BLOOM~\cite{le2022bloom} & 22-07 & 176B & BigScience & - & Open & 46\\
 FLAN-T5~\cite{chung2024scaling} & 22-10 & 11B & Google & T5 & Open & 60\\
 FLAN-PaLM~\cite{chung2024scaling} & 22-10 & 8B, 62B, 540B & Google & PaLM & Open & 60\\
 ChatGLM~\cite{zeng2022glm} & 22-10 & 130B & ZHIPU & GLM & Open & 2 \\
 GPT-3.5~\cite{ouyang2022training} & 22-11 & - & OpenAI & GPT & Closed & - \\
 XGLM~\cite{lin2022few} & 22-11 & 7.5B & Meta AI & - & Open & 30\\ 
 LLaMA~\cite{touvron2023llama} & 23-02 & 7B, 13B, 33B, 65B & Meta & - & Open &-\\
 GPT-4~\cite{achiam2023gpt} & 23-03 & - & OpenAI & - & Closed & -\\
 Alpaca~\cite{taori2023stanford} & 23-03 & 7B & StandFord & LLaMA & Open & - \\
 PANGU-$\sum$~\cite{ren2023pangu}  & 23-03 & 1085B & Huawei & - & Closed & 26 \\
 Pythia~\cite{biderman2023pythia} & 23-04 & 12B  & EleutherAI & - & Open &- \\
 ParroT~\cite{jiao2023parrot} & 23-04 & 7B & Tencent  & Alpaca & Open & - \\
 BigTrans~\cite{yang2023bigtrans} & 23-05 & 13B & - & LLaMA & Open & 102 \\
 PaLM-2~\cite{anil2023palm} & 23-05 & 340B & Google & PaLM & Closed & - \\
 Vicuna~\cite{chiang2023vicuna} & 23-06 & 13B & LMSYS & LLaMA & Open & - \\
 InternLM~\cite{team2023internlm}  & 23-06 & 104B & Shanghai AI Laboratory & - & Open & -\\
 BayLing~\cite{zhang2023bayling} & 23-06 & 7B, 13B & ICT/CAS & - &Open & - \\
 PolyLM~\cite{wei2023polylm} & 23-07 & 13B & DAMO Academy & -&Open & 18\\
 LLaMA-2~\cite{touvron2023llama} & 23-07 & 7B, 13B, 34B, 70B & Meta & LLaMA & Open &-\\
 Baichuan-2~\cite{yang2023baichuan} & 23-09 & 7B, 13B & Baichuan & - & Open & - \\
 Qwen~\cite{bai2023qwen} & 23-09 & 7B, 14B & Alibaba & - & Open & - \\
 OpenChat~\cite{wang2023openchat} & 23-09 & 7B, 13B & Tsinghua & LLaMA &Open & - \\
 Mistral~\cite{jiang2023mistral} & 23-10 & 7B & Mistral & - & Open & - \\
 Gemini~\cite{team2023gemini} & 23-12 & - & DeepMind & - & Closed & 40+ \\
 TigerBot~\cite{chen2023tigerbot} & 23-12 & 7B, 13B, 70B, 180B & Tiger & - & Open & - \\
 YAYI-2~\cite{luo2023yayi} & 23-12 & 30B & IACAS & - & Open & - \\
 DeepSeek~\cite{bi2024deepseek} & 24-01 & 67B & DeepSeek AI & - & Open & - \\
 GLM-4~\cite{glm4} & 24-01 & - & ZHIPU & GLM & Close &-\\
 Orion~\cite{chen2024orion} & 24-01 & 14B & OrionStar&- & Open & 8+ \\
 TeleChat~\cite{abdin2024phi} & 24-01 & 7B, 12B & China Telecom & - & Open & - \\
 Aya~\cite{ustun-etal-2024-aya} & 24-02 & 13B & Cohere & mT5 & Open & 101 \\
 Claude3~\cite{anthropic2024claude} & 24-03 & - & Anthropic & - & Closed & -\\
 InternLM-2~\cite{cai2024internlm2} & 24-03 & 7B, 20B & Shanghai AI Lab & Open & - \\
 LLaMA-3~\cite{llama3modelcard} & 24-04 & 8B, 70B & Meta & -&Open & 30+ \\
 Phi-3~\cite{abdin2024phi} & 24-04 & 7B, 14B & Microsoft & - & Open & - \\
\bottomrule
\end{tabular}}

\caption{An overview of representative LLMs (trainable parameters greater than 7B) that have certain multilingual capabilities in recent three years, including their release time, parameters, affiliation, base model, availability, and support languages.}
\label{table:model}
\end{table*}

% \begin{table*}[!t]
% \centering
% \resizebox{\columnwidth}{!}{
% \begin{tabular}{ccccccc}
% \toprule
%  Name & Languages Coverage & Data Size &  Data Sources & Affiliation & Other \\

%  \midrule
%  OSCAR & 166 & 6.3TB & Common Crawl & University of Orego & -\\
%  mC4 & 101 & 9.7TB & Common Crawl & Google & mT5\\
%  CulturaX & 167 &  27 TB & mC4, OSCAR & -\\

% \bottomrule
% \end{tabular}}

% \caption{An overview of training dataset in recent 3 years.}
% \label{table:model}
% \end{table*}

Based on the training paradigm, this survey divides existing multilingual LLMs into two categories: (1) the foundational LLMs trained from scratch and (2) the continually trained LLMs on top of the foundational models.
%the first category consists of foundational LLMs trained from scratch, while the second category comprises LLMs trained continually based on foundational models. 
In Table~\ref{table:model}, we present the statistics of the representative LLMs with certain multilingual capabilities in the past three years.

\subsection{Training from Scratch}
\label{sec:train-from-scratch}
%A simple yet effective approach to obtain multilingual models is to first mix all available multilingual data, and then balance the importance of each language in the model through the training paradigm of monolingual pre-training models combined with language sampling algorithms.
To obtain a language model with multilingual capability, a common practice is to leverage all available data in different languages for training. The language sampling algorithms are usually applied to control the importance of each language.
As shown in Figure~\ref{fig:timeline}, multilingual pre-trained language PLMs before 2022 mainly consisted of two structures: (1) encoder-only designed for natural language understanding tasks and (2) encoder-decoder designed for natural language generation tasks.
The parameters of these PLMs are not comparable with the existing LLMs. Recent studies show that according to the scaling law~\cite{kaplan2020scaling,muennighoff2024scaling,aghajanyan2023scaling}, the scale of the parameter has significant impacts on the performance of the models, i.e., the larger models lead to better performance. 
The researchers also observe that when the parameter scale of the language models exceeds a certain level, they not only achieve significant performance improvements but also find some special capabilities not observed within small-scale PLMs~\cite{kaplan2020scaling,muennighoff2024scaling}. 
To distinguish the difference in parameter scale, the research community specifies the term ``LLMs'' compared with PLMs based on a significant scale~\cite{zhao2023survey}.

The multilingual capabilities of PLMs are primarily obtained by combining large amounts of multilingual data in the pre-training stage.
Training on multilingual corpora through pre-training objectives such as mask language modeling (MLM)~\cite{devlin2018bert,ouyang2020ernie} and next token prediction~\cite{radford2018improving,shliazhko2022mgpt,brown2020language,achiam2023gpt,touvron2023llama,zeng2022glm,chowdhery2023palm} endows these models with multilingual capabilities (e.g., BERT$\rightarrow$mBERT and T5$\rightarrow$mT5).
In particular, mLongT5~\cite{uthus2023mlongt5} integrates the mT5 training dataset and the architecture of LongT5 and replaces the principle sentence generation (PSG) with mixture-of-denoisers (MoD) making it ideal for multilingual pre-training.
% (mBERT~\cite{devlin2018bert} Ernie-M~\cite{ouyang2020ernie} mGPT~\cite{shliazhko2022mgpt} GPT3~\cite{brown2020language} GPT3.5 GPT4~\cite{achiam2023gpt}  Mistral~\cite{jiang2023mistral}	mT5~\cite{xue2020mt5}	llama~\cite{touvron2023llama} OPT~\cite{zhang2022opt}	BLOOM~\cite{le2023bloom}	GPT-NeoX\cite{black2022gpt} 	GLM~\cite{zeng2022glm}	PaLM~\cite{chowdhery2023palm}	Skywork~\cite{wei2023skywork} Qwen~\cite{bai2023qwen}	Baichuan2~\cite{yang2023baichuan}	InternLM~\cite{team2023internlm} TigerBot~\cite{chen2023tigerbot} Yayi2~\cite{luo2023yayi}).
% Moreover, OPT~\cite{zhang2022opt} is not intentionally trained to be multilingual, but the multilingual capabilities anecdotally found in succeeding in simple translations in German, Spanish, French, and Chinese.

In addition to constructing multilingual data in the pre-training stage, the cross-lingual capabilities can also be improved by some specific methods (e.g., XLM~\cite{conneau2019cross}, XLM-R~\cite{conneau2019unsupervised}, XLM-V~\cite{liang2023xlm}, XGLM~\cite{lin2022few}, PaLM-2~\cite{anil2023palm}, PolyLM~\cite{wei2023polylm}, mLongT5~\cite{uthus2023mlongt5}).
% In XLM~\cite{conneau2019cross}, it is proven that causal language modeling and MLM approaches provided powerful cross-lingual features that can be used for pre-trained models. 
For instance, the series of XLM~\cite{conneau2019cross,conneau2019unsupervised,liang2023xlm} proposes a translation language modeling (TLM) pre-training task, utilizing parallel multilingual data to improve cross-lingual model pre-training, and it extends MLM by using batches of parallel sentences instead of consecutive sentences.
% Based on the cross-language methods used in XLM and RoBERTa, XLM-R~\cite{conneau2019unsupervised} further increases the size of the training data and achieves better performance.
% Furthermore, XLM-V~\cite{liang2023xlm} scales the multilingual vocabulary to a large size by reducing token sharing between languages and allocating appropriate vocabulary capacity for each language. This refinement led to superior performance over XLM-R across all tasks.
Moreover, the sampling strategies are important to balance the proportion of multiple languages in training data, which facilitates models to learn multilingual representation~\cite{lin2022few,anil2023palm,chung2023unimax}.
% XGLM~\cite{lin2022few} up-samples the low-resource languages to render a more balanced language distribution. 
% And the proportion of non-English training data is increased in PaLM-2~\cite{anil2023palm}, which facilitates to learn multilingual tasks. 
% In addition, PaLM-2 is trained on parallel data covering hundreds of languages, the incorporation of parallel multilingual data significantly improves the multilingual ability of the model.  
PoLyLM~\cite{wei2023polylm} integrates bilingual alignment within the training corpus and implements a curriculum learning strategy to maintain a balanced representation of various languages during the pre-training phase.
The model proposes a multilingual self-instruct approach, enabling the automatic generation of diverse multilingual instructions for fine-tuning the model.

An inevitable part of the multilingual learning paradigm for PLMs is to deal with vocabulary~\cite{dabre2020survey,garcia2021towards,huang2022entropy}.
The ``out-of-vocabulary (OOV)'' tokens hurts the translation performance naturally for multilingual learning~\cite{zhang2022robust}.
To alleviate this issue, ByT5~\cite{xue2022byt5} is a token-free model that operates directly on raw text (bytes or characters) which simplifies the process of different languages.
% umT5~\cite{chung2023unimax} introduces a language sampling strategy, delivering more uniform coverage of high-resource languages while avoiding overfitting on low-resource languages.
% mLongT5~\cite{uthus2023mlongt5} integrates the mT5 training dataset and the architecture of LongT5 and replaces the principle sentences generation (PSG) with mixture-of-denoisers (MoD) making it ideal for multilingual pre-training.

\subsection{Continual Training}
\label{sec:cl}
Another way to improve the multilingual capabilities of LLMs is continual training, which involves updating the model with new data rather than training a model from scratch. 
The main idea is to transfer knowledge from the foundation model and inject additional multilingual capability via the updated data, which does not require excessive computational or data resources and thus reduces the training cost.

%By transferring knowledge from the base model and enhancing its capabilities with additional data, this approach does not require excessive computational or data resources. 
%It also helps save training costs, enabling more researchers to engage in training LLMs.

This kind of training method utilizes existing monolingual or multilingual models for enhancing the specific languages ability (e.g., Chinese-LLaMA~\cite{cui2023efficient}, Chinese-Alpaca~\cite{cui2023efficient}, Chinese-Mixtral~\cite{Chinese-Mixtral-8x7B}, CPM-2~\cite{zhang2021cpm}, LLaMAntino~\cite{basile2023llamantino}, FinGPT~\cite{luukkonen2023fingpt}, Sabiá~\cite{pires2023sabia}, Bode~\cite{garcia2024introducing}) or extending to a wider range of languages, such as BigTrans~\cite{yang2023bigtrans}, Glot500~\cite{imanigooghari2023glot500} and others~\cite{ebrahimi2021adapt,alabi2022adapting,muennighoff2022crosslingual}.
A branch of work is to directly continually train the foundation models without any other techniques~\cite{luukkonen2023fingpt,yang2023bigtrans,alabi2022adapting}.
For instance, Muennighoff~\etal~\cite{muennighoff2022crosslingual} introduce xP3, a corpus consisting of tasks in 46 languages, and utilizes a cross-lingual multitask fine-tuning approach to fine-tune the BLOOM and mT5 on the newly created corpora, along with the English-only P3 corpus, resulting in the production of BLOOMZ and mT0Z.
Considering the importance of vocabulary for multilingual models, prior work employs vocabulary extension to improve performance for low-resource languages~\cite{imanigooghari2023glot500}.
Another branch of work is to introduce external parameters and strategies to learn new languages for multilingual models~\cite{ebrahimi2021adapt}.
% In particular, Chinese-LLaMA~\cite{cui2023efficient} extends the original LLaMA vocabulary with an additional 20,000 Chinese tokens to improve the Chinese ability, then employ the Low-Rank Adaptation (LoRA) approach to facilitate efficient training on Chinese corpus.
For instance, Chinese-LLaMA~\cite{cui2023efficient} integrates an additional 20K Chinese tokens into the original vocabulary to improve the Chinese ability, then employs the Low-Rank Adaptation (LoRA) approach to facilitate efficient training on Chinese corpus. 
CPM-2~\cite{zhang2021cpm} divides the pre-training into three phases, including the monolingual pre-training, bilingual pre-training, and mixture-of-expert pre-training. 
Multi-phase continual training, incorporating knowledge inheritance, offers a substantial reduction in computation costs compared to training from scratch.
% FinGPT~\cite{luukkonen2023fingpt} further refines the BLOOM model through continued pre-training on a combination of its original training data and Finnish language data, resulting in the creation of BLUUMI, a 176 billion-parameter model.
% To broaden the language support of LLMs, BigTrans~\cite{yang2023bigtrans} augments LLaMA with multilingual translation capabilities across more than 100 languages, which is achieved by further training the model with an extensive parallel dataset covering 102 natural languages.
% Glot500~\cite{imanigooghari2023glot500} employs vocabulary extension and continual training based on XLM-R with the MLM objective improving performance for low-resource languages.
% Ebrahimi~\etal~\cite{ebrahimi2021adapt} leverage the adapters on New Testament, a resource available for over 1600 languages.
Moreover, Wang~\etal~\cite{wang2022expanding} systematically investigate strategies aimed at reducing dependence on traditional language resources by leveraging bilingual lexicons to fine-tune pre-trained multilingual models for underrepresented languages.
% Alabi~\etal~\cite{alabi2022adapting} adapt XLM-R to 17 African languages using multilingual adaptive fine-tuning methods.

\subsection{Limitations and Future Directions on Training Paradigm}
Although the existing LLMs demonstrate a certain superiority in non-English language, either by training from scratch or continuing training on top of a foundation model with extended language data, these approaches based on text knowledge only still face challenges in addressing more complex scenarios as listed below.

%Whether training a new model from scratch or continually training a foundational model with new data for language extension, LLMs demonstrate certain superiority on non-English language. However, these approaches based on text knowledge of new languages still face challenges in addressing more complex scenarios.

\textbf{Low-Resource scenarios.} 
The current training paradigm relies on large-scale data for the target language and task, e.g., supervised fine-tuning based on annotated data or continual pre-training based on unsupervised data.
Both approaches require a substantial amount of data collection and thus constraints the effectiveness under low-resources or multi-scenario demands.
%Current training paradigm still relies on large-scale data for the target language and task. It is supervised fine-tuning based on annotated data or continual pre-training based on unsupervised data, both requiring a substantial amount of data collection. The effectiveness under resource constraints and multi-scenario demands remains to be explored.

\textbf{Knowledge conflict.} 
As the foundation model expands the supported languages, its multilingual knowledge accumulates into the model through continuous new data available. 
The newly acquired knowledge from the new data would interfere with old knowledge stored in the parameters, and thus lead to a rapid decline in model performance. This is known as catastrophic forgetting~\cite{kirkpatrick2017overcoming}, which might result in even completely forgetting previously learned knowledge.
%As the foundation model continually expands its language support, multilingual knowledge accumulates into the model through continuous data stream training. When the model continually acquires knowledge from new data, the newly learned knowledge interferes with old knowledge, leading to a rapid decline in model performance, even completely covering or forgetting previously learned knowledge, resulting in catastrophic forgetting~\cite{kirkpatrick2017overcoming}.

\textbf{Knowledge type.} 
The existing multilingual NLP datasets, such as XNLI~\cite{conneau2018xnli} and XQuAD~\cite{artetxe2020cross}, are constructed based on the translated texts from the original English version datasets~\cite{rajpurkar2016squad,yin2021docnli}. This is because of the high cost of obtaining multilingual data, especially in languages with limited annotations/translators.
As a result, the essence of existing datasets lies in translation-based aligned knowledge, lacking cultural and domain knowledge in non-English contexts. This type of knowledge learned by the models reduces the distribution shift ability in the real-world scenario, leading to sub-optimal performance in specialized fields (e.g., finance and legislation).

%Due to the high cost of obtaining multilingual data, especially in languages with limited annotations, existing multilingual datasets, such as XNLI~\cite{conneau2018xnli} and XQuAD~\cite{artetxe2020cross}, are based on translated texts from English datasets. Their essence still lies in translation-based aligned knowledge, lacking cultural and domain knowledge in non-English contexts. The translation-based relevant knowledge reduces the distribution shift in the practical scenario, performing suboptimally in academic datasets or specialized fields such as finance and legislation.

%Despite the aboved methods, existing LLMs perform suboptimally, especially on non-Latin languages and those with limited resources. To address the inadequacy of multilingual capabilities of current LLMs, 
To address the aforementioned problems and aim to improve the multilingual capabilities of existing LLMs,
this survey proposes the following points for future exploration:

\begin{itemize}
    \item \textit{Training Strategies.} LLMs are data-driven and follow the scaling law which suggests that increasing the model size can guarantee better generalization, higher accuracy, and improved ability to capture complex patterns from data. However, such a paradigm results in the existing training strategies primarily focusing on expanding diverse data, while overlooking other multilingual training techniques, such as sampling algorithms~\cite{arivazhagan2019massively,wang2020balancing,wu2021uncertainty}, training objective optimization~\cite{yang2021multilingual,pan2021contrastive}, representation space~\cite{oncevay2020bridging,cheng2022multilingual,stap2023viewing}, and others~\cite{zhang2020improving,wang2022parameter}. 
    The potential of these traditional multilingual training strategies has not been further explored in the current data effort paradigm in existing multilingual literature.
    %Due to the performance improvements brought by these techniques being replaced by data effort, traditional multilingual training strategies have not been further explored. This survey suggests that multilingual researchers still need to further optimize for the multilingual representation space.
    \item \textit{Architecture.} Existing LLMs mainly rely on monolingual backbone models, such as the standard Transformer, rather than undergoing architectural changes specifically targeting multilingual capabilities. %instead, they mainly rely on monolingual backbone models, such as the standard Transformer. 
    For the other specific tasks, for example, LLMs designed for long texts introduce variations in positional encoding~\cite{beltagy2020longformer,su2024roformer}, and those designed for multi-modal tasks enhance alignment between different modalities~\cite{peng2023kosmos,liu2024visual} have been explored in some extend. Some advanced architectures like Mamba~\cite{gu2023mamba} have also shown remarkable performance and may become the backbone architecture for the next generation of LLMs. 
    Thus, for multilingual tasks, it is essential to tailor the architecture with specific variations rather than merely augmenting data on the standard Transformer.
    \item \textit{Sustainability.} When trying to utilize new data to enhance existing LLMs, a practical approach is to allow LLMs to continually learn from the updated data, akin to how the human brain operates, instead of creating new models~\cite{huang2022entropy,huang2023learn,huang2023knowledge}. Essentially, when new language data comes up, we aim to continually extend the language support of LLMs and improve their corresponding language capabilities while preserving the performance of those languages that the model has already demonstrated good performance, as forming a \textit{lifelong/incremental learning} paradigm.
\end{itemize}
\section{Multilingual Inference Strategies}\label{sec:infer}

This chapter investigates the development of robust multilingual inference strategies that are crucial for deploying language models across varied linguistic environments.

% We explore five principal dimensions of multilingual inference and evaluation: multilingual chain-of-thought inference, prompt strategies and alignment, code-switching capabilities, and systematic evaluation methods. Each dimension will be elaborated upon in the subsequent sections.

\subsection{Direct Inference in Multilingual Models}
\begin{table*}[ht]
    \centering
    \resizebox{\columnwidth}{!}{
    \begin{tabular}{cccccccc}
    \toprule
         \multirow{2}{*}{Model} &  \multirow{2}{*}{Inference}&   XCOPA&  XStoryCloze&  BELEBELE&  XLSum&  XQuAD&   TyDiQA-GP\\
         \cmidrule(r){3-8}
         &  &  Acc&  Acc&  Acc&  Rouge-L&  F1&   F1\\
         \midrule
         \multirow{2}{*}{PaLM2-S$^\dagger$} &  En-Pivot&  87.3&   96.4&  76.7&    23.7&   67.2&  81.6\\
         &  Direct&  \textbf{89.7}&  \textbf{96.8}&  \textbf{77.8}&  \textbf{26.8}&   \textbf{70.7}&  \textbf{83.8}\\
         \multirow{2}{*}{PaLM2-L$^\dagger$} &   En-Pivot&  89.6&   97.8&  84.3&   25.4&  78.7&   81.0\\
         &  Direct&  \textbf{93.4}&   \textbf{99.1}&  \textbf{88.4}&  \textbf{28.0}&  \textbf{85.9}&  \textbf{83.0}\\
         \bottomrule
    \end{tabular}}
    
    \caption{The performance of LLMs using different inference strategies in the multilingual scenario. The best score is highlighted in \textbf{bold}. ``$\dagger$'' indicates the results are quoted from~\cite{intrator2024breaking}.}
    \label{tab:infer_1}
\end{table*}
With the advent of LLMs, efforts to enhance the diversity of training corpora lead to the inclusion of multiple languages alongside English. 
This approach endows LLMs with inherent multilingual capabilities, which enables these models to engage in direct multilingual inference.
% , expanding their applicability across different linguistic contexts without the need for pre-translation.
%It allows for the processing of
In other words, the models can process the input in their native language without requiring translation into a pivot language.
This capability is valuable as it maintains the authenticity of the linguistic and cultural nuances present in the original text, preventing semantic distortion or information loss that might otherwise occur during translation. 
% However, early LLMs still exhibits considerable limitations. For instance, models like BLOOM~\cite{le2022bloom} shows potential in multilingual inference but are hampered by constrained inference performance.
% LLaMA~\cite{touvron2023llama} is primarily trained on English data, performing poorly in other languages, and models such as Baichuan~\cite{yang2023baichuan}, InternLM~\cite{team2023internlm}, and ChatGLM~\cite{du2022glm} are trained exclusively on Chinese and English.

Recently, significant advancements have been made in enhancing the multilingual capabilities of LLMs, greatly increasing the number of languages covered and the performance in non-English inference. 
Advanced models such as GPT-4~\cite{achiam2023gpt} and PaLM-2~\cite{anil2023palm} demonstrate remarkable multilingual capabilities and support hundreds of languages.
The support of multiple languages enables direct multilingual inference to become feasible and come up with a current focal point of LLM research.
%not only feasible but also a current focal point of LLM research.
As shown in Table~\ref{tab:infer_1}, we investigate the comparison between direct inference and the pre-translation method which translates the prompt into a high-resource language (e.g., English and Chinese) before inference.
The direct inference achieves better performance compared to the pre-translation method based on both PaLM2-S and PaLM2-L.
The result demonstrates that these two LLMs have multilingual capabilities without the need to use English as a pivot for other language tasks.
Moreover, without the requirement of the translation step, the direct inference approach reduces computational overhead and simplifies the processing pipeline with higher efficiency. %enabling faster response times and reduced resource consumption.
The observations also confirm the benefits of direct inference, including the preservation of linguistic authenticity, enhanced processing efficiency, and improved performance in low-resource languages.

\subsection{Pre-Translation Inference}
\label{sec:trans}
The direct inference may not work for all LLMs, depending on their multilingual capacities.
The existing LLMs usually perform better on those high-resource languages than the low-resource ones, because of the imbalance ratio within the training data.
%Due to extensive training data in high-resource languages, the abilities of these languages are better than those on low-resource languages.
To enhance the performance on low-resource languages, pre-translation inference standardizes the input with various languages by translating them into a pivot high-resource language (e.g., English or Chinese) before querying the LLMs, which is based on the proficiency of the pivot language within the LLMs~\cite{huang2023not,qin2023cross,etxaniz2023multilingual}.
% Recent work {CITE NEEDED https://arxiv.org/abs/2403.10258}has explored the performance comparison between direct inference using native languages and inference after translating into English in large language models for multilingual tasks.
% The practice of pre-translating inputs to English before inference with LLMs leverages models' proficiency in English {CITE NEEDED ArXiv:2305.07004; Cross-lingual Prompting: Improving Zero-shot Chain-of-Thought Reasoningacross Languages. ; ArXiv:2308.01223}due to extensive training data in this language. This method standardizes inputs across languages by translating them into English, aiming to enhance model performance by reducing linguistic inconsistencies.
%However, the limitation of this approach is obvious. 
However, the guarantee of this method relies on high-quality translation services available, which are not necessarily true for most languages. 
As a result, the translation errors would accumulate and the final output of LLMs would become wrong.
Besides, translation on a pivot language obscures the cultural and linguistic nuances of the original text, which might also lead to inaccurate results.
%On one hand, the accuracy of translation is critical. 
%And translation errors lead to inaccuracies in the final output. 
%On the other hand, translation on a pivot language obscures the cultural and linguistic nuances of the original text. 
%This method assumes high-quality translation services are available, which may not be the case for many languages.
% Liu~\etal~\cite{liu2024translation} explore the performance comparison between direct inference using native languages and inference after translating into English in LLMs for multilingual tasks. 
% Although translation can enhance the multilingual reasoning performance of English-centric LLMs, for high-resource languages and advanced LLMs, reasoning in native languages is usually more effective. 
% Moreover, for tasks that require a deep understanding of culture and language, using the original language directly is also more effective. The paper emphasizes the importance of developing true multilingual models, especially for handling diverse and culturally related tasks.
As an exploratory work, Liu~\etal~\cite{liu2024translation} explore the performance comparison between direct inference using native languages and inference after translating into English for multilingual tasks.
Although translation can enhance the multilingual reasoning performance of English-centric LLMs, for high-resource languages and advanced LLMs, reasoning in native languages is more effective. 
% Table~\ref{tab:infer_2} illustrates such a phenomenon. We can observe that the ``Trans-Google'' is better than the ``Trans-NLLB'', which is because the translation accuracy of Google is higher than that of NLLB and results in better final performance from LLMs.
% Besides, we can see the performance improvement of pre-translation inference \modif{(e.g., the performance between ``XLT'' and ``Trans-Google'')} based on the ChatGPT is weaker than that of the Mistral and LLaMA-2.
Besides, pre-translating to English is a practical approach in terms of current LLMs as they are predominantly trained on English data from their pilot experimental results.
However, with the development of direct multilingual inference, the intermediary step may not be necessary, which could allow more authentic interactions with LLMs under multilingual scenarios.

% Recent work {CITE NEEDED https://arxiv.org/abs/2403.10258}has explored the performance comparison between direct inference using native languages and inference after translating into English in large language models for multilingual tasks. Studies have found that although translation can enhance the multilingual reasoning performance of English-centric LLMs, for high-resource languages and advanced LLMs, reasoning in native languages is usually more effective. Moreover, for tasks that require a deep understanding of culture and language, using the original language directly is also more effective. The paper emphasizes the importance of developing true multilingual models, especially for handling diverse and culturally related tasks.

\subsection{Multilingual CoT}
\label{sec:cot}
The Chain of Thought (CoT) is an effective approach to enhance the performance of LLMs in complex reasoning~\cite{sap2020commonsense,yu2023nature,liu2023logiqa}, which has been extensively explored in existing studies, mainly focusing on English~\cite{wang2022self,lyu2023faithful,wei2022chain}.
Liu~\etal~\cite{liu2024translation} propose several inference strategies to investigate the effectiveness of prompting the LLMs in the multilingual scenario.
For instance, ``Native-CoT'' indicates that asking questions in the native language with the instruction ``Let’s think step by step.'' is translated into the native language. 
``En-CoT'' indicates that asking questions in the native language
but instructing reasoning in English with ``Let’s think step by step in English.'' and ``XLT''~\cite{huang2023not} indicates that translating questions into English and solving them step-by-step. 
``Trans'' indicates that using the translation systems to convert questions into English and then solving them step-by-step. 
The experimental results demonstrate that the overall performance of the CoT instruction in English (En-CoT) is better than the one with instruction in native languages (Native-CoT).
The results demonstrate that the CoT is still effective while using English to form the instructions for non-English queries.
%And applying CoT to other languages remains at the forefront of research. 
On the other hand, multilingual CoT attempts to enhance the reasoning capabilities of LLMs across multiple languages~\cite{shi2022language}. 
The multilingual CoT approach is especially beneficial for complex reasoning tasks deeply embedded in specific cultural contexts, enabling more natural and intuitive problem-solving~\cite{kim2023cot,suzgun2023challenging}.
The common practice of multilingual CoT is to prompt the LLMs to establish a step-by-step reasoning process in the original language of queries, which can preserve linguistic and cultural nuances. 
% \modif{It is especially beneficial for complex reasoning tasks deeply embedded in specific cultural contexts, enabling more natural and intuitive problem-solving~\cite{kim2023cot,suzgun2023challenging}.}
In addition, the results corresponding to different sizes of LLaMA-2-Chat attend to examine the impact of scaling law with different multilingual inference strategies.

The understanding of the syntax and semantics of various languages, such as idiomatic expressions and culture-specific references is challenging for existing LLMs. 
A recent study~\cite{chai2024xcot} introduces a cross-lingual CoT reasoning instruction fine-tuning framework, which randomly replaces language fragments with those from low-resource languages, and mixes the original and target languages within a single query.
Besides, this study creates multilingual CoT instruction training data, which can be used to supervise fine-tuning LLMs to reduce the performance gap across different languages.
However, developing datasets for training and evaluating multilingual CoT capabilities requires a robust representation of linguistic diversity and detailed language-specific nuances, which should be further addressed.

% The CoT approach has been extensively explored and studied in English{CITE NEEDED}, showing that its use in LLMs can significantly enhance model performance in complex reasoning tasks{CITE NEEDED} However, applying this method to other languages remains at the forefront of research. 

% Recent research {CITE NEEDED https://arxiv.org/abs/2401.07037} has introduced a cross-lingual CoT reasoning instruction fine-tuning framework. It uses CoT datasets from high-resource languages, such as English, randomly replaces language fragments within with those from low-resource languages, and mixes the original and target languages within a single query. This creates multilingual CoT instruction training data, which is used to supervise fine-tuning of  LLMs. This approach aims to reduce the performance gap of LLMs across different languages.

% {TABEL NEEDED https://arxiv.org/abs/2401.07037}

\subsection{Code-Switching}
\label{sec:cs}
\begin{table*}[!t]
    \centering
    \small
    \begin{tabular}{cccccc}
    \toprule
        \multirow{2}{*}{Module} & \multirow{2}{*}{Model} &   SA~\cite{aguilar2020lince} & LID~\cite{aguilar2020lince}& MT~\cite{srivastava2022overview} &  Summarization~\cite{mehnaz2021gupshup} \\
         & & Es-En (F1)  & Hi-En (F1) & Hi-En (BLEU) & Hi-En (ROUGE-L) \\
         \midrule
         \multirow{3}{*}{0-shot} & XGLM-7.5B & 68.52  &  0.27 & 1.43 & 5.92 \\
         & LLaMA-7B & 51.28  &  0.44 & 1.44 & 3.37 \\
         & GPT-3.5-turbo & 75.64  &  78.17 & 27.64 & 25.07 \\
         \midrule
        \multirow{3}{*}{5-shot} & XGLM-7.5B & 61.06  & 16.91 & 3.28 & 5.18 \\
         & LLaMA-7B & 58.77  & 15.55 & 5.14 & 6.01 \\
         & GPT-3.5-turbo & 76.21  &  80.19 & 28.90 & 26.77 \\
         \midrule
        \multirow{2}{*}{SFT} & XGLM-7.5B & 80.32  & 80.06 & 28.11 &- \\
         & LLaMA-7B & 77.21  &  55.32 & 16.58 & - \\
 \bottomrule
    \end{tabular}
    \caption{The overall results on code-switching benchmark for LLMs. We report the 0-shot, 5-shot, and SFT performance. SFT only adapts to open-resource LLMs. ``SA'', ``LID'', and ``MT'' denote the Sentiment Analysis, the Language Identification, and the Machine Translation, respectively.}
    \label{tab:infer_3}
\end{table*}

Code-switching refers to the phenomenon where communicators switch between two or more languages during linguistic interactions, based on contextual needs. This phenomenon is common in bilingual or multilingual communities, especially in spoken communication~\cite{zhang2023speak,huang2024zero}. For example, in conversations involving Chinese-English or Spanish-English direction, speakers may freely switch between languages depending on the fluency, precision of expression, or emotional needs of the dialogue~\cite{dougruoz2021survey,winata2021multilingual,aguilar2020lince}. Solving code-switching texts is an important and challenging task as the language IDs are not specified before LLMs inference~\cite{zhang2023multilingual,koto2024zero}.

As shown in Table~\ref{tab:infer_3}, we investigate the performance of LLMs on code-switching tasks.
The results demonstrate that open-source LLMs cannot deal with the code-switching problem that needs to process multiple languages simultaneously under a single query.
Multiple languages increase the complexity of processing and understanding language for models and LLMs need to continue to be supervised fine-tuned to accommodate simultaneous processing in this scenario.
In addition, to improve the ability to deal with code-switching texts, researchers develop new pre-training techniques to enhance the understanding capacity of this linguistic phenomenon in LLMs~\cite{peng2023prompting}.
For instance, Das~\etal~\cite{das-etal-2023-improving} modify the MLM~\cite{devlin2018bert} to enhance model reasoning performance in code-switching environments. By employing code-switching corpora, it masks tokens at the boundaries between two languages, thus forcing the model to learn semantic nuances at the points of language transition. Additionally, this work introduces modifications to the model structure by adding residual connections from intermediate layers to the final layer. An auxiliary loss function based on the representations of intermediate layers is also proposed, which enhances the cross-lingual ability to understand multiple languages in a single sentence.
However, the cost of annotated code-switching data is expensive and it is challenging to investigate the non-parametric or semi-parametric method to adapt the code-switching problems.

\subsection{Multilingual Retrieval Augmented Generation}
\label{sec:rag}
% \modif{should be modified together with the XIR later.}
Retrieval-Augmented Generation (RAG) is a methodology that integrates text generation with external knowledge retrieval, dynamically enhancing the quality of model response and accuracy by accessing relevant information~\cite{gao2023retrieval,he2023exploring}. 
This approach enables the model to utilize up-to-date or specialized knowledge in text generation, thereby increasing its practicality and reliability.
A main branch of the multilingual RAG adopts to retrieve knowledge from the open domain and applies it to the in-context (i.e., augmented prompt)~\cite{zhang-etal-2023-leveraging,shi-etal-2022-xricl,agrawal-etal-2023-context,li2023classification,li-etal-2023-crosslingual,winata2023multilingual,garcia2023unreasonable,ramos-etal-2023-lmcap,kim2023boosting}.
In particular, Thakur~\etal~\cite{thakur2024nomiracl} propose NoMIRACL, a dataset across 18 languages, to evaluate the hallucination of LLM when given a piece of text in external retrieved knowledge.
It consists of two parts, the non-relevant and the relevant subset, corresponding to the question and the passage which are non-related and related, respectively. 
The model is expected to answer ``I don't know'' in the non-relevant subset, otherwise it is determined as a hallucination.
In addition, hallucinations and off-target issues occur when incorporating LLMs with low-resource machine translation~\cite{sennrich2023mitigating}, where the RAG can mitigate these issues via improving the translation quality for low-resource directions~\cite{vernikos2024don,fu2024relay,zeng2024teaching,he2024exploring}.
Overall, the RAG can effectively facilitate LLMs to generate more reliable responses, alleviating the issues of hallucination and factual error without fine-tuning.
However, achieving substantial enhancements solely through the RAG method for low-resource languages, where LLMs struggle, poses a considerable challenge.
Meanwhile, it is also challenging to build retrievers that can be used in low-resource languages~\cite{conia-etal-2023-increasing}.

\subsection{Limitations and Future Directions on Inference Strategies}
The multilingual inference strategies exhibit a diverse range of characteristics.
Although existing methods contribute to the performance of LLMs to some extent in the multilingual scenario, they still have many limitations as follows.

\textbf{Universal inference paradigm.}
% \modif{Due to the universality of LLMs, an aim is to keep them as task-agnostic and language-agnostic as feasible.} 
Previous PLMs are task-specific and language-specific, indicating that it is essential to specify the language ID.
For instance, we need to assign the language ID of input and output respectively when leveraging the M2M-100 for machine translation~\cite{fan2021beyond}.
However, it is not required in the existing paradigm of LLMs, such as GPT-4.
Due to the universality of LLMs, an aim is to keep them as task-agnostic and language-agnostic as feasible.
Current LLMs employ a standardized approach for inference across all languages (either direct inference, pre-translation inference, CoT, or RAG). 
However, as shown in the experiments and analyses above, the inference strategies exhibit varied performances when faced with different LLMs, tasks, and languages. 
Therefore, a flexible and universally applicable paradigm is desirable to uniformly handle the environments of all languages (i.e., high-/medium-/low-resource languages).
    
\textbf{Language-Specific characteristics.}
Existing multilingual inference strategies are primarily adapted from monolingual (e.g., English) strategies, lacking exploration of language-specific characteristics. For example, the prompt engineering for a low-resource language is derived from English instructions, but these prompts are more aligned with the habits of English speakers instead of native speakers. Therefore, the approach might pose challenges in harnessing the potential of native languages during inference with LLMs.
    
\textbf{Emergence ability.}
One important advantage of LLMs is the emergence ability that is informally defined as ``the ability that does not exist in small models but appears in large models''~\cite{wei2022emergent}. However, this advantage is not embodied significantly in multilingual scenarios. Thus, the performance of LLMs is far behind that of the English scenario, and difficult to outperform small language models.
It is not only due to limitations during the training phase but also potentially influenced by inappropriate strategies during the inference phase.

\textbf{Model collaboration.}
Existing inference pipelines are designed for single LLMs. 
Due to data scarcity in low-resource languages, it is challenging to address tasks for all languages within a single model~\cite{magueresse2020low,ranathunga2023neural}.
Considering that the smaller models excel at specific tasks or handling specific languages, which do not require large-scale data to optimize~\cite{miceli2017regularization,dabre2019exploiting,bapna2019simple}.
It is beneficial to leverage the strengths of both large and small models and solve tasks for different resource languages through model collaboration.
The model collaboration that effectively bridges the large and small models is a direction yet to be explored for enhancing inference performance in the multilingual scenario.
% \modif{Considering that the smaller models excel at specific tasks or handling specific languages, the model collaboration that effectively bridges the large and small models to form a is a direction yet to be explored for enhancing inference performance in the multilingual scenario.}

\section{Multilingual Information Retrieval}
\label{sec:XIR}

The task of Information Retrieval (IR) is to find the relevant documents that satisfy the information needs of the users (in the form of {\it queries}) from a large collection~\citep{manning_raghavan_schutze_2008}.
Multilingual IR studies the IR tasks in more than only English,
and is generally categorized into the {\it monolingual}, {\it cross-lingual} and {\it multilingual} scenarios based on the language(s) of the queries and documents:\ 
Given a query in the language ($L_1$),
{\it monolingual retrieval} aims to find relevant information from the same language as the query ($L_1$), 
{\it cross-lingual retrieval} finds relevant information from a different language ($L_2$), 
and {\it multilingual retrieval} finds relevant information from multiple different languages ($L_i, i \in \{1, ..., n\}$). 
There are many surveys and discussions related to this section, 
including multilingual retrieval progress prior to neural IR~\citep{fluhr19952,oard1998cross,nie1999cross,hollink2004monolingual,gao2007cross,grefenstette2012cross,peters2012multilingual,nie2022cross,lawrie2022hc4},
progress on general neural IR prior to the emergence of LLMs~\citep{lin2022pretrained},
and interplay of LLM and IR~\cite{ai202380,zhu2024largelanguagemodelsinformation,liu2024information}. 
This section will focus on the multilingual aspect,
especially the new opportunities of multilingual retrieval brought by LLMs (``LLM for mIR''),
as oppose to the RAG methods (``mIR for LLM'') introduced in the Section~\ref{sec:rag} above.

\subsection{Synthetic Training Data}
\label{sec:XIR-data}
Synthetic datasets for multilingual retrieval are traditionally created in two approaches:\ 
machine translation~\cite{bonifacio2022mmarcomultilingualversionms,haq-etal-2024-indicirsuite},
and 
natural semantic structures~\cite{izacard2021contriever}, e.g., title--passage~\cite{wang2024textembeddingsweaklysupervisedcontrastive}, inner-connected Wikipedia links~\cite{schwenk-etal-2021-ccmatrix,el-kishky-etal-2020-ccaligned}, and so on.
% parallel corpora~\cite{schwenk-etal-2021-ccmatrix,el-kishky-etal-2020-ccaligned},

Spearheaded by InPars~\citep{bonifacio2022inpars,inparsv2,abonizio2023inpars} and Promptagator~\citep{dai2023promptagator},
LLMs bring the third approach, which generates large-scale synthetic data for training retrieval models in an affordable way.
While the above two works focus on English, 
following this line,
SwimIR~\cite{thakur-etal-2024-leveraging} builds large-scale training data for both cross-lingual and monolingual retrieval tasks.
{JH-POLO}~\cite{mayfield2023syntheticcrosslanguageinformationretrieval} build cross-lingual retrieval training data by generating English queries based on non-English positive and negative passages using LLM.
It has also been found that synthetic data generated by LLM could improve the performance of LLM-based embedding models:\
mE5-Mistral~\cite{wang2024improvingtextembeddingslarge} generate synthetic data by GPT-3.5 and GPT-4 on 93 languages.
Gecko~\cite{lee2024geckoversatiletextembeddings} built Few-shot Prompted Retrieval dataset (FRet).
Both works adopt a two-stage generation pipeline:\
in the first stage, the LLM generates the query and the task description(s) given passage(s); in the second stage, while both works produce positive and (hard) negative passages,
Gecko asks LLM to score the positive and negative passages from candidates returned by a retriever,
while mE5-Mistral lets the LLM generate the positive and negative passages based on given requirements. 
Additionally,
Arctic-Embed~\cite{merrick2024arcticembedscalableefficientaccurate} find that hard negative mined from existing corpora is in higher quality compared to the ones generated by LLM.

\subsection{Multilingual Retrievers}
\label{sec:XIR-retriever}
According to the taxonomy by \citet{lin2021few},
the retrievers are categorized into the unsupervised sparse, supervised sparse and, supervised dense models,
where the dense models could be further categorized into the single-vector and multi-vector models.
This section briefly introduces models under each category that are extended into multilingual scenarios,
and the impact of LLMs on the category if applicable.

The {\it unsupervised sparse models} refers to the bag-of-words exact matching ranking algorithms, e.g., BM25~\citep{robertson2009probabilistic}.
It requires language-specific analyzers to extend the models into the multilingual scenarios,
which are designed based on expert knowledge on the target language. 
% {\bf citation}
This category of models serves a strong baseline itself on the mono-lingual retrieval task~\citep{zhang-etal-2021-mr}, or on the cross-lingual and multilingual retrieval when used together with the query or document translation~\citep{hull1996querying,zhou2012translation,lin2023simpleeffectiveneuralranking}.

The {\it supervised sparse models} 
learn the weight per term from the training data rather than using the collection statistics as unsupervised sparse models~\cite{dai2019contextawaresentencepassagetermimportance,gao-etal-2021-coil,10.1145/3404835.3463098,10.48550/arxiv.2109.10086,10.1145/3477495.3531857,10.1145/3477495.3531833}.
Works extending this category into the multilingual scenarios are mainly surrounding \splade
on monolingual retrieval~\cite{lassance2023extendingenglishirmethods} 
or cross-lingual retrieval~\cite{conneau2020unsupervisedcrosslingualrepresentationlearning,nair2023blade}.
All above works are based on BERT-level multilingual pretrained models.
% :\
\ignore{ 
\citet{lassance2023extendingenglishirmethods} adopted SPLADE model to the monolingual tasks,
where they pretrain the model from scratch on the subset of MIRACL corpora~\citep{zhang2023miracl}:\
Wikipedia documents on 15 non-English languages,
pre-finetune on the mMARCO datasets~\citep{bonifacio2022mmarcomultilingualversionms},
and finally finetune on the MIRACL training data~\citep{zhang2023miracl}.
\citet{nair2022learning} proposed SPLADE-X for the cross-lingual retrieval task,
where they found bilingual training is beneficial on the cross-lingual performance, 
and that the the \splade based models are complementary to PSQ, an cross-lingual unsupervised sparse models based on statistical machine translation.
In both works mentioned above, the vocabulary size leads to a challenge in the multilingual sparse model training:\ 
It requires 80G-level GPU memory when training multilingual SPLADE with the vocabulary size of XLM-R~\citep{conneau2020unsupervisedcrosslingualrepresentationlearning}.
To address the issue, \citet{nair2023blade} proposed BLADE, a multilingual SPLADE model with vocabulary pruning followed by intermediate cross-lingual pretraining,
which significantly improves the training and inference efficiency.
}

The {\it supervised dense models} category, including both the single-vector and multi-vector dense retrieval models,
resides most of the work on the multilingual IR and LLM-related embedding models.
The dense models was firstly extended into the multilingual scenarios by switching the backbones into mBERT or XLM-R~\citep{zhang-etal-2021-mr,asai-etal-2021-xor,zhang2023toward}.
Many methods that are found effective on English also show similar performance improvements on the multilingual scenarios,
for example, 
extending the pretraining and pre-finetuning corpora into multilingual unsupervised data or translated corpora~\cite{bonifacio2022mmarcomultilingualversionms,zhang2023toward},
knowledge distillation from rerankers~\cite{zhang2023toward}
or even from English retriever~\cite{li-etal-2022-learning-cross}.
% For the single-vector dense retrieval models,
% \citet{asai-etal-2021-xor} provide the multilingual dense retrievers (mDPR) baseline results on the cross-lingual retrieval task. 
% \citet{zhang-etal-2021-mr} investigate mDPR on monolingual retrieval tasks and find it is complementary to the unsupervised sparse models.
% \citet{zhang2023toward} explore the the design space of training mDPR and provide recommendations on both zero-shot and supervised scenarios for monolingual retrieval tasks.
Recently, more multilingual embedding models are proposed with large-scale pre-training and pre-finetuning.
Open-source models include  
mE5~\cite{wang2024multilinguale5textembeddings},
mGTE~\cite{zhang2024mgtegeneralizedlongcontexttext},
and BGE~\cite{chen2024bgem3embeddingmultilingualmultifunctionality},
whereas
\cite{openaiembeddings,cohereembeddings,voyageembeddings}
provide black-box API for the multilingual embedding models.

Many LLM-based embedding models emerge these years:
Focusing on the retrieval task, 
RepLLAMA~\cite{ma2024finetuningllama} shows that LLM-based embedding models could be fine-tuned to
achieve better in-domain effectiveness and out-of-domain generalizability.
PromptReps~\cite{zhuang2024promptrepspromptinglargelanguage} shows that the 
LLMs could be prompted to generate dense and sparse representations
to achieve competitive zero-shot performance on the passage retrieval tasks.
More works emerge with the target of unifying multiple embedding tasks into a single model~\cite{muennighoff2022mteb}.
The modification on the model design includes:\ 
enabling bi-directional attention~\cite{lee2024nvembedimprovedtechniquestraining,springer2024repetitionimproveslanguagemodel,behnamghader2024llm2veclargelanguagemodels},
adding instructional tuning~\cite{lee2024nvembedimprovedtechniquestraining,muennighoff2024generativerepresentationalinstructiontuning},
adapting Matryoshka representations~\cite{openaiembeddings,kusupati2024matryoshkarepresentationlearning},
and merging multiple models~\cite{xiao-etal-2024-lm}.
% \crys{and so on.}
Training wise,
AnglE~\cite{emb2024mxbai} proposed angle-optimized text embedding models to eliminate the gradient saturation zone of cosine function. 
% To the best of our knowledge,
Yet most of the work along this line focuses on English embedding models,
only Gecko~\cite{lee2024geckoversatiletextembeddings} supports the multilingual inputs. 
% model in large models (> 1B parameters). 
Additionally,
NLLB-E5~\cite{acharya2024nllbe5scalablemultilingualretrieval} propose to integrate NLLB models with E5 to enable the multilingual ability of English embedding models.

Works extending {\it multi-vector dense} retrieval models into multilingual scenarios mainly focus on ColBERT~\cite{khattab2020colbert},
dubbed mColBERT~\cite{bonifacio2022mmarcomultilingualversionms,zhang2023miracl} or ColBERT-X~\cite{nair2022transfer}.
The extension covers monolingual retrieval~\citep{bonifacio2022mmarcomultilingualversionms,louis2024colbertxmmodularmultivectorrepresentation,zhang2023miracl}, cross-lingual retrieval~\citep{nair2022transfer,yang2024translate,huang2023improving}, and multilingual retrieval~\citep{lawrie2023neuralapproachestomultilingualinformationretrieval,yang2024multilingual}.
\ignore{ 
Specifically,
\citet{yang2024translate} distill knowledge from rerankers via translations of training data. 
\citet{huang2023improving} proposed to distill via optimal transport, 
which greatly enhances the cross-lingual alignments and improve the cross-lingual retrieval performance.
Moving from cross-lingual to multilingual retrieval,
\citet{lawrie2023neuralapproachestomultilingualinformationretrieval} finds that training with mixed-language batch data shows significant performance improvement comparing to single-language batch data or English-only training on ColBERT-X.
}
% yang2024multilingual
Other than adopting the default model structure,
ColBERT-XM~\cite{louis2024colbertxmmodularmultivectorrepresentation} adds modular language-specific adapter in XMOD architecture,
which shows competitive zero-shot monolingual retrieval performance.
% 
% Recently,
% While supporting being used as single-vector models,
BGE-M3~\citep{chen-etal-2024-m3} unites the single-vector dense, multi-vector dense and sparse models by distilling the knowledge from the {ensemble} of the three types of models into each single models,
demonstrating strong on both monolingual and cross-lingual retrieval tasks.

\subsection{Multilingual Rerankers}
\label{sec:XIR-reranker}
% early works deploy mBERT into rerankers
Early works multilingual neural rerankers are mostly based on mBERT~\cite{macavaney2020teaching}, 
exploring the translation configuration on the training and inference stages~\cite{shi2019crosslingualrelevancetransferdocument, shi-etal-2020-cross}.
Later works found that mMiniLM and mT5 serve as better backbones as multilingual rerankers~\citep{bonifacio2022mmarcomultilingualversionms,jeronymo2023neuralmindunicamp2022trecneuclir}.
% \crys{mMiniLM, mT5, Jina, BGE}

Recently, a line of works explored using LLM as the zero-shot rerankers in retrieval initiated by 
RankGPT~\cite{sun-etal-2023-chatgpt} and LRL~\cite{ma2023zeroshotlistwisedocumentreranking}, dubbed listwise rerankers,
followed by works on distilling from the GPTs with enhanced training techniques~\citep{pradeep2023rankvicunazeroshotlistwisedocument,pradeep2023rankzephyreffectiverobustzeroshot},
and on building listwise reranker without relying on the close-sourced GPT-models~\citep{zhang2023rankwithoutgptbuildinggptindependentlistwise}.
One criticism on the listwise rerankers is the sequential inference step caused by the sliding-window aggregation step that merges the ranked documents from different batch. 
To address the issue, different inference and aggregation strategies have been proposed, including Setwise~\citep{zhuang2024setwise} and TourRank~\citep{chen2024tourrankutilizinglargelanguage}.
Limited works explored the listwise rerankers on the multilingual IR tasks,
the only one to our best knowledge is \citet{adeyemi2023zeroshotcrosslingualrerankinglarge},
which evaluated GPT-based listwise reranker on CIRAL~\citep{mofetoluwa2024ciral}, a cross-lingual retrieval benchmark on African languages, 
finding that GPT-4 yields competitive zero-shot performance on the task and even on par with the zero-shot results with machine-translated documents on some of the languages.

\subsection{Challenges and Future Directions on Multilingual Information Retrieval}
\label{sec:XIR-future}
% Beyond the challenges and future directions shared with general retrieval and LLM integration,
% such as high latency of LLMs-depends retrieval models, and new retrieval paradigm brought by generative models~\cite{li2024matchinggenerationsurveygenerative,lee2024longcontextlanguagemodelssubsume}. Multilingual retrieval with LLMs also faces unique challenges. 
% For instance,
% the current open-source auto-regressive LLMs often have limited supported languages compared to the encoder-only or encoder-decoder models,
% hindering direct application of LLM-based retriever models into multilingual scenarios.

The aforementioned existing studies demonstrate how LLM-based methods enhance the effectiveness and out-of-domain generalization in information retrieval tasks. 
Beyond integrating LLMs as components in the established retrieval-reranking pipeline,  
LLMs also open new possibilities to the paradigm of search process and broader information access:\
For example, \citet{li2024matchinggenerationsurveygenerative} and \citet{lee2024longcontextlanguagemodelssubsume} both propose a unified generative framework that integrates retrieval and question answering,
where retrieval is considered as an inherent part together with the other generative tasks rather than as a separate component as in RAG.

That said, challenges present in deploying LLM in assessible search systems,
which include inherently high latency for indexing and searching, as well as high computational demand during inference and fine-tuning.
Current work that distills knowledge from LLM to smaller models strike a balance between effectiveness and efficiency~\cite{tamber2023scalingdownlittingup}.

Beyond the general challenges of LLM-based retrieval,
The open-source state-of-the-art LLM usually have fewer supported languages compared to encoder-only or encoder-decoder models:\ 
LLaMA-3~\cite{grattafiori2024llama3herdmodels} and Command-R+\footnote{\url{https://docs.cohere.com/v2/docs/command-r-plus}} support around 20 languages,
while encoder-only models such as mBERT and XLM-R support 100 languages,
and translation-targeting encoder-decoder such as NLLB support 200 languages.
Current retrieval methods are applied to LLMs, which mainly regard LLMs as a knowledge store. However, in low-resource languages, LLMs lack generation capabilities and have not been trained with large-scale data, thus they are difficult to serve as a reliable source of knowledge.
How can the above methods be applied to languages not yet supported by LLM but available in smaller language models?

\section{Security of Multilingual Large Language Models}\label{sec:llmsafety}

% Multilingual large language models have gained significant attention due to their remarkable capabilities in natural language understanding and generation. 
With the wide deployment of LLMs in various applications, increasing security concerns have emerged. This chapter introduces the security aspects of LLMs in the multilingual scenario, specifically exploring attack methods and the existing research on defense mechanisms.
Since there are no clear definitions to distinguish whether an LLM is a multilingual model or not, this survey not only focuses on security issues specific to different languages but also provides a perspective on common security issues.
The investigated methods work equally across all languages and can be easily transferred to multilingual scenarios, providing inspiring thoughts on future research.

\newcommand{\rot}{\rotatebox[origin=c]{45}}
\begin{table*}[!t]  
\centering  
%\small 
\resizebox{\textwidth}{!}{
\begin{tabular}{l | c c c c c c c }
\toprule

\multirow{2}{*}{\textbf{Model}} & \multicolumn{7}{c}{\textbf{Prompt-Based Jailbreak}} \\  
% \cmidrule(lr){2-2} \cmidrule(lr){3-9}  \cmidrule(lr){10-12}  
 & JailBroken~\cite{wei2024jailbroken} & GPTFUZZER~\cite{yu2023gptfuzzer} & AutoDAN~\cite{liu2023autodan} & DeepInception~\cite{li2023deepinception} & ICA~\cite{wei2023jailbreak} & PAIR~\cite{chao2023jailbreaking} & ReNeLLM~\cite{ding2023wolf}  \\  
\midrule  
GPT-3.5-turbo  &  100\% &35\% &45\% & 66\% & 0\% & 19\%  &87\%  \\  
GPT-4-0613      & 58\%& 0\% & 2\%& 35\% & 1\% & 20\% & 38\% \\  
LLaMA-2-7B-Chat   & 6\% & 31\% & 51\%& 8\% & 0\% & 27\% & 31\%  \\  
LLaMA-2-13B-Chat  & 4\%& 41\% & 72\%& 0\% & 0\% &  13\%  & 69\% \\  
Vicuna-7B-v1.5    & 100\% & 93\% &100\% & 29\% & 51\% &  99\% & 77\%  \\  
Vicuna-13B-v1.5   & 100\% & 94\% & 97\% & 17\% & 81\% & 95\% & 87\%  \\  
ChatGLM-3         & 95\% & 85\%& 89\% & 33\% & 54\% & 96\% & 86\%  \\  
Qwen-7B-Chat     & 100\% & 82\% & 99\% & 58\% & 36\% & 77\% & 70\%  \\  
Intern-7B       & 100\% & 92\% & 98\% & 36\% & 23\% & 86\% &  67\%  \\  
Mistral-7B      & 100\%& 99\%& 98\% & 40\% &  75\% & 95\% & 90\%   \\  
\midrule  
 & \multicolumn{1}{c|}{\textbf{GCG series}}  & \multicolumn{6}{c}{\textbf{Multilingual Jailbreak}}  \\  
% \cmidrule(lr){2-2} \cmidrule(lr){3-9}  \cmidrule(lr){10-12}  
& \multicolumn{1}{c|}{GCG~\cite{zou2023universal}}     
& \multicolumn{2}{c}{Multilingual~\cite{deng2023multilingual}}  & \multicolumn{2}{c}{Cipher~\cite{yuan2024gpt4}} & \multicolumn{2}{c}{CodeChameleon~\cite{lv2024codechameleon}} \\  
\midrule  
GPT-3.5-turbo   & \multicolumn{1}{c|}{12\%} & \multicolumn{2}{c}{100\%} &  \multicolumn{2}{c}{80\%} & \multicolumn{2}{c}{90\% }\\  
GPT-4-0613      & \multicolumn{1}{c|}{0\%}  & \multicolumn{2}{c}{63\%}  & \multicolumn{2}{c}{75\%}& \multicolumn{2}{c}{72\%} \\  
LLaMA-2-7B-Chat  & \multicolumn{1}{c|}{46\%} &  \multicolumn{2}{c}{2\%} & \multicolumn{2}{c}{61\%} & \multicolumn{2}{c}{80\%}  \\  
LLaMA-2-13B-Chat & \multicolumn{1}{c|}{46\%} & \multicolumn{2}{c}{0\%} & \multicolumn{2}{c}{90\%}& \multicolumn{2}{c}{67\%}\\  
Vicuna-7B-v1.5   & \multicolumn{1}{c|}{94\%} &  \multicolumn{2}{c}{94\%} & \multicolumn{2}{c}{28\%} & \multicolumn{2}{c}{80\%} \\  
Vicuna-13B-v1.5  & \multicolumn{1}{c|}{94\%} &  \multicolumn{2}{c}{100\%} & \multicolumn{2}{c}{76\%} & \multicolumn{2}{c}{73\%} \\  
ChatGLM-3        & \multicolumn{1}{c|}{34\%} &  \multicolumn{2}{c}{100\%} & \multicolumn{2}{c}{78\%} & \multicolumn{2}{c}{92\% }\\  
Qwen-7B-Chat    & \multicolumn{1}{c|}{48\%} &  \multicolumn{2}{c}{99\%} & \multicolumn{2}{c}{58\%} & \multicolumn{2}{c}{84\%} \\  
Intern-7B        & \multicolumn{1}{c|}{10\%} & \multicolumn{2}{c}{99\%} & \multicolumn{2}{c}{99\%} & \multicolumn{2}{c}{71\%} \\  
Mistral-7B      & \multicolumn{1}{c|}{82\%} &  \multicolumn{2}{c}{100\%} & \multicolumn{2}{c}{97\%} & \multicolumn{2}{c}{95\%} \\  
% \textbf{Avg.} &  & & & & & &  & & & & & \\  
\bottomrule 

\end{tabular}}
\caption{An overview of \textbf{Greedy Coordinate Gradient}, \textbf{Prompt-Based} and \textbf{Multilingual} attack methods jailbreaking for LLMs on the AdvBench~\cite{zhou2024easyjailbreak}. The evaluation method is consistent with the EasyJailbreak~\cite{zhou2024easyjailbreak} framework, which uses GPT-4-turbo-1106 as the scoring model and the evaluation prompts from GPTFUZZER~\cite{yu2023gptfuzzer}.}
\label{table:attack}
\end{table*}

\subsection{Attack Methods}
\label{sec:attack}
% \modif{IMPORTANT: lack of the definition of Jailbreak}
To explore the security of LLMs, a red team attack on LLMs is required.
A red-team attack is a cybersecurity exercise where a group of ethical hackers, known as the red team, simulate real-world cyberattacks on an organization's systems, networks, or infrastructure. The goal of a red-team attack is to identify vulnerabilities, weaknesses, and potential security breaches that malicious actors could exploit~\cite{mirkovic2008testing}.
A common practice is the ``jailbreak'' attack which typically refers to the unauthorized access or modification of the underlying code or functionality of models. Essentially, it involves breaking out of the constraints or limitations imposed by the design or usage policies of LLMs. It includes techniques to bypass security measures or enable functionalities that are not intended or permitted by the developers.

According to the criteria in existing studies~\cite{wu2024llms,zhou2024easyjailbreak},
%Referring to the classification criteria of Wu~\etal~\cite{wu2024llms} and Zhou~\etal~\cite{zhou2024easyjailbreak}, 
the jailbreak methods on LLMs can be divided into three types: Greedy Coordinate Gradient (GCG) jailbreak~\cite{sitawarin2024pal,zou2023universal}, prompt-based jailbreak~\cite{wei2024jailbroken,liu2023jailbreaking,shen2023do,deng2024masterkey,li2024salad,puttaparthi2023comprehensive,liu2023autodan,jin2024guard} and multilingual jailbreak~\cite{shen2024language,deng2023multilingual,puttaparthi2023comprehensive, yong2023low,xu2023cognitive,li2024cross,yuan2024gpt4,huang2023catastrophic}. In this section, we focus on these three jailbreak methods, especially the multilingual jailbreak.
The first two methods involve generic attacks on LLMs, and the latter emphasizes jailbreaking through multiple languages.
All these methods are aimed at bypassing the security measures of LLMs to generate malicious information.
Since most of the jailbreaking methods are customized, the effectiveness of each method varies depending on the specific LLMs.
To provide a comprehensive comparison, we investigate the performance of different jailbreaking methods across various LLMs based on a unified evaluation framework, as shown in Table~\ref{table:attack}.
% \modif{We provide an overview performance of attack methods on common LLMs in Table~\ref{table:attack}.}

\subsubsection{Greedy Coordinate Gradient Jailbreak} 
% The \textbf{G}reedy \textbf{C}oordinate \textbf{G}radient ~\cite{zou2023universal} is an attack method based on greedy algorithms and gradient metric design hints. It first creates a seed prompt, and then iteratively replaces the tokens. The best prompt for attacking the model is determined by calculating the gradient.
% Sitawarin~\etal~\cite{sitawarin2024pal} propose GCG++ to increase the attack success rate (ASR) of the original GCG method on LLaMA-2-7B.
% which improves the score from 58\% to 94\%. 
% \modif{Based on the original GCG, the GCG++ replaces the cross entropy loss with the proposed CWloss and adds a simple format-ware target string (\textit{i.e.},  “[ASSISTANT]:”) before the input prompt. }
The \textbf{G}reedy \textbf{C}oordinate \textbf{G}radient ~\cite{zou2023universal} is an attack method based on greedy algorithms and gradient metric design hints. It first creates a seed prompt, and then iteratively replaces the tokens. The best prompt for attacking the model is determined by calculating the gradient.
Sitawarin~\etal~\cite{sitawarin2024pal} propose GCG++ to increase the attack success rate (ASR) of the original GCG method on LLaMA-2-7B.
% , which improves the score from 58\% to 94\%. 
% \modif{Based on the original GCG, the GCG++ replaces the cross entropy loss with the proposed CWloss and adds a simple format-ware target string (\textit{i.e.},  “[ASSISTANT]:”) before the input prompt. }
On the basis of the original GCG, they replace the cross entropy loss with multi-class loss to avoid vanishing gradients in the $Softmax$, which is effective in attacking LLM. 
Another discovery is that LLMs such as LLaMA-2-7B are format sensitive, having a strong prior for predicting a space token at the beginning of the model response. 
GCG++ adds a simple format-ware target string (\textit{i.e.},  ``[ASSISTENT]:'' before the input prompt to force the model to output harmful content.
Through the two changes, GCG++ increases the GCG attack success rate on LLaMA-2-7B from 56\% to 80\%.
% This detail was overlooked in the previous GCG, and GCG++ considered this \textit{format-aware target string}, such as using "Sure" instead of "Sure" to force the model to output harmful content. Through two changes, GCG++ increases GCG’s attack success rate on Llama-2-7B from 56\% to 80\%.

\subsubsection{Prompt-Based Jailbreak}

\begin{table*}[!t]  
\centering  
\small 
\begin{tabular}{l | c c c}
\toprule
\multirow{2}{*}{\textbf{Characteristic}} & \multicolumn{3}{c}{\textbf{Attack Method Types}} \\  
% \cmidrule(lr){2-2} \cmidrule(lr){3-9}  \cmidrule(lr){10-12}  
& GCG series  & LLM-Based & Rule-Based \\  
\midrule  
\textit{The LLM type for optimizing prompt}  & Target LLM  & Agent LLM & - \\  
\textit{The status of Jailbreak prompt} & Dynamic &  Dynamic & Static \\  
\textit{The form of generating new prompts} & Auto & Auto & Manual \\
% \midrule  
% \textbf{Avg.} &  & & & & & &\\  

\bottomrule 
\end{tabular}
\caption{An overview of comparison among \textbf{GCG}, \textbf{Prompt-based} and \textbf{Rule-based} jailbreak methods. Target LLM and Agent LLM refer to the attacked LLM and other LLMs that are different from target LLM, respectively.}
\label{table:compare}
\end{table*}

The prompt-based methods aim to mislead LLM into generating harmful content based on specifically designed templates. 
Among them, LLM-based~\cite{liu2023autodan,mehrotra2023tree,sitawarin2024pal,chao2023jailbreaking,deng2024masterkey,yu2023gptfuzzer,ding2023wolf,bhardwaj2023red} and rule-based~\cite{wei2024jailbroken,li2023deepinception,liu2023jailbreaking,shen2023do,puttaparthi2023comprehensive} are the two common used methods for designing attacking prompts. 
The differences between GCG, LLM-based, and rule-based are presented in Table~\ref{table:compare} with various aspects.

Rule-based jailbreak methods are those works that design specific prompts to formulate rules for LLMs~\cite{wei2024jailbroken, li2023deepinception}, or directly collect prompts from various channels such as websites and forums~\cite{liu2023jailbreaking, shen2023do, puttaparthi2023comprehensive}. Rule-based jailbreak methods rely more on human participation because these rules are customized based on empirical experience in specific scenarios. 
Thus, the produced prompts are effective in a narrow area but lack versatility and are static, which can be easily defended.

LLM-based jailbreak methods leverage the instances of LLMs to optimize existing attack prompts or generate new jailbreak prompts.
Different from the GCG series approaches, the principle of LLM-based jailbreak methods acts as using an agent LLM to attack a target LLM. Existing studies ~\cite{liu2023autodan, mehrotra2023tree, sitawarin2024pal, chao2023jailbreaking} usually take the agent LLM as an optimizer, scoring module or evaluator to optimize prompts for attackers.
% Some other methods~\cite{deng2024masterkey, yu2023gptfuzzer, ding2023wolf, bhardwaj2023red} utilize the generation capabilities of LLM to perform multiple operations, such as generating new prompts, rewriting or shortening feasible attack prompts, to produce feasible attack prompts, which enable the original attack prompt effective to evade the security alignment policy.
%These simple operations based on LLM can allow the originally ineffective attack prompt to evade the security alignment policy and make the attack successful.
Some other methods~\cite{deng2024masterkey, yu2023gptfuzzer, ding2023wolf, bhardwaj2023red} use the generation capabilities of LLM to perform multiple operations such as rewriting and shortening feasible attack prompts. They also wrap harmful prompts with scenarios to avoid safe alignment. These operations based on LLMs can enable the original attack prompt to evade the security alignment policy.

\subsubsection{Multilingual Jailbreak} 
% \modif{Some existing LLMs already have strong multilingual capabilities, which belong to the broad category of multilingual LLMs. }
Due to the unequal security alignment of LLMs across different languages, with higher security in high-resource languages compared to low-resource languages, the multilingual jailbreak methods~\cite{shen2024language,deng2023multilingual,yong2023low,xu2023cognitive,li2024cross} attend to cheat the LLMs via the alignment vulnerabilities between high-resource and low-resource language.
%utilize high-resource and low-resource language alignment vulnerabilities to guide LLMs to generate harmful content. 
Some other methods~\cite{yuan2024gpt4,huang2023catastrophic,lv2024codechameleon,wei2024jailbroken} use special encoding or ciphers to disguise harmful prompts to achieve jailbreak, which can be considered as a type of special language.

%Nowadays, there are already several research works on the potential risks of LLMs. But these works have primarily focused on English. 
The existing studies on the potential security risks of LLMs, primarily focused on the usage of English scenarios.
Under the multilingual scenario, Deng~\etal~\cite{deng2023multilingual} reveal the presence of multilingual jailbreak challenges within LLMs, who divide potential risks into two scenarios: (1) \textit{unintentional} and (2) \textit{intentional}. 
The unintentional scenario involves users inadvertently bypassing the safety alignment by querying LLMs with non-English prompts, while the intentional scenario refers to users intentionally combining malicious instructions with multilingual prompts. 
Compared to high-resource languages, low-resource languages are approximately three times more likely to encounter harmful content in GPT-3.5 and GPT-4, and thus the LLMs are much easier to attack via low-resource languages.
This could also be the reason why the model attack approaches are designed with low-resource languages within these two scenarios.

For the \textit{unintentional} scenario, Shen~\etal~\cite{shen2024language} indicate that the bottleneck in cross-lingual alignment lies within the training stage of LLMs. Thus, they study the effect of instruction tuning with RLHF and supervised fine-tuning (SFT) on the HH-RLHF dataset~\cite{bai2022constitutional}. The results show that although training with high-resource languages can improve model alignment, the effect of training with low-resource languages is still negligible. 
For the \textit{intentional} scenario, various prompts are designed to bypass the security defenses of the LLMs via high-resource languages.
In particular, Yong~\etal~\cite{yong2023low} successfully circumvent the safeguard of GPT-4 by translating unsafe English prompts into low-resource languages. They propose the LRL-Combined Attacks approach to achieve 79\% ASR on the AdvBench dataset~\cite{zou2023universal}. 
Xu~\etal~\cite{xu2023cognitive} study a new type of black box jailbreak attack, Cognitive Overload, which is specifically designed for the cognitive structure and processes of LLMs. With Google Cloud API, they translate original English harmful instructions from AdvBench and MasterKey~\cite{deng2024masterkey} into 52 other languages and propose multilingual cognitive overload~\cite{sweller1988cognitive,sweller2011cognitive} to hinder in-context learning and inference process when the knowledge exceeds the limited capacity of LLMs~\cite{szulewski2021theory}.
By using Google Cloud API to translate English to other 52 languages and nllb-200-distilled-1.3B to translate other non-English responses back to English, they conducted a multilingual version of their attack method. With multilingual Cognitive Overload, their results show that LLMs are more vulnerable to non-English adversarial prompts.
The observation indicates that the further a language is from English, the more effective the malicious prompt conveyed is in attacking LLMs.
Moreover, Li~\etal~\cite{li2024cross} conduct an extensive empirical study on multilingual jailbreak attacks, which develops a novel semantic preservation algorithm to create multilingual jailbreak prompts. With these prompts, they reveal patterns in multilingual jailbreak attacks and implement fine-tuning mitigation methods for defending against cross-lingual jailbreak attacks.

\textbf{Password jailbreak.}
From the perspective of writing scripts, special encoding or password can be considered as a language conversion, i.e., an English prompt can be converted into a password representation by encryption.
%Special encoding or password is essentially a language conversion. By encryption, English is converted into a password representation. 
Though the password representation can also be converted to English or other languages via decryption, the encrypted prompts cannot be understood directly by humans but by the LLMs, and thus could lead to secure alignment fails.
%Through decryption, the password representation can also be converted to English or other languages. Although we cannot directly understand the encrypted prompts, LLMs can understand its meaning, so secure alignment fails. 
For example, Huang~\etal~\cite{huang2023catastrophic} propose the \textit{generation exploitation} attack to manipulate variations of decoding methods (e.g., greedy decoding and sampling-based decoding), which can easily disrupt model alignment because existing alignment procedures are based on default decoding settings.
Vulnerabilities are exposed when configurations of decoding change slightly. 
Yuan~\etal~\cite{yuan2024gpt4} propose a novel framework CipherChat to examine the generalizability of safety alignment to non-natural languages, where the user can chat with LLMs through cipher prompts.
%We can use CipherChat to chat with LLMs through cipher prompts topped with system role descriptions and few-shot enciphered demonstrations. 
% Experimental results show that UTF and ASCII succeed almost 100\% of the time in bypassing the safety alignment of GPT-4 in several safety domains. Notably, there is a ``secret cipher'' named SelfCipher that outperforms existing human cipher in almost all cases. 
Furthermore, Lv~\etal~\cite{lv2024codechameleon} introduce a hypothesis for the safety mechanism of LLMs: intent security recognition followed by response generation. Following this hypothesis, they propose CodeChameleon to transform queries into decryptable formats with custom Python functions. This approach enables the modification of the original query and achieves state-of-the-art ASR on 7 LLMs.
%Through personalized encryption they propose a method, CodeChameleon, which can achieve the state-of-the-art ASR on 7 LLMs, especially 86.6\% on GPT-4. 
Besides, Wei~\etal~\cite{wei2024jailbroken} propose Jailbroken, which obfuscates the queries using base64 to bypass the safety training of LLMs. They observe that LLMs can understand base64 but cannot defend against attacks under base64.
%which is not only a template-based method, but also a multilingual jailbreak. In the setting of Jailbroken, they attack some LLMs successfully with Base64-encoded inputs. Almost no model can perform secure alignment with Base64-encoded contents. 

\textbf{Other jailbreak.} %Apart from the three main methods of jailbreak research mentioned above, there are also many other directions of jailbreak research now. 
Apart from the aforementioned methods,
Shayegani~\etal~\cite{shayegani2023jailbreak} combine an image targeted towards toxic embeddings with generic prompts to accomplish the jailbreak, which utilizes four embedding space target strategies to poison the vision encoder. Their attacks achieve 87\% and 63.3\% ASR on two vision-language models, LLaVA~\cite{liu2024visual} and LLaMA-Adapter V2~\cite{gao2023llama}, respectively. 
Rando~\etal~\cite{rando2023universal} embed a ``jailbreak backdoor'' into the LLM by poisoning the RLHF training data. Then, users can easily achieve jailbreak by using a trigger word like the ``sudo'' command in the Linux system. 
Wolf~\etal~\cite{wolf2023fundamental} propose a theoretical approach Behavior Expectation Bounds (BEB) that investigates several limitations of alignment in LLMs, which exposes fundamental problems and emphasizes the necessity of designing reliable mechanisms to ensure AI safety.

\begin{table*}[!t]  
\centering  
\small 

\begin{tabular}{l | c c c}
\toprule
\multirow{2}{*}{\textbf{Model}} & \multicolumn{3}{c}{Attacker: \textbf{PAIR}~\cite{chao2023jailbreaking}} \\  
% \cmidrule(lr){2-2} \cmidrule(lr){3-9}  \cmidrule(lr){10-12}  
& No Defense  & SmoothLLM~\cite{robey2023smoothllm} & Perplexity Filter~\cite{jain2023baseline} \\  
\midrule  
GPT-3.5-turbo   & 76\% &  12\% & 15\% \\  
GPT-4-0125-preview      & 50\%  & 25\%& 43\% \\  
LLaMA-2-7B-chat & 4\% & 1\% & 4\% \\  
Vicuna-13B-v1.5  & 82\% & 47\% & 81\% \\  
\midrule
 & \multicolumn{3}{c}{Attacker: \textbf{GCG}~\cite{zou2023universal}}\\
\midrule
GPT-3.5-turbo   &34\% & 1\% & 1\%  \\  
GPT-4-0125-preview       & 1\%& 3\% & 0\% \\  
LLaMA-2-7B-chat  & 2\%& 1\% & 0\%  \\  
Vicuna-13B-v1.5   & 58\% & 1\% & 1\%  \\  
\midrule
& \multicolumn{3}{c}{Attacker: \textbf{JBC}~\cite{jailbreakchat}} \\
\midrule
GPT-3.5-turbo    & 0\% & 0\% & 0\%  \\  
GPT-4-0125-preview    & 0\% & 0\% & 0\%  \\  
LLaMA-2-7B-chat &  0\% & 0\% & 0\%  \\  
Vicuna-13B-v1.5 & 79\% & 64\% & 79\%  \\  
% \midrule  
% \textbf{Avg.} &  & & & & & &\\  

\bottomrule 
\end{tabular}

\caption{An overview of defense methods under jailbreaking on listed LLMs. The evaluation method for the Attack Success Rate (ASR) indicator is consistent with JailbreakBench~\cite{chao2024jailbreakbench}. The responses of LLMs are evaluated using LLaMAGuard-7B.}
\label{table:defense}
\end{table*}

\subsection{Defense Methods}
\label{sec:defense}
Only a few studies attempt to address the defense methods in the security LLMs, which can be categorized into open-source and close-source LLMs.
Based on the available access to the open-source LLMs, the existing studies~\cite{deng2023multilingual,li2024cross,robey2023smoothllm} enhance the security by fine-tuning the foundation models with security instructions. %but still have a large room to improve.
For closed-source LLMs, previous works~\cite{li2023rain, wu2024llms, robey2023smoothllm, jain2023baseline} defend the risks by auditing the input prompts with various security judgment strategies.
% \modif{However, these simple mechanisms cannot achieve satisfactory performance as shown in Table~\ref{table:defense}.} 
However, these simple mechanisms cannot achieve satisfactory performance, as shown in Table~\ref{table:defense}.
The results demonstrate that it is difficult to completely eliminate the generation of unsafe content, regardless of the defense mechanism. And ``SmoothLLM''~\cite{robey2023smoothllm} performs better than ``Perplexity Filter''~\cite{jain2023baseline} but still generates over 10\% unsafe content in most scenarios. 
It indicates that current defense methods for LLMs are in the early stages and require further exploration.

\subsubsection{Defense Methods for Open-Source LLM} 
Robey~\etal~\cite{robey2023smoothllm} propose SmoothLLM, a compatible defense method that acts as a wrapper to smooth the outputs of LLMs by perturbing original attack inputs. Although effective in prohibiting the attacking intent, it also disrupts the semantics of the original input and leads to performance degradation.
Deng~\etal~\cite{deng2023multilingual} propose a data augment defense framework SELF-DEFENSE to defend the attacks for both unintentional and intentional scenarios. 
The key idea is to translate some English seed examples into a target language and then merge the language-specific corpora into the original dataset.
Li~\etal~\cite{li2024cross} implement a fine-tuning mitigation method based on their multilingual jailbreak prompts dataset, reducing the attack success rate by 96.2\%.
Zhou~\etal~\cite{zhou2024robust} introduce Robust Prompt Optimization (RPO) to LLM security by minimizing the confrontation loss.
With less impact on benign cues, the success rate of attacking GPT-4 with GUARD~\cite{jin2024guard} dropped from 96\% to 4\%.

\subsubsection{Defense Methods for Closed-Source LLM} 
For closed-source LLM, the defense algorithms mainly focus on dealing with the model inputs.
Jain~\etal~\cite{jain2023baseline} introduce a filter method, judging the danger of the input based on the perplexity.
Wu~\etal~\cite{wu2024llms} propose a simple and effective defense method SELFDEFEND, which incorporates a shadow stack to check whether harmful prompts exist in the input.
%They design a shadow stack to check whether harmful prompts exist in the input prompt. By empowering LLM to autonomously identify toxic elements within sentences, SELFDEFEND can confidently reject potentially harmful inputs with a simple ``No''.
Li~\etal~\cite{li2023rain} introduce a Rewindable Auto-regressive INference (RAIN) approach, which has the same evaluation setting as SELFDEFEND but focuses on the output of LLMs.
The key component is to leverage another LLM to score the output content and determine whether the output contains harmful content, which achieves good results on LLaMA-base models.

\subsection{Limitations and Future Directions on LLM Security}
In previous sections, we provide a detailed introduction to the research achievements in attack and defense in terms of LLMs security. Most of the current research on LLM security is tested on those popular models (i.e., GPT-3.5, GPT-4, and LLaMA) with multilingual capabilities. 
% \modif{Thus, we have taken these attack or defense methods into the security of multilingual LLM. }
%Previous studies have summarized general attack or defense methods. 
Based on the summary of existing approaches for general attack and defense, we explore the following two aspects as future directions:
(1) \textit{Jailbreak by targeting the multilingual ability of LLMs.}
(2) \textit{How to improve the robustness of LLMs in multilingualism.}

\subsubsection{Jailbreak by Targeting the Multilingual Ability of LLMs}
% \modif{we conduct security research from the following directions: \textit{low-resource language attack, the asymmetry between multiple languages, custom encryption or language rules and misleading information in specific languages.}}
% After summarizing previous work, we can make full use of the multi-language capabilities of large models to conduct security research from the following directions: \textit{low-resource language attack, the asymmetry between multiple languages, custom encryption or language rules and misleading information in specific languages.}

\textbf{Low-Resource language attack.}  
Although LLMs already have strong multilingual capabilities, they cannot remedy the weakness of low-resource languages in the corpus. The ability of LLMs to identify harmful contents in low-resource languages becomes much worse. Existing studies~\cite{yong2023low,deng2023multilingual,shen2024language} explore a lot in low-resource languages, while only focusing on GPT-4 with incorporated translation models. Such a pipeline would inevitably introduce noise. Meanwhile, the evaluation standards should be improved to identify the security capacity of LLMs.

\textbf{The asymmetry between different languages.}  
% \modif{The XXX problems XXX}
Language security is highly culturally specific, and each language has its own security vulnerabilities, leading to asymmetry. 
For example, the word ``uso'' has insulting connotations only in Basque, thus it needs specific supervised tuning for each language, which is difficult to transfer from other languages. 
This contradicts the traditional multilingual methods, where knowledge between different languages can be mutually beneficial~\cite{johnson2017google,huang2023knowledge}.
% Xu~\etal~\cite{xu2024exploring} empirically substantiate the existence of multilingual human values in LLMs, which covers 7 types of human values, 16 languages, and 3 LLM series.
Xu~\etal~\cite{xu2024exploring} empirically validate the presence of multilingual human values within LLMs, encompassing many categories of human values across multiple languages and LLM series.
However, the problems of asymmetry between different languages have risen, and no existing studies about it. 
%Referring to their experimental results, there are problems of asymmetry between different languages. However, there is currently no related work that deeply studies how to accurately discover and exploit the vulnerabilities caused by these asymmetries.

\textbf{Customized encryption and language rules.}  
To bypass the security mechanism of LLMs, an innovative way to create a new language ``X'' by customizing some language or encryption rules~\cite{lv2024codechameleon, huang2023catastrophic, yuan2024gpt4} from the original input. This is based on the translation paradigm, which can be viewed as a mapping function to transform character set A to character set B.
Thus, exploiting the potential loopholes when aligning LLMs in different languages becomes an important problem.
%Translation is essentially a mapping from character set A to character set B. Therefore, an innovative way is to create a new language ``X'' by customizing some language or encryption rules~\cite{lv2024codechameleon, huang2023catastrophic, yuan2024gpt4} can bypass the security mechanism of LLMs. It is necessary to exploit the loopholes that may exist when aligning LLMs in different languages.

\textbf{Misleading information in specific languages.} 
The prior method attempts to assign a word with a specific meaning by embedding a backdoor or directly defining it through dialogue~\cite{rando2023universal,jin2024guard}.
For instance, Rando~\etal~\cite{li2023deepinception} insert the word ``SUDO'' at the end of the query, causing the LLM to misunderstand and bypass the security defenses.
% \modif{The problem of XXX}
It provides insight for attacking the LLMs through multilingual languages.
People can introduce words or phrases with specific meanings into queries provided to LLMs.
It can lead to the model misunderstanding the meaning of queries, which results in the generation of harmful information.
In particular, the polysemous phenomenon that is prevalent in most languages can be used to mislead LLMs into producing prohibited content.
% Introducing words or phrases with multiple meanings into queries provided to large models can lead to the model misunderstanding the query's meaning, resulting in the generation of harmful information.

% \modif{On one hand, the proposed method attempts to give a word in a specific language a special meaning, by embedding the backdoor or define it directly through the dialogue~\cite{rando2023universal}. On the other hand, the polysemous phenomenon that is prevalent in most languages can be used to mislead LLMs into producing prohibited content. None of these ideas have been worked on yet.}

\subsubsection{Robustness of LLMs with Multilingualism}

%According to our investigation, 
Existing studies intend to attack LLMs to find some leaks, rather than to improve the defense ability and robustness of LLMs. However, an attack or jailbreak cannot ensure the development of LLM security. In this section, we will discuss the potential solutions for improving the security and robustness of LLMs.

% \modif{\textbf{Adversarial training}}
\textbf{Adversarial training with multilingual samples.}
Inspired by Mazeika~\etal~\cite{mazeika2024harmbench}, the proposed method enhances defense and robustness using adversarial training. Unlike fine-tuning on fixed datasets with harmful prompts, they propose a Robust Refusal Dynamic Defense to fine-tune the LLM on a dynamic pool of samples, which are continuously updated through a strong optimization-based jailbreak method.
It is feasible to incorporate multilingual data into the sample pool using multilingual jailbreak methods which are more practical in attacks, making the data in the sample pool more diverse and advantageous for adversarial training.
Besides, adversarial training can be introduced in the pre-training stage or fine-tuning stage of LLMs. 
As a result, the safety and robustness of LLMs with multilingualism can be enhanced at the root by adversarial training.

% \modif{\textbf{Fine-tuning with multilingual security datasets.}}
\textbf{Multilingual security alignment with data augmentation.}
% In most cases, the methods to train LLMs are based on a generic open-source pedestal model, such as Vicuna, LLaMA, BLOOMZ, etc. 
% Considering the high cost of pre-training from scratch, we can build the security dataset.
Similar to the approach of enhancing LLMs instruction-following capabilities and human values proposed by SFT and RLHF~\cite{ouyang2022training}, effective data augmentation is the direct way to improve multilingual alignment.
Thus, it is beneficial to explore ways to construct multilingual security datasets facing the localization of security information across languages.
Depending on the business or scientific requirements, the datasets can contain different harmful question pairs and languages and then fine-tune the generic LLM with customized security datasets to construct a more robust version of LLMs.

\textbf{Pre-processing prompts or post-processing outputs.}  Another approach with fewer resources is to add proxy security efforts in small models outside of LLMs. Before constructing prompts for the LLM, we can consider maintaining a harmful vocabulary list or using another LLM as an evaluator to extract dangerous elements and judge the security of prompts, which can help detect unstable factors.
The judgment outputs from LLMs with scores indicate whether the risks are identified in the original inputs and need to be rewritten, similar to RAIN~\cite{li2023rain}.
The general goal is to force LLMs to reject harmful answers or replace them with a rule-based template.

\section{Multi-Domain LLMs in Multilingual Scenarios} \label{sec:multidomain} 

The remarkable capabilities of LLMs have facilitated their application across diverse domains, including finance~\cite{liu2021finbert,yang2023fingpt,liu2023fingpt,chen2023disc,wu2023bloomberggpt,yang2023investlm}, medicine~\cite{tian2023chimed, zhang2023alpacareinstructiontuned, luo2024taiyi,yang2023mentalllama, liu2023chatcounselor, huatuogpt-2023, xiong2023doctorglm}, law~\cite{yue2023disclawllm,SAILER,lawyer-llama-report,HanFei,cui2023chatlaw}, education~\cite{dan2023educhat,Taoli-LLama}, transportation~\cite{wang2024transgpt} etc. 
These domain-specific LLMs have demonstrated superb capability and promising perspective within associated domains. 
However, these LLMs are predominantly focused on English, while fewer cater to medium or low-resource languages, which dramatically hinders the utility of LLMs for a global audience. 
In this chapter, we present the pioneering multilingual studies conducted in the medical and legal domains\footnote{To the best of our knowledge, at the time we conduct our survey, there are few pertinent works in other domains such as finance and education in multilingual scenarios.}, and we conclude by offering a comprehensive discussion on the emerging limitations and challenges.

\begin{table*}[!t]  
\centering  
\small 
\resizebox{\textwidth}{!}{
\begin{tabular}{cccccccc}
\toprule

\multirow{3}{*}{Model} & \multirow{3}{*}{Params} & \multirow{3}{*}{Base Model} & \multicolumn{5}{c}{Training Corpus} \\ \cmidrule{4-8}
& & & Name &  \makecell[c]{Non-English\\proportion}  & Translated & Size & Languages\\ 
\midrule  
\renewcommand{\arraystretch}{4} BioMistral~\cite{labrak2024biomistral}   & 7B &  Mistral-7B & PMC  & 1.25\% & $\times$ & 3B & 10 \\
\renewcommand{\arraystretch}{4} MMedLM2~\cite{qiu2024towards}      & 7B  & InterLM2-7B & MMedC &  58\% & $\times$ & 25.5B & 6 \\  
\renewcommand{\arraystretch}{4} Apollo~\cite{wang2024apollo}  & 0.5/1.8/2/6/7B & Qwen/Gemma/Yi & Apollo & 33.72\% & $\times$ & 2.5B & 6\\ 
\renewcommand{\arraystretch}{4} Medical mT5~\cite{garcia2024medical} & 0.77/3B & mT5 & -  & 66.67\% & $\times$ & 3B & 4\\
\renewcommand{\arraystretch}{4} {L2M3~\cite{gangavarapu2024introducing}}  & {70B} & {\makecell[c]{Meditron\&\\SeamlessM4T}}& -  & 80\% & \checkmark &  0.9*5B & 5\\ 
\bottomrule 

\end{tabular}}
\caption{An overview of the existing competitive LLMs with multilingualism in the medical domain. \textbf{Languages} denotes the number of languages included in the training corpus. \textbf{Non-English proportion} denotes the proportion of non-English languages in the training corpus. The training dataset of L2M3 is obtained by translating English into four low-resource languages: Telugu, Hindi, Swahili, and Arabic.}
\label{table:multidomain_1}
\end{table*}

\begin{table*}[!t]  
\centering  
\small 
\resizebox{\textwidth}{!}{
\begin{tabular}{ c c c c c c }

\toprule
\multirow{2}{*}{Model} &  \multicolumn{5}{c}{Benchmark} \\ \cmidrule{2-6}
 & Name & Translated & Task & Metrics & Languages \\ 
\midrule  
\renewcommand{\arraystretch}{4} BioMistral\cite{labrak2024biomistral}   &  - & \checkmark &  Multi-choice QA & ACC & en,ar,zh,fr,de,pt,ru,es  \\ 
\renewcommand{\arraystretch}{4} MMedLM2\cite{qiu2024towards}      & MMedBench & \checkmark &  \makecell[c]{Multi-choice QA\\Rationale eval} & \makecell[c]{ACC\\ROUGE-1/BLEU-1} & en,zh,ja,fr,ru,es \\  
\renewcommand{\arraystretch}{4} Apollo\cite{wang2024apollo}   & XMedBench & $\times$ &  Multi-choice QA & ACC & en,zh,fr,es,ar,hi\\ 
\renewcommand{\arraystretch}{4}Medical mT5\cite{garcia2024medical}  & - & $\times$ &   \makecell[c]{Seqence labeling\\Abstractive QA} & F1 score & en, es, it, fr\\ 
 
\bottomrule 
\end{tabular}}
\caption{An overview of the existing cross-lingual medical benchmarks. For the Languages column, en, ar, zh, fr, de, pt, ru, es, ja, and hi represent English, Arabic, Chinese, French, German, Portuguese, Russian, Spanish, Japanese, Hindi, and Italian.}
\label{table:multidomain_2}
\end{table*}
 
\subsection{Medical Domain}
\label{sec:medical}
% \subsubsection{Current Research State}
Previous works have made substantial strides in integrating LLMs into the medical domain~\cite{tian2023chimed, zhang2023alpacareinstructiontuned,chen2023meditron}.
In particular, Med-PaLM2~\cite{singhal2023large} notably achieves success by passing the US Medical Licensing Examination. 
% and Meditron~\cite{chen2023meditron} expands medical pre-training to 70B parameters.
A common practice to adapt LLMs to the medical domain is to continually train the foundation model with domain corpus as introduced in Section~\ref{sec:cl}~\cite{luo2022biogpt}.
% \modif{For instance, BioGPT~\cite{luo2022biogpt} and ClinicalGPT~\cite{wang2023clinicalgpt} are fine-tuned based on GPT-2~\cite{radford2019language} and BLOOM~\cite{le2022bloom} respectively for medical applications.
% The other models such as PMC-LLaMA~\cite{wu2023pmc}, MedAlpaca~\cite{han2023medalpaca} and ChatDoctors~\cite{yunxiang2023chatdoctor} have been progressively fine-tuned LLaMA-2 on a medical related corpus, achieving dramatic medical performance enhancement over the original foundation model.}
For instance, some models such as BioGPT~\cite{luo2022biogpt}, ClinicalGPT~\cite{wang2023clinicalgpt}, PMC-LLaMA~\cite{wu2023pmc} and MedAlpaca~\cite{han2023medalpaca} have been progressively fine-tuned foundation models on the medical related corpus, achieving dramatic medical performance enhancement over the original foundation model.
In particular, ChatDoctors~\cite{yunxiang2023chatdoctor} resembles more of a multidisciplinary doctor, built on mixed data instead of solely medical data, capable of conducting patient-doctor dialogues, focusing on comprehensive inquiry services in the real-world scenario.
In addition, in the realm of traditional Chinese medicine, HuaTuoGPT2~\cite{chen2023huatuogpt} that is fine-tuned on the Baichuan~\cite{yang2023baichuan} with four types of data~(distilled instructions/conversations from ChatGPT~\cite{ouyang2022training} and real-world instructions/conversations from doctors) and capable of mimicking the diagnostic abilities of doctors provides useful medical information.
The experimental results demonstrate the capabilities of HuaTuoGPT2 surpassing those of the GPT-4~\cite{achiam2023gpt} in the 2023 national medical licensing examination of traditional Chinese medicine.
However, the aforementioned models primarily focus on a limited set of high-resource languages and show varying degrees of performance degradation when extended to other languages.
Such a phenomenon renders them unreliable, incapable, and insecure for application in linguistically diverse medical environments.

\subsubsection{Medical LLMs in Multilingual Scenarios}
To alleviate the issue of multilingualism in the medical domain, prior studies attempt to introduce multilingual medical corpus to enhance the multilingual ability of foundation models as shown in Table~\ref{table:multidomain_1} and Table~\ref{table:multidomain_2},
Specifically, KBioXLM~\cite{geng2023kbioxlm} adapts XLM-R~\cite{conneau2020unsupervised} to the medical domain, which encompasses diverse medical knowledge.
However, KBioXLM explores both the training corpus and evaluation data through translation, adapted to only two languages.
To further address the limited availability of data beyond English, L2M3~\cite{gangavarapu2024introducing} integrates Meditron-70B with the Meta SeamlessM4T machine translation system and separately fine-tune two components on four extremely low-resource languages. Garcia~\etal~\cite{garcia2024medical} explore the medical LLM on encoder-decoder architecture based on mT5\cite{xue2020mt5}.
In the monolingual setting, both in-context learning (ICL) and SFT demonstrate considerable improvements compared to the foundation models~\cite{han2023medalpaca,wu2023pmc,chen2023meditron}.
MMedLM2~\cite{qiu2024towards} underscores medical multilingualism by presenting a 25.5B massive cross-lingual training corpus covering a set of 6 languages.
By training with this corpus, the multilingual ability of the model improves in all 6 languages,
substantially surpassing the prior models and even rivaling GPT-3.5 and Gemini-1.0-pro.
Meanwhile, to further assess the multilingual generalization of medical LLMs, BioMistral~\cite{labrak2024biomistral} introduces the first large-scale multilingual medical benchmark of LLMs into 7 languages. 
Similar to MMedLM2~\cite{qiu2024towards}, the Apollo~\cite{wang2024apollo} introduces a training dataset ``Apollo Corpora'', a benchmark ``XMedBench'' and a collection of models ranging from 0.5B to 7B. 
% By meticulously collecting data exclusively from the local language and strictly prohibiting any forms of translation, the training corpus is rich with high-quality medical knowledge in each language. 
% Besides, the Apollo-7B achieves the best performance among prior multilingual medical LLMs with 70B parameters. 
% \modif{The comparison on the XMedBench} demonstrates that only a limited capacity boost can be gained when continually training the model on PubMed-Central corpus.
% \modif{In conclusion, Apollo has paved a big step in both the accuracy and multilingual ability of LLMs.} 
To investigate the effectiveness of LLMs in the multilingual medical domain, they build a benchmark ``XMedBench''.
The comparison on the XMedBench demonstrates that prior multilingual medical LLMs are limited by the PubMed-Central corpus which is constructed based on the translation technique.
To alleviate the issue of training data, they propose a training corpus ``Apollo Corpora'' which is rich with high-quality medical knowledge in each language.
The corpus is meticulously collected from the local language and strictly prohibits any form of translation.
As a result, the Apollo-7B is fine-tuned on the ``Apollo Corpora'' and can achieve better performance compared with prior multilingual medical LLMs with 70B parameters.

\subsubsection{Discussion and Challenges}
Existing studies achieve remarkable progress in medical LLMs for the multilingual scenario, yet numerous challenges are perverse. 
First, given the presence of language-specific knowledge that is highly pertinent to the local cultural, historical, political, and regional backgrounds, Wang~\etal~\cite{wang2024apollo} examine whether these language-specific medical knowledge stimulates or deteriorates with each other and investigate whether the model always outperforms its counterpart trained solely on the monolingual language corpus when further trained on the whole multilingual corpus.
For instance, the ability of the Spanish model trained in 6 languages surpasses that of a model only trained on a monolingual corpus in Spanish. 
Despite underlying conflicts between different language-specific medical knowledge and potential biases due to varying data, the performance boost suggests that cross-lingual joint training promotes the performance of medical LLMs, shedding light on the potential of cross-lingual pre-training.
Thus, further exploration into the effectiveness of the real-world and pseudo data is yet to be undertaken.

Second, the ongoing scarcity of medical data in various languages persistently hampers further advancement. 
Although translation can mitigate some of these issues, it may not be effective due to the complex medical terminology and challenges in precise translation. 
Moreover, each language might carry unique cultural and contextual differences, resulting in abundant language-specific medical knowledge intricately linked to cultural, historical, political, and regional backgrounds. 
For example, traditional Chinese medicine embodies a rich history deeply intertwined with its cultural heritage, constituting a distinct medical system. 
Enabling our systems to grasp such language-specific medical knowledge and provide tailored medical assistance for specific groups poses a formidable challenge. 
Hence, exploring the underlying mechanism of language-specific medical knowledge integration is a promising research direction.

\begin{table*}[t]  
\centering  
\small 
\resizebox{\textwidth}{!}{
\begin{tabular}{ c  c }

\toprule
Categories & Details \\
\midrule  
Component & Pile of Law(292 GB), Eurlex Resources2 (179 GB), Native Multi Legal Pile (112 GB), Legal MC43 (106 GB)\\ 

Text Type Distribution & case law(51.4\%), legal-mc4(16.6\%), legislation(12.6\%), contracts(9.23\%), other(10.2\%) \\

Top 5 Languages & Portuguese(15.93\%), German(6.29\%), Spanish(6.1\%), French(3.32\%), Italian(2.92\%) \\

Last 5 Languages & Maltese(0.43\%), Lithuanian(0.43\%), Latvian(0.42\%), Croatian(0.3\%), Irish(0.08\%) \\
 
Non-English proportion & 48.27\%\\

Total Words & 86.36B \\
\bottomrule 

\end{tabular}}
\caption{The proportion of different languages in MultiLegalPile (689GB multilingual legal corpus).}
\label{table:multidomain_3}
\end{table*}

\subsection{Legal Domain}
\label{sec:legal}
\subsubsection{Legal LLMs in Multilingual Scenarios}
Similar to the medical domain, the applications of LLMs in the legal domain principally center on English. Several precedent attempts such as Chatlaw~\cite{cui2023chatlaw}, Lawyer LLaMA~\cite{huang2023lawyer}, SaulLM~\cite{colombo2024saullm}, and LegalBERT~\cite{chalkidis2020legal} expand the general ability of LLMs to the legal domain.
The universal performance degradation has been observed when expanding to other languages. 

To address the specific problems in the legal domain, the proposed models need to adapt the legal features, which are factual, ambiguous, structured, and timely~\cite{bansal2019review,gordon2009rules}, compared to other domains. 
Brugger~\etal~\cite{brugger2023multilegalsbd} take a preliminary step in the multilingual legal Sentence Boundary Detection (SBD) covering 6 languages. %In spite of the complex sentence structure and recondite legal terminology, transformer-based methods show decent performance in both monolingual and multilingual settings compared with conventional approaches (such as NLTK, CRF, and BiLSTM-CRF) and reveal decent zero-shot cross-lingual transferability. 
Christen~\etal~\cite{christen2023resolving} conduct similar research on the multilingual Negative Scope Resolution (detecting words affected by negation cue) task and Baumgartner~\etal~\cite{baumgartner2024towards} extend legal judgment prediction to German, French and Italian.
%, stimulating the explainability and transparency during the legal reasoning process. Yet only negligible amount of LLM-based methods and works have been proposed.
Furthermore, for a comprehensive evaluation, Niklaus~\etal~\cite{niklaus2023lextreme} collect 11 natural language understanding legal datasets covering a total of 24 languages and 8 subdivisions, which is the first cross-lingual legal benchmark (LEXTREME) showing there is still improvement room even for the popular models like ChatGPT. 
After the ChatGPT series models were proposed, Nguyen~\etal~\cite{nguyen2024gpts} implement a preliminary empirical comparison between ChatGPT and GPT-4 on the ``COLIEE'' benchmark~\cite{rabelo2020summary,rabelo2022overview} encompassing Japanese and English, where GPT-4 consistently surpasses its predecessor but still falls behind human performance. %On evaluating the fairness and performance divergence between different legal groups, legal systems (European Council, USA, Switzerland, and China), and group attributes i.e gender, region, racial, age, \cite{chalkidis2022fairlex} performs continuing training with classification-based tasks and results in four mini transformer encoder-based models. These vanilla models do demonstrate different levels of unfairness among different groups, with absolute differences up to 21.5\%, yet the bias-mitigation techniques fail to realize universally efficient bias reduction.

%Component & \makecell[c]{Pile of Law(292 GB), Eurlex Resources2 (179 GB)\\Native Multi Legal Pile (112 GB), Legal MC43 (106 GB)}\\ 
%Text Type Distribution & \makecell[c]{case law(51.4\%), legal-mc4(16.6\%), legislation(12.6\%)\\contracts(9.23\%), other(10.2\%)} \\
%Non-English proportion & \makecell[c]{48.27\%\\} \\
%Top 5 Languages & \makecell[c]{Portuguese(15.93\%), German(6.29\%), Spanish(6.1\%)\\French(3.32\%), Italian(2.92\%)} \\
%Last 5 Languages & \makecell[c]{Maltese(0.43\%), Lithuanian(0.43\%), Latvian(0.42\%)\\Croatian(0.3\%), Irish(0.08\%)} \\

% Only negligible amount of LLM-based methods and works have been proposed until 
For a more in-depth exploration of LLMs in multilingual legal issues, Niklaus~\etal~\cite{niklaus2023multilegalpile} construct a multilingual legal-domain corpus with 689GB, MultiLegalPile, whose detail is shown in Table~\ref{table:multidomain_3}. 
They train two PLMs based on XLM-R~\cite{conneau2020unsupervised} and the Longformer~\cite{beltagy2020longformer} with this corpus, achieving the state-of-the-art performance on the LEXTREME~\cite{niklaus2023lextreme} compared to XLM-R in most languages. 
% benchmark and dramatic competence elevation over the original vanilla XLM-R in most languages. 
Moreover, the English version of the model achieves state-of-the-art performance on 5 out of 7 tasks in LexGLUE~\cite{chalkidis2022lexglue}, underscoring the exceptional cross-lingual ability and legal-domain expertise. 
Trautmann~\etal~\cite{trautmann2022legal} focus on employing legal prompt engineering (LPE) to enhance the capabilities of LLMs, thereby mitigating the challenges posed by the scarcity of cross-lingual legal data and the substantial computational resources required. 
Although improvements over baselines are observed, bridging the performance gap to match the supervised methods remains a significant endeavor.

\subsubsection{Discussion and Challenges}
% Despite current attempts, there remains a dearth of powerful multilingual LLMs capable of comprehensively and accurately conducting law-related tasks, underscoring the need for future endeavors in this area. Moreover, several challenges remain. In addition to the data-scarcity problem, the issue of language-specific knowledge aggregates in the legal domain even further, as the legal systems or jurisdictions in different regions greatly vary from each other. As multilingual LLMs already face difficulties in learning to represent semantic features in low-resource languages, it would be even more difficult to accurately capture and grasp legal nuances in various jurisdictions. Apart from the spatial problem, the temporal challenge persists as laws are constantly altered, amended, or abolished, requiring the models to stay up to date.
Despite current efforts, there is still a lack of robust LLMs capable of effectively and comprehensively performing law-related multilingual tasks, highlighting the need for further exploration in this domain. 
% Additionally, several challenges persist. 
Beyond the issue of data scarcity, the accumulation of language-specific legal knowledge compounds the complexity, as legal systems and jurisdictions vary significantly across regions.
Given that existing LLMs already struggle with representing semantic features in low-resource languages, accurately capturing legal nuances across diverse jurisdictions poses an even greater challenge. 
Moreover, the temporal dimension adds complexity, as laws undergo constant revision, amendment, or abolition, necessitating that models remain continuously updated.
%In essence, the multilingual legal domain represents a novel and pioneering field with vast potential and promising prospects. Moreover, it plays a crucial role in advancing global legal systems and providing timely legal support. As research and development efforts continue, the possibilities for innovation and progress in this field are endless, offering a bright future for multilingual legal LLMs.

\subsection{Limitations and Future Directions on Multi-Domain}
% translation problem(import bias and so forth), in-domain knowledge differs from languages, amount of available corpus

Despite the remarkable advancements in multilingual LLMs, persistent limitations and challenges necessitate further exploration.
This survey provides a brief discussion of the current limitations and potential improvement in the following two parts.

\textbf{Data scarcity and translation issues.}
A powerful multilingual LLM, especially in specific domains, is predominantly hindered by the scarcity of domain data. 
Although knowledge transfer provides some relief, the issue of under-representation persists, particularly for low-resource languages, and becomes further compounded when extending to specific domains.
Machine translation techniques offer a potential solution to mitigate this issue, however, they lead to new challenges~\cite{nicholas2023lost}. 
On one hand, machine translation systems introduce errors, particularly when handling domain-specific terminology across multiple languages. 
Terms or phrases that native speakers do not use may also be included in the translated corpus. 
On the other hand, the models suffer from comprehensively understanding and accounting for the local and cultural context of the target language, complicating the task of in-depth and high-level feature learning and capturing. 

\textbf{Language-Specific knowledge integration}
In specific contexts such as the legal or financial domain, each language harbors distinctive knowledge influenced by diverse historical, cultural, and regional backgrounds.
Beyond linguistic semantics, the challenge arises in capturing these nuances among various languages and integrating language-specific domain knowledge into LLMs. 
For example, disparities in legal definitions between European Council and USA jurisdictions, as well as the contrast between traditional Chinese medicine and Western medicine, indicate these challenges.
Current LLMs face challenges in effectively understanding such language-specific knowledge, hindering their capacity to provide customized domain-specific assistance for diverse populations.
Further research is necessary to explore how LLMs can integrate and leverage this particular type of knowledge.

% \paragraph{Bias Mitigation Across Languages}

% As the potential of LLMs expands, it is crucial to ensure equitable utilization without bias or prejudice, particularly when more language groups are included. We can either transfer the current English-centered mitigation approaches to other languages or perform specific multilingual bias mitigation\cite{reusens2023investigating} by learning to identify inappropriate content such as prejudice, hate speech, disinformation and discrimination in various languages and eliminate them during different parts of LLM such as word representation, pre-training process, fine-tuning stage and prompt engineering techniques.\\

% The following content is generated by Gemini
These limitations highlight the need for further research efforts in the following directions:
\begin{itemize}
\item Development of strategies to construct high-quality, domain-specific multilingual datasets that preserve cultural context.
% \item Enhancement of the multilingual language modeling ability to incorporate domain-specific knowledge.
\item Exploration of techniques for LLMs to effectively integrate and leverage language-specific in-domain knowledge.
% \item Design of bias detection and mitigation methods that are specifically tailored for multilingual LLM applications.
\end{itemize}
By addressing these two challenges, researchers can pave the way for the development of truly robust and equitable LLMs that serve a global audience in multilingual scenarios.

\section{Multilingual Data Resource}
\label{sec:dataset}

\begin{table*}[!t]
\centering
\resizebox{1.0\textwidth}{!}{
\begin{tabular}{lcccll}
\toprule
 Name & Release Time & Languages & Size &  Sources & Affiliation \\

 \midrule
 Amazon intent~\cite{amazon_massive_intent}  & 2022 & 49 & 2.02 GB & Amazon & Amazon  \\
 Amazon reviews~\cite{amazon_reviews_multi} & 2020 & 6 & - & Amazon & Amazon   \\
 Aya~\cite{singh-etal-2024-aya} & 2024 & 114 & 156 GB & Mixed & Cohere\\
 Bactrian-x~\cite{li2023bactrianx}  &   2023  &   51  &  -  &   -  &   MBZUAI  \\
 Biblenlp~\cite{biblenlp-corpus-mmteb} & 2024 & 861 & 581 MB & Bible & -   \\
 Bloom-lm~\cite{le2022bloom}  & 2022 & 364 & - & - & SIL International   \\
 CC100~\cite{cc100_1}~\cite{cc100_2} & 2020 & 109 & 185 GB & CC & Facebook \\
 CulturaX~\cite{nguyen2023culturax} & 2023 & 167 &  27 TB & mC4, OSCAR &  University of Oregon \\
 GPT-4 Prompts~\cite{GPT-4-Prompts}  & 2024 & 5 & 1.05 GB & GPT-4 & -   \\
 Guanaco~\cite{guannaco}  & 2023 & 4 & 400 MB & - & University of Washington  \\
 HPLT~\cite{hplt_monolingual_v1_2}  & 2023 & 75 & 22 TB & CC, Internet Archive & HPLT project  \\
 IWSLT 2017~\cite{iwslt2017} & 2017 & 10 & 4.24 GB & TED Talks  & FBK \\
 mC4~\cite{mc4} & 2019 & 101 & 9.7 TB & CC & Google\\
 Mewsli-x~\cite{mewsli-x} & 2021 & 50 & 285 MB & WikiNews and Wikipedia & DeepMind  \\
 Minds14~\cite{minds14} & 2021 & 12 & 471 MB & Banking Assistant & PolyAI Limited\\
 MLDR~\cite{MLDR} & 2024 & 13 & - & Wikipeida, Wudao, mC4 & BAAI  \\
 MMedC~\cite{MMedC} & 2024 & 6 & 105 GB & Multiple & SJTU   \\
 MQA~\cite{mqa} & 2021 & 36 & 122 GB & CC & CLiPS Research \\
 Multi-sentiments~\cite{multilingual-sentiments} & 2022 & 12 & 141 MB & Multiple Sources & -   \\
 Multiconer~\cite{multiconer2-data}~\cite{multiconer2-report} & 2023 & 12 & 338 MB & Wikipedia & Amazon  \\ 
 Open Subtitles~\cite{lison2016opensubtitles2016} & 2023 & 58 & 273 MB & Subtitles & -   \\
 OSCAR~\cite{OSCAR} & 2020 & 166 & 6.3 TB & CC & University of Orego  \\
 Para-pat~\cite{para_pat}  & 2020 & 15 & 2.57 GB & Google Cloud & University of Sheffield   \\
 Project Gutenberg~\cite{project_gutenberg}  & 2023 & 11 & 14.4 GB & eBook & Project Gutenberg  \\
 ShareGPT~\cite{ShareGPT52K} & 2023 & 5 & 1.08 GB & GPT / Human & Ronso  \\
 SREDFM~\cite{SREDFM} & 2023 & 18 & 8.29 GB & Wikidata, Wikipedia & Babelscape  \\
 TED Talks~\cite{ted_talks}  & 2018 & 55 & - & TED Talks & CMU   \\
 TED-talks-iwslt~\cite{ted_talks_iwslt} & 2012 & 104 & 25 GB & TED Talks & FBK  \\
 Toxi-text~\cite{toxi-text-3M} & 2023 & 55 & 1.96 GB & - & -   \\
 UD~\cite{universal_dependencies} & 2023 & 102 & 2.19 GB & - & Universal Dependencies  \\
 Wikiann~\cite{wikiann1}~\cite{wikiann2} & 2019 & 173 & 143 MB & Wikipedia & RPI\\
 Wikipedia~\cite{wikipedia} & 2024 & 322 & 71.8 GB & Wikipedia & Wikimedia Foundation \\
 Wit Base~\cite{witbase}  & 2021 & 105 & 5.15 GB & Wikipedia & Google \\
 xP3~\cite{xP3} & 2022 & 277 & - & - & Cohere For AI  \\

% M-arc~\cite{m_arc} & 2024 & 35 & 72.9MB & a variety of sources & University of Oregon, etc.   \\

\bottomrule
\end{tabular}}

\caption{An overview and statistic detail of the representative multilingual data resource. We only include the large-size datasets with much more supported languages.}
\label{table:dataset}
\end{table*}
LLMs are data-driven, thus the impressive learning capabilities of LLMs stem from their massive model sizes and extensive training datasets, which have been proven in high-resource languages.
However, English stands as the closest approximation to the lingua franca, wielding dominance across various domains. With the largest number of total speakers, its prominence extends far and wide~\cite{pereltsvaig2020languages}, where English reigns as the primary language of the internet~\cite{schneider2018english,pennycook2002english}.
Meanwhile, English is the main language used in the higher economic status countries of the world, such as America, Britain, and other Western countries~\cite{joshi-etal-2020-state}.
Therefore, existing data resources focus on the English-centric, which came at the expense of regional and indigenous languages, exacerbating language endangerment and economic marginalization~\cite{dodge2021documenting}.
Due to the lack of resources, this situation deeply restricts the development of multilingual models, and it is a vicious circle~\cite{caswell-etal-2020-language}.
Moreover, low-resource languages suffer from lower quality, due to mislabeling or inadequate representation of native usage. This situation is especially prevalent with web-crawled data which predominantly consists of pornographic, nonsensical, or non-linguistic content~\cite{kreutzer2022quality}.

As shown in Table~\ref{table:dataset}, we collect as much large-scale multilingual data resource as is reliable. We can observe that the scale of existing multilingual resources is much smaller than that of English monolingual resources (only four datasets with the TB level). And the low-/medium-resource languages typically derive data from a narrower range of sources compared to their high-resource counterparts~\cite{nekoto-etal-2020-participatory}. The available data mainly originates from sources, such as Wikipedia, the Bible, and Common Crawl.
In addition, these data suffer from bias and fairness, which we will present in Section~\ref{sec:bias}. For instance, a gender bias exists in Wikipedia, with studies revealing a persistently low percentage of women editing articles~\cite{callahan2011cultural}.
The vicious circle of multilingual data includes the issue of open source issues. Due to the high cost of high-quality multilingual resources, researchers and companies are reluctant to share resources in the open-resource community (i.e., ``data island'')~\cite{peris2023privacy,yao2024survey,yan2024protecting,kandpal2022deduplicating}. This situation results in multilingual research staying at the data level and ignoring the competition in the model paradigm. 
The united governments, companies, and researchers must start a virtuous cycle of multilingual data resources.
Access to abundant, meticulously collected datasets in a language empowers researchers and developers to construct models and benchmarks. The abundance of models and benchmarks, in turn, fosters increased publication, facilitates communication, and promotes real-world application scenarios for companies.
These outputs have the potential to attract more users, while government-mandated guidelines help generate non-toxic data, which can be repurposed for further research and development.
% Along with the progress in natural language processing, LLMs have been frequently made accessible to the public to foster deeper investigation and applications. However, when it comes to training datasets for these LLMs, especially the recent state-of-the-art models, they are often not fully disclosed. 
% Creating training data for high-performing LLMs involves extensive cleaning and deduplication to ensure the necessary level of quality. The lack of transparency for training data has thus hampered research on attributing and addressing hallucination and bias issues in LLMs, hindering replication efforts and further advancements in the community. 
% These challenges become even more pronounced in multilingual learning scenarios, where the available multilingual text datasets are often inadequately collected and cleaned. 
\section{Multilingual Benchmark and Evaluation}
\label{sec:benchmark}

\begin{table*}[!t]
\centering
\resizebox{1.0\textwidth}{!}{
\begin{tabular}{lccccc}
\toprule
 Name & Release Time & Languages & Parallel & Type & Affiliation \\

 \midrule

% NLI
M-Hellaswag~\cite{dac2023okapi}  &   2023  &   35  &  \checkmark  &   Commonsense NLI  &   University of Oregon   \\
XNLI~\cite{conneau2018xnli}  &   2018  &   15  &  \checkmark  &   NLI &   NYU   \\
Multilingual-Fig-QA~\cite{kabra-etal-2023-multi}  &   2023  &   7  &   $\times$  &   NLI &  CMU  \\

% RAG
NoMIRACL~\cite{thakur2023nomiracl}~\cite{thakur2024nomiracl} & 2023 & 16 & $\times$ & RAG & University of Waterloo \\
MIRAGE-Bench~\cite{thakur2024miragebenchautomaticmultilingualbenchmark} & 2024 & 19 & $\times$ & RAG & University of Waterloo \\

% summarization
Cross-Sum~\cite{bhattacharjee2023crosssum}  &   2021  &   45  &  \checkmark  &   Summarization  &   BUET   \\
XL-SUM~\cite{hasan2021xlsum}  &   2021  &   44  &   $\times$  &   Summarization  &   BUET  \\
Pmindiasum~\cite{urlana2023pmindiasum}  &   2023  &   14  &  \checkmark  &   Summarization  &   IIIT Hyderabad  \\
SEAHORSE~\cite{clark2023seahorse}  &   2023  &   6  &   $\times$  &   Summarization  & DeepMind  \\ 
M3LS~\cite{verma-etal-2023-large}  &   2023  &   20  &   $\times$  &   Summarization  &   IITs   \\

% QA
BELEBELE~\cite{bandarkar2023belebele}  &   2023  &   122  &  \checkmark  &   Question Answering  &   Meta  \\
BioInstructQA~\cite{BioInstructQA}   &   2024  &   7  &  \checkmark  &   Question Answering  &   Avignon Université \\
MLQA~\cite{lewis2020mlqa}  &   2020  &   7  &  \checkmark  &   Question Answering  &   Facebook   \\
TyDiQA~\cite{clark2020tydi}  &   2020  &   11  &   $\times$  &   Question Answering  &   Google   \\
XOR-TyDi~\cite{asai-etal-2021-xor} &   2021  &   11 &  $\times$ &  Question Answering &   University of Washington \\
XQuAD~\cite{artetxe2020cross}  &   2020  &   10  &  \checkmark  &   Question Answering  &   University of the Basque Country \\
MaXM~\cite{changpinyo2023maxm}  &   2023  &   7  &  \checkmark  &   Visual Question Answering  &   Google  \\

% IR
CIRAL~\cite{mofetoluwa2024ciral}  &   2024  &   4  &  $\times$ &   Information Retrieval  &   University of Waterloo   \\
MIRACL~\cite{miracl} & 2023 & 18 & - & Information Retrieval & University of Waterloo  \\
LAReQA~\cite{roy-etal-2020-lareqa} &   2020 &   11  &  \checkmark &   Information Retrieval  &   Google Research   \\
STSB-multi-mt~\cite{huggingface:dataset:stsb_multi_mt}  &   2021  &   10  &  \checkmark  &   Text Similarity  & -     \\

% NER
MasakhaNER~\cite{adelani2021masakhaner}  &   2021  &   10  &  \checkmark  &   Named Entity Recognition  &   Masakhane   \\
Multi-CoNER~\cite{malmasi2022multiconer}  &   2022  &   11  &   $\times$  &   Named Entity Recognition  &   Amazon  \\

% commonsense reasoning
XCOPA~\cite{ponti-etal-2020-xcopa}  &   2020  &   12  &  \checkmark  &   Commonsense Reasoning  &   Cambridge University \\
XCSQA~\cite{lin2021common}  &   2021  &   11  &  \checkmark  &   Commonsense Reasoning  &  USC   \\
XStoryCloze~\cite{lin2022few}  &   2022  &   11  &  \checkmark  &   Commonsense Reasoning  &   Meta  \\
XWinograd~\cite{tikhonov-ryabinin-2021-heads}  &   2021  &   6  &   $\times$  &   Commonsense Reasoning  &   Yandex  \\
XCSR~\cite{lin-etal-2021-common}   &   2021  &   16  &  \checkmark  &   Commonsense Reasoning  &   USC   \\

% machine translation
% DiaBLa~\cite{bawden2019diabl}  &   2019  &   2  &  \checkmark  &   Machine Translation  &   University of Edinburgh  \\
% FLORES-101~\cite{goyal-etal-2022-flores}  &   2022  &   101  &  \checkmark  &   Machine Translation  &   Meta   \\
FLORES-200~\cite{nllb2022}~\cite{flores200_2}~\cite{flores200_3}  &   2022  &   200  &  \checkmark  &   Machine Translation &   Meta  \\
OPUS-100~\cite{zhang2020improving}  &   2020  &   100  &  \checkmark  &   Machine Translation &   University of Edinburgh \\
Tatoeba-mt~\cite{tiedemann-2020-tatoeba}   &   2020  &   93  &  \checkmark  &   Machine Translation &   Tatoeba.org  \\
% WMT  &   -  &   -  &  \checkmark  &   Machine Translation  &   WMT, etc.   \\

% code generation
Humaneval-XL~\cite{peng2024humaneval}   &   2024  &   23  &  \checkmark  &   Code Generation   &   University of Copenhagen    \\
% Mconala  &   2022  &   3  &  \checkmark  &   Code generation  &   CMU   \\
ODEX~\cite{wang2023executionbased}  &   2022  &   14  &  \checkmark  &   Code Generation  &   CMU   \\

% classification
MARC~\cite{keung2020multilingual}  &   2020  &   6  &   $\times$  &   Text Classification  &   Amazon   \\
Masakhanews~\cite{adelani2023masakhanews}  &   2023  &   16  &   $\times$  &   News Topic Classification   &   Masakhane   \\
MULTIEURLEXDOC~\cite{chalkidis-etal-2021-multieurlex}  &   2021  &   23  &  \checkmark  &   Legal Topic Classification  &   University of Copenhagen  \\
Sib200~\cite{adelani2023sib200}  &   2024  &   205  &  \checkmark  &   Topic Classification  &  UCL   \\

 Afrisent~\cite{muhammad2023afrisenti}  &   2023  &   14  &   $\times$  &   Sentiment Analysis  &   U.Porto   \\
 
% AGIEval~\cite{zhong2023agieval}  &   2023  &   2  &   $\times$  &   Examination  &   Microsoft  \\
ASPEN~\cite{razumovskaia2024little}  &   2022  &   31  &  \checkmark  &   Story Planning  &   Cambridge University  \\
% Bactrian-x~\cite{li2023bactrianx}  &   2023  &   51  &  \checkmark  &   Training Set  &   MBZUAI  \\

% Clidsum~\cite{wang-etal-2022-clidsum}  &   2022  &   3  &  \checkmark  &   Summary Generation  &   Wechat   \\
% Crocosum~\cite{zhang2023crocosum}  &   2023  &   2  &   $\times$  &   Summary Generation  &   Brown University  \\

Crossmodal-3600~\cite{thapliyal2022crossmodal3600}  &   2022  &   36  &  \checkmark  &   Image Captioning  &   Google   \\
% Culturax~\cite{nguyen2023culturax}  &   2023  &   167  &   $\times$  &   Training Set  &   University of Oregon   \\
Exams~\cite{hardalov2020exams}  &   2020  &   15  &  \checkmark  &   Examination  &   Sofia University  \\

Fairlex~\cite{chalkidis2022fairlex}  &   2022  &   5  &   $\times$  &   Fairness  &   University of Copenhagen    \\
GEOMLAMA~\cite{yin2022geomlama}  &   2022  &   5  &  \checkmark  &   Knowledge Diversity  &   University of California   \\

M-MMLU~\cite{dac2023okapi}  &   2023  &   35  &  \checkmark  &   NLU  &   University of Oregon   \\
M3Exam~\cite{zhang2023m3exam}  &   2023  &   9  &   $\times$  &   Examination  &   DAMO Academy   \\
% Madlad-400  &   2023  &   419  &   $\times$  &   Training Set  &   DeepMind  \\

MASSIVE~\cite{fitzgerald2022massive}  &   2022  &   51  &  \checkmark  &   Intent Recognition, Slot Filling  &   Amazon \\
Mela~\cite{zhang2024mela}  &   2022  &   10  &   $\times$  &   Linguistic Acceptability  &  SJTU   \\
MGSM~\cite{shi2022language}  &   2022  &   11  &  \checkmark  &   Mathematical Reasoning  &   Google \\
% Mildsum  &   2023  &   2  &  \checkmark  &   Legal Judgment  &   IITs \\
% MixATIS++  &   2020  &   9  &  \checkmark  &   Training Set  &   UMD \\

MMedBench~\cite{qiu2024building}  &   2023  &   6  &   $\times$  &   Medical  &   SJTU   \\
% MTOP  &   2021  &   6  &  \checkmark  &   Training Set  &   Facebook   \\
% MultiCochrane  &   2023  &   4  &  \checkmark  &   Medical Text Simplification  &   UT  \\ 
% Normdial  &   2023  &   2  &   $\times$  &   Training Set  &   Columbia   \\
% Normsage  &   2024  &   2  &   $\times$  &   Extract culture specific norms   &   UIUC \\

Paws-X~\cite{yang-etal-2019-paws}  &   2019  &   7  &  \checkmark  &   Paraphrase Identification  &   Google  \\

% Readme++  &   2023  &   5  &   $\times$  &   Training Set  &   Georgia Tech  \\
SMiLER~\cite{seganti-etal-2021-multilingual}  &   2021  &   14  &   $\times$  &   Entity and Relation Extraction  &   Samsung \\
Tydip~\cite{srinivasan-choi-2022-tydip}  &   2022  &   9  &   $\times$  &   Identify Politeness Levels  &   UT-Austin   \\

% WIKI-DOC  &   2023  &   23  &  \checkmark  &   Training Set  &   AWS AI Labs, etc.   \\
% WikiAnn  &   2017  &   282  &   $\times$  &   Name Tagging and Linking  &   RPI   \\ 

X-CLAIM~\cite{mittal-etal-2023-lost}  &   2023  &   6  &   $\times$  &   Realworld Claims  &   MBZUAI \\
% X-PARADE  &   2024  &   4  &   $\times$  &   Natural Language Inference  &   UT   \\
X-RiSAWOZ~\cite{moradshahi2023xrisawoz}  &   2023  &   6  &  \checkmark  &   Dialogue Utterances  &   Stanford University \\

xDial-Eval~\cite{zhang-etal-2023-xdial}  &   2023  &   10  &  \checkmark  &   Dialogues   &   NUS   \\

% XKaggle-DBQA  &   2022  &   4  &  \checkmark  &   Text-to-SQL Semantic Parsing  &  Waterloo  \\
% XLINGHEALTH  &   2023  &   4  &  \checkmark  &   Question Answering  &   Georgia Tech   \\

PRESTO~\cite{goel2023presto}  &   2023  &   6  &   $\times$  &   Conversational Parsing  &   University of Rochester  \\
Universal Dependencies~\cite{universal_dependencies}  &   2022  &   146  &   $\times$  &   Parser  &   -   \\
XSEMPLR~\cite{zhang2023xsemplr}  &   2023  &   22  &   $\times$  &   Semantic Parsing  &   PSU   \\

Bucc-bitext-mining~\cite{bucc-bitext-mining}   &   2022  &   5  &   $\times$  &   Mixed  &   HuggingFace  \\
MEGA~\cite{ahuja2023mega}  &   2023  &   70  &   $\times$  &   Mixed  &  UW \\
MEGAVerse~\cite{ahuja2024megaverse}  &   2023  &   83  &   $\times$  &   Mixed  &   Microsoft   \\
NusaX~\cite{winata2023nusax}  &   2023  &   10  &  \checkmark  &   Mixed   &   Bloomberg   \\
XGLUE~\cite{liang2020xglue}  &   2020  &   -  &   $\times$  &   Mixed  &   Microsoft   \\
XTREME~\cite{hu2020xtreme}  &   2020  &   40  &  \checkmark  &   Mixed  &  CMU   \\
XTREME-R~\cite{ruder-etal-2021-xtreme}  &   2021  &   50  &  \checkmark  &   Mixed  &   DeepMind  \\

% MathVista   &   2024  &   3  &   $\times$  &   Mathematical reasoning  &   UCLA  \\
% Alpaca-CoT   &   2023  &   3  &   $\times$  &   Instruction-Finetuning  &   Chinese Academy of Sciences  \\
\bottomrule
\end{tabular}}

\caption{An overview of multilingual benchmarks with more than four supported languages after 2018 in which mBERT~\cite{devlin2018bert} was proposed. \textbf{NLU} and \textbf{NLI} denote the natural language understanding and the natural language inference task, respectively.}
\label{table:benchmark}
\end{table*}

With the emergence of novel models and algorithms, researchers inevitably scrutinize the capabilities by evaluating their performance on specific and challenging tasks~\cite{chang2024survey}.
Recently, LLMs have garnered significant interest in the academia and industry. 
The remarkable performance of LLMs has been proven on a variety of tasks, showing strong universality compared with prior PLMs limited to specific tasks.
As shown in Table~\ref{table:benchmark}, to the best of our knowledge, we list the representative multilingual benchmarks after 2018 in which mBERT~\cite{devlin2018bert} was proposed.
From the statistical results, we can observe many kinds of existing benchmarks, but there remain some issues with these benchmarks described as follows.
\begin{itemize}
    \item \textit{Lack of task types.} Most of the multilingual benchmarks focus on understanding tasks and lack generation tasks that mainly consist of machine translation and summarization. However, LLMs are generative paradigms, and generation tasks closer to the real world should be introduced to evaluate the effectiveness of LLMs in multilingual scenarios. In addition, the evaluation dimensions of LLMs need to be richer today, and there is a lack of evaluation tasks for LLMs, such as safety, agents, social simulation, etc.
    \item \textit{Language culture and domain.} Existing multilingual benchmarks often rely on machine-translated text, which may contain errors or terms not commonly used by native language speakers. The benchmark of native usage with language culture habits is urgently required. Besides, the challenges within different language environments vary significantly, necessitating a thorough exploration of multilingual and multi-domain issues.
    \item \textit{Unified framework.} The number of benchmarks is sufficient but there is a lack of an authoritative and unified evaluation framework, which can be updated over time, and the evaluation aspect is more comprehensive based on the proposed framework. This phenomenon can be attributed to the dominance of English and individuals' focus on their native languages. To address this issue, collaborative efforts from the multilingual community are essential.
    \item \textit{Data leakage.} The main differences between LLMs lie in the training data and model size. The existing systems primarily prioritize real focus on user experience (open test), rather than assessing effectiveness through closed test sets that align with fair training data. Thus, there is a potential for data leakage as models may have inadvertently learned from the test set, especially the closed LLMs. It requires the evaluation methods to adopt a more secure strategy to reduce the risk of data leakage.
    \item \textit{Evaluation methods.} There are limitations in existing evaluation methods, particularly in generating tasks such as BLEU and ROUGE, thus manual evaluation is more reliable than automatic evaluation. 
    However, due to the diversity of multilingual tasks, performing manual evaluation requires numerous language experts, which leads to increased costs and makes it challenging to achieve multilingual tasks. Therefore, reliable automatic evaluation methods are needed, which would also advance the development of evaluation techniques.
\end{itemize}

\section{Bias and Fairness}
\label{sec:bias}

\subsection{Bias Categories}
The bias of LLMs in the multilingual scenario can be divided into two categories: language bias and demographic bias~\cite{xu2024survey}. Intuitively, the former is due to the imbalance of available training corpus for different languages~\cite{wei2023polylm, nicholas2023lost}, where English possesses the most text corpus~\cite{dodge2021documenting, penedo2024fineweb, gao2020pile, soldaini2024dolma}, resulting in the performance degradation of the LLMs when generalized to other language settings~\cite{zhang2023don, zhang2024m3exam}. 
The demographic bias occurs due to embedded biases and misinformation on the internet, leading to LLMs unavoidably inheriting demographic biases across gender, race, and political backgrounds~\cite{ferrara2023should, wang2022assessing, levy2023comparing, yeh2023evaluating}. This exacerbates existing inequalities, perpetuates stereotypes, and reinforces discrimination.

The core of multilingual LLM research is to improve the language modeling ability in other medium/low-resource languages while maintaining competence in English.
Concerning demographic biases, previous attempts to mitigate bias and align LLMs with human values have primarily concentrated on English~\cite{hutchinson2020social, nadeem2021stereoset, wan2023kelly}.
Consequently, bias and ethical issues persist in other languages, potentially leading to significant negative impacts for non-English-speaking users.

\begin{table*}[!t]  
\centering  
\small 
\resizebox{\textwidth}{!}{
\begin{tabular}{ccccc}
\toprule

\textbf{Name} & \textbf{Languages} & \textbf{Type of Bias}  & \textbf{Debias Object} & \textbf{Metrics} \\ 
\midrule  
\renewcommand{\arraystretch}{4} RTP-LX~\cite{de2024rtp}   & 28 &  \makecell[c]{Bias, Identity attack, Insult\\Microaggression, Self-harm\\Sexual content, Toxicity, Violence}  & \makecell[c]{Small LM\\Large LM} & \makecell[c]{Exact Label Match\\Interrater Reliability~\cite{de2024rtp}}\\
 
\renewcommand{\arraystretch}{4} MGB~\cite{yu2024your}  & zh,en,de,pt,es& Gender Bias &  Masked LM & MBE, MGL, LSG, MSG~\cite{yu2024your} \\  
 
\renewcommand{\arraystretch}{4} MIBs~\cite{zhao2020gender}  & en,es,de,fr& Gender Bias, Occupation Bias & Word Embedding   & inBias~\cite{zhao2020gender}\\  
\renewcommand{\arraystretch}{4} MozArt~\cite{piqueras2022pretrained} & en,de,es,fr  & Gender Bias, Language & Masked LM & Close Test\\
\bottomrule 
 
\end{tabular}}
\caption{An overview of the multilingual bias evaluation datasets. The 28 languages supported by RTP-LX are Arabic, Hebrew, 
Indonesian, Danish, Norwegian, Swedish, Dutch, English, German, Russian, Ukrainian, Czech, Polish, Serbian, Bosnian, Croatian, Montenegrin, Spanish, Portuguese, French, Italian, Hindi, Thai, Kiswahili, Chinese, Japanese, Korean, Turkish, Finnish, and Hungarian.}
\label{table:bias}
\end{table*}

\subsection{Multilingual Debias}
Language bias of LLMs persists in the multilingual scenario as a consequence of the dominance of English resources and the insufficiency of other languages on the internet. 
To enhance the model ability on low-resource languages, a common practice is to incorporate large-scale data~\cite{gao2020pile,dodge2021documenting} for training. 
The extensive training data facilitates language transfer, especially among typologically similar languages.
Furthermore, the strategies such as curriculum learning~\cite{wei2023polylm} and up-sampling~\cite{lin2022few,shliazhko2022mgpt,labrak2024biomistral} progressively increase the proportion of non-English resource.
These techniques expose LLMs to a wider range of languages while maximizing the utilization of existing data.

To mitigate demographic bias in the multilingual scenario, Zhao~\etal~\cite{zhao2020gender} extend word embedding bias to the cross-lingual and Piqueras~\etal~\cite{piqueras2022pretrained} evaluate group bias of three pre-trained LM (mBERT, XLM-R, and mT5) on four languages (En, Es, De, and Fr). 
Besides, Vashishtha~\etal~\cite{vashishtha2023evaluating} extend debiasing strategies such as counterfactual data augmentation and self-debias to non-English languages, revealing a greater potential for debiasing and generalization among linguistically similar languages.
However, they only investigate a few Indian languages, without comprehensive mitigation strategies for broader language groups.
For a more comprehensive evaluation, De~\etal~\cite{de2024rtp} introduce RTP-LX, a dataset designed for identifying culture-specific toxic languages with much wider coverage (28 languages and 8 different classes).
Experimental results on up-to-date LLMs (Mistral~\cite{jiang2023mistral}, Gemma~\cite{team2024gemma}, GPT-4~\cite{yuan2024gpt4}, etc) demonstrate that even the popular models still struggle to judge history or content-dependent toxic content.
Moreover, Lin~\etal\cite{lin2022few} and Shliazhko~\etal~\cite{shliazhko2022mgpt} analyze safety and bias in multilingual PLMs (XGLM and mGPT), respectively. They observe that multilingual PLMs pronounce gender bias in certain occupations, while few-shot learning has minimal impact on performance improvement.
To analyze gender bias robustly, Yu~\etal~\cite{yu2024your} propose a novel model-based approach to generate sentence pairs.
Based on mBERT~\cite{devlin2018bert}, Reusens~\etal~\cite{reusens2023investigating} investigate cross-lingual transferability of debias techniques on 4 languages and stimulate cross-lingual debiasing effectiveness with additional pre-training.

\subsection{Limitations and Discussion}
To mitigate the cross-lingual bias is a promising research question and it has profound implications for the welfare, fairness, and esteem of numerous social and racial groups. 
Due to the scarcity of high-quality data in low-resource languages and the absence of pertinent evaluation benchmarks, effective bias detection and elimination remain largely unexplored, underscoring the necessity and imperative for future research. 
Besides, machine translation can mitigate the issue of low-resource languages via pseudo data generation~\cite{prabhumoye2018style,hoang2018iterative} but it omits the cultural context or fails to capture the cultural nuances of a specific language~\cite{brislin1970back}.
The local and cultural backgrounds are critical to prejudice and hate speech.
Thus, leveraging raw corpus in the local original context is important for detecting toxic content by native speakers.

\section{Conclusion and Future Directions}
\label{sec:con}

In this paper, we summarize the existing representative research efforts on LLMs in the multilingual scenario from multiple perspectives. %and hope it can facilitate future research. 
We first rethink the transitions between previous and current research on pre-trained language models. Based on the main aspect, the survey is divided into several sections from the view of training paradigms, inference strategies, information retrieval, security, multi-domain, data resources, and benchmark evaluation. 
Besides, we appeal to the research community for bias and fairness when exploring multilingual models.
We also discuss several urgent challenges related to each investigated aspect and provide reflections and potential solutions for future work. 
Finally, considering the rapid growth of LLM research, we establish a continuously updated repository to provide relevant literature with the latest advancements of LLMs in the multilingual scenario.

In conclusion, LLMs have greatly contributed to the advancement of multilingual applications, progressing toward the goal of user-oriented. 
However, the existing technologies and algorithms in various multilingual tasks still fall short of expectations, which makes it difficult to meet practical standards. 
Aiming for language-fair AI, extensive research efforts are required to adapt LLMs for multilingual tasks much more feasible.
We summarize the suggestions for both academic and industry as they build, study, and regulate LLMs as follows:
\begin{itemize}
    \item \textit{Sustainable language adaptation.} 
    The limited data resources of various languages restrict the number of supported languages with the initial model pre-training. An ideal situation is to use the newly available language data to improve the performance and supported languages of LLMs.
    %Due to the limitation of data availability, the practical scenario is that new data is continuously used to improve the performance and supported languages of LLMs. 
    Although mammalian brains could protect previously acquired knowledge through cortical loops to avoid catastrophic forgetting, the neural network models lack this capability.
    Therefore, sustainably achieving language adaptation for LLMs is not trivial.
    The longstanding goal of LLMs with multilingualism is to achieve good performance among multiple languages for all tasks in an incremental learning paradigm. %which is still unexplored.
    % \item Universal multilingual paradigm. Due to the limitation of computing resources, most researchers, particularly in economically underdeveloped regions, often face challenges in training LLMs, especially in non-English-speaking countries. Therefore,
    \item \textit{Universal multilingual paradigm.} The existing studies mainly focus on leveraging parameter-tuning techniques and prompt engineering to explore the potential multilingual capabilities of LLMs. Aiming for a universal multilingual paradigm based on LLMs, it is beneficial to investigate the potential mechanisms without additional training to effectively address language-specific issues, such as code-switching, multilingual jailbreaking, cross-domain adaptation, etc.
    \item \textit{Comprehensive and authoritative evaluation.} 
    The majority popular mainly focus on English and their native languages because of regional and linguistic restrictions, which poses a challenge in bridging the language barrier. 
    %Due to regional and linguistic restrictions, people mainly focus on English and their native languages, which poses a challenge in bridging the language barrier.  
    To mitigate the language barrier issue, an urgent requirement for the multilingual community is to construct a comprehensive and authoritative benchmark to evaluate the multilingual capabilities of LLMs with various aspects, including language culture, multilingual security, multilingual reasoning, domain knowledge in native languages, etc.
    %Therefore, there is an urgent need for collaborative efforts from the multilingual community to construct a comprehensive and authoritative benchmark to evaluate the multilingual capabilities of LLMs from multiple dimensions, including but not limited to language culture, security, reasoning, domain knowledge in native languages and others.    
    This can be achieved by a reasonable combination of multiple benchmarks or guidelines initiated by linguistic experts from corresponding language communities.
    \item \textit{Bias impact with multilingualism.} 
    Existing LLMs inherit biases from the training corpus because of a lack of feasible data management/processing, which poses generation risks.
    Besides, the high proportion of Western languages in the training corpus exacerbates the bias issues from a culturally insensitive generation aspect.
    How to enable LLMs to avoid generating biased/risky content and to possess the ability to generate cultural concepts within different languages are important and meaningful to achieve language-fair technology.
\end{itemize}

\section*{Acknowledgment}
We sincerely thank the valuable feedback from every domain expert, including Junpeng Liu (Dalian University of Technology, China), Jinsong Su (Xiamen University, China), Deyi Xiong (Tianjin University, China), and the anonymous others. In particular, Chaoqun Liu (Nanyang Technological University, Singapore) provides valuable thoughts and contributes part of the implementation of the multilingual inference strategies. %Among the co-authors, Hongliang Li, You Li, Yuanchi Zhang, Weijian Yi, and Yulong Mao are contributors to each section. Jian-Yun Nie and Yang Liu are supervisors.

\bibliography{reference}

\begin{thebibliography}{495}
\providecommand{\natexlab}[1]{#1}
\providecommand{\url}[1]{\texttt{#1}}
\expandafter\ifx\csname urlstyle\endcsname\relax
  \providecommand{\doi}[1]{doi: #1}\else
  \providecommand{\doi}{doi: \begingroup \urlstyle{rm}\Url}\fi

\bibitem[Ouyang et~al.(2022)Ouyang, Wu, Jiang, Almeida, Wainwright, Mishkin, Zhang, Agarwal, Slama, Ray, et~al.]{ouyang2022training}
Long Ouyang, Jeffrey Wu, Xu~Jiang, Diogo Almeida, Carroll Wainwright, Pamela Mishkin, Chong Zhang, Sandhini Agarwal, Katarina Slama, Alex Ray, et~al.
\newblock Training language models to follow instructions with human feedback.
\newblock \emph{Advances in neural information processing systems}, 35:\penalty0 27730--27744, 2022.

\bibitem[Achiam et~al.(2023)Achiam, Adler, Agarwal, Ahmad, Akkaya, Aleman, Almeida, Altenschmidt, Altman, Anadkat, et~al.]{achiam2023gpt}
Josh Achiam, Steven Adler, Sandhini Agarwal, Lama Ahmad, Ilge Akkaya, Florencia~Leoni Aleman, Diogo Almeida, Janko Altenschmidt, Sam Altman, Shyamal Anadkat, et~al.
\newblock Gpt-4 technical report.
\newblock \emph{arXiv preprint arXiv:2303.08774}, 2023.

\bibitem[Touvron et~al.(2023)Touvron, Lavril, Izacard, Martinet, Lachaux, Lacroix, Rozi{\`e}re, Goyal, Hambro, Azhar, et~al.]{touvron2023llama}
Hugo Touvron, Thibaut Lavril, Gautier Izacard, Xavier Martinet, Marie-Anne Lachaux, Timoth{\'e}e Lacroix, Baptiste Rozi{\`e}re, Naman Goyal, Eric Hambro, Faisal Azhar, et~al.
\newblock Llama: Open and efficient foundation language models.
\newblock \emph{arXiv preprint arXiv:2302.13971}, 2023.

\bibitem[Liu et~al.(2023{\natexlab{a}})Liu, Yuan, Fu, Jiang, Hayashi, and Neubig]{liu2023pre}
Pengfei Liu, Weizhe Yuan, Jinlan Fu, Zhengbao Jiang, Hiroaki Hayashi, and Graham Neubig.
\newblock Pre-train, prompt, and predict: A systematic survey of prompting methods in natural language processing.
\newblock \emph{ACM Computing Surveys}, 55\penalty0 (9):\penalty0 1--35, 2023{\natexlab{a}}.

\bibitem[Manakhimova et~al.(2023)Manakhimova, Avramidis, Macketanz, Lapshinova-Koltunski, Bagdasarov, and M{\"o}ller]{manakhimova2023linguistically}
Shushen Manakhimova, Eleftherios Avramidis, Vivien Macketanz, Ekaterina Lapshinova-Koltunski, Sergei Bagdasarov, and Sebastian M{\"o}ller.
\newblock Linguistically motivated evaluation of the 2023 state-of-the-art machine translation: Can chatgpt outperform nmt?
\newblock In \emph{Proceedings of the Eighth Conference on Machine Translation}, pages 224--245, 2023.

\bibitem[Hendy et~al.(2023)Hendy, Abdelrehim, Sharaf, Raunak, Gabr, Matsushita, Kim, Afify, and Awadalla]{hendy2023good}
Amr Hendy, Mohamed Abdelrehim, Amr Sharaf, Vikas Raunak, Mohamed Gabr, Hitokazu Matsushita, Young~Jin Kim, Mohamed Afify, and Hany~Hassan Awadalla.
\newblock How good are gpt models at machine translation? a comprehensive evaluation.
\newblock \emph{arXiv preprint arXiv:2302.09210}, 2023.

\bibitem[Khatri et~al.(2023)Khatri, Srivastava, and Vig]{khatri2023can}
Jyotsana Khatri, Vivek Srivastava, and Lovekesh Vig.
\newblock Can you translate for me? code-switched machine translation with large language models.
\newblock In \emph{Proceedings of the 13th International Joint Conference on Natural Language Processing and the 3rd Conference of the Asia-Pacific Chapter of the Association for Computational Linguistics (Volume 2: Short Papers)}, pages 83--92, 2023.

\bibitem[Luo et~al.(2023{\natexlab{a}})Luo, Xie, and Ananiadou]{luo2023chatgpt}
Zheheng Luo, Qianqian Xie, and Sophia Ananiadou.
\newblock Chatgpt as a factual inconsistency evaluator for abstractive text summarization.
\newblock \emph{arXiv preprint arXiv:2303.15621}, 2023{\natexlab{a}}.

\bibitem[Wang et~al.(2023{\natexlab{a}})Wang, Liang, Meng, Sun, Shi, Li, Xu, Qu, and Zhou]{wang2023chatgpt}
Jiaan Wang, Yunlong Liang, Fandong Meng, Zengkui Sun, Haoxiang Shi, Zhixu Li, Jinan Xu, Jianfeng Qu, and Jie Zhou.
\newblock Is chatgpt a good nlg evaluator? a preliminary study.
\newblock In \emph{Proceedings of EMNLP Workshop}, page~1, 2023{\natexlab{a}}.

\bibitem[Sudirjo et~al.(2023)Sudirjo, Diantoro, Al-Gasawneh, Azzaakiyyah, and Ausat]{sudirjo2023application}
Frans Sudirjo, Karno Diantoro, Jassim~Ahmad Al-Gasawneh, Hizbul~Khootimah Azzaakiyyah, and Abu Muna~Almaududi Ausat.
\newblock Application of chatgpt in improving customer sentiment analysis for businesses.
\newblock \emph{Jurnal Teknologi Dan Sistem Informasi Bisnis}, 5\penalty0 (3):\penalty0 283--288, 2023.

\bibitem[Fatouros et~al.(2023)Fatouros, Soldatos, Kouroumali, Makridis, and Kyriazis]{fatouros2023transforming}
Georgios Fatouros, John Soldatos, Kalliopi Kouroumali, Georgios Makridis, and Dimosthenis Kyriazis.
\newblock Transforming sentiment analysis in the financial domain with chatgpt.
\newblock \emph{Machine Learning with Applications}, 14:\penalty0 100508, 2023.

\bibitem[Fan et~al.(2023)Fan, Kang, Ma, Chen, Wei, Fan, and Yang]{fan2023fate}
Tao Fan, Yan Kang, Guoqiang Ma, Weijing Chen, Wenbin Wei, Lixin Fan, and Qiang Yang.
\newblock Fate-llm: A industrial grade federated learning framework for large language models.
\newblock \emph{arXiv preprint arXiv:2310.10049}, 2023.

\bibitem[Goyal et~al.(2024)Goyal, Rastogi, Rajagopal, Yuan, Zhao, Chintagunta, Naik, and Ward]{goyal2024healai}
Sagar Goyal, Eti Rastogi, Sree~Prasanna Rajagopal, Dong Yuan, Fen Zhao, Jai Chintagunta, Gautam Naik, and Jeff Ward.
\newblock Healai: A healthcare llm for effective medical documentation.
\newblock In \emph{Proceedings of the 17th ACM International Conference on Web Search and Data Mining}, pages 1167--1168, 2024.

\bibitem[Vaswani et~al.(2017)Vaswani, Shazeer, Parmar, Uszkoreit, Jones, Gomez, Kaiser, and Polosukhin]{vaswani2017attention}
Ashish Vaswani, Noam Shazeer, Niki Parmar, Jakob Uszkoreit, Llion Jones, Aidan~N Gomez, {\L}ukasz Kaiser, and Illia Polosukhin.
\newblock Attention is all you need.
\newblock \emph{Advances in neural information processing systems}, 30, 2017.

\bibitem[Lai et~al.(2023)Lai, Ngo, Veyseh, Man, Dernoncourt, Bui, and Nguyen]{lai2023chatgpt}
Viet Lai, Nghia Ngo, Amir Pouran~Ben Veyseh, Hieu Man, Franck Dernoncourt, Trung Bui, and Thien Nguyen.
\newblock Chatgpt beyond english: Towards a comprehensive evaluation of large language models in multilingual learning.
\newblock In \emph{Findings of the Association for Computational Linguistics: EMNLP 2023}, pages 13171--13189, 2023.

\bibitem[Ding et~al.(2024)Ding, Qin, Zhao, Luo, Li, Chen, Xia, Hu, Luu, and Joty]{ding2024data}
Bosheng Ding, Chengwei Qin, Ruochen Zhao, Tianze Luo, Xinze Li, Guizhen Chen, Wenhan Xia, Junjie Hu, Anh~Tuan Luu, and Shafiq Joty.
\newblock Data augmentation using llms: Data perspectives, learning paradigms and challenges.
\newblock \emph{arXiv preprint arXiv:2403.02990}, 2024.

\bibitem[Yang et~al.(2023{\natexlab{a}})Yang, Li, Zhang, and Zong]{yang2023bigtrans}
Wen Yang, Chong Li, Jiajun Zhang, and Chengqing Zong.
\newblock Bigtrans: Augmenting large language models with multilingual translation capability over 100 languages.
\newblock \emph{arXiv preprint arXiv:2305.18098}, 2023{\natexlab{a}}.

\bibitem[Le~Scao et~al.(2022)Le~Scao, Fan, Akiki, Pavlick, Ili{\'c}, Hesslow, Castagn{\'e}, Luccioni, Yvon, Gall{\'e}, et~al.]{le2022bloom}
Teven Le~Scao, Angela Fan, Christopher Akiki, Ellie Pavlick, Suzana Ili{\'c}, Daniel Hesslow, Roman Castagn{\'e}, Alexandra~Sasha Luccioni, Fran{\c{c}}ois Yvon, Matthias Gall{\'e}, et~al.
\newblock Bloom: A 176b-parameter open-access multilingual language model, 2022.

\bibitem[Wang et~al.(2023{\natexlab{b}})Wang, Tu, Chen, Yuan, Huang, Jiao, and Lyu]{wang2023all}
Wenxuan Wang, Zhaopeng Tu, Chang Chen, Youliang Yuan, Jen-tse Huang, Wenxiang Jiao, and Michael~R Lyu.
\newblock All languages matter: On the multilingual safety of large language models.
\newblock \emph{arXiv preprint arXiv:2310.00905}, 2023{\natexlab{b}}.

\bibitem[Zhang et~al.(2023{\natexlab{a}})Zhang, Fang, Zhang, Ma, Zhou, Huang, Bu, Gui, Chen, Chen, et~al.]{zhang2023bayling}
Shaolei Zhang, Qingkai Fang, Zhuocheng Zhang, Zhengrui Ma, Yan Zhou, Langlin Huang, Mengyu Bu, Shangtong Gui, Yunji Chen, Xilin Chen, et~al.
\newblock Bayling: Bridging cross-lingual alignment and instruction following through interactive translation for large language models.
\newblock \emph{arXiv preprint arXiv:2306.10968}, 2023{\natexlab{a}}.

\bibitem[Wang et~al.(2023{\natexlab{c}})Wang, Cheng, Zhan, Li, Song, and Liu]{wang2023openchat}
Guan Wang, Sijie Cheng, Xianyuan Zhan, Xiangang Li, Sen Song, and Yang Liu.
\newblock Openchat: Advancing open-source language models with mixed-quality data.
\newblock In \emph{The Twelfth International Conference on Learning Representations}, 2023{\natexlab{c}}.

\bibitem[Zhang et~al.(2024{\natexlab{a}})Zhang, Wang, Liu, Wang, Wang, Li, Sun, and Liu]{zhang2024enhancing}
Yuanchi Zhang, Yile Wang, Zijun Liu, Shuo Wang, Xiaolong Wang, Peng Li, Maosong Sun, and Yang Liu.
\newblock Enhancing multilingual capabilities of large language models through self-distillation from resource-rich languages.
\newblock \emph{arXiv preprint arXiv:2402.12204}, 2024{\natexlab{a}}.

\bibitem[Chang et~al.(2023)Chang, Arnett, Tu, and Bergen]{chang2023multilinguality}
Tyler~A Chang, Catherine Arnett, Zhuowen Tu, and Benjamin~K Bergen.
\newblock When is multilinguality a curse? language modeling for 250 high-and low-resource languages.
\newblock \emph{arXiv preprint arXiv:2311.09205}, 2023.

\bibitem[Li et~al.(2024{\natexlab{a}})Li, Shi, Liu, Yang, Liu, and Du]{li2024quantifying}
Zihao Li, Yucheng Shi, Zirui Liu, Fan Yang, Ninghao Liu, and Mengnan Du.
\newblock Quantifying multilingual performance of large language models across languages.
\newblock \emph{arXiv preprint arXiv:2404.11553}, 2024{\natexlab{a}}.

\bibitem[Kirkpatrick et~al.(2017)Kirkpatrick, Pascanu, Rabinowitz, Veness, Desjardins, Rusu, Milan, Quan, Ramalho, Grabska-Barwinska, et~al.]{kirkpatrick2017overcoming}
James Kirkpatrick, Razvan Pascanu, Neil Rabinowitz, Joel Veness, Guillaume Desjardins, Andrei~A Rusu, Kieran Milan, John Quan, Tiago Ramalho, Agnieszka Grabska-Barwinska, et~al.
\newblock Overcoming catastrophic forgetting in neural networks.
\newblock \emph{Proceedings of the national academy of sciences}, 114\penalty0 (13):\penalty0 3521--3526, 2017.

\bibitem[Luo et~al.(2022)Luo, Sun, Xia, Qin, Zhang, Poon, and Liu]{luo2022biogpt}
Renqian Luo, Liai Sun, Yingce Xia, Tao Qin, Sheng Zhang, Hoifung Poon, and Tie-Yan Liu.
\newblock Biogpt: generative pre-trained transformer for biomedical text generation and mining.
\newblock \emph{Briefings in bioinformatics}, 23\penalty0 (6):\penalty0 bbac409, 2022.

\bibitem[Liu et~al.(2021)Liu, Huang, Huang, Li, and Zhao]{liu2021finbert}
Zhuang Liu, Degen Huang, Kaiyu Huang, Zhuang Li, and Jun Zhao.
\newblock Finbert: A pre-trained financial language representation model for financial text mining.
\newblock In \emph{Proceedings of the twenty-ninth international conference on international joint conferences on artificial intelligence}, pages 4513--4519, 2021.

\bibitem[Xu et~al.(2024{\natexlab{a}})Xu, Hu, Zhao, Qiu, Ye, and Gu]{xu2024survey}
Yuemei Xu, Ling Hu, Jiayi Zhao, Zihan Qiu, Yuqi Ye, and Hanwen Gu.
\newblock A survey on multilingual large language models: Corpora, alignment, and bias.
\newblock \emph{arXiv preprint arXiv:2404.00929}, 2024{\natexlab{a}}.

\bibitem[Qin et~al.(2024)Qin, Chen, Zhou, Chen, Li, Liao, Li, Che, and Yu]{qin2024multilingual}
Libo Qin, Qiguang Chen, Yuhang Zhou, Zhi Chen, Yinghui Li, Lizi Liao, Min Li, Wanxiang Che, and Philip~S Yu.
\newblock Multilingual large language model: A survey of resources, taxonomy and frontiers.
\newblock \emph{arXiv preprint arXiv:2404.04925}, 2024.

\bibitem[Brown et~al.(2020)Brown, Mann, Ryder, Subbiah, Kaplan, Dhariwal, Neelakantan, Shyam, Sastry, Askell, et~al.]{brown2020language}
Tom Brown, Benjamin Mann, Nick Ryder, Melanie Subbiah, Jared~D Kaplan, Prafulla Dhariwal, Arvind Neelakantan, Pranav Shyam, Girish Sastry, Amanda Askell, et~al.
\newblock Language models are few-shot learners.
\newblock \emph{Advances in neural information processing systems}, 33:\penalty0 1877--1901, 2020.

\bibitem[Xue et~al.(2020)Xue, Constant, Roberts, Kale, Al-Rfou, Siddhant, Barua, and Raffel]{xue2020mt5}
Linting Xue, Noah Constant, Adam Roberts, Mihir Kale, Rami Al-Rfou, Aditya Siddhant, Aditya Barua, and Colin Raffel.
\newblock mt5: A massively multilingual pre-trained text-to-text transformer.
\newblock \emph{arXiv preprint arXiv:2010.11934}, 2020.

\bibitem[Xue et~al.(2022)Xue, Barua, Constant, Al-Rfou, Narang, Kale, Roberts, and Raffel]{xue2022byt5}
Linting Xue, Aditya Barua, Noah Constant, Rami Al-Rfou, Sharan Narang, Mihir Kale, Adam Roberts, and Colin Raffel.
\newblock Byt5: Towards a token-free future with pre-trained byte-to-byte models.
\newblock \emph{Transactions of the Association for Computational Linguistics}, 10:\penalty0 291--306, 2022.

\bibitem[Rae et~al.(2021)Rae, Borgeaud, Cai, Millican, Hoffmann, Song, Aslanides, Henderson, Ring, Young, et~al.]{rae2021scaling}
Jack~W Rae, Sebastian Borgeaud, Trevor Cai, Katie Millican, Jordan Hoffmann, Francis Song, John Aslanides, Sarah Henderson, Roman Ring, Susannah Young, et~al.
\newblock Scaling language models: Methods, analysis \& insights from training gopher.
\newblock \emph{arXiv preprint arXiv:2112.11446}, 2021.

\bibitem[Thoppilan et~al.(2022)Thoppilan, De~Freitas, Hall, Shazeer, Kulshreshtha, Cheng, Jin, Bos, Baker, Du, et~al.]{thoppilan2022lamda}
Romal Thoppilan, Daniel De~Freitas, Jamie Hall, Noam Shazeer, Apoorv Kulshreshtha, Heng-Tze Cheng, Alicia Jin, Taylor Bos, Leslie Baker, Yu~Du, et~al.
\newblock Lamda: Language models for dialog applications.
\newblock \emph{arXiv preprint arXiv:2201.08239}, 2022.

\bibitem[Zhang et~al.(2022{\natexlab{a}})Zhang, Roller, Goyal, Artetxe, Chen, Chen, Dewan, Diab, Li, Lin, et~al.]{zhang2022opt}
Susan Zhang, Stephen Roller, Naman Goyal, Mikel Artetxe, Moya Chen, Shuohui Chen, Christopher Dewan, Mona Diab, Xian Li, Xi~Victoria Lin, et~al.
\newblock Opt: Open pre-trained transformer language models.
\newblock \emph{arXiv preprint arXiv:2205.01068}, 2022{\natexlab{a}}.

\bibitem[Chowdhery et~al.(2023)Chowdhery, Narang, Devlin, Bosma, Mishra, Roberts, Barham, Chung, Sutton, Gehrmann, et~al.]{chowdhery2023palm}
Aakanksha Chowdhery, Sharan Narang, Jacob Devlin, Maarten Bosma, Gaurav Mishra, Adam Roberts, Paul Barham, Hyung~Won Chung, Charles Sutton, Sebastian Gehrmann, et~al.
\newblock Palm: Scaling language modeling with pathways.
\newblock \emph{Journal of Machine Learning Research}, 24\penalty0 (240):\penalty0 1--113, 2023.

\bibitem[Shliazhko et~al.(2022)Shliazhko, Fenogenova, Tikhonova, Mikhailov, Kozlova, and Shavrina]{shliazhko2022mgpt}
Oleh Shliazhko, Alena Fenogenova, Maria Tikhonova, Vladislav Mikhailov, Anastasia Kozlova, and Tatiana Shavrina.
\newblock mgpt: Few-shot learners go multilingual.
\newblock \emph{arXiv preprint arXiv:2204.07580}, 2022.

\bibitem[Lin et~al.(2022{\natexlab{a}})Lin, Mihaylov, Artetxe, Wang, Chen, Simig, Ott, Goyal, Bhosale, Du, et~al.]{lin2022few}
Xi~Victoria Lin, Todor Mihaylov, Mikel Artetxe, Tianlu Wang, Shuohui Chen, Daniel Simig, Myle Ott, Naman Goyal, Shruti Bhosale, Jingfei Du, et~al.
\newblock Few-shot learning with multilingual generative language models.
\newblock In \emph{Proceedings of the 2022 Conference on Empirical Methods in Natural Language Processing}, pages 9019--9052, 2022{\natexlab{a}}.

\bibitem[Ren et~al.(2023)Ren, Zhou, Meng, Huang, Wang, Wang, Li, Zhang, Podolskiy, Arshinov, et~al.]{ren2023pangu}
Xiaozhe Ren, Pingyi Zhou, Xinfan Meng, Xinjing Huang, Yadao Wang, Weichao Wang, Pengfei Li, Xiaoda Zhang, Alexander Podolskiy, Grigory Arshinov, et~al.
\newblock Pangu: Towards trillion parameter language model with sparse heterogeneous computing.
\newblock \emph{arXiv preprint arXiv:2303.10845}, 10:\penalty0 11--15, 2023.

\bibitem[Biderman et~al.(2023)Biderman, Schoelkopf, Anthony, Bradley, O’Brien, Hallahan, Khan, Purohit, Prashanth, Raff, et~al.]{biderman2023pythia}
Stella Biderman, Hailey Schoelkopf, Quentin~Gregory Anthony, Herbie Bradley, Kyle O’Brien, Eric Hallahan, Mohammad~Aflah Khan, Shivanshu Purohit, USVSN~Sai Prashanth, Edward Raff, et~al.
\newblock Pythia: A suite for analyzing large language models across training and scaling.
\newblock In \emph{International Conference on Machine Learning}, pages 2397--2430. PMLR, 2023.

\bibitem[Anil et~al.(2023)Anil, Dai, Firat, Johnson, Lepikhin, Passos, Shakeri, Taropa, Bailey, Chen, et~al.]{anil2023palm}
Rohan Anil, Andrew~M Dai, Orhan Firat, Melvin Johnson, Dmitry Lepikhin, Alexandre Passos, Siamak Shakeri, Emanuel Taropa, Paige Bailey, Zhifeng Chen, et~al.
\newblock Palm 2 technical report.
\newblock \emph{arXiv preprint arXiv:2305.10403}, 2023.

\bibitem[Team(2023)]{team2023internlm}
InternLM Team.
\newblock Internlm: A multilingual language model with progressively enhanced capabilities, 2023.

\bibitem[Wei et~al.(2023{\natexlab{a}})Wei, Wei, Lin, Li, Zhang, Ren, Li, Wan, Cao, Xie, et~al.]{wei2023polylm}
Xiangpeng Wei, Haoran Wei, Huan Lin, Tianhao Li, Pei Zhang, Xingzhang Ren, Mei Li, Yu~Wan, Zhiwei Cao, Binbin Xie, et~al.
\newblock Polylm: An open source polyglot large language model.
\newblock \emph{arXiv preprint arXiv:2307.06018}, 2023{\natexlab{a}}.

\bibitem[Yang et~al.(2023{\natexlab{b}})Yang, Xiao, Wang, Zhang, Bian, Yin, Lv, Pan, Wang, Yan, et~al.]{yang2023baichuan}
Aiyuan Yang, Bin Xiao, Bingning Wang, Borong Zhang, Ce~Bian, Chao Yin, Chenxu Lv, Da~Pan, Dian Wang, Dong Yan, et~al.
\newblock Baichuan 2: Open large-scale language models.
\newblock \emph{arXiv preprint arXiv:2309.10305}, 2023{\natexlab{b}}.

\bibitem[Bai et~al.(2023)Bai, Bai, Chu, Cui, Dang, Deng, Fan, Ge, Han, Huang, et~al.]{bai2023qwen}
Jinze Bai, Shuai Bai, Yunfei Chu, Zeyu Cui, Kai Dang, Xiaodong Deng, Yang Fan, Wenbin Ge, Yu~Han, Fei Huang, et~al.
\newblock Qwen technical report.
\newblock \emph{arXiv preprint arXiv:2309.16609}, 2023.

\bibitem[Jiang et~al.(2023)Jiang, Sablayrolles, Mensch, Bamford, Chaplot, Casas, Bressand, Lengyel, Lample, Saulnier, et~al.]{jiang2023mistral}
Albert~Q Jiang, Alexandre Sablayrolles, Arthur Mensch, Chris Bamford, Devendra~Singh Chaplot, Diego de~las Casas, Florian Bressand, Gianna Lengyel, Guillaume Lample, Lucile Saulnier, et~al.
\newblock Mistral 7b.
\newblock \emph{arXiv preprint arXiv:2310.06825}, 2023.

\bibitem[Team et~al.(2023)Team, Anil, Borgeaud, Wu, Alayrac, Yu, Soricut, Schalkwyk, Dai, Hauth, et~al.]{team2023gemini}
Gemini Team, Rohan Anil, Sebastian Borgeaud, Yonghui Wu, Jean-Baptiste Alayrac, Jiahui Yu, Radu Soricut, Johan Schalkwyk, Andrew~M Dai, Anja Hauth, et~al.
\newblock Gemini: a family of highly capable multimodal models.
\newblock \emph{arXiv preprint arXiv:2312.11805}, 2023.

\bibitem[Chen et~al.(2023{\natexlab{a}})Chen, Cai, Wu, Li, Xin, and Fu]{chen2023tigerbot}
Ye~Chen, Wei Cai, Liangmin Wu, Xiaowei Li, Zhanxuan Xin, and Cong Fu.
\newblock Tigerbot: An open multilingual multitask llm.
\newblock \emph{arXiv preprint arXiv:2312.08688}, 2023{\natexlab{a}}.

\bibitem[Luo et~al.(2023{\natexlab{b}})Luo, Kong, Xu, Cao, Hao, Qu, Chen, Zhu, Zhao, Zhang, et~al.]{luo2023yayi}
Yin Luo, Qingchao Kong, Nan Xu, Jia Cao, Bao Hao, Baoyu Qu, Bo~Chen, Chao Zhu, Chenyang Zhao, Donglei Zhang, et~al.
\newblock Yayi 2: Multilingual open-source large language models.
\newblock \emph{arXiv preprint arXiv:2312.14862}, 2023{\natexlab{b}}.

\bibitem[Bi et~al.(2024)Bi, Chen, Chen, Chen, Dai, Deng, Ding, Dong, Du, Fu, et~al.]{bi2024deepseek}
Xiao Bi, Deli Chen, Guanting Chen, Shanhuang Chen, Damai Dai, Chengqi Deng, Honghui Ding, Kai Dong, Qiushi Du, Zhe Fu, et~al.
\newblock Deepseek llm: Scaling open-source language models with longtermism.
\newblock \emph{arXiv preprint arXiv:2401.02954}, 2024.

\bibitem[Chen et~al.(2024{\natexlab{a}})Chen, Huang, Li, Li, Liu, Pan, Xu, Zhang, Zhang, and Han]{chen2024orion}
Du~Chen, Yi~Huang, Xiaopu Li, Yongqiang Li, Yongqiang Liu, Haihui Pan, Leichao Xu, Dacheng Zhang, Zhipeng Zhang, and Kun Han.
\newblock Orion-14b: Open-source multilingual large language models.
\newblock \emph{arXiv preprint arXiv:2401.12246}, 2024{\natexlab{a}}.

\bibitem[Abdin et~al.(2024)Abdin, Jacobs, Awan, Aneja, Awadallah, Awadalla, Bach, Bahree, Bakhtiari, Behl, et~al.]{abdin2024phi}
Marah Abdin, Sam~Ade Jacobs, Ammar~Ahmad Awan, Jyoti Aneja, Ahmed Awadallah, Hany Awadalla, Nguyen Bach, Amit Bahree, Arash Bakhtiari, Harkirat Behl, et~al.
\newblock Phi-3 technical report: A highly capable language model locally on your phone.
\newblock \emph{arXiv preprint arXiv:2404.14219}, 2024.

\bibitem[Anthropic(2024)]{anthropic2024claude}
AI~Anthropic.
\newblock The claude 3 model family: Opus, sonnet, haiku.
\newblock \emph{Claude-3 Model Card}, 2024.

\bibitem[Cai et~al.(2024)Cai, Cao, Chen, Chen, Chen, Chen, Chen, Chen, Chen, Chu, et~al.]{cai2024internlm2}
Zheng Cai, Maosong Cao, Haojiong Chen, Kai Chen, Keyu Chen, Xin Chen, Xun Chen, Zehui Chen, Zhi Chen, Pei Chu, et~al.
\newblock Internlm2 technical report.
\newblock \emph{arXiv preprint arXiv:2403.17297}, 2024.

\bibitem[AI@Meta(2024)]{llama3modelcard}
AI@Meta.
\newblock Llama 3 model card, 2024.
\newblock URL \url{https://github.com/meta-llama/llama3/blob/main/MODEL_CARD.md}.

\bibitem[Chung et~al.(2024)Chung, Hou, Longpre, Zoph, Tay, Fedus, Li, Wang, Dehghani, Brahma, et~al.]{chung2024scaling}
Hyung~Won Chung, Le~Hou, Shayne Longpre, Barret Zoph, Yi~Tay, William Fedus, Yunxuan Li, Xuezhi Wang, Mostafa Dehghani, Siddhartha Brahma, et~al.
\newblock Scaling instruction-finetuned language models.
\newblock \emph{Journal of Machine Learning Research}, 25\penalty0 (70):\penalty0 1--53, 2024.

\bibitem[Zeng et~al.(2022)Zeng, Liu, Du, Wang, Lai, Ding, Yang, Xu, Zheng, Xia, et~al.]{zeng2022glm}
Aohan Zeng, Xiao Liu, Zhengxiao Du, Zihan Wang, Hanyu Lai, Ming Ding, Zhuoyi Yang, Yifan Xu, Wendi Zheng, Xiao Xia, et~al.
\newblock Glm-130b: An open bilingual pre-trained model.
\newblock \emph{arXiv preprint arXiv:2210.02414}, 2022.

\bibitem[Taori et~al.(2023)Taori, Gulrajani, Zhang, Dubois, Li, Guestrin, Liang, and Hashimoto]{taori2023stanford}
Rohan Taori, Ishaan Gulrajani, Tianyi Zhang, Yann Dubois, Xuechen Li, Carlos Guestrin, Percy Liang, and Tatsunori~B Hashimoto.
\newblock Stanford alpaca: An instruction-following llama model, 2023.

\bibitem[Jiao et~al.(2023)Jiao, Huang, Wang, He, Liang, Wang, Shi, and Tu]{jiao2023parrot}
Wenxiang Jiao, Jen-tse Huang, Wenxuan Wang, Zhiwei He, Tian Liang, Xing Wang, Shuming Shi, and Zhaopeng Tu.
\newblock Parrot: Translating during chat using large language models tuned with human translation and feedback.
\newblock In \emph{Findings of the Association for Computational Linguistics: EMNLP 2023}, pages 15009--15020, 2023.

\bibitem[Chiang et~al.(2023)Chiang, Li, Lin, Sheng, Wu, Zhang, Zheng, Zhuang, Zhuang, Gonzalez, et~al.]{chiang2023vicuna}
Wei-Lin Chiang, Zhuohan Li, Zi~Lin, Ying Sheng, Zhanghao Wu, Hao Zhang, Lianmin Zheng, Siyuan Zhuang, Yonghao Zhuang, Joseph~E Gonzalez, et~al.
\newblock Vicuna: An open-source chatbot impressing gpt-4 with 90\%* chatgpt quality.
\newblock \emph{See https://vicuna. lmsys. org (accessed 14 April 2023)}, 2\penalty0 (3):\penalty0 6, 2023.

\bibitem[ZHIPU(2024)]{glm4}
ZHIPU.
\newblock Zhipu ai devday glm-4, 2024.

\bibitem[{\"U}st{\"u}n et~al.(2024){\"U}st{\"u}n, Aryabumi, Yong, Ko, D{'}souza, Onilude, Bhandari, Singh, Ooi, Kayid, Vargus, Blunsom, Longpre, Muennighoff, Fadaee, Kreutzer, and Hooker]{ustun-etal-2024-aya}
Ahmet {\"U}st{\"u}n, Viraat Aryabumi, Zheng Yong, Wei-Yin Ko, Daniel D{'}souza, Gbemileke Onilude, Neel Bhandari, Shivalika Singh, Hui-Lee Ooi, Amr Kayid, Freddie Vargus, Phil Blunsom, Shayne Longpre, Niklas Muennighoff, Marzieh Fadaee, Julia Kreutzer, and Sara Hooker.
\newblock Aya model: An instruction finetuned open-access multilingual language model.
\newblock In Lun-Wei Ku, Andre Martins, and Vivek Srikumar, editors, \emph{Proceedings of the 62nd Annual Meeting of the Association for Computational Linguistics (Volume 1: Long Papers)}, pages 15894--15939, Bangkok, Thailand, August 2024. Association for Computational Linguistics.
\newblock \doi{10.18653/v1/2024.acl-long.845}.
\newblock URL \url{https://aclanthology.org/2024.acl-long.845}.

\bibitem[Intrator et~al.(2024)Intrator, Halfon, Goldenberg, Tsarfaty, Eyal, Rivlin, Matias, and Aizenberg]{intrator2024breaking}
Yotam Intrator, Matan Halfon, Roman Goldenberg, Reut Tsarfaty, Matan Eyal, Ehud Rivlin, Yossi Matias, and Natalia Aizenberg.
\newblock Breaking the language barrier: Can direct inference outperform pre-translation in multilingual llm applications?
\newblock \emph{arXiv preprint arXiv:2403.04792}, 2024.

\bibitem[Liu et~al.(2024{\natexlab{a}})Liu, Zhang, Zhao, Luu, and Bing]{liu2024translation}
Chaoqun Liu, Wenxuan Zhang, Yiran Zhao, Anh~Tuan Luu, and Lidong Bing.
\newblock Is translation all you need? a study on solving multilingual tasks with large language models.
\newblock \emph{arXiv preprint arXiv:2403.10258}, 2024{\natexlab{a}}.

\bibitem[Huang et~al.(2023{\natexlab{a}})Huang, Tang, Zhang, Zhao, Song, Xia, and Wei]{huang2023not}
Haoyang Huang, Tianyi Tang, Dongdong Zhang, Wayne~Xin Zhao, Ting Song, Yan Xia, and Furu Wei.
\newblock Not all languages are created equal in llms: Improving multilingual capability by cross-lingual-thought prompting.
\newblock In \emph{Findings of the Association for Computational Linguistics: EMNLP 2023}, pages 12365--12394, 2023{\natexlab{a}}.

\bibitem[Shi et~al.(2022{\natexlab{a}})Shi, Suzgun, Freitag, Wang, Srivats, Vosoughi, Chung, Tay, Ruder, Zhou, et~al.]{shi2022language}
Freda Shi, Mirac Suzgun, Markus Freitag, Xuezhi Wang, Suraj Srivats, Soroush Vosoughi, Hyung~Won Chung, Yi~Tay, Sebastian Ruder, Denny Zhou, et~al.
\newblock Language models are multilingual chain-of-thought reasoners.
\newblock \emph{arXiv preprint arXiv:2210.03057}, 2022{\natexlab{a}}.

\bibitem[Kim et~al.(2023{\natexlab{a}})Kim, Joo, Kim, Jang, Ye, Shin, and Seo]{kim2023cot}
Seungone Kim, Se~Joo, Doyoung Kim, Joel Jang, Seonghyeon Ye, Jamin Shin, and Minjoon Seo.
\newblock The cot collection: Improving zero-shot and few-shot learning of language models via chain-of-thought fine-tuning.
\newblock In \emph{Proceedings of the 2023 Conference on Empirical Methods in Natural Language Processing}, pages 12685--12708, 2023{\natexlab{a}}.

\bibitem[Suzgun et~al.(2023)Suzgun, Scales, Sch{\"a}rli, Gehrmann, Tay, Chung, Chowdhery, Le, Chi, Zhou, et~al.]{suzgun2023challenging}
Mirac Suzgun, Nathan Scales, Nathanael Sch{\"a}rli, Sebastian Gehrmann, Yi~Tay, Hyung~Won Chung, Aakanksha Chowdhery, Quoc Le, Ed~Chi, Denny Zhou, et~al.
\newblock Challenging big-bench tasks and whether chain-of-thought can solve them.
\newblock In \emph{Findings of the Association for Computational Linguistics: ACL 2023}, pages 13003--13051, 2023.

\bibitem[Chai et~al.(2024)Chai, Yang, Sun, Guo, Liu, Wang, Liang, Bai, Li, Peng, et~al.]{chai2024xcot}
Linzheng Chai, Jian Yang, Tao Sun, Hongcheng Guo, Jiaheng Liu, Bing Wang, Xiannian Liang, Jiaqi Bai, Tongliang Li, Qiyao Peng, et~al.
\newblock xcot: Cross-lingual instruction tuning for cross-lingual chain-of-thought reasoning.
\newblock \emph{arXiv preprint arXiv:2401.07037}, 2024.

\bibitem[Zhang et~al.(2023{\natexlab{b}})Zhang, Cahyawijaya, Cruz, Winata, and Aji]{zhang2023multilingual}
Ruochen Zhang, Samuel Cahyawijaya, Jan Christian~Blaise Cruz, Genta Winata, and Alham Aji.
\newblock Multilingual large language models are not (yet) code-switchers.
\newblock In \emph{Proceedings of the 2023 Conference on Empirical Methods in Natural Language Processing}, pages 12567--12582, 2023{\natexlab{b}}.

\bibitem[Koto et~al.(2024)Koto, Beck, Talat, Gurevych, and Baldwin]{koto2024zero}
Fajri Koto, Tilman Beck, Zeerak Talat, Iryna Gurevych, and Timothy Baldwin.
\newblock Zero-shot sentiment analysis in low-resource languages using a multilingual sentiment lexicon.
\newblock \emph{arXiv preprint arXiv:2402.02113}, 2024.

\bibitem[Peng et~al.(2023{\natexlab{a}})Peng, Yan, Watanabe, and Harwath]{peng2023prompting}
Puyuan Peng, Brian Yan, Shinji Watanabe, and David Harwath.
\newblock Prompting the hidden talent of web-scale speech models for zero-shot task generalization.
\newblock In \emph{Proceedings of the Annual Conference of the International Speech Communication Association, INTERSPEECH}, volume 2023, pages 396--400, 2023{\natexlab{a}}.

\bibitem[Zhang et~al.(2023{\natexlab{c}})Zhang, Liu, Yanqing, Qiao, Chang, Zhao, Zhu, Zhu, Peng, Li, Liu, Ma, Piao, Tao, Yang, and Jiang]{zhang-etal-2023-leveraging}
Min Zhang, Limin Liu, Zhao Yanqing, Xiaosong Qiao, Su~Chang, Xiaofeng Zhao, Junhao Zhu, Ming Zhu, Song Peng, Yinglu Li, Yilun Liu, Wenbing Ma, Mengyao Piao, Shimin Tao, Hao Yang, and Yanfei Jiang.
\newblock Leveraging multilingual knowledge graph to boost domain-specific entity translation of {C}hat{GPT}.
\newblock In Masaru Yamada and Felix do~Carmo, editors, \emph{Proceedings of Machine Translation Summit XIX, Vol. 2: Users Track}, pages 77--87, Macau SAR, China, September 2023{\natexlab{c}}. Asia-Pacific Association for Machine Translation.
\newblock URL \url{https://aclanthology.org/2023.mtsummit-users.7}.

\bibitem[Shi et~al.(2022{\natexlab{b}})Shi, Zhang, Bai, and Lin]{shi-etal-2022-xricl}
Peng Shi, Rui Zhang, He~Bai, and Jimmy Lin.
\newblock {XRICL}: Cross-lingual retrieval-augmented in-context learning for cross-lingual text-to-{SQL} semantic parsing.
\newblock In Yoav Goldberg, Zornitsa Kozareva, and Yue Zhang, editors, \emph{Findings of the Association for Computational Linguistics: EMNLP 2022}, pages 5248--5259, Abu Dhabi, United Arab Emirates, December 2022{\natexlab{b}}. Association for Computational Linguistics.
\newblock \doi{10.18653/v1/2022.findings-emnlp.384}.
\newblock URL \url{https://aclanthology.org/2022.findings-emnlp.384}.

\bibitem[Agrawal et~al.(2023)Agrawal, Zhou, Lewis, Zettlemoyer, and Ghazvininejad]{agrawal-etal-2023-context}
Sweta Agrawal, Chunting Zhou, Mike Lewis, Luke Zettlemoyer, and Marjan Ghazvininejad.
\newblock In-context examples selection for machine translation.
\newblock In Anna Rogers, Jordan Boyd-Graber, and Naoaki Okazaki, editors, \emph{Findings of the Association for Computational Linguistics: ACL 2023}, pages 8857--8873, Toronto, Canada, July 2023. Association for Computational Linguistics.
\newblock \doi{10.18653/v1/2023.findings-acl.564}.
\newblock URL \url{https://aclanthology.org/2023.findings-acl.564}.

\bibitem[Li et~al.(2023{\natexlab{a}})Li, Nie, and Liang]{li2023classification}
Xiaoqian Li, Ercong Nie, and Sheng Liang.
\newblock From classification to generation: Insights into crosslingual retrieval augmented icl, 2023{\natexlab{a}}.

\bibitem[Li et~al.(2023{\natexlab{b}})Li, Nie, and Liang]{li-etal-2023-crosslingual}
Xiaoqian Li, Ercong Nie, and Sheng Liang.
\newblock Crosslingual retrieval augmented in-context learning for {B}angla.
\newblock In Firoj Alam, Sudipta Kar, Shammur~Absar Chowdhury, Farig Sadeque, and Ruhul Amin, editors, \emph{Proceedings of the First Workshop on Bangla Language Processing (BLP-2023)}, pages 136--151, Singapore, December 2023{\natexlab{b}}. Association for Computational Linguistics.
\newblock \doi{10.18653/v1/2023.banglalp-1.15}.
\newblock URL \url{https://aclanthology.org/2023.banglalp-1.15}.

\bibitem[Winata et~al.(2023{\natexlab{a}})Winata, Huang, Vadlamannati, and Chandarana]{winata2023multilingual}
Genta~Indra Winata, Liang-Kang Huang, Soumya Vadlamannati, and Yash Chandarana.
\newblock Multilingual few-shot learning via language model retrieval, 2023{\natexlab{a}}.

\bibitem[Garcia et~al.(2023)Garcia, Bansal, Cherry, Foster, Krikun, Feng, Johnson, and Firat]{garcia2023unreasonable}
Xavier Garcia, Yamini Bansal, Colin Cherry, George Foster, Maxim Krikun, Fangxiaoyu Feng, Melvin Johnson, and Orhan Firat.
\newblock The unreasonable effectiveness of few-shot learning for machine translation, 2023.

\bibitem[Ramos et~al.(2023)Ramos, Martins, and Elliott]{ramos-etal-2023-lmcap}
Rita Ramos, Bruno Martins, and Desmond Elliott.
\newblock {LMC}ap: Few-shot multilingual image captioning by retrieval augmented language model prompting.
\newblock In Anna Rogers, Jordan Boyd-Graber, and Naoaki Okazaki, editors, \emph{Findings of the Association for Computational Linguistics: ACL 2023}, pages 1635--1651, Toronto, Canada, July 2023. Association for Computational Linguistics.
\newblock \doi{10.18653/v1/2023.findings-acl.104}.
\newblock URL \url{https://aclanthology.org/2023.findings-acl.104}.

\bibitem[Kim et~al.(2023{\natexlab{b}})Kim, Ki, Kim, and Lee]{kim2023boosting}
Sunkyoung Kim, Dayeon Ki, Yireun Kim, and Jinsik Lee.
\newblock Boosting cross-lingual transferability in multilingual models via in-context learning, 2023{\natexlab{b}}.

\bibitem[Thakur et~al.(2024{\natexlab{a}})Thakur, Bonifacio, Zhang, Ogundepo, Kamalloo, Alfonso-Hermelo, Li, Liu, Chen, Rezagholizadeh, and Lin]{thakur2024nomiracl}
Nandan Thakur, Luiz Bonifacio, Xinyu Zhang, Odunayo Ogundepo, Ehsan Kamalloo, David Alfonso-Hermelo, Xiaoguang Li, Qun Liu, Boxing Chen, Mehdi Rezagholizadeh, and Jimmy Lin.
\newblock Nomiracl: Knowing when you don't know for robust multilingual retrieval-augmented generation, 2024{\natexlab{a}}.

\bibitem[Sennrich et~al.(2023)Sennrich, Vamvas, and Mohammadshahi]{sennrich2023mitigating}
Rico Sennrich, Jannis Vamvas, and Alireza Mohammadshahi.
\newblock Mitigating hallucinations and off-target machine translation with source-contrastive and language-contrastive decoding.
\newblock \emph{arXiv preprint arXiv:2309.07098}, 2023.

\bibitem[Vernikos and Popescu-Belis(2024)]{vernikos2024don}
Giorgos Vernikos and Andrei Popescu-Belis.
\newblock Don't rank, combine! combining machine translation hypotheses using quality estimation.
\newblock \emph{arXiv preprint arXiv:2401.06688}, 2024.

\bibitem[Fu et~al.(2024)Fu, Feng, Huang, Huo, Li, Wang, Qin, and Liu]{fu2024relay}
Chengpeng Fu, Xiaocheng Feng, Yichong Huang, Wenshuai Huo, Baohang Li, Hui Wang, Bin Qin, and Ting Liu.
\newblock Relay decoding: Concatenating large language models for machine translation.
\newblock \emph{arXiv preprint arXiv:2405.02933}, 2024.

\bibitem[Zeng et~al.(2024)Zeng, Meng, Yin, and Zhou]{zeng2024teaching}
Jiali Zeng, Fandong Meng, Yongjing Yin, and Jie Zhou.
\newblock Teaching large language models to translate with comparison.
\newblock In \emph{Proceedings of the AAAI Conference on Artificial Intelligence}, volume~38, pages 19488--19496, 2024.

\bibitem[He et~al.(2024)He, Liang, Jiao, Zhang, Yang, Wang, Tu, Shi, and Wang]{he2024exploring}
Zhiwei He, Tian Liang, Wenxiang Jiao, Zhuosheng Zhang, Yujiu Yang, Rui Wang, Zhaopeng Tu, Shuming Shi, and Xing Wang.
\newblock Exploring human-like translation strategy with large language models.
\newblock \emph{Transactions of the Association for Computational Linguistics}, 12:\penalty0 229--246, 2024.

\bibitem[Conia et~al.(2023)Conia, Li, Lee, Minhas, Ilyas, and Li]{conia-etal-2023-increasing}
Simone Conia, Min Li, Daniel Lee, Umar Minhas, Ihab Ilyas, and Yunyao Li.
\newblock Increasing coverage and precision of textual information in multilingual knowledge graphs.
\newblock In Houda Bouamor, Juan Pino, and Kalika Bali, editors, \emph{Proceedings of the 2023 Conference on Empirical Methods in Natural Language Processing}, pages 1612--1634, Singapore, December 2023. Association for Computational Linguistics.
\newblock \doi{10.18653/v1/2023.emnlp-main.100}.
\newblock URL \url{https://aclanthology.org/2023.emnlp-main.100}.

\bibitem[Bonifacio et~al.(2022{\natexlab{a}})Bonifacio, Abonizio, Fadaee, and Nogueira]{bonifacio2022inpars}
Luiz Bonifacio, Hugo Abonizio, Marzieh Fadaee, and Rodrigo Nogueira.
\newblock Inpars: Unsupervised dataset generation for information retrieval.
\newblock In \emph{Proceedings of the 45th International ACM SIGIR Conference on Research and Development in Information Retrieval}, SIGIR '22, page 2387–2392, New York, NY, USA, 2022{\natexlab{a}}. Association for Computing Machinery.
\newblock ISBN 9781450387323.
\newblock \doi{10.1145/3477495.3531863}.
\newblock URL \url{https://doi.org/10.1145/3477495.3531863}.

\bibitem[Jeronymo et~al.(2023{\natexlab{a}})Jeronymo, Bonifacio, Abonizio, Fadaee, Lotufo, Zavrel, and Nogueira]{inparsv2}
Vitor Jeronymo, Luiz Bonifacio, Hugo Abonizio, Marzieh Fadaee, Roberto Lotufo, Jakub Zavrel, and Rodrigo Nogueira.
\newblock {InPars-v2}: Large language models as efficient dataset generators for information retrieval, 2023{\natexlab{a}}.
\newblock URL \url{https://arxiv.org/abs/2301.01820}.

\bibitem[Abonizio et~al.(2023)Abonizio, Bonifacio, Jeronymo, Lotufo, Zavrel, and Nogueira]{abonizio2023inpars}
Hugo Abonizio, Luiz Bonifacio, Vitor Jeronymo, Roberto Lotufo, Jakub Zavrel, and Rodrigo Nogueira.
\newblock Inpars toolkit: A unified and reproducible synthetic data generation pipeline for neural information retrieval, 2023.

\bibitem[Dai et~al.(2023)Dai, Zhao, Ma, Luan, Ni, Lu, Bakalov, Guu, Hall, and Chang]{dai2023promptagator}
Zhuyun Dai, Vincent~Y Zhao, Ji~Ma, Yi~Luan, Jianmo Ni, Jing Lu, Anton Bakalov, Kelvin Guu, Keith Hall, and Ming-Wei Chang.
\newblock Promptagator: Few-shot dense retrieval from 8 examples.
\newblock In \emph{The Eleventh International Conference on Learning Representations}, 2023.
\newblock URL \url{https://openreview.net/forum?id=gmL46YMpu2J}.

\bibitem[Thakur et~al.(2024{\natexlab{b}})Thakur, Ni, Hernandez~Abrego, Wieting, Lin, and Cer]{thakur-etal-2024-leveraging}
Nandan Thakur, Jianmo Ni, Gustavo Hernandez~Abrego, John Wieting, Jimmy Lin, and Daniel Cer.
\newblock Leveraging {LLM}s for synthesizing training data across many languages in multilingual dense retrieval.
\newblock In Kevin Duh, Helena Gomez, and Steven Bethard, editors, \emph{Proceedings of the 2024 Conference of the North American Chapter of the Association for Computational Linguistics: Human Language Technologies (Volume 1: Long Papers)}, pages 7699--7724, Mexico City, Mexico, June 2024{\natexlab{b}}. Association for Computational Linguistics.
\newblock \doi{10.18653/v1/2024.naacl-long.426}.
\newblock URL \url{https://aclanthology.org/2024.naacl-long.426}.

\bibitem[Mayfield et~al.(2023)Mayfield, Yang, Lawrie, Barham, Weller, Mason, Nair, and Miller]{mayfield2023syntheticcrosslanguageinformationretrieval}
James Mayfield, Eugene Yang, Dawn Lawrie, Samuel Barham, Orion Weller, Marc Mason, Suraj Nair, and Scott Miller.
\newblock Synthetic cross-language information retrieval training data, 2023.
\newblock URL \url{https://arxiv.org/abs/2305.00331}.

\bibitem[Wang et~al.(2024{\natexlab{a}})Wang, Yang, Huang, Yang, Majumder, and Wei]{wang2024improvingtextembeddingslarge}
Liang Wang, Nan Yang, Xiaolong Huang, Linjun Yang, Rangan Majumder, and Furu Wei.
\newblock Improving text embeddings with large language models, 2024{\natexlab{a}}.
\newblock URL \url{https://arxiv.org/abs/2401.00368}.

\bibitem[Lee et~al.(2024{\natexlab{a}})Lee, Dai, Ren, Chen, Cer, Cole, Hui, Boratko, Kapadia, Ding, Luan, Duddu, Abrego, Shi, Gupta, Kusupati, Jain, Jonnalagadda, Chang, and Naim]{lee2024geckoversatiletextembeddings}
Jinhyuk Lee, Zhuyun Dai, Xiaoqi Ren, Blair Chen, Daniel Cer, Jeremy~R. Cole, Kai Hui, Michael Boratko, Rajvi Kapadia, Wen Ding, Yi~Luan, Sai Meher~Karthik Duddu, Gustavo~Hernandez Abrego, Weiqiang Shi, Nithi Gupta, Aditya Kusupati, Prateek Jain, Siddhartha~Reddy Jonnalagadda, Ming-Wei Chang, and Iftekhar Naim.
\newblock Gecko: Versatile text embeddings distilled from large language models, 2024{\natexlab{a}}.
\newblock URL \url{https://arxiv.org/abs/2403.20327}.

\bibitem[Merrick et~al.(2024)Merrick, Xu, Nuti, and Campos]{merrick2024arcticembedscalableefficientaccurate}
Luke Merrick, Danmei Xu, Gaurav Nuti, and Daniel Campos.
\newblock Arctic-embed: Scalable, efficient, and accurate text embedding models, 2024.
\newblock URL \url{https://arxiv.org/abs/2405.05374}.

\bibitem[Wang et~al.(2024{\natexlab{b}})Wang, Yang, Huang, Yang, Majumder, and Wei]{wang2024multilinguale5textembeddings}
Liang Wang, Nan Yang, Xiaolong Huang, Linjun Yang, Rangan Majumder, and Furu Wei.
\newblock Multilingual e5 text embeddings: A technical report, 2024{\natexlab{b}}.
\newblock URL \url{https://arxiv.org/abs/2402.05672}.

\bibitem[Zhang et~al.(2024{\natexlab{b}})Zhang, Zhang, Long, Xie, Dai, Tang, Lin, Yang, Xie, Huang, Zhang, Li, and Zhang]{zhang2024mgtegeneralizedlongcontexttext}
Xin Zhang, Yanzhao Zhang, Dingkun Long, Wen Xie, Ziqi Dai, Jialong Tang, Huan Lin, Baosong Yang, Pengjun Xie, Fei Huang, Meishan Zhang, Wenjie Li, and Min Zhang.
\newblock mgte: Generalized long-context text representation and reranking models for multilingual text retrieval, 2024{\natexlab{b}}.
\newblock URL \url{https://arxiv.org/abs/2407.19669}.

\bibitem[Chen et~al.(2024{\natexlab{b}})Chen, Xiao, Zhang, Luo, Lian, and Liu]{chen2024bgem3embeddingmultilingualmultifunctionality}
Jianlv Chen, Shitao Xiao, Peitian Zhang, Kun Luo, Defu Lian, and Zheng Liu.
\newblock Bge m3-embedding: Multi-lingual, multi-functionality, multi-granularity text embeddings through self-knowledge distillation, 2024{\natexlab{b}}.
\newblock URL \url{https://arxiv.org/abs/2402.03216}.

\bibitem[ope()]{openaiembeddings}
New embedding models and api updates.
\newblock \url{https://openai.com/index/new-embedding-models-and-api-updates/}.
\newblock Accessed: 2024-01-25.

\bibitem[coh()]{cohereembeddings}
Introducing embed v3.
\newblock \url{https://cohere.com/blog/introducing-embed-v3}.
\newblock Accessed: 2023-11-02.

\bibitem[voy()]{voyageembeddings}
voyage-multilingual-2: Multilingual embedding model.
\newblock \url{https://blog.voyageai.com/2024/06/10/voyage-multilingual-2-multilingual-embedding-model/}.
\newblock Accessed: 2024-06-10.

\bibitem[Ma et~al.(2024)Ma, Wang, Yang, Wei, and Lin]{ma2024finetuningllama}
Xueguang Ma, Liang Wang, Nan Yang, Furu Wei, and Jimmy Lin.
\newblock Fine-tuning llama for multi-stage text retrieval.
\newblock In \emph{Proceedings of the 47th International ACM SIGIR Conference on Research and Development in Information Retrieval}, SIGIR '24, page 2421–2425, New York, NY, USA, 2024. Association for Computing Machinery.
\newblock ISBN 9798400704314.
\newblock \doi{10.1145/3626772.3657951}.
\newblock URL \url{https://doi.org/10.1145/3626772.3657951}.

\bibitem[Zhuang et~al.(2024{\natexlab{a}})Zhuang, Ma, Koopman, Lin, and Zuccon]{zhuang2024promptrepspromptinglargelanguage}
Shengyao Zhuang, Xueguang Ma, Bevan Koopman, Jimmy Lin, and Guido Zuccon.
\newblock Promptreps: Prompting large language models to generate dense and sparse representations for zero-shot document retrieval, 2024{\natexlab{a}}.
\newblock URL \url{https://arxiv.org/abs/2404.18424}.

\bibitem[Muennighoff et~al.(2022{\natexlab{a}})Muennighoff, Tazi, Magne, and Reimers]{muennighoff2022mteb}
Niklas Muennighoff, Nouamane Tazi, Lo{\"\i}c Magne, and Nils Reimers.
\newblock Mteb: Massive text embedding benchmark.
\newblock \emph{arXiv preprint arXiv:2210.07316}, 2022{\natexlab{a}}.
\newblock \doi{10.48550/ARXIV.2210.07316}.
\newblock URL \url{https://arxiv.org/abs/2210.07316}.

\bibitem[Lee et~al.(2024{\natexlab{b}})Lee, Roy, Xu, Raiman, Shoeybi, Catanzaro, and Ping]{lee2024nvembedimprovedtechniquestraining}
Chankyu Lee, Rajarshi Roy, Mengyao Xu, Jonathan Raiman, Mohammad Shoeybi, Bryan Catanzaro, and Wei Ping.
\newblock Nv-embed: Improved techniques for training llms as generalist embedding models, 2024{\natexlab{b}}.
\newblock URL \url{https://arxiv.org/abs/2405.17428}.

\bibitem[Springer et~al.(2024)Springer, Kotha, Fried, Neubig, and Raghunathan]{springer2024repetitionimproveslanguagemodel}
Jacob~Mitchell Springer, Suhas Kotha, Daniel Fried, Graham Neubig, and Aditi Raghunathan.
\newblock Repetition improves language model embeddings, 2024.
\newblock URL \url{https://arxiv.org/abs/2402.15449}.

\bibitem[BehnamGhader et~al.(2024)BehnamGhader, Adlakha, Mosbach, Bahdanau, Chapados, and Reddy]{behnamghader2024llm2veclargelanguagemodels}
Parishad BehnamGhader, Vaibhav Adlakha, Marius Mosbach, Dzmitry Bahdanau, Nicolas Chapados, and Siva Reddy.
\newblock Llm2vec: Large language models are secretly powerful text encoders, 2024.
\newblock URL \url{https://arxiv.org/abs/2404.05961}.

\bibitem[Muennighoff et~al.(2024{\natexlab{a}})Muennighoff, Su, Wang, Yang, Wei, Yu, Singh, and Kiela]{muennighoff2024generativerepresentationalinstructiontuning}
Niklas Muennighoff, Hongjin Su, Liang Wang, Nan Yang, Furu Wei, Tao Yu, Amanpreet Singh, and Douwe Kiela.
\newblock Generative representational instruction tuning, 2024{\natexlab{a}}.
\newblock URL \url{https://arxiv.org/abs/2402.09906}.

\bibitem[Kusupati et~al.(2024)Kusupati, Bhatt, Rege, Wallingford, Sinha, Ramanujan, Howard-Snyder, Chen, Kakade, Jain, and Farhadi]{kusupati2024matryoshkarepresentationlearning}
Aditya Kusupati, Gantavya Bhatt, Aniket Rege, Matthew Wallingford, Aditya Sinha, Vivek Ramanujan, William Howard-Snyder, Kaifeng Chen, Sham Kakade, Prateek Jain, and Ali Farhadi.
\newblock Matryoshka representation learning, 2024.
\newblock URL \url{https://arxiv.org/abs/2205.13147}.

\bibitem[Xiao et~al.(2024)Xiao, Liu, Zhang, and Xing]{xiao-etal-2024-lm}
Shitao Xiao, Zheng Liu, Peitian Zhang, and Xingrun Xing.
\newblock {LM}-cocktail: Resilient tuning of language models via model merging.
\newblock In Lun-Wei Ku, Andre Martins, and Vivek Srikumar, editors, \emph{Findings of the Association for Computational Linguistics ACL 2024}, pages 2474--2488, Bangkok, Thailand and virtual meeting, August 2024. Association for Computational Linguistics.
\newblock \doi{10.18653/v1/2024.findings-acl.145}.
\newblock URL \url{https://aclanthology.org/2024.findings-acl.145}.

\bibitem[Lee et~al.(2024{\natexlab{c}})Lee, Shakir, Koenig, and Lipp]{emb2024mxbai}
Sean Lee, Aamir Shakir, Darius Koenig, and Julius Lipp.
\newblock Open source strikes bread - new fluffy embeddings model, 2024{\natexlab{c}}.
\newblock URL \url{https://www.mixedbread.ai/blog/mxbai-embed-large-v1}.

\bibitem[Acharya et~al.(2024)Acharya, Murthy, Kumar, and Sen]{acharya2024nllbe5scalablemultilingualretrieval}
Arkadeep Acharya, Rudra Murthy, Vishwajeet Kumar, and Jaydeep Sen.
\newblock {NLLB-E5}: A scalable multilingual retrieval model, 2024.
\newblock URL \url{https://arxiv.org/abs/2409.05401}.

\bibitem[Bonifacio et~al.(2022{\natexlab{b}})Bonifacio, Jeronymo, Abonizio, Campiotti, Fadaee, Lotufo, and Nogueira]{bonifacio2022mmarcomultilingualversionms}
Luiz Bonifacio, Vitor Jeronymo, Hugo~Queiroz Abonizio, Israel Campiotti, Marzieh Fadaee, Roberto Lotufo, and Rodrigo Nogueira.
\newblock mmarco: A multilingual version of the ms marco passage ranking dataset, 2022{\natexlab{b}}.
\newblock URL \url{https://arxiv.org/abs/2108.13897}.

\bibitem[Zhang et~al.(2023{\natexlab{d}})Zhang, Thakur, Ogundepo, Kamalloo, Alfonso-Hermelo, Li, Liu, Rezagholizadeh, and Lin]{zhang2023miracl}
Xinyu Zhang, Nandan Thakur, Odunayo Ogundepo, Ehsan Kamalloo, David Alfonso-Hermelo, Xiaoguang Li, Qun Liu, Mehdi Rezagholizadeh, and Jimmy Lin.
\newblock {MIRACL: A Multilingual Retrieval Dataset Covering 18 Diverse Languages}.
\newblock \emph{Transactions of the Association for Computational Linguistics}, 11:\penalty0 1114--1131, 09 2023{\natexlab{d}}.
\newblock ISSN 2307-387X.
\newblock \doi{10.1162/tacl_a_00595}.
\newblock URL \url{https://doi.org/10.1162/tacl\_a\_00595}.

\bibitem[Nair et~al.(2022)Nair, Yang, Lawrie, Duh, McNamee, Murray, Mayfield, and Oard]{nair2022transfer}
Suraj Nair, Eugene Yang, Dawn Lawrie, Kevin Duh, Paul McNamee, Kenton Murray, James Mayfield, and Douglas~W. Oard.
\newblock Transfer learning approaches for building cross-language dense retrieval models.
\newblock In \emph{Advances in Information Retrieval: 44th European Conference on IR Research, ECIR 2022, Stavanger, Norway, April 10–14, 2022, Proceedings, Part I}, page 382–396, Berlin, Heidelberg, 2022. Springer-Verlag.
\newblock ISBN 978-3-030-99735-9.
\newblock \doi{10.1007/978-3-030-99736-6_26}.
\newblock URL \url{https://doi.org/10.1007/978-3-030-99736-6_26}.

\bibitem[Yang et~al.(2024{\natexlab{a}})Yang, Lawrie, Mayfield, Oard, and Miller]{yang2024translate}
Eugene Yang, Dawn Lawrie, James Mayfield, Douglas~W. Oard, and Scott Miller.
\newblock Translate-distill: Learning cross-language dense retrieval by translation and distillation.
\newblock In \emph{Advances in Information Retrieval: 46th European Conference on Information Retrieval, ECIR 2024, Glasgow, UK, March 24–28, 2024, Proceedings, Part II}, page 50–65, Berlin, Heidelberg, 2024{\natexlab{a}}. Springer-Verlag.
\newblock ISBN 978-3-031-56059-0.
\newblock \doi{10.1007/978-3-031-56060-6_4}.
\newblock URL \url{https://doi.org/10.1007/978-3-031-56060-6_4}.

\bibitem[Huang et~al.(2023{\natexlab{b}})Huang, Yu, and Allan]{huang2023improving}
Zhiqi Huang, Puxuan Yu, and James Allan.
\newblock Improving cross-lingual information retrieval on low-resource languages via optimal transport distillation.
\newblock In \emph{Proceedings of the Sixteenth ACM International Conference on Web Search and Data Mining}, WSDM '23, page 1048–1056, New York, NY, USA, 2023{\natexlab{b}}. Association for Computing Machinery.
\newblock ISBN 9781450394079.
\newblock \doi{10.1145/3539597.3570468}.
\newblock URL \url{https://doi.org/10.1145/3539597.3570468}.

\bibitem[Lawrie et~al.(2023)Lawrie, Yang, Oard, and Mayfield]{lawrie2023neuralapproachestomultilingualinformationretrieval}
Dawn Lawrie, Eugene Yang, Douglas~W. Oard, and James Mayfield.
\newblock Neural approaches to multilingual information retrieval.
\newblock In \emph{Advances in Information Retrieval: 45th European Conference on Information Retrieval, ECIR 2023, Dublin, Ireland, April 2–6, 2023, Proceedings, Part I}, page 521–536, Berlin, Heidelberg, 2023. Springer-Verlag.
\newblock ISBN 978-3-031-28243-0.
\newblock \doi{10.1007/978-3-031-28244-7_33}.
\newblock URL \url{https://doi.org/10.1007/978-3-031-28244-7_33}.

\bibitem[Yang et~al.(2024{\natexlab{b}})Yang, Lawrie, and Mayfield]{yang2024multilingual}
Eugene Yang, Dawn Lawrie, and James Mayfield.
\newblock Distillation for multilingual information retrieval.
\newblock In \emph{Proceedings of the 47th International ACM SIGIR Conference on Research and Development in Information Retrieval}, SIGIR '24, page 2368–2373, New York, NY, USA, 2024{\natexlab{b}}. Association for Computing Machinery.
\newblock ISBN 9798400704314.
\newblock \doi{10.1145/3626772.3657955}.
\newblock URL \url{https://doi.org/10.1145/3626772.3657955}.

\bibitem[Louis et~al.(2024)Louis, Saxena, van Dijck, and Spanakis]{louis2024colbertxmmodularmultivectorrepresentation}
Antoine Louis, Vageesh Saxena, Gijs van Dijck, and Gerasimos Spanakis.
\newblock Colbert-xm: A modular multi-vector representation model for zero-shot multilingual information retrieval, 2024.
\newblock URL \url{https://arxiv.org/abs/2402.15059}.

\bibitem[Chen et~al.(2024{\natexlab{c}})Chen, Xiao, Zhang, Luo, Lian, and Liu]{chen-etal-2024-m3}
Jianlyu Chen, Shitao Xiao, Peitian Zhang, Kun Luo, Defu Lian, and Zheng Liu.
\newblock {M}3-embedding: Multi-linguality, multi-functionality, multi-granularity text embeddings through self-knowledge distillation.
\newblock In Lun-Wei Ku, Andre Martins, and Vivek Srikumar, editors, \emph{Findings of the Association for Computational Linguistics ACL 2024}, pages 2318--2335, Bangkok, Thailand and virtual meeting, August 2024{\natexlab{c}}. Association for Computational Linguistics.
\newblock \doi{10.18653/v1/2024.findings-acl.137}.
\newblock URL \url{https://aclanthology.org/2024.findings-acl.137}.

\bibitem[Jeronymo et~al.(2023{\natexlab{b}})Jeronymo, Lotufo, and Nogueira]{jeronymo2023neuralmindunicamp2022trecneuclir}
Vitor Jeronymo, Roberto Lotufo, and Rodrigo Nogueira.
\newblock Neuralmind-unicamp at 2022 trec neuclir: Large boring rerankers for cross-lingual retrieval, 2023{\natexlab{b}}.
\newblock URL \url{https://arxiv.org/abs/2303.16145}.

\bibitem[Sun et~al.(2023)Sun, Yan, Ma, Wang, Ren, Chen, Yin, and Ren]{sun-etal-2023-chatgpt}
Weiwei Sun, Lingyong Yan, Xinyu Ma, Shuaiqiang Wang, Pengjie Ren, Zhumin Chen, Dawei Yin, and Zhaochun Ren.
\newblock Is {C}hat{GPT} good at search? investigating large language models as re-ranking agents.
\newblock In Houda Bouamor, Juan Pino, and Kalika Bali, editors, \emph{Proceedings of the 2023 Conference on Empirical Methods in Natural Language Processing}, pages 14918--14937, Singapore, December 2023. Association for Computational Linguistics.
\newblock \doi{10.18653/v1/2023.emnlp-main.923}.
\newblock URL \url{https://aclanthology.org/2023.emnlp-main.923}.

\bibitem[Ma et~al.(2023)Ma, Zhang, Pradeep, and Lin]{ma2023zeroshotlistwisedocumentreranking}
Xueguang Ma, Xinyu Zhang, Ronak Pradeep, and Jimmy Lin.
\newblock Zero-shot listwise document reranking with a large language model, 2023.
\newblock URL \url{https://arxiv.org/abs/2305.02156}.

\bibitem[Zhuang et~al.(2024{\natexlab{b}})Zhuang, Zhuang, Koopman, and Zuccon]{zhuang2024setwise}
Shengyao Zhuang, Honglei Zhuang, Bevan Koopman, and Guido Zuccon.
\newblock A setwise approach for effective and highly efficient zero-shot ranking with large language models.
\newblock In \emph{Proceedings of the 47th International ACM SIGIR Conference on Research and Development in Information Retrieval}, SIGIR '24, page 38–47, New York, NY, USA, 2024{\natexlab{b}}. Association for Computing Machinery.
\newblock ISBN 9798400704314.
\newblock \doi{10.1145/3626772.3657813}.
\newblock URL \url{https://doi.org/10.1145/3626772.3657813}.

\bibitem[Pradeep et~al.(2023{\natexlab{a}})Pradeep, Sharifymoghaddam, and Lin]{pradeep2023rankvicunazeroshotlistwisedocument}
Ronak Pradeep, Sahel Sharifymoghaddam, and Jimmy Lin.
\newblock Rankvicuna: Zero-shot listwise document reranking with open-source large language models, 2023{\natexlab{a}}.
\newblock URL \url{https://arxiv.org/abs/2309.15088}.

\bibitem[Pradeep et~al.(2023{\natexlab{b}})Pradeep, Sharifymoghaddam, and Lin]{pradeep2023rankzephyreffectiverobustzeroshot}
Ronak Pradeep, Sahel Sharifymoghaddam, and Jimmy Lin.
\newblock Rankzephyr: Effective and robust zero-shot listwise reranking is a breeze!, 2023{\natexlab{b}}.
\newblock URL \url{https://arxiv.org/abs/2312.02724}.

\bibitem[Zhang et~al.(2023{\natexlab{e}})Zhang, Hofstätter, Lewis, Tang, and Lin]{zhang2023rankwithoutgptbuildinggptindependentlistwise}
Xinyu Zhang, Sebastian Hofstätter, Patrick Lewis, Raphael Tang, and Jimmy Lin.
\newblock Rank-without-gpt: Building gpt-independent listwise rerankers on open-source large language models, 2023{\natexlab{e}}.
\newblock URL \url{https://arxiv.org/abs/2312.02969}.

\bibitem[Chen et~al.(2024{\natexlab{d}})Chen, Liu, Zhang, Sun, Shi, Mao, and Yin]{chen2024tourrankutilizinglargelanguage}
Yiqun Chen, Qi~Liu, Yi~Zhang, Weiwei Sun, Daiting Shi, Jiaxin Mao, and Dawei Yin.
\newblock Tourrank: Utilizing large language models for documents ranking with a tournament-inspired strategy, 2024{\natexlab{d}}.
\newblock URL \url{https://arxiv.org/abs/2406.11678}.

\bibitem[Adeyemi et~al.(2023)Adeyemi, Oladipo, Pradeep, and Lin]{adeyemi2023zeroshotcrosslingualrerankinglarge}
Mofetoluwa Adeyemi, Akintunde Oladipo, Ronak Pradeep, and Jimmy Lin.
\newblock Zero-shot cross-lingual reranking with large language models for low-resource languages, 2023.
\newblock URL \url{https://arxiv.org/abs/2312.16159}.

\bibitem[Sitawarin et~al.(2024)Sitawarin, Mu, Wagner, and Araujo]{sitawarin2024pal}
Chawin Sitawarin, Norman Mu, David Wagner, and Alexandre Araujo.
\newblock Pal: Proxy-guided black-box attack on large language models, 2024.

\bibitem[Zou et~al.(2023)Zou, Wang, Kolter, and Fredrikson]{zou2023universal}
Andy Zou, Zifan Wang, J~Zico Kolter, and Matt Fredrikson.
\newblock Universal and transferable adversarial attacks on aligned language models.
\newblock \emph{arXiv preprint arXiv:2307.15043}, 2023.

\bibitem[Wei et~al.(2024)Wei, Haghtalab, and Steinhardt]{wei2024jailbroken}
Alexander Wei, Nika Haghtalab, and Jacob Steinhardt.
\newblock Jailbroken: How does llm safety training fail?
\newblock \emph{Advances in Neural Information Processing Systems}, 36, 2024.

\bibitem[Liu et~al.(2023{\natexlab{b}})Liu, Deng, Xu, Li, Zheng, Zhang, Zhao, Zhang, and Liu]{liu2023jailbreaking}
Yi~Liu, Gelei Deng, Zhengzi Xu, Yuekang Li, Yaowen Zheng, Ying Zhang, Lida Zhao, Tianwei Zhang, and Yang Liu.
\newblock Jailbreaking chatgpt via prompt engineering: An empirical study.
\newblock \emph{arXiv preprint arXiv:2305.13860}, 2023{\natexlab{b}}.

\bibitem[Shen et~al.(2023)Shen, Chen, Backes, Shen, and Zhang]{shen2023do}
Xinyue Shen, Zeyuan Chen, Michael Backes, Yun Shen, and Yang Zhang.
\newblock "do anything now": Characterizing and evaluating in-the-wild jailbreak prompts on large language models, 2023.

\bibitem[Deng et~al.(2024)Deng, Liu, Li, Wang, Zhang, Li, Wang, Zhang, and Liu]{deng2024masterkey}
Gelei Deng, Yi~Liu, Yuekang Li, Kailong Wang, Ying Zhang, Zefeng Li, Haoyu Wang, Tianwei Zhang, and Yang Liu.
\newblock Masterkey: Automated jailbreaking of large language model chatbots.
\newblock In \emph{Proc. ISOC NDSS}, 2024.

\bibitem[Li et~al.(2024{\natexlab{b}})Li, Dong, Wang, Hu, Zuo, Lin, Qiao, and Shao]{li2024salad}
Lijun Li, Bowen Dong, Ruohui Wang, Xuhao Hu, Wangmeng Zuo, Dahua Lin, Yu~Qiao, and Jing Shao.
\newblock Salad-bench: A hierarchical and comprehensive safety benchmark for large language models.
\newblock \emph{arXiv preprint arXiv:2402.05044}, 2024{\natexlab{b}}.

\bibitem[Liu et~al.(2023{\natexlab{c}})Liu, Xu, Chen, and Xiao]{liu2023autodan}
Xiaogeng Liu, Nan Xu, Muhao Chen, and Chaowei Xiao.
\newblock Autodan: Generating stealthy jailbreak prompts on aligned large language models.
\newblock \emph{arXiv preprint arXiv:2310.04451}, 2023{\natexlab{c}}.

\bibitem[Jin et~al.(2024)Jin, Chen, Zhou, Chen, Zhang, and Wang]{jin2024guard}
Haibo Jin, Ruoxi Chen, Andy Zhou, Jinyin Chen, Yang Zhang, and Haohan Wang.
\newblock Guard: Role-playing to generate natural-language jailbreakings to test guideline adherence of large language models.
\newblock \emph{arXiv preprint arXiv:2402.03299}, 2024.

\bibitem[Shen et~al.(2024)Shen, Tan, Chen, Chen, Zhang, Xu, Zheng, Koehn, and Khashabi]{shen2024language}
Lingfeng Shen, Weiting Tan, Sihao Chen, Yunmo Chen, Jingyu Zhang, Haoran Xu, Boyuan Zheng, Philipp Koehn, and Daniel Khashabi.
\newblock The language barrier: Dissecting safety challenges of llms in multilingual contexts, 2024.

\bibitem[Deng et~al.(2023)Deng, Zhang, Pan, and Bing]{deng2023multilingual}
Yue Deng, Wenxuan Zhang, Sinno~Jialin Pan, and Lidong Bing.
\newblock Multilingual jailbreak challenges in large language models.
\newblock \emph{arXiv preprint arXiv:2310.06474}, 2023.

\bibitem[Puttaparthi et~al.(2023)Puttaparthi, Deo, Gul, Tang, Shang, and Yu]{puttaparthi2023comprehensive}
Poorna Chander~Reddy Puttaparthi, Soham~Sanjay Deo, Hakan Gul, Yiming Tang, Weiyi Shang, and Zhe Yu.
\newblock Comprehensive evaluation of chatgpt reliability through multilingual inquiries, 2023.

\bibitem[Yong et~al.(2023)Yong, Menghini, and Bach]{yong2023low}
Zheng-Xin Yong, Cristina Menghini, and Stephen~H Bach.
\newblock Low-resource languages jailbreak gpt-4.
\newblock \emph{arXiv preprint arXiv:2310.02446}, 2023.

\bibitem[Xu et~al.(2023)Xu, Wang, Zhou, Li, Xiao, and Chen]{xu2023cognitive}
Nan Xu, Fei Wang, Ben Zhou, Bang~Zheng Li, Chaowei Xiao, and Muhao Chen.
\newblock Cognitive overload: Jailbreaking large language models with overloaded logical thinking.
\newblock \emph{arXiv preprint arXiv:2311.09827}, 2023.

\bibitem[Li et~al.(2024{\natexlab{c}})Li, Liu, Liu, Shi, Ren, Zheng, Liu, and Xue]{li2024cross}
Jie Li, Yi~Liu, Chongyang Liu, Ling Shi, Xiaoning Ren, Yaowen Zheng, Yang Liu, and Yinxing Xue.
\newblock A cross-language investigation into jailbreak attacks in large language models.
\newblock \emph{arXiv preprint arXiv:2401.16765}, 2024{\natexlab{c}}.

\bibitem[Yuan et~al.(2024)Yuan, Jiao, Wang, tse Huang, He, Shi, and Tu]{yuan2024gpt4}
Youliang Yuan, Wenxiang Jiao, Wenxuan Wang, Jen tse Huang, Pinjia He, Shuming Shi, and Zhaopeng Tu.
\newblock Gpt-4 is too smart to be safe: Stealthy chat with llms via cipher, 2024.

\bibitem[Huang et~al.(2023{\natexlab{c}})Huang, Gupta, Xia, Li, and Chen]{huang2023catastrophic}
Yangsibo Huang, Samyak Gupta, Mengzhou Xia, Kai Li, and Danqi Chen.
\newblock Catastrophic jailbreak of open-source llms via exploiting generation.
\newblock \emph{arXiv preprint arXiv:2310.06987}, 2023{\natexlab{c}}.

\bibitem[Robey et~al.(2023)Robey, Wong, Hassani, and Pappas]{robey2023smoothllm}
Alexander Robey, Eric Wong, Hamed Hassani, and George~J Pappas.
\newblock Smoothllm: Defending large language models against jailbreaking attacks.
\newblock \emph{arXiv preprint arXiv:2310.03684}, 2023.

\bibitem[Zhou et~al.(2024{\natexlab{a}})Zhou, Li, and Wang]{zhou2024robust}
Andy Zhou, Bo~Li, and Haohan Wang.
\newblock Robust prompt optimization for defending language models against jailbreaking attacks.
\newblock \emph{arXiv preprint arXiv:2401.17263}, 2024{\natexlab{a}}.

\bibitem[Jain et~al.(2023)Jain, Schwarzschild, Wen, Somepalli, Kirchenbauer, Chiang, Goldblum, Saha, Geiping, and Goldstein]{jain2023baseline}
Neel Jain, Avi Schwarzschild, Yuxin Wen, Gowthami Somepalli, John Kirchenbauer, Ping-yeh Chiang, Micah Goldblum, Aniruddha Saha, Jonas Geiping, and Tom Goldstein.
\newblock Baseline defenses for adversarial attacks against aligned language models.
\newblock \emph{arXiv preprint arXiv:2309.00614}, 2023.

\bibitem[Wu et~al.(2024)Wu, Wang, Liu, and Liu]{wu2024llms}
Daoyuan Wu, Shuai Wang, Yang Liu, and Ning Liu.
\newblock Llms can defend themselves against jailbreaking in a practical manner: A vision paper, 2024.

\bibitem[Li et~al.(2023{\natexlab{c}})Li, Wei, Zhao, Zhang, and Zhang]{li2023rain}
Yuhui Li, Fangyun Wei, Jinjing Zhao, Chao Zhang, and Hongyang Zhang.
\newblock Rain: Your language models can align themselves without finetuning.
\newblock In \emph{The Twelfth International Conference on Learning Representations}, 2023{\natexlab{c}}.

\bibitem[Geng et~al.(2023)Geng, Yan, Cao, Li, Li, Li, Zhou, Yang, and Zhang]{geng2023kbioxlm}
Lei Geng, Xu~Yan, Ziqiang Cao, Juntao Li, Wenjie Li, Sujian Li, Xinjie Zhou, Yang Yang, and Jun Zhang.
\newblock Kbioxlm: A knowledge-anchored biomedical multilingual pretrained language model.
\newblock \emph{arXiv preprint arXiv:2311.11564}, 2023.

\bibitem[Labrak et~al.(2024)Labrak, Bazoge, Morin, Gourraud, Rouvier, and Dufour]{labrak2024biomistral}
Yanis Labrak, Adrien Bazoge, Emmanuel Morin, Pierre-Antoine Gourraud, Mickael Rouvier, and Richard Dufour.
\newblock Biomistral: A collection of open-source pretrained large language models for medical domains.
\newblock \emph{arXiv preprint arXiv:2402.10373}, 2024.

\bibitem[Qiu et~al.(2024{\natexlab{a}})Qiu, Wu, Zhang, Lin, Wang, Zhang, Wang, and Xie]{qiu2024towards}
Pengcheng Qiu, Chaoyi Wu, Xiaoman Zhang, Weixiong Lin, Haicheng Wang, Ya~Zhang, Yanfeng Wang, and Weidi Xie.
\newblock Towards building multilingual language model for medicine.
\newblock \emph{arXiv preprint arXiv:2402.13963}, 2024{\natexlab{a}}.

\bibitem[Wang et~al.(2024{\natexlab{c}})Wang, Chen, Chen, Hu, Wang, Wu, Gao, Wan, Li, and Wang]{wang2024apollo}
Xidong Wang, Nuo Chen, Junyin Chen, Yan Hu, Yidong Wang, Xiangbo Wu, Anningzhe Gao, Xiang Wan, Haizhou Li, and Benyou Wang.
\newblock Apollo: Lightweight multilingual medical llms towards democratizing medical ai to 6b people.
\newblock \emph{arXiv preprint arXiv:2403.03640}, 2024{\natexlab{c}}.

\bibitem[Gangavarapu(2024)]{gangavarapu2024introducing}
Agasthya Gangavarapu.
\newblock Introducing l2m3, a multilingual medical large language model to advance health equity in low-resource regions.
\newblock \emph{arXiv preprint arXiv:2404.08705}, 2024.

\bibitem[Garc{\'\i}a-Ferrero et~al.(2024)Garc{\'\i}a-Ferrero, Agerri, Salazar, Cabrio, de~la Iglesia, Lavelli, Magnini, Molinet, Ramirez-Romero, Rigau, et~al.]{garcia2024medical}
Iker Garc{\'\i}a-Ferrero, Rodrigo Agerri, Aitziber~Atutxa Salazar, Elena Cabrio, Iker de~la Iglesia, Alberto Lavelli, Bernardo Magnini, Benjamin Molinet, Johana Ramirez-Romero, German Rigau, et~al.
\newblock Medical mt5: An open-source multilingual text-to-text llm for the medical domain.
\newblock \emph{arXiv preprint arXiv:2404.07613}, 2024.

\bibitem[Niklaus et~al.(2023{\natexlab{a}})Niklaus, Matoshi, Rani, Galassi, St{\"u}rmer, Chalkidis, et~al.]{niklaus2023lextreme}
Joel Niklaus, Veton Matoshi, Pooja Rani, Andrea Galassi, Matthias St{\"u}rmer, Ilias Chalkidis, et~al.
\newblock Lextreme: A multi-lingual and multi-task benchmark for the legal domain.
\newblock In \emph{Findings of the Association for Computational Linguistics: EMNLP 2023}, pages 3016--3054. Association for Computational Linguistics, 2023{\natexlab{a}}.

\bibitem[Brugger et~al.(2023)Brugger, St{\"u}rmer, and Niklaus]{brugger2023multilegalsbd}
Tobias Brugger, Matthias St{\"u}rmer, and Joel Niklaus.
\newblock Multilegalsbd: A multilingual legal sentence boundary detection dataset.
\newblock In \emph{Proceedings of the Nineteenth International Conference on Artificial Intelligence and Law}, pages 42--51, 2023.

\bibitem[Christen et~al.(2023)Christen, Shaitarova, St{\"u}rmer, and Niklaus]{christen2023resolving}
Ramona Christen, Anastassia Shaitarova, Matthias St{\"u}rmer, and Joel Niklaus.
\newblock Resolving legalese: A multilingual exploration of negation scope resolution in legal documents.
\newblock \emph{arXiv preprint arXiv:2309.08695}, 2023.

\bibitem[Baumgartner et~al.(2024)Baumgartner, St{\"u}rmer, Grabmair, Niklaus, et~al.]{baumgartner2024towards}
Nina Baumgartner, Matthias St{\"u}rmer, Matthias Grabmair, Joel Niklaus, et~al.
\newblock Towards explainability and fairness in swiss judgement prediction: Benchmarking on a multilingual dataset.
\newblock \emph{arXiv preprint arXiv:2402.17013}, 2024.

\bibitem[Niklaus et~al.(2023{\natexlab{b}})Niklaus, Matoshi, St{\"u}rmer, Chalkidis, and Ho]{niklaus2023multilegalpile}
Joel Niklaus, Veton Matoshi, Matthias St{\"u}rmer, Ilias Chalkidis, and Daniel~E Ho.
\newblock Multilegalpile: A 689gb multilingual legal corpus.
\newblock \emph{arXiv preprint arXiv:2306.02069}, 2023{\natexlab{b}}.

\bibitem[Chalkidis et~al.(2023)Chalkidis, Garneau, Goanț{\u{a}}, Katz, and S{\o}gaard]{chalkidis2023lexfiles}
Ilias Chalkidis, Nicolas Garneau, C{\u{a}}t{\u{a}}lina Goanț{\u{a}}, Daniel Katz, and Anders S{\o}gaard.
\newblock Lexfiles and legallama: Facilitating english multinational legal language model development.
\newblock In \emph{Proceedings of the 61st Annual Meeting of the Association for Computational Linguistics (Volume 1: Long Papers)}, pages 15513--15535, 2023.

\bibitem[Trautmann et~al.(2022)Trautmann, Petrova, and Schilder]{trautmann2022legal}
Dietrich Trautmann, Alina Petrova, and Frank Schilder.
\newblock Legal prompt engineering for multilingual legal judgement prediction.
\newblock \emph{arXiv preprint arXiv:2212.02199}, 2022.

\bibitem[mteb({\natexlab{a}})]{amazon_massive_intent}
mteb.
\newblock Amazon massive intent, {\natexlab{a}}.
\newblock URL \url{https://huggingface.co/datasets/mteb/amazon_massive_intent}.

\bibitem[Keung et~al.(2020{\natexlab{a}})Keung, Lu, Szarvas, and Smith]{amazon_reviews_multi}
Phillip Keung, Yichao Lu, György Szarvas, and Noah~A. Smith.
\newblock The multilingual amazon reviews corpus.
\newblock In \emph{Proceedings of the 2020 Conference on Empirical Methods in Natural Language Processing}, 2020{\natexlab{a}}.

\bibitem[Singh et~al.(2024)Singh, Vargus, D{'}souza, Karlsson, Mahendiran, Ko, Shandilya, Patel, Mataciunas, O{'}Mahony, Zhang, Hettiarachchi, Wilson, Machado, Moura, Krzemi{\'n}ski, Fadaei, Ergun, Okoh, Alaagib, Mudannayake, Alyafeai, Chien, Ruder, Guthikonda, Alghamdi, Gehrmann, Muennighoff, Bartolo, Kreutzer, {\"U}st{\"u}n, Fadaee, and Hooker]{singh-etal-2024-aya}
Shivalika Singh, Freddie Vargus, Daniel D{'}souza, B{\"o}rje Karlsson, Abinaya Mahendiran, Wei-Yin Ko, Herumb Shandilya, Jay Patel, Deividas Mataciunas, Laura O{'}Mahony, Mike Zhang, Ramith Hettiarachchi, Joseph Wilson, Marina Machado, Luisa Moura, Dominik Krzemi{\'n}ski, Hakimeh Fadaei, Irem Ergun, Ifeoma Okoh, Aisha Alaagib, Oshan Mudannayake, Zaid Alyafeai, Vu~Chien, Sebastian Ruder, Surya Guthikonda, Emad Alghamdi, Sebastian Gehrmann, Niklas Muennighoff, Max Bartolo, Julia Kreutzer, Ahmet {\"U}st{\"u}n, Marzieh Fadaee, and Sara Hooker.
\newblock Aya dataset: An open-access collection for multilingual instruction tuning.
\newblock In Lun-Wei Ku, Andre Martins, and Vivek Srikumar, editors, \emph{Proceedings of the 62nd Annual Meeting of the Association for Computational Linguistics (Volume 1: Long Papers)}, pages 11521--11567, Bangkok, Thailand, August 2024. Association for Computational Linguistics.
\newblock \doi{10.18653/v1/2024.acl-long.620}.
\newblock URL \url{https://aclanthology.org/2024.acl-long.620}.

\bibitem[Li et~al.(2023{\natexlab{d}})Li, Koto, Wu, Aji, and Baldwin]{li2023bactrianx}
Haonan Li, Fajri Koto, Minghao Wu, Alham~Fikri Aji, and Timothy Baldwin.
\newblock Bactrian-x: Multilingual replicable instruction-following models with low-rank adaptation, 2023{\natexlab{d}}.

\bibitem[eBible.org()]{biblenlp-corpus-mmteb}
eBible.org.
\newblock ebible.
\newblock URL \url{https://github.com/BibleNLP}.

\bibitem[Conneau et~al.(2020{\natexlab{a}})Conneau, Khandelwal, Goyal, Chaudhary, Wenzek, Guzm{\'a}n, Grave, Ott, Zettlemoyer, and Stoyanov]{cc100_1}
Alexis Conneau, Kartikay Khandelwal, Naman Goyal, Vishrav Chaudhary, Guillaume Wenzek, Francisco Guzm{\'a}n, Edouard Grave, Myle Ott, Luke Zettlemoyer, and Veselin Stoyanov.
\newblock Unsupervised cross-lingual representation learning at scale.
\newblock In Dan Jurafsky, Joyce Chai, Natalie Schluter, and Joel Tetreault, editors, \emph{Proceedings of the 58th Annual Meeting of the Association for Computational Linguistics}, pages 8440--8451, Online, July 2020{\natexlab{a}}. Association for Computational Linguistics.
\newblock \doi{10.18653/v1/2020.acl-main.747}.
\newblock URL \url{https://aclanthology.org/2020.acl-main.747}.

\bibitem[Wenzek et~al.(2020)Wenzek, Lachaux, Conneau, Chaudhary, Guzm{\'a}n, Joulin, and Grave]{cc100_2}
Guillaume Wenzek, Marie-Anne Lachaux, Alexis Conneau, Vishrav Chaudhary, Francisco Guzm{\'a}n, Armand Joulin, and Edouard Grave.
\newblock {CCN}et: Extracting high quality monolingual datasets from web crawl data.
\newblock In Nicoletta Calzolari, Fr{\'e}d{\'e}ric B{\'e}chet, Philippe Blache, Khalid Choukri, Christopher Cieri, Thierry Declerck, Sara Goggi, Hitoshi Isahara, Bente Maegaard, Joseph Mariani, H{\'e}l{\`e}ne Mazo, Asuncion Moreno, Jan Odijk, and Stelios Piperidis, editors, \emph{Proceedings of the Twelfth Language Resources and Evaluation Conference}, pages 4003--4012, Marseille, France, May 2020. European Language Resources Association.
\newblock URL \url{https://aclanthology.org/2020.lrec-1.494}.

\bibitem[Nguyen et~al.(2023)Nguyen, Nguyen, Lai, Man, Ngo, Dernoncourt, Rossi, and Nguyen]{nguyen2023culturax}
Thuat Nguyen, Chien~Van Nguyen, Viet~Dac Lai, Hieu Man, Nghia~Trung Ngo, Franck Dernoncourt, Ryan~A. Rossi, and Thien~Huu Nguyen.
\newblock Culturax: A cleaned, enormous, and multilingual dataset for large language models in 167 languages, 2023.

\bibitem[erfanzar()]{GPT-4-Prompts}
erfanzar.
\newblock Multi-turn conversational prompts from chatgpt-4.
\newblock URL \url{https://huggingface.co/datasets/erfanzar/GPT-4-Prompts}.

\bibitem[Dettmers et~al.(2023)Dettmers, Pagnoni, Holtzman, and Zettlemoyer]{guannaco}
Tim Dettmers, Artidoro Pagnoni, Ari Holtzman, and Luke Zettlemoyer.
\newblock Qlora: Efficient finetuning of quantized llms, 2023.

\bibitem[de~Gibert et~al.(2024)de~Gibert, Nail, Arefyev, Bañón, van~der Linde, Ji, Zaragoza-Bernabeu, Aulamo, Ramírez-Sánchez, Kutuzov, Pyysalo, Oepen, and Tiedemann]{hplt_monolingual_v1_2}
Ona de~Gibert, Graeme Nail, Nikolay Arefyev, Marta Bañón, Jelmer van~der Linde, Shaoxiong Ji, Jaume Zaragoza-Bernabeu, Mikko Aulamo, Gema Ramírez-Sánchez, Andrey Kutuzov, Sampo Pyysalo, Stephan Oepen, and Jörg Tiedemann.
\newblock A new massive multilingual dataset for high-performance language technologies, 2024.

\bibitem[Cettolo et~al.(2017)Cettolo, Federico, Bentivogli, Niehues, St{\"u}ker, Sudoh, Yoshino, and Federmann]{iwslt2017}
Mauro Cettolo, Marcello Federico, Luisa Bentivogli, Jan Niehues, Sebastian St{\"u}ker, Katsuhito Sudoh, Koichiro Yoshino, and Christian Federmann.
\newblock Overview of the {IWSLT} 2017 evaluation campaign.
\newblock In \emph{Proceedings of the 14th International Conference on Spoken Language Translation}, pages 2--14, Tokyo, Japan, December 14-15 2017. International Workshop on Spoken Language Translation.
\newblock URL \url{https://aclanthology.org/2017.iwslt-1.1}.

\bibitem[Raffel et~al.(2019)Raffel, Shazeer, Roberts, Lee, Narang, Matena, Zhou, Li, and Liu]{mc4}
Colin Raffel, Noam Shazeer, Adam Roberts, Katherine Lee, Sharan Narang, Michael Matena, Yanqi Zhou, Wei Li, and Peter~J. Liu.
\newblock Exploring the limits of transfer learning with a unified text-to-text transformer.
\newblock \emph{arXiv e-prints}, 2019.

\bibitem[Ruder et~al.(2021{\natexlab{a}})Ruder, Constant, Botha, Siddhant, Firat, Fu, Liu, Hu, Garrette, Neubig, and Johnson]{mewsli-x}
Sebastian Ruder, Noah Constant, Jan Botha, Aditya Siddhant, Orhan Firat, Jinlan Fu, Pengfei Liu, Junjie Hu, Dan Garrette, Graham Neubig, and Melvin Johnson.
\newblock {XTREME}-{R}: Towards more challenging and nuanced multilingual evaluation.
\newblock In \emph{Proceedings of the 2021 Conference on Empirical Methods in Natural Language Processing}, pages 10215--10245, Online and Punta Cana, Dominican Republic, November 2021{\natexlab{a}}. Association for Computational Linguistics.
\newblock \doi{10.18653/v1/2021.emnlp-main.802}.
\newblock URL \url{https://aclanthology.org/2021.emnlp-main.802}.

\bibitem[Gerz et~al.(2021)Gerz, Su, Kusztos, Mondal, Lis, Singhal, Mrksic, Wen, and Vulic]{minds14}
Daniela Gerz, Pei{-}Hao Su, Razvan Kusztos, Avishek Mondal, Michal Lis, Eshan Singhal, Nikola Mrksic, Tsung{-}Hsien Wen, and Ivan Vulic.
\newblock Multilingual and cross-lingual intent detection from spoken data.
\newblock \emph{CoRR}, abs/2104.08524, 2021.
\newblock URL \url{https://arxiv.org/abs/2104.08524}.

\bibitem[Zhang et~al.(2022{\natexlab{b}})Zhang, Thakur, Ogundepo, Kamalloo, Alfonso-Hermelo, Li, Liu, Rezagholizadeh, and Lin]{miracl}
Xinyu Zhang, Nandan Thakur, Odunayo Ogundepo, Ehsan Kamalloo, David Alfonso-Hermelo, Xiaoguang Li, Qun Liu, Mehdi Rezagholizadeh, and Jimmy Lin.
\newblock Making a miracl: Multilingual information retrieval across a continuum of languages, 2022{\natexlab{b}}.

\bibitem[Chen et~al.(2024{\natexlab{e}})Chen, Xiao, Zhang, Luo, Lian, and Liu]{MLDR}
Jianlv Chen, Shitao Xiao, Peitian Zhang, Kun Luo, Defu Lian, and Zheng Liu.
\newblock Bge m3-embedding: Multi-lingual, multi-functionality, multi-granularity text embeddings through self-knowledge distillation, 2024{\natexlab{e}}.

\bibitem[Qiu et~al.(2024{\natexlab{b}})Qiu, Wu, Zhang, Lin, Wang, Zhang, Wang, and Xie]{MMedC}
Pengcheng Qiu, Chaoyi Wu, Xiaoman Zhang, Weixiong Lin, Haicheng Wang, Ya~Zhang, Yanfeng Wang, and Weidi Xie.
\newblock Towards building multilingual language model for medicine, 2024{\natexlab{b}}.

\bibitem[De~Bruyn et~al.(2021)De~Bruyn, Lotfi, Buhmann, and Daelemans]{mqa}
Maxime De~Bruyn, Ehsan Lotfi, Jeska Buhmann, and Walter Daelemans.
\newblock {MFAQ}: a multilingual {FAQ} dataset.
\newblock In \emph{Proceedings of the 3rd Workshop on Machine Reading for Question Answering}, pages 1--13, Punta Cana, Dominican Republic, November 2021. Association for Computational Linguistics.
\newblock URL \url{https://aclanthology.org/2021.mrqa-1.1}.

\bibitem[tyqiangz()]{multilingual-sentiments}
tyqiangz.
\newblock Multilingual sentiment datasets.
\newblock URL \url{https://github.com/tyqiangz/multilingual-sentiment-datasets}.

\bibitem[Fetahu et~al.(2023{\natexlab{a}})Fetahu, Chen, Kar, Rokhlenko, and Malmasi]{multiconer2-data}
Besnik Fetahu, Zhiyu Chen, Sudipta Kar, Oleg Rokhlenko, and Shervin Malmasi.
\newblock Multiconer v2: a large multilingual dataset for fine-grained and noisy named entity recognition.
\newblock \emph{arXiv preprint arXiv:2310.13213}, 2023{\natexlab{a}}.

\bibitem[Fetahu et~al.(2023{\natexlab{b}})Fetahu, Kar, Chen, Rokhlenko, and Malmasi]{multiconer2-report}
Besnik Fetahu, Sudipta Kar, Zhiyu Chen, Oleg Rokhlenko, and Shervin Malmasi.
\newblock {SemEval-2023 Task 2: Fine-grained Multilingual Named Entity Recognition (MultiCoNER 2)}.
\newblock In \emph{Proceedings of the 17th International Workshop on Semantic Evaluation (SemEval-2023)}. Association for Computational Linguistics, 2023{\natexlab{b}}.

\bibitem[Thakur et~al.(2023)Thakur, Bonifacio, Zhang, Ogundepo, Kamalloo, Alfonso-Hermelo, Li, Liu, Chen, Rezagholizadeh, and Lin]{thakur2023nomiracl}
Nandan Thakur, Luiz Bonifacio, Xinyu Zhang, Odunayo Ogundepo, Ehsan Kamalloo, David Alfonso-Hermelo, Xiaoguang Li, Qun Liu, Boxing Chen, Mehdi Rezagholizadeh, and Jimmy Lin.
\newblock Nomiracl: Knowing when you don't know for robust multilingual retrieval-augmented generation.
\newblock \emph{ArXiv}, abs/2312.11361, 2023.

\bibitem[Lison and Tiedemann(2016)]{lison2016opensubtitles2016}
Pierre Lison and J{\"o}rg Tiedemann.
\newblock Opensubtitles2016: Extracting large parallel corpora from movie and tv subtitles, 2016.

\bibitem[{Abadji} et~al.(2022){Abadji}, {Ortiz Suarez}, {Romary}, and {Sagot}]{OSCAR}
Julien {Abadji}, Pedro {Ortiz Suarez}, Laurent {Romary}, and Beno{\^\i}t {Sagot}.
\newblock {Towards a Cleaner Document-Oriented Multilingual Crawled Corpus}.
\newblock \emph{arXiv e-prints}, art. arXiv:2201.06642, January 2022.

\bibitem[Soares et~al.(2020)Soares, Stevenson, Bartolome, and Zaretskaya]{para_pat}
Felipe Soares, Mark Stevenson, Diego Bartolome, and Anna Zaretskaya.
\newblock {P}ara{P}at: The multi-million sentences parallel corpus of patents abstracts.
\newblock In \emph{Proceedings of The 12th Language Resources and Evaluation Conference}, pages 3769--3774, Marseille, France, May 2020. European Language Resources Association.
\newblock ISBN 979-10-95546-34-4.
\newblock URL \url{https://www.aclweb.org/anthology/2020.lrec-1.465}.

\bibitem[Gutenberg()]{project_gutenberg}
Project Gutenberg.
\newblock Project gutenberg.
\newblock URL \url{https://www.gutenberg.org/}.

\bibitem[RyokoAI()]{ShareGPT52K}
RyokoAI.
\newblock Sharegpt52k.
\newblock URL \url{https://huggingface.co/datasets/RyokoAI/ShareGPT52K}.

\bibitem[Huguet~Cabot et~al.(2023)Huguet~Cabot, Tedeschi, Ngonga~Ngomo, and Navigli]{SREDFM}
Pere-Llu{\'\i}s Huguet~Cabot, Simone Tedeschi, Axel-Cyrille Ngonga~Ngomo, and Roberto Navigli.
\newblock Red$^{\rm fm}$: a filtered and multilingual relation extraction dataset.
\newblock In \emph{Proc. of the 61st Annual Meeting of the Association for Computational Linguistics: ACL 2023}, Toronto, Canada, July 2023. Association for Computational Linguistics.
\newblock URL \url{https://arxiv.org/abs/2306.09802}.

\bibitem[Qi et~al.(2018)Qi, Sachan, Felix, Padmanabhan, and Neubig]{ted_talks}
Ye~Qi, Devendra Sachan, Matthieu Felix, Sarguna Padmanabhan, and Graham Neubig.
\newblock When and why are pre-trained word embeddings useful for neural machine translation?
\newblock In \emph{Proceedings of the 2018 Conference of the North {A}merican Chapter of the Association for Computational Linguistics: Human Language Technologies, Volume 2 (Short Papers)}, pages 529--535, New Orleans, Louisiana, June 2018. Association for Computational Linguistics.
\newblock \doi{10.18653/v1/N18-2084}.
\newblock URL \url{https://aclanthology.org/N18-2084}.

\bibitem[Cettolo et~al.(2012)Cettolo, Girardi, and Federico]{ted_talks_iwslt}
Mauro Cettolo, Christian Girardi, and Marcello Federico.
\newblock {WIT}3: Web inventory of transcribed and translated talks.
\newblock In \emph{Proceedings of the 16th Annual conference of the European Association for Machine Translation}, pages 261--268, Trento, Italy, May 28{--}30 2012. European Association for Machine Translation.
\newblock URL \url{https://www.aclweb.org/anthology/2012.eamt-1.60}.

\bibitem[FredZhang()]{toxi-text-3M}
FredZhang.
\newblock toxi-text-3m.
\newblock URL \url{https://huggingface.co/datasets/FredZhang7/toxi-text-3M}.

\bibitem[Nivre et~al.(2020)Nivre, de~Marneffe, Ginter, Haji{\v{c}}, Manning, Pyysalo, Schuster, Tyers, and Zeman]{universal_dependencies}
Joakim Nivre, Marie-Catherine de~Marneffe, Filip Ginter, Jan Haji{\v{c}}, Christopher~D. Manning, Sampo Pyysalo, Sebastian Schuster, Francis Tyers, and Daniel Zeman.
\newblock {U}niversal {D}ependencies v2: An evergrowing multilingual treebank collection.
\newblock In Nicoletta Calzolari, Fr{\'e}d{\'e}ric B{\'e}chet, Philippe Blache, Khalid Choukri, Christopher Cieri, Thierry Declerck, Sara Goggi, Hitoshi Isahara, Bente Maegaard, Joseph Mariani, H{\'e}l{\`e}ne Mazo, Asuncion Moreno, Jan Odijk, and Stelios Piperidis, editors, \emph{Proceedings of the Twelfth Language Resources and Evaluation Conference}, pages 4034--4043, Marseille, France, May 2020. European Language Resources Association.
\newblock ISBN 979-10-95546-34-4.
\newblock URL \url{https://aclanthology.org/2020.lrec-1.497}.

\bibitem[Pan et~al.(2017)Pan, Zhang, May, Nothman, Knight, and Ji]{wikiann1}
Xiaoman Pan, Boliang Zhang, Jonathan May, Joel Nothman, Kevin Knight, and Heng Ji.
\newblock Cross-lingual name tagging and linking for 282 languages.
\newblock In \emph{Proceedings of the 55th Annual Meeting of the Association for Computational Linguistics (Volume 1: Long Papers)}, pages 1946--1958, Vancouver, Canada, July 2017. Association for Computational Linguistics.
\newblock \doi{10.18653/v1/P17-1178}.
\newblock URL \url{https://www.aclweb.org/anthology/P17-1178}.

\bibitem[Rahimi et~al.(2019)Rahimi, Li, and Cohn]{wikiann2}
Afshin Rahimi, Yuan Li, and Trevor Cohn.
\newblock Massively multilingual transfer for {NER}.
\newblock In \emph{Proceedings of the 57th Annual Meeting of the Association for Computational Linguistics}, pages 151--164, Florence, Italy, July 2019. Association for Computational Linguistics.
\newblock URL \url{https://www.aclweb.org/anthology/P19-1015}.

\bibitem[Foundation()]{wikipedia}
Wikimedia Foundation.
\newblock Wikimedia downloads.
\newblock URL \url{https://dumps.wikimedia.org}.

\bibitem[Srinivasan et~al.(2021)Srinivasan, Raman, Chen, Bendersky, and Najork]{witbase}
Krishna Srinivasan, Karthik Raman, Jiecao Chen, Michael Bendersky, and Marc Najork.
\newblock Wit: Wikipedia-based image text dataset for multimodal multilingual machine learning.
\newblock \emph{arXiv preprint arXiv:2103.01913}, 2021.

\bibitem[Muennighoff et~al.(2022{\natexlab{b}})Muennighoff, Wang, Sutawika, Roberts, Biderman, Scao, Bari, Shen, Yong, Schoelkopf, et~al.]{xP3}
Niklas Muennighoff, Thomas Wang, Lintang Sutawika, Adam Roberts, Stella Biderman, Teven~Le Scao, M~Saiful Bari, Sheng Shen, Zheng-Xin Yong, Hailey Schoelkopf, et~al.
\newblock Crosslingual generalization through multitask finetuning.
\newblock \emph{arXiv preprint arXiv:2211.01786}, 2022{\natexlab{b}}.

\bibitem[Muhammad et~al.(2023)Muhammad, Abdulmumin, Ayele, Ousidhoum, Adelani, Yimam, Ahmad, Beloucif, Mohammad, Ruder, Hourrane, Brazdil, Ali, David, Osei, Bello, Ibrahim, Gwadabe, Rutunda, Belay, Messelle, Balcha, Chala, Gebremichael, Opoku, and Arthur]{muhammad2023afrisenti}
Shamsuddeen~Hassan Muhammad, Idris Abdulmumin, Abinew~Ali Ayele, Nedjma Ousidhoum, David~Ifeoluwa Adelani, Seid~Muhie Yimam, Ibrahim~Sa'id Ahmad, Meriem Beloucif, Saif~M. Mohammad, Sebastian Ruder, Oumaima Hourrane, Pavel Brazdil, Felermino Dário Mário~António Ali, Davis David, Salomey Osei, Bello~Shehu Bello, Falalu Ibrahim, Tajuddeen Gwadabe, Samuel Rutunda, Tadesse Belay, Wendimu~Baye Messelle, Hailu~Beshada Balcha, Sisay~Adugna Chala, Hagos~Tesfahun Gebremichael, Bernard Opoku, and Steven Arthur.
\newblock Afrisenti: A twitter sentiment analysis benchmark for african languages, 2023.

\bibitem[Razumovskaia et~al.(2024)Razumovskaia, Maynez, Louis, Lapata, and Narayan]{razumovskaia2024little}
Evgeniia Razumovskaia, Joshua Maynez, Annie Louis, Mirella Lapata, and Shashi Narayan.
\newblock Little red riding hood goes around the globe:crosslingual story planning and generation with large language models, 2024.

\bibitem[Bandarkar et~al.(2023)Bandarkar, Liang, Muller, Artetxe, Shukla, Husa, Goyal, Krishnan, Zettlemoyer, and Khabsa]{bandarkar2023belebele}
Lucas Bandarkar, Davis Liang, Benjamin Muller, Mikel Artetxe, Satya~Narayan Shukla, Donald Husa, Naman Goyal, Abhinandan Krishnan, Luke Zettlemoyer, and Madian Khabsa.
\newblock The belebele benchmark: a parallel reading comprehension dataset in 122 language variants.
\newblock \emph{arXiv preprint arXiv:2308.16884}, 2023.

\bibitem[BioMistral()]{BioInstructQA}
BioMistral.
\newblock Bioinstructqa.
\newblock URL \url{https://huggingface.co/datasets/BioMistral/BioInstructQA}.

\bibitem[mteb({\natexlab{b}})]{bucc-bitext-mining}
mteb.
\newblock Mteb benchmark, {\natexlab{b}}.
\newblock URL \url{https://huggingface.co/datasets/mteb/bucc-bitext-mining}.

\bibitem[Bhattacharjee et~al.(2023)Bhattacharjee, Hasan, Ahmad, Li, Kang, and Shahriyar]{bhattacharjee2023crosssum}
Abhik Bhattacharjee, Tahmid Hasan, Wasi~Uddin Ahmad, Yuan-Fang Li, Yong-Bin Kang, and Rifat Shahriyar.
\newblock Crosssum: Beyond english-centric cross-lingual summarization for 1,500+ language pairs, 2023.

\bibitem[Thapliyal et~al.(2022)Thapliyal, Pont-Tuset, Chen, and Soricut]{thapliyal2022crossmodal3600}
Ashish~V. Thapliyal, Jordi Pont-Tuset, Xi~Chen, and Radu Soricut.
\newblock Crossmodal-3600: A massively multilingual multimodal evaluation dataset, 2022.

\bibitem[Hardalov et~al.(2020)Hardalov, Mihaylov, Zlatkova, Dinkov, Koychev, and Nakov]{hardalov2020exams}
Momchil Hardalov, Todor Mihaylov, Dimitrina Zlatkova, Yoan Dinkov, Ivan Koychev, and Preslav Nakov.
\newblock Exams: A multi-subject high school examinations dataset for cross-lingual and multilingual question answering, 2020.

\bibitem[Chalkidis et~al.(2022{\natexlab{a}})Chalkidis, Pasini, Zhang, Tomada, Schwemer, and S{\o}gaard]{chalkidis2022fairlex}
Ilias Chalkidis, Tommaso Pasini, Sheng Zhang, Letizia Tomada, Sebastian Schwemer, and Anders S{\o}gaard.
\newblock Fairlex: A multilingual benchmark for evaluating fairness in legal text processing.
\newblock In \emph{Proceedings of the 60th Annual Meeting of the Association for Computational Linguistics (Volume 1: Long Papers)}, pages 4389--4406, 2022{\natexlab{a}}.

\bibitem[Team(2022)]{nllb2022}
NLLB Team.
\newblock No language left behind: Scaling human-centered machine translation.
\newblock \emph{arXiv preprint arXiv:2207.04672}, 2022.

\bibitem[Goyal et~al.(2022)Goyal, Gao, Chaudhary, Chen, Wenzek, Ju, Krishnan, Ranzato, Guzm{\'a}n, and Fan]{flores200_2}
Naman Goyal, Cynthia Gao, Vishrav Chaudhary, Peng-Jen Chen, Guillaume Wenzek, Da~Ju, Sanjana Krishnan, Marc’Aurelio Ranzato, Francisco Guzm{\'a}n, and Angela Fan.
\newblock The flores-101 evaluation benchmark for low-resource and multilingual machine translation.
\newblock \emph{Transactions of the Association for Computational Linguistics}, 10:\penalty0 522--538, 2022.

\bibitem[Guzm\'{a}n et~al.(2019)Guzm\'{a}n, Chen, Ott, Pino, Lample, Koehn, Chaudhary, and Ranzato]{flores200_3}
Francisco Guzm\'{a}n, Peng-Jen Chen, Myle Ott, Juan Pino, Guillaume Lample, Philipp Koehn, Vishrav Chaudhary, and Marc'Aurelio Ranzato.
\newblock Two new evaluation datasets for low-resource machine translation: Nepali-english and sinhala-english.
\newblock \emph{arXiv preprint arXiv:1902.01382}, 2019.

\bibitem[Yin et~al.(2022)Yin, Bansal, Monajatipoor, Li, and Chang]{yin2022geomlama}
Da~Yin, Hritik Bansal, Masoud Monajatipoor, Liunian~Harold Li, and Kai-Wei Chang.
\newblock Geomlama: Geo-diverse commonsense probing on multilingual pre-trained language models, 2022.

\bibitem[Peng et~al.(2024)Peng, Chai, and Li]{peng2024humaneval}
Qiwei Peng, Yekun Chai, and Xuhong Li.
\newblock Humaneval-xl: A multilingual code generation benchmark for cross-lingual natural language generalization.
\newblock \emph{arXiv preprint arXiv:2402.16694}, 2024.

\bibitem[Dac~Lai et~al.(2023)Dac~Lai, Van~Nguyen, Ngo, Nguyen, Dernoncourt, Rossi, and Nguyen]{dac2023okapi}
Viet Dac~Lai, Chien Van~Nguyen, Nghia~Trung Ngo, Thuat Nguyen, Franck Dernoncourt, Ryan~A Rossi, and Thien~Huu Nguyen.
\newblock Okapi: Instruction-tuned large language models in multiple languages with reinforcement learning from human feedback.
\newblock \emph{arXiv e-prints}, pages arXiv--2307, 2023.

\bibitem[Zhang et~al.(2023{\natexlab{f}})Zhang, Aljunied, Gao, Chia, and Bing]{zhang2023m3exam}
Wenxuan Zhang, Sharifah~Mahani Aljunied, Chang Gao, Yew~Ken Chia, and Lidong Bing.
\newblock M3exam: A multilingual, multimodal, multilevel benchmark for examining large language models, 2023{\natexlab{f}}.

\bibitem[Verma et~al.(2023)Verma, Jangra, Verma, and Saha]{verma-etal-2023-large}
Yash Verma, Anubhav Jangra, Raghvendra Verma, and Sriparna Saha.
\newblock Large scale multi-lingual multi-modal summarization dataset.
\newblock In Andreas Vlachos and Isabelle Augenstein, editors, \emph{Proceedings of the 17th Conference of the European Chapter of the Association for Computational Linguistics}, pages 3620--3632, Dubrovnik, Croatia, May 2023. Association for Computational Linguistics.
\newblock \doi{10.18653/v1/2023.eacl-main.263}.
\newblock URL \url{https://aclanthology.org/2023.eacl-main.263}.

\bibitem[Keung et~al.(2020{\natexlab{b}})Keung, Lu, Szarvas, and Smith]{keung2020multilingual}
Phillip Keung, Yichao Lu, György Szarvas, and Noah~A. Smith.
\newblock The multilingual amazon reviews corpus, 2020{\natexlab{b}}.

\bibitem[Adelani et~al.(2021)Adelani, Abbott, Neubig, D'souza, Kreutzer, Lignos, Palen-Michel, Buzaaba, Rijhwani, Ruder, Mayhew, Azime, Muhammad, Emezue, Nakatumba-Nabende, Ogayo, Aremu, Gitau, Mbaye, Alabi, Yimam, Gwadabe, Ezeani, Niyongabo, Mukiibi, Otiende, Orife, David, Ngom, Adewumi, Rayson, Adeyemi, Muriuki, Anebi, Chukwuneke, Odu, Wairagala, Oyerinde, Siro, Bateesa, Oloyede, Wambui, Akinode, Nabagereka, Katusiime, Awokoya, MBOUP, Gebreyohannes, Tilaye, Nwaike, Wolde, Faye, Sibanda, Ahia, Dossou, Ogueji, DIOP, Diallo, Akinfaderin, Marengereke, and Osei]{adelani2021masakhaner}
David~Ifeoluwa Adelani, Jade Abbott, Graham Neubig, Daniel D'souza, Julia Kreutzer, Constantine Lignos, Chester Palen-Michel, Happy Buzaaba, Shruti Rijhwani, Sebastian Ruder, Stephen Mayhew, Israel~Abebe Azime, Shamsuddeen Muhammad, Chris~Chinenye Emezue, Joyce Nakatumba-Nabende, Perez Ogayo, Anuoluwapo Aremu, Catherine Gitau, Derguene Mbaye, Jesujoba Alabi, Seid~Muhie Yimam, Tajuddeen Gwadabe, Ignatius Ezeani, Rubungo~Andre Niyongabo, Jonathan Mukiibi, Verrah Otiende, Iroro Orife, Davis David, Samba Ngom, Tosin Adewumi, Paul Rayson, Mofetoluwa Adeyemi, Gerald Muriuki, Emmanuel Anebi, Chiamaka Chukwuneke, Nkiruka Odu, Eric~Peter Wairagala, Samuel Oyerinde, Clemencia Siro, Tobius~Saul Bateesa, Temilola Oloyede, Yvonne Wambui, Victor Akinode, Deborah Nabagereka, Maurice Katusiime, Ayodele Awokoya, Mouhamadane MBOUP, Dibora Gebreyohannes, Henok Tilaye, Kelechi Nwaike, Degaga Wolde, Abdoulaye Faye, Blessing Sibanda, Orevaoghene Ahia, Bonaventure F.~P. Dossou, Kelechi Ogueji, Thierno~Ibrahima DIOP, Abdoulaye
  Diallo, Adewale Akinfaderin, Tendai Marengereke, and Salomey Osei.
\newblock Masakhaner: Named entity recognition for african languages, 2021.

\bibitem[Adelani et~al.(2023{\natexlab{a}})Adelani, Masiak, Azime, Alabi, Tonja, Mwase, Ogundepo, Dossou, Oladipo, Nixdorf, Emezue, sana~al azzawi, Sibanda, David, Ndolela, Mukiibi, Ajayi, Moteu, Odhiambo, Owodunni, Obiefuna, Mohamed, Muhammad, Ababu, Salahudeen, Yigezu, Gwadabe, Abdulmumin, Taye, Awoyomi, Shode, Adelani, Abdulganiyu, Omotayo, Adeeko, Afolabi, Aremu, Samuel, Siro, Kimotho, Ogbu, Mbonu, Chukwuneke, Fanijo, Ojo, Awosan, Kebede, Sakayo, Nyatsine, Sidume, Yousuf, Oduwole, Tshinu, Kimanuka, Diko, Nxakama, Nigusse, Johar, Mohamed, Hassan, Mehamed, Ngabire, Jules, Ssenkungu, and Stenetorp]{adelani2023masakhanews}
David~Ifeoluwa Adelani, Marek Masiak, Israel~Abebe Azime, Jesujoba Alabi, Atnafu~Lambebo Tonja, Christine Mwase, Odunayo Ogundepo, Bonaventure F.~P. Dossou, Akintunde Oladipo, Doreen Nixdorf, Chris~Chinenye Emezue, sana~al azzawi, Blessing Sibanda, Davis David, Lolwethu Ndolela, Jonathan Mukiibi, Tunde Ajayi, Tatiana Moteu, Brian Odhiambo, Abraham Owodunni, Nnaemeka Obiefuna, Muhidin Mohamed, Shamsuddeen~Hassan Muhammad, Teshome~Mulugeta Ababu, Saheed~Abdullahi Salahudeen, Mesay~Gemeda Yigezu, Tajuddeen Gwadabe, Idris Abdulmumin, Mahlet Taye, Oluwabusayo Awoyomi, Iyanuoluwa Shode, Tolulope Adelani, Habiba Abdulganiyu, Abdul-Hakeem Omotayo, Adetola Adeeko, Abeeb Afolabi, Anuoluwapo Aremu, Olanrewaju Samuel, Clemencia Siro, Wangari Kimotho, Onyekachi Ogbu, Chinedu Mbonu, Chiamaka Chukwuneke, Samuel Fanijo, Jessica Ojo, Oyinkansola Awosan, Tadesse Kebede, Toadoum~Sari Sakayo, Pamela Nyatsine, Freedmore Sidume, Oreen Yousuf, Mardiyyah Oduwole, Tshinu Tshinu, Ussen Kimanuka, Thina Diko, Siyanda Nxakama, Sinodos
  Nigusse, Abdulmejid Johar, Shafie Mohamed, Fuad~Mire Hassan, Moges~Ahmed Mehamed, Evrard Ngabire, Jules Jules, Ivan Ssenkungu, and Pontus Stenetorp.
\newblock Masakhanews: News topic classification for african languages, 2023{\natexlab{a}}.

\bibitem[FitzGerald et~al.(2022)FitzGerald, Hench, Peris, Mackie, Rottmann, Sanchez, Nash, Urbach, Kakarala, Singh, Ranganath, Crist, Britan, Leeuwis, Tur, and Natarajan]{fitzgerald2022massive}
Jack FitzGerald, Christopher Hench, Charith Peris, Scott Mackie, Kay Rottmann, Ana Sanchez, Aaron Nash, Liam Urbach, Vishesh Kakarala, Richa Singh, Swetha Ranganath, Laurie Crist, Misha Britan, Wouter Leeuwis, Gokhan Tur, and Prem Natarajan.
\newblock Massive: A 1m-example multilingual natural language understanding dataset with 51 typologically-diverse languages, 2022.

\bibitem[Changpinyo et~al.(2023)Changpinyo, Xue, Yarom, Thapliyal, Szpektor, Amelot, Chen, and Soricut]{changpinyo2023maxm}
Soravit Changpinyo, Linting Xue, Michal Yarom, Ashish~V. Thapliyal, Idan Szpektor, Julien Amelot, Xi~Chen, and Radu Soricut.
\newblock Maxm: Towards multilingual visual question answering, 2023.

\bibitem[Ahuja et~al.(2023)Ahuja, Diddee, Hada, Ochieng, Ramesh, Jain, Nambi, Ganu, Segal, Axmed, Bali, and Sitaram]{ahuja2023mega}
Kabir Ahuja, Harshita Diddee, Rishav Hada, Millicent Ochieng, Krithika Ramesh, Prachi Jain, Akshay Nambi, Tanuja Ganu, Sameer Segal, Maxamed Axmed, Kalika Bali, and Sunayana Sitaram.
\newblock Mega: Multilingual evaluation of generative ai, 2023.

\bibitem[Ahuja et~al.(2024)Ahuja, Aggarwal, Gumma, Watts, Sathe, Ochieng, Hada, Jain, Axmed, Bali, and Sitaram]{ahuja2024megaverse}
Sanchit Ahuja, Divyanshu Aggarwal, Varun Gumma, Ishaan Watts, Ashutosh Sathe, Millicent Ochieng, Rishav Hada, Prachi Jain, Maxamed Axmed, Kalika Bali, and Sunayana Sitaram.
\newblock Megaverse: Benchmarking large language models across languages, modalities, models and tasks, 2024.

\bibitem[Zhang et~al.(2024{\natexlab{c}})Zhang, Liu, Huang, Mao, Wang, and Hu]{zhang2024mela}
Ziyin Zhang, Yikang Liu, Weifang Huang, Junyu Mao, Rui Wang, and Hai Hu.
\newblock Mela: Multilingual evaluation of linguistic acceptability, 2024{\natexlab{c}}.

\bibitem[Lewis et~al.(2020{\natexlab{a}})Lewis, Oğuz, Rinott, Riedel, and Schwenk]{lewis2020mlqa}
Patrick Lewis, Barlas Oğuz, Ruty Rinott, Sebastian Riedel, and Holger Schwenk.
\newblock Mlqa: Evaluating cross-lingual extractive question answering, 2020{\natexlab{a}}.

\bibitem[Qiu et~al.(2024{\natexlab{c}})Qiu, Wu, Zhang, Lin, Wang, Zhang, Wang, and Xie]{qiu2024building}
Pengcheng Qiu, Chaoyi Wu, Xiaoman Zhang, Weixiong Lin, Haicheng Wang, Ya~Zhang, Yanfeng Wang, and Weidi Xie.
\newblock Towards building multilingual language model for medicine, 2024{\natexlab{c}}.

\bibitem[Malmasi et~al.(2022)Malmasi, Fang, Fetahu, Kar, and Rokhlenko]{malmasi2022multiconer}
Shervin Malmasi, Anjie Fang, Besnik Fetahu, Sudipta Kar, and Oleg Rokhlenko.
\newblock Multiconer: A large-scale multilingual dataset for complex named entity recognition, 2022.

\bibitem[Chalkidis et~al.(2021)Chalkidis, Fergadiotis, and Androutsopoulos]{chalkidis-etal-2021-multieurlex}
Ilias Chalkidis, Manos Fergadiotis, and Ion Androutsopoulos.
\newblock {M}ulti{EURLEX} - a multi-lingual and multi-label legal document classification dataset for zero-shot cross-lingual transfer.
\newblock In Marie-Francine Moens, Xuanjing Huang, Lucia Specia, and Scott Wen-tau Yih, editors, \emph{Proceedings of the 2021 Conference on Empirical Methods in Natural Language Processing}, pages 6974--6996, Online and Punta Cana, Dominican Republic, November 2021. Association for Computational Linguistics.
\newblock \doi{10.18653/v1/2021.emnlp-main.559}.
\newblock URL \url{https://aclanthology.org/2021.emnlp-main.559}.

\bibitem[Kabra et~al.(2023)Kabra, Liu, Khanuja, Aji, Winata, Cahyawijaya, Aremu, Ogayo, and Neubig]{kabra-etal-2023-multi}
Anubha Kabra, Emmy Liu, Simran Khanuja, Alham~Fikri Aji, Genta Winata, Samuel Cahyawijaya, Anuoluwapo Aremu, Perez Ogayo, and Graham Neubig.
\newblock Multi-lingual and multi-cultural figurative language understanding.
\newblock In Anna Rogers, Jordan Boyd-Graber, and Naoaki Okazaki, editors, \emph{Findings of the Association for Computational Linguistics: ACL 2023}, pages 8269--8284, Toronto, Canada, July 2023. Association for Computational Linguistics.
\newblock \doi{10.18653/v1/2023.findings-acl.525}.
\newblock URL \url{https://aclanthology.org/2023.findings-acl.525}.

\bibitem[Winata et~al.(2023{\natexlab{b}})Winata, Aji, Cahyawijaya, Mahendra, Koto, Romadhony, Kurniawan, Moeljadi, Prasojo, Fung, Baldwin, Lau, Sennrich, and Ruder]{winata2023nusax}
Genta~Indra Winata, Alham~Fikri Aji, Samuel Cahyawijaya, Rahmad Mahendra, Fajri Koto, Ade Romadhony, Kemal Kurniawan, David Moeljadi, Radityo~Eko Prasojo, Pascale Fung, Timothy Baldwin, Jey~Han Lau, Rico Sennrich, and Sebastian Ruder.
\newblock Nusax: Multilingual parallel sentiment dataset for 10 indonesian local languages, 2023{\natexlab{b}}.

\bibitem[Wang et~al.(2023{\natexlab{d}})Wang, Zhou, Fried, and Neubig]{wang2023executionbased}
Zhiruo Wang, Shuyan Zhou, Daniel Fried, and Graham Neubig.
\newblock Execution-based evaluation for open-domain code generation, 2023{\natexlab{d}}.

\bibitem[Zhang et~al.(2020)Zhang, Williams, Titov, and Sennrich]{zhang2020improving}
Biao Zhang, Philip Williams, Ivan Titov, and Rico Sennrich.
\newblock Improving massively multilingual neural machine translation and zero-shot translation.
\newblock In \emph{Proceedings of the 58th Annual Meeting of the Association for Computational Linguistics}, pages 1628--1639, 2020.

\bibitem[Yang et~al.(2019)Yang, Zhang, Tar, and Baldridge]{yang-etal-2019-paws}
Yinfei Yang, Yuan Zhang, Chris Tar, and Jason Baldridge.
\newblock {PAWS}-{X}: A cross-lingual adversarial dataset for paraphrase identification.
\newblock In Kentaro Inui, Jing Jiang, Vincent Ng, and Xiaojun Wan, editors, \emph{Proceedings of the 2019 Conference on Empirical Methods in Natural Language Processing and the 9th International Joint Conference on Natural Language Processing (EMNLP-IJCNLP)}, pages 3687--3692, Hong Kong, China, November 2019. Association for Computational Linguistics.
\newblock \doi{10.18653/v1/D19-1382}.
\newblock URL \url{https://aclanthology.org/D19-1382}.

\bibitem[Urlana et~al.(2023)Urlana, Chen, Zhao, Cohen, Shrivastava, and Haddow]{urlana2023pmindiasum}
Ashok Urlana, Pinzhen Chen, Zheng Zhao, Shay~B. Cohen, Manish Shrivastava, and Barry Haddow.
\newblock Pmindiasum: Multilingual and cross-lingual headline summarization for languages in india, 2023.

\bibitem[Goel et~al.(2023)Goel, Ammar, Gupta, Vashishtha, Sano, Surani, Chang, Choe, Greene, He, Nitisaroj, Trukhina, Paul, Shah, Shah, and Yu]{goel2023presto}
Rahul Goel, Waleed Ammar, Aditya Gupta, Siddharth Vashishtha, Motoki Sano, Faiz Surani, Max Chang, HyunJeong Choe, David Greene, Kyle He, Rattima Nitisaroj, Anna Trukhina, Shachi Paul, Pararth Shah, Rushin Shah, and Zhou Yu.
\newblock Presto: A multilingual dataset for parsing realistic task-oriented dialogs, 2023.

\bibitem[Clark et~al.(2023)Clark, Rijhwani, Gehrmann, Maynez, Aharoni, Nikolaev, Sellam, Siddhant, Das, and Parikh]{clark2023seahorse}
Elizabeth Clark, Shruti Rijhwani, Sebastian Gehrmann, Joshua Maynez, Roee Aharoni, Vitaly Nikolaev, Thibault Sellam, Aditya Siddhant, Dipanjan Das, and Ankur~P. Parikh.
\newblock Seahorse: A multilingual, multifaceted dataset for summarization evaluation, 2023.

\bibitem[Adelani et~al.(2023{\natexlab{b}})Adelani, Liu, Shen, Vassilyev, Alabi, Mao, Gao, and Lee]{adelani2023sib200}
David~Ifeoluwa Adelani, Hannah Liu, Xiaoyu Shen, Nikita Vassilyev, Jesujoba~O. Alabi, Yanke Mao, Haonan Gao, and Annie En-Shiun Lee.
\newblock Sib-200: A simple, inclusive, and big evaluation dataset for topic classification in 200+ languages and dialects, 2023{\natexlab{b}}.

\bibitem[Seganti et~al.(2021)Seganti, Firl{\k{a}}g, Skowronska, Sat{\l}awa, and Andruszkiewicz]{seganti-etal-2021-multilingual}
Alessandro Seganti, Klaudia Firl{\k{a}}g, Helena Skowronska, Micha{\l} Sat{\l}awa, and Piotr Andruszkiewicz.
\newblock Multilingual entity and relation extraction dataset and model.
\newblock In Paola Merlo, Jorg Tiedemann, and Reut Tsarfaty, editors, \emph{Proceedings of the 16th Conference of the European Chapter of the Association for Computational Linguistics: Main Volume}, pages 1946--1955, Online, April 2021. Association for Computational Linguistics.
\newblock \doi{10.18653/v1/2021.eacl-main.166}.
\newblock URL \url{https://aclanthology.org/2021.eacl-main.166}.

\bibitem[May(2021)]{huggingface:dataset:stsb_multi_mt}
Philip May.
\newblock Machine translated multilingual sts benchmark dataset., 2021.
\newblock URL \url{https://github.com/PhilipMay/stsb-multi-mt}.

\bibitem[Tiedemann(2020)]{tiedemann-2020-tatoeba}
J{\"o}rg Tiedemann.
\newblock The tatoeba translation challenge {--} realistic data sets for low resource and multilingual {MT}.
\newblock In \emph{Proceedings of the Fifth Conference on Machine Translation}, pages 1174--1182, Online, November 2020. Association for Computational Linguistics.
\newblock URL \url{https://aclanthology.org/2020.wmt-1.139}.

\bibitem[Srinivasan and Choi(2022)]{srinivasan-choi-2022-tydip}
Anirudh Srinivasan and Eunsol Choi.
\newblock {T}y{D}i{P}: A dataset for politeness classification in nine typologically diverse languages.
\newblock In Yoav Goldberg, Zornitsa Kozareva, and Yue Zhang, editors, \emph{Findings of the Association for Computational Linguistics: EMNLP 2022}, pages 5723--5738, Abu Dhabi, United Arab Emirates, December 2022. Association for Computational Linguistics.
\newblock \doi{10.18653/v1/2022.findings-emnlp.420}.
\newblock URL \url{https://aclanthology.org/2022.findings-emnlp.420}.

\bibitem[Clark et~al.(2020)Clark, Choi, Collins, Garrette, Kwiatkowski, Nikolaev, and Palomaki]{clark2020tydi}
Jonathan~H Clark, Eunsol Choi, Michael Collins, Dan Garrette, Tom Kwiatkowski, Vitaly Nikolaev, and Jennimaria Palomaki.
\newblock Tydi qa: A benchmark for information-seeking question answering in ty pologically di verse languages.
\newblock \emph{Transactions of the Association for Computational Linguistics}, 8:\penalty0 454--470, 2020.

\bibitem[Mittal et~al.(2023)Mittal, Sundriyal, and Nakov]{mittal-etal-2023-lost}
Shubham Mittal, Megha Sundriyal, and Preslav Nakov.
\newblock Lost in translation, found in spans: Identifying claims in multilingual social media.
\newblock In Houda Bouamor, Juan Pino, and Kalika Bali, editors, \emph{Proceedings of the 2023 Conference on Empirical Methods in Natural Language Processing}, pages 3887--3902, Singapore, December 2023. Association for Computational Linguistics.
\newblock \doi{10.18653/v1/2023.emnlp-main.236}.
\newblock URL \url{https://aclanthology.org/2023.emnlp-main.236}.

\bibitem[Moradshahi et~al.(2023)Moradshahi, Shen, Bali, Choudhury, de~Chalendar, Goel, Kim, Kodali, Kumaraguru, Semmar, Semnani, Seo, Seshadri, Shrivastava, Sun, Yadavalli, You, Xiong, and Lam]{moradshahi2023xrisawoz}
Mehrad Moradshahi, Tianhao Shen, Kalika Bali, Monojit Choudhury, Gaël de~Chalendar, Anmol Goel, Sungkyun Kim, Prashant Kodali, Ponnurangam Kumaraguru, Nasredine Semmar, Sina~J. Semnani, Jiwon Seo, Vivek Seshadri, Manish Shrivastava, Michael Sun, Aditya Yadavalli, Chaobin You, Deyi Xiong, and Monica~S. Lam.
\newblock X-risawoz: High-quality end-to-end multilingual dialogue datasets and few-shot agents, 2023.

\bibitem[Ponti et~al.(2020)Ponti, Glava{\v{s}}, Majewska, Liu, Vuli{\'c}, and Korhonen]{ponti-etal-2020-xcopa}
Edoardo~Maria Ponti, Goran Glava{\v{s}}, Olga Majewska, Qianchu Liu, Ivan Vuli{\'c}, and Anna Korhonen.
\newblock {XCOPA}: A multilingual dataset for causal commonsense reasoning.
\newblock In Bonnie Webber, Trevor Cohn, Yulan He, and Yang Liu, editors, \emph{Proceedings of the 2020 Conference on Empirical Methods in Natural Language Processing (EMNLP)}, pages 2362--2376, Online, November 2020. Association for Computational Linguistics.
\newblock \doi{10.18653/v1/2020.emnlp-main.185}.
\newblock URL \url{https://aclanthology.org/2020.emnlp-main.185}.

\bibitem[Lin et~al.(2021{\natexlab{a}})Lin, Lee, Qiao, and Ren]{lin2021common}
Bill~Yuchen Lin, Seyeon Lee, Xiaoyang Qiao, and Xiang Ren.
\newblock Common sense beyond english: Evaluating and improving multilingual language models for commonsense reasoning.
\newblock In \emph{Proceedings of the 59th Annual Meeting of the Association for Computational Linguistics and the 11th International Joint Conference on Natural Language Processing (Volume 1: Long Papers)}, pages 1274--1287, 2021{\natexlab{a}}.

\bibitem[Zhang et~al.(2023{\natexlab{g}})Zhang, D{'}Haro, Tang, Shi, Tang, and Li]{zhang-etal-2023-xdial}
Chen Zhang, Luis D{'}Haro, Chengguang Tang, Ke~Shi, Guohua Tang, and Haizhou Li.
\newblock x{D}ial-eval: A multilingual open-domain dialogue evaluation benchmark.
\newblock In Houda Bouamor, Juan Pino, and Kalika Bali, editors, \emph{Findings of the Association for Computational Linguistics: EMNLP 2023}, pages 5579--5601, Singapore, December 2023{\natexlab{g}}. Association for Computational Linguistics.
\newblock \doi{10.18653/v1/2023.findings-emnlp.371}.
\newblock URL \url{https://aclanthology.org/2023.findings-emnlp.371}.

\bibitem[Liang et~al.(2020)Liang, Duan, Gong, Wu, Guo, Qi, Gong, Shou, Jiang, Cao, Fan, Zhang, Agrawal, Cui, Wei, Bharti, Qiao, Chen, Wu, Liu, Yang, Campos, Majumder, and Zhou]{liang2020xglue}
Yaobo Liang, Nan Duan, Yeyun Gong, Ning Wu, Fenfei Guo, Weizhen Qi, Ming Gong, Linjun Shou, Daxin Jiang, Guihong Cao, Xiaodong Fan, Ruofei Zhang, Rahul Agrawal, Edward Cui, Sining Wei, Taroon Bharti, Ying Qiao, Jiun-Hung Chen, Winnie Wu, Shuguang Liu, Fan Yang, Daniel Campos, Rangan Majumder, and Ming Zhou.
\newblock Xglue: A new benchmark dataset for cross-lingual pre-training, understanding and generation, 2020.

\bibitem[Hasan et~al.(2021)Hasan, Bhattacharjee, Islam, Samin, Li, Kang, Rahman, and Shahriyar]{hasan2021xlsum}
Tahmid Hasan, Abhik Bhattacharjee, Md~Saiful Islam, Kazi Samin, Yuan-Fang Li, Yong-Bin Kang, M.~Sohel Rahman, and Rifat Shahriyar.
\newblock Xl-sum: Large-scale multilingual abstractive summarization for 44 languages, 2021.

\bibitem[Conneau et~al.(2018)Conneau, Rinott, Lample, Williams, Bowman, Schwenk, and Stoyanov]{conneau2018xnli}
Alexis Conneau, Ruty Rinott, Guillaume Lample, Adina Williams, Samuel Bowman, Holger Schwenk, and Veselin Stoyanov.
\newblock Xnli: Evaluating cross-lingual sentence representations.
\newblock In \emph{Proceedings of the 2018 Conference on Empirical Methods in Natural Language Processing}, pages 2475--2485, 2018.

\bibitem[Artetxe et~al.(2020)Artetxe, Ruder, and Yogatama]{artetxe2020cross}
Mikel Artetxe, Sebastian Ruder, and Dani Yogatama.
\newblock On the cross-lingual transferability of monolingual representations.
\newblock In \emph{Proceedings of the 58th Annual Meeting of the Association for Computational Linguistics}, pages 4623--4637, 2020.

\bibitem[Zhang et~al.(2023{\natexlab{h}})Zhang, Wang, Wang, and Zhang]{zhang2023xsemplr}
Yusen Zhang, Jun Wang, Zhiguo Wang, and Rui Zhang.
\newblock Xsemplr: Cross-lingual semantic parsing in multiple natural languages and meaning representations, 2023{\natexlab{h}}.

\bibitem[Hu et~al.(2020)Hu, Ruder, Siddhant, Neubig, Firat, and Johnson]{hu2020xtreme}
Junjie Hu, Sebastian Ruder, Aditya Siddhant, Graham Neubig, Orhan Firat, and Melvin Johnson.
\newblock Xtreme: A massively multilingual multi-task benchmark for evaluating cross-lingual generalization, 2020.

\bibitem[Ruder et~al.(2021{\natexlab{b}})Ruder, Constant, Botha, Siddhant, Firat, Fu, Liu, Hu, Garrette, Neubig, and Johnson]{ruder-etal-2021-xtreme}
Sebastian Ruder, Noah Constant, Jan Botha, Aditya Siddhant, Orhan Firat, Jinlan Fu, Pengfei Liu, Junjie Hu, Dan Garrette, Graham Neubig, and Melvin Johnson.
\newblock {XTREME}-{R}: Towards more challenging and nuanced multilingual evaluation.
\newblock In Marie-Francine Moens, Xuanjing Huang, Lucia Specia, and Scott Wen-tau Yih, editors, \emph{Proceedings of the 2021 Conference on Empirical Methods in Natural Language Processing}, pages 10215--10245, Online and Punta Cana, Dominican Republic, November 2021{\natexlab{b}}. Association for Computational Linguistics.
\newblock \doi{10.18653/v1/2021.emnlp-main.802}.
\newblock URL \url{https://aclanthology.org/2021.emnlp-main.802}.

\bibitem[Tikhonov and Ryabinin(2021)]{tikhonov-ryabinin-2021-heads}
Alexey Tikhonov and Max Ryabinin.
\newblock {I}t{'}s {A}ll in the {H}eads: {U}sing {A}ttention {H}eads as a {B}aseline for {C}ross-{L}ingual {T}ransfer in {C}ommonsense {R}easoning.
\newblock In Chengqing Zong, Fei Xia, Wenjie Li, and Roberto Navigli, editors, \emph{Findings of the Association for Computational Linguistics: ACL-IJCNLP 2021}, pages 3534--3546, Online, August 2021. Association for Computational Linguistics.
\newblock \doi{10.18653/v1/2021.findings-acl.310}.
\newblock URL \url{https://aclanthology.org/2021.findings-acl.310}.

\bibitem[Lin et~al.(2021{\natexlab{b}})Lin, Lee, Qiao, and Ren]{lin-etal-2021-common}
Bill~Yuchen Lin, Seyeon Lee, Xiaoyang Qiao, and Xiang Ren.
\newblock Common sense beyond {E}nglish: Evaluating and improving multilingual language models for commonsense reasoning.
\newblock In \emph{Proceedings of the 59th Annual Meeting of the Association for Computational Linguistics and the 11th International Joint Conference on Natural Language Processing (Volume 1: Long Papers)}, pages 1274--1287, Online, August 2021{\natexlab{b}}. Association for Computational Linguistics.
\newblock \doi{10.18653/v1/2021.acl-long.102}.
\newblock URL \url{https://aclanthology.org/2021.acl-long.102}.

\bibitem[Huang et~al.(2023{\natexlab{d}})Huang, Li, Liu, Sun, and Liu]{huang2023learn}
Kaiyu Huang, Peng Li, Junpeng Liu, Maosong Sun, and Yang Liu.
\newblock Learn and consolidate: Continual adaptation for zero-shot and multilingual neural machine translation.
\newblock In \emph{Proceedings of the 2023 Conference on Empirical Methods in Natural Language Processing}, pages 13938--13951, 2023{\natexlab{d}}.

\bibitem[Johnson et~al.(2017)Johnson, Schuster, Le, Krikun, Wu, Chen, Thorat, Vi{\'e}gas, Wattenberg, Corrado, et~al.]{johnson2017google}
Melvin Johnson, Mike Schuster, Quoc~V Le, Maxim Krikun, Yonghui Wu, Zhifeng Chen, Nikhil Thorat, Fernanda Vi{\'e}gas, Martin Wattenberg, Greg Corrado, et~al.
\newblock Google’s multilingual neural machine translation system: Enabling zero-shot translation.
\newblock \emph{Transactions of the Association for Computational Linguistics}, 5:\penalty0 339--351, 2017.

\bibitem[Huang et~al.(2023{\natexlab{e}})Huang, Li, Ma, Yao, and Liu]{huang2023knowledge}
Kaiyu Huang, Peng Li, Jin Ma, Ting Yao, and Yang Liu.
\newblock Knowledge transfer in incremental learning for multilingual neural machine translation.
\newblock In \emph{Proceedings of the 61st Annual Meeting of the Association for Computational Linguistics (Volume 1: Long Papers)}, pages 15286--15304, 2023{\natexlab{e}}.

\bibitem[Lin et~al.(2021{\natexlab{c}})Lin, Wu, Wang, and Li]{lin2021learning}
Zehui Lin, Liwei Wu, Mingxuan Wang, and Lei Li.
\newblock Learning language specific sub-network for multilingual machine translation.
\newblock In \emph{Proceedings of the 59th Annual Meeting of the Association for Computational Linguistics and the 11th International Joint Conference on Natural Language Processing (Volume 1: Long Papers)}, pages 293--305, 2021{\natexlab{c}}.

\bibitem[Escolano et~al.(2021)Escolano, Costa-Juss{\`a}, and Fonollosa]{escolano2021bilingual}
Carlos Escolano, Marta~R Costa-Juss{\`a}, and Jos{\'e}~AR Fonollosa.
\newblock From bilingual to multilingual neural-based machine translation by incremental training.
\newblock \emph{Journal of the Association for Information Science and Technology}, 72\penalty0 (2):\penalty0 190--203, 2021.

\bibitem[Min et~al.(2023)Min, Ross, Sulem, Veyseh, Nguyen, Sainz, Agirre, Heintz, and Roth]{min2023recent}
Bonan Min, Hayley Ross, Elior Sulem, Amir Pouran~Ben Veyseh, Thien~Huu Nguyen, Oscar Sainz, Eneko Agirre, Ilana Heintz, and Dan Roth.
\newblock Recent advances in natural language processing via large pre-trained language models: A survey.
\newblock \emph{ACM Computing Surveys}, 56\penalty0 (2):\penalty0 1--40, 2023.

\bibitem[Li et~al.(2020)Li, Zhou, He, Wang, Yang, and Li]{li2020sentence}
Bohan Li, Hao Zhou, Junxian He, Mingxuan Wang, Yiming Yang, and Lei Li.
\newblock On the sentence embeddings from pre-trained language models.
\newblock In \emph{Proceedings of the 2020 Conference on Empirical Methods in Natural Language Processing (EMNLP)}, pages 9119--9130, 2020.

\bibitem[Hu et~al.(2023)Hu, Liu, Zhao, Hou, Nie, and Li]{hu2023survey}
Linmei Hu, Zeyi Liu, Ziwang Zhao, Lei Hou, Liqiang Nie, and Juanzi Li.
\newblock A survey of knowledge enhanced pre-trained language models.
\newblock \emph{IEEE Transactions on Knowledge and Data Engineering}, 2023.

\bibitem[Devlin et~al.(2018)Devlin, Chang, Lee, and Toutanova]{devlin2018bert}
Jacob Devlin, Ming-Wei Chang, Kenton Lee, and Kristina Toutanova.
\newblock Bert: Pre-training of deep bidirectional transformers for language understanding.
\newblock \emph{arXiv preprint arXiv:1810.04805}, 2018.

\bibitem[Radford et~al.(2018)Radford, Narasimhan, Salimans, Sutskever, et~al.]{radford2018improving}
Alec Radford, Karthik Narasimhan, Tim Salimans, Ilya Sutskever, et~al.
\newblock Improving language understanding by generative pre-training, 2018.

\bibitem[Lewis et~al.(2020{\natexlab{b}})Lewis, Liu, Goyal, Ghazvininejad, Mohamed, Levy, Stoyanov, and Zettlemoyer]{lewis2020bart}
Mike Lewis, Yinhan Liu, Naman Goyal, Marjan Ghazvininejad, Abdelrahman Mohamed, Omer Levy, Veselin Stoyanov, and Luke Zettlemoyer.
\newblock Bart: Denoising sequence-to-sequence pre-training for natural language generation, translation, and comprehension.
\newblock In \emph{Proceedings of the 58th Annual Meeting of the Association for Computational Linguistics}, pages 7871--7880, 2020{\natexlab{b}}.

\bibitem[Li et~al.(2022{\natexlab{a}})Li, Tang, Zhao, Nie, and Wen]{li2022pretrained}
Junyi Li, Tianyi Tang, Wayne~Xin Zhao, Jian-Yun Nie, and Ji-Rong Wen.
\newblock Pretrained language models for text generation: A survey.
\newblock \emph{arXiv preprint arXiv:2201.05273}, 2022{\natexlab{a}}.

\bibitem[Radford et~al.(2019)Radford, Wu, Child, Luan, Amodei, Sutskever, et~al.]{radford2019language}
Alec Radford, Jeffrey Wu, Rewon Child, David Luan, Dario Amodei, Ilya Sutskever, et~al.
\newblock Language models are unsupervised multitask learners.
\newblock \emph{OpenAI blog}, 1\penalty0 (8):\penalty0 9, 2019.

\bibitem[Kaplan et~al.(2020)Kaplan, McCandlish, Henighan, Brown, Chess, Child, Gray, Radford, Wu, and Amodei]{kaplan2020scaling}
Jared Kaplan, Sam McCandlish, Tom Henighan, Tom~B Brown, Benjamin Chess, Rewon Child, Scott Gray, Alec Radford, Jeffrey Wu, and Dario Amodei.
\newblock Scaling laws for neural language models.
\newblock \emph{arXiv preprint arXiv:2001.08361}, 2020.

\bibitem[Fedus et~al.(2022)Fedus, Zoph, and Shazeer]{fedus2022switch}
William Fedus, Barret Zoph, and Noam Shazeer.
\newblock Switch transformers: Scaling to trillion parameter models with simple and efficient sparsity.
\newblock \emph{Journal of Machine Learning Research}, 23\penalty0 (120):\penalty0 1--39, 2022.

\bibitem[Conneau and Lample(2019)]{conneau2019cross}
Alexis Conneau and Guillaume Lample.
\newblock Cross-lingual language model pretraining.
\newblock \emph{Advances in neural information processing systems}, 32, 2019.

\bibitem[Aharoni et~al.(2019)Aharoni, Johnson, and Firat]{aharoni2019massively}
Roee Aharoni, Melvin Johnson, and Orhan Firat.
\newblock Massively multilingual neural machine translation.
\newblock In \emph{Proceedings of the 2019 Conference of the North American Chapter of the Association for Computational Linguistics: Human Language Technologies, Volume 1 (Long and Short Papers)}, pages 3874--3884, 2019.

\bibitem[Firat et~al.(2017)Firat, Cho, Sankaran, Yarman~Vural, and Bengio]{firat2017multi}
Orhan Firat, Kyunghyun Cho, Baskaran Sankaran, Fatos~T Yarman~Vural, and Yoshua Bengio.
\newblock Multi-way, multilingual neural machine translation.
\newblock \emph{Computer Speech and Language}, 45\penalty0 (C):\penalty0 236--252, 2017.

\bibitem[Loginova et~al.(2018)Loginova, Varanasi, and Neumann]{loginova2018towards}
Ekaterina Loginova, Stalin Varanasi, and G{\"u}nter Neumann.
\newblock Towards multilingual neural question answering.
\newblock In \emph{New Trends in Databases and Information Systems: ADBIS 2018 Short Papers and Workshops, AI* QA, BIGPMED, CSACDB, M2U, BigDataMAPS, ISTREND, DC, Budapest, Hungary, September, 2-5, 2018, Proceedings 22}, pages 274--285. Springer, 2018.

\bibitem[Loginova et~al.(2021)Loginova, Varanasi, and Neumann]{loginova2021towards}
Ekaterina Loginova, Stalin Varanasi, and G{\"u}nter Neumann.
\newblock Towards end-to-end multilingual question answering.
\newblock \emph{Information Systems Frontiers}, 23\penalty0 (1):\penalty0 227--241, 2021.

\bibitem[Ruder and Sil(2021)]{ruder2021multi}
Sebastian Ruder and Avirup Sil.
\newblock Multi-domain multilingual question answering.
\newblock In \emph{Proceedings of the 2021 Conference on Empirical Methods in Natural Language Processing: Tutorial Abstracts}, pages 17--21, 2021.

\bibitem[Fang et~al.(2022)Fang, Wang, Xu, Xu, Sun, Zhu, and Zeng]{fang2022leveraging}
Yuwei Fang, Shuohang Wang, Yichong Xu, Ruochen Xu, Siqi Sun, Chenguang Zhu, and Michael Zeng.
\newblock Leveraging knowledge in multilingual commonsense reasoning.
\newblock In \emph{Findings of the Association for Computational Linguistics: ACL 2022}, pages 3237--3246, 2022.

\bibitem[Bang et~al.(2023)Bang, Cahyawijaya, Lee, Dai, Su, Wilie, Lovenia, Ji, Yu, Chung, et~al.]{bang2023multitask}
Yejin Bang, Samuel Cahyawijaya, Nayeon Lee, Wenliang Dai, Dan Su, Bryan Wilie, Holy Lovenia, Ziwei Ji, Tiezheng Yu, Willy Chung, et~al.
\newblock A multitask, multilingual, multimodal evaluation of chatgpt on reasoning, hallucination, and interactivity.
\newblock In \emph{Proceedings of the 13th International Joint Conference on Natural Language Processing and the 3rd Conference of the Asia-Pacific Chapter of the Association for Computational Linguistics (Volume 1: Long Papers)}, pages 675--718, 2023.

\bibitem[Yang et~al.(2023{\natexlab{c}})Yang, Li, Lin, Wang, Lin, Liu, and Wang]{yang2023dawn}
Zhengyuan Yang, Linjie Li, Kevin Lin, Jianfeng Wang, Chung-Ching Lin, Zicheng Liu, and Lijuan Wang.
\newblock The dawn of lmms: Preliminary explorations with gpt-4v (ision).
\newblock \emph{arXiv preprint arXiv:2309.17421}, 9\penalty0 (1):\penalty0 1, 2023{\natexlab{c}}.

\bibitem[Muennighoff et~al.(2024{\natexlab{b}})Muennighoff, Rush, Barak, Le~Scao, Tazi, Piktus, Pyysalo, Wolf, and Raffel]{muennighoff2024scaling}
Niklas Muennighoff, Alexander Rush, Boaz Barak, Teven Le~Scao, Nouamane Tazi, Aleksandra Piktus, Sampo Pyysalo, Thomas Wolf, and Colin~A Raffel.
\newblock Scaling data-constrained language models.
\newblock \emph{Advances in Neural Information Processing Systems}, 36, 2024{\natexlab{b}}.

\bibitem[Aghajanyan et~al.(2023)Aghajanyan, Yu, Conneau, Hsu, Hambardzumyan, Zhang, Roller, Goyal, Levy, and Zettlemoyer]{aghajanyan2023scaling}
Armen Aghajanyan, Lili Yu, Alexis Conneau, Wei-Ning Hsu, Karen Hambardzumyan, Susan Zhang, Stephen Roller, Naman Goyal, Omer Levy, and Luke Zettlemoyer.
\newblock Scaling laws for generative mixed-modal language models.
\newblock In \emph{International Conference on Machine Learning}, pages 265--279. PMLR, 2023.

\bibitem[Zhao et~al.(2023)Zhao, Zhou, Li, Tang, Wang, Hou, Min, Zhang, Zhang, Dong, et~al.]{zhao2023survey}
Wayne~Xin Zhao, Kun Zhou, Junyi Li, Tianyi Tang, Xiaolei Wang, Yupeng Hou, Yingqian Min, Beichen Zhang, Junjie Zhang, Zican Dong, et~al.
\newblock A survey of large language models.
\newblock \emph{arXiv preprint arXiv:2303.18223}, 2023.

\bibitem[Ouyang et~al.(2020)Ouyang, Wang, Pang, Sun, Tian, Wu, and Wang]{ouyang2020ernie}
Xuan Ouyang, Shuohuan Wang, Chao Pang, Yu~Sun, Hao Tian, Hua Wu, and Haifeng Wang.
\newblock Ernie-m: Enhanced multilingual representation by aligning cross-lingual semantics with monolingual corpora.
\newblock \emph{arXiv preprint arXiv:2012.15674}, 2020.

\bibitem[Uthus et~al.(2023)Uthus, Onta{\~n}{\'o}n, Ainslie, and Guo]{uthus2023mlongt5}
David Uthus, Santiago Onta{\~n}{\'o}n, Joshua Ainslie, and Mandy Guo.
\newblock mlongt5: A multilingual and efficient text-to-text transformer for longer sequences.
\newblock \emph{arXiv preprint arXiv:2305.11129}, 2023.

\bibitem[Conneau et~al.(2019)Conneau, Khandelwal, Goyal, Chaudhary, Wenzek, Guzm{\'a}n, Grave, Ott, Zettlemoyer, and Stoyanov]{conneau2019unsupervised}
Alexis Conneau, Kartikay Khandelwal, Naman Goyal, Vishrav Chaudhary, Guillaume Wenzek, Francisco Guzm{\'a}n, Edouard Grave, Myle Ott, Luke Zettlemoyer, and Veselin Stoyanov.
\newblock Unsupervised cross-lingual representation learning at scale.
\newblock \emph{arXiv preprint arXiv:1911.02116}, 2019.

\bibitem[Liang et~al.(2023)Liang, Gonen, Mao, Hou, Goyal, Ghazvininejad, Zettlemoyer, and Khabsa]{liang2023xlm}
Davis Liang, Hila Gonen, Yuning Mao, Rui Hou, Naman Goyal, Marjan Ghazvininejad, Luke Zettlemoyer, and Madian Khabsa.
\newblock Xlm-v: Overcoming the vocabulary bottleneck in multilingual masked language models.
\newblock \emph{arXiv preprint arXiv:2301.10472}, 2023.

\bibitem[Chung et~al.(2023)Chung, Constant, Garcia, Roberts, Tay, Narang, and Firat]{chung2023unimax}
Hyung~Won Chung, Noah Constant, Xavier Garcia, Adam Roberts, Yi~Tay, Sharan Narang, and Orhan Firat.
\newblock Unimax: Fairer and more effective language sampling for large-scale multilingual pretraining.
\newblock \emph{arXiv preprint arXiv:2304.09151}, 2023.

\bibitem[Dabre et~al.(2020)Dabre, Chu, and Kunchukuttan]{dabre2020survey}
Raj Dabre, Chenhui Chu, and Anoop Kunchukuttan.
\newblock A survey of multilingual neural machine translation.
\newblock \emph{ACM Computing Surveys (CSUR)}, 53\penalty0 (5):\penalty0 1--38, 2020.

\bibitem[Garcia et~al.(2021)Garcia, Constant, Parikh, and Firat]{garcia2021towards}
Xavier Garcia, Noah Constant, Ankur Parikh, and Orhan Firat.
\newblock Towards continual learning for multilingual machine translation via vocabulary substitution.
\newblock In \emph{Proceedings of the 2021 Conference of the North American Chapter of the Association for Computational Linguistics: Human Language Technologies}, pages 1184--1192, 2021.

\bibitem[Huang et~al.(2022)Huang, Li, Ma, and Liu]{huang2022entropy}
Kaiyu Huang, Peng Li, Jin Ma, and Yang Liu.
\newblock Entropy-based vocabulary substitution for incremental learning in multilingual neural machine translation.
\newblock In \emph{Proceedings of the 2022 Conference on Empirical Methods in Natural Language Processing}, pages 10537--10550, 2022.

\bibitem[Zhang et~al.(2022{\natexlab{c}})Zhang, Chaudhary, Goyal, Cross, Wenzek, Bansal, and Guzman]{zhang2022robust}
Shiyue Zhang, Vishrav Chaudhary, Naman Goyal, James Cross, Guillaume Wenzek, Mohit Bansal, and Francisco Guzman.
\newblock How robust is neural machine translation to language imbalance in multilingual tokenizer training?
\newblock \emph{arXiv preprint arXiv:2204.14268}, 2022{\natexlab{c}}.

\bibitem[Cui et~al.(2023{\natexlab{a}})Cui, Yang, and Yao]{cui2023efficient}
Yiming Cui, Ziqing Yang, and Xin Yao.
\newblock Efficient and effective text encoding for chinese llama and alpaca.
\newblock \emph{arXiv preprint arXiv:2304.08177}, 2023{\natexlab{a}}.

\bibitem[HIT-SCIR(2024)]{Chinese-Mixtral-8x7B}
HIT-SCIR.
\newblock Chinese-mixtral-8x7b: An open-source mixture-of-experts llm.
\newblock \url{https://github.com/HIT-SCIR/Chinese-Mixtral-8x7B}, 2024.

\bibitem[Zhang et~al.(2021{\natexlab{a}})Zhang, Gu, Han, Chen, Xiao, Sun, Yao, Qi, Guan, Ke, et~al.]{zhang2021cpm}
Zhengyan Zhang, Yuxian Gu, Xu~Han, Shengqi Chen, Chaojun Xiao, Zhenbo Sun, Yuan Yao, Fanchao Qi, Jian Guan, Pei Ke, et~al.
\newblock Cpm-2: Large-scale cost-effective pre-trained language models.
\newblock \emph{AI Open}, 2:\penalty0 216--224, 2021{\natexlab{a}}.

\bibitem[Basile et~al.(2023)Basile, Musacchio, Polignano, Siciliani, Fiameni, and Semeraro]{basile2023llamantino}
Pierpaolo Basile, Elio Musacchio, Marco Polignano, Lucia Siciliani, Giuseppe Fiameni, and Giovanni Semeraro.
\newblock Llamantino: Llama 2 models for effective text generation in italian language.
\newblock \emph{arXiv preprint arXiv:2312.09993}, 2023.

\bibitem[Luukkonen et~al.(2023)Luukkonen, Komulainen, Luoma, Eskelinen, Kanerva, Kupari, Ginter, Laippala, Muennighoff, Piktus, et~al.]{luukkonen2023fingpt}
Risto Luukkonen, Ville Komulainen, Jouni Luoma, Anni Eskelinen, Jenna Kanerva, Hanna-Mari~Kristiina Kupari, Filip Ginter, Veronika Laippala, Niklas Muennighoff, Aleksandra Piktus, et~al.
\newblock Fingpt: Large generative models for a small language.
\newblock In \emph{The 2023 Conference on Empirical Methods in Natural Language Processing}, 2023.

\bibitem[Pires et~al.(2023)Pires, Abonizio, Almeida, and Nogueira]{pires2023sabia}
Ramon Pires, Hugo Abonizio, Thales~Sales Almeida, and Rodrigo Nogueira.
\newblock Sabi{\'a}: Portuguese large language models.
\newblock In \emph{Brazilian Conference on Intelligent Systems}, pages 226--240. Springer, 2023.

\bibitem[Garcia et~al.(2024)Garcia, Paiola, Morelli, Candido, J{\'u}nior, Jodas, Afonso, Guilherme, Penteado, and Papa]{garcia2024introducing}
Gabriel~Lino Garcia, Pedro~Henrique Paiola, Luis~Henrique Morelli, Giovani Candido, Arnaldo~C{\^a}ndido J{\'u}nior, Danilo~Samuel Jodas, Luis Afonso, Ivan~Rizzo Guilherme, Bruno~Elias Penteado, and Jo{\~a}o~Paulo Papa.
\newblock Introducing bode: A fine-tuned large language model for portuguese prompt-based task.
\newblock \emph{arXiv preprint arXiv:2401.02909}, 2024.

\bibitem[ImaniGooghari et~al.(2023)ImaniGooghari, Lin, Kargaran, Severini, Sabet, Kassner, Ma, Schmid, Martins, Yvon, et~al.]{imanigooghari2023glot500}
Ayyoob ImaniGooghari, Peiqin Lin, Amir~Hossein Kargaran, Silvia Severini, Masoud~Jalili Sabet, Nora Kassner, Chunlan Ma, Helmut Schmid, Andr{\'e}~FT Martins, Fran{\c{c}}ois Yvon, et~al.
\newblock Glot500: Scaling multilingual corpora and language models to 500 languages.
\newblock \emph{arXiv preprint arXiv:2305.12182}, 2023.

\bibitem[Ebrahimi and Kann(2021)]{ebrahimi2021adapt}
Abteen Ebrahimi and Katharina Kann.
\newblock How to adapt your pretrained multilingual model to 1600 languages.
\newblock \emph{arXiv preprint arXiv:2106.02124}, 2021.

\bibitem[Alabi et~al.(2022)Alabi, Adelani, Mosbach, and Klakow]{alabi2022adapting}
Jesujoba~O Alabi, David~Ifeoluwa Adelani, Marius Mosbach, and Dietrich Klakow.
\newblock Adapting pre-trained language models to african languages via multilingual adaptive fine-tuning.
\newblock \emph{arXiv preprint arXiv:2204.06487}, 2022.

\bibitem[Muennighoff et~al.(2022{\natexlab{c}})Muennighoff, Wang, Sutawika, Roberts, Biderman, Scao, Bari, Shen, Yong, Schoelkopf, et~al.]{muennighoff2022crosslingual}
Niklas Muennighoff, Thomas Wang, Lintang Sutawika, Adam Roberts, Stella Biderman, Teven~Le Scao, M~Saiful Bari, Sheng Shen, Zheng-Xin Yong, Hailey Schoelkopf, et~al.
\newblock Crosslingual generalization through multitask finetuning.
\newblock \emph{arXiv preprint arXiv:2211.01786}, 2022{\natexlab{c}}.

\bibitem[Wang et~al.(2022{\natexlab{a}})Wang, Ruder, and Neubig]{wang2022expanding}
Xinyi Wang, Sebastian Ruder, and Graham Neubig.
\newblock Expanding pretrained models to thousands more languages via lexicon-based adaptation.
\newblock \emph{arXiv preprint arXiv:2203.09435}, 2022{\natexlab{a}}.

\bibitem[Rajpurkar et~al.(2016)Rajpurkar, Zhang, Lopyrev, and Liang]{rajpurkar2016squad}
Pranav Rajpurkar, Jian Zhang, Konstantin Lopyrev, and Percy Liang.
\newblock Squad: 100,000+ questions for machine comprehension of text.
\newblock In \emph{Proceedings of the 2016 Conference on Empirical Methods in Natural Language Processing}, pages 2383--2392, 2016.

\bibitem[Yin et~al.(2021)Yin, Radev, and Xiong]{yin2021docnli}
Wenpeng Yin, Dragomir Radev, and Caiming Xiong.
\newblock Docnli: A large-scale dataset for document-level natural language inference.
\newblock In \emph{Findings of the Association for Computational Linguistics: ACL-IJCNLP 2021}, pages 4913--4922, 2021.

\bibitem[Arivazhagan et~al.(2019)Arivazhagan, Bapna, Firat, Lepikhin, Johnson, Krikun, Chen, Cao, Foster, Cherry, et~al.]{arivazhagan2019massively}
Naveen Arivazhagan, Ankur Bapna, Orhan Firat, Dmitry Lepikhin, Melvin Johnson, Maxim Krikun, Mia~Xu Chen, Yuan Cao, George Foster, Colin Cherry, et~al.
\newblock Massively multilingual neural machine translation in the wild: Findings and challenges.
\newblock \emph{arXiv preprint arXiv:1907.05019}, 2019.

\bibitem[Wang et~al.(2020)Wang, Tsvetkov, and Neubig]{wang2020balancing}
Xinyi Wang, Yulia Tsvetkov, and Graham Neubig.
\newblock Balancing training for multilingual neural machine translation.
\newblock In \emph{Proceedings of the 58th Annual Meeting of the Association for Computational Linguistics}, pages 8526--8537, 2020.

\bibitem[Wu et~al.(2021)Wu, Li, Zhang, Li, Haffari, and Liu]{wu2021uncertainty}
Minghao Wu, Yitong Li, Meng Zhang, Liangyou Li, Gholamreza Haffari, and Qun Liu.
\newblock Uncertainty-aware balancing for multilingual and multi-domain neural machine translation training.
\newblock In \emph{Proceedings of the 2021 Conference on Empirical Methods in Natural Language Processing}, pages 7291--7305, 2021.

\bibitem[Yang et~al.(2021)Yang, Yin, Ma, Huang, Zhang, Li, and Wei]{yang2021multilingual}
Jian Yang, Yuwei Yin, Shuming Ma, Haoyang Huang, Dongdong Zhang, Zhoujun Li, and Furu Wei.
\newblock Multilingual agreement for multilingual neural machine translation.
\newblock In \emph{Proceedings of the 59th Annual Meeting of the Association for Computational Linguistics and the 11th International Joint Conference on Natural Language Processing (Volume 2: Short Papers)}, pages 233--239, 2021.

\bibitem[Pan et~al.(2021)Pan, Wang, Wu, and Li]{pan2021contrastive}
Xiao Pan, Mingxuan Wang, Liwei Wu, and Lei Li.
\newblock Contrastive learning for many-to-many multilingual neural machine translation.
\newblock In \emph{Proceedings of the 59th Annual Meeting of the Association for Computational Linguistics and the 11th International Joint Conference on Natural Language Processing (Volume 1: Long Papers)}, pages 244--258, 2021.

\bibitem[Oncevay et~al.(2020)Oncevay, Haddow, and Birch]{oncevay2020bridging}
Arturo Oncevay, Barry Haddow, and Alexandra Birch.
\newblock Bridging linguistic typology and multilingual machine translation with multi-view language representations.
\newblock In \emph{Proceedings of the 2020 Conference on Empirical Methods in Natural Language Processing (EMNLP)}, pages 2391--2406, 2020.

\bibitem[Cheng et~al.(2022)Cheng, Bapna, Firat, Cao, Wang, and Macherey]{cheng2022multilingual}
Yong Cheng, Ankur Bapna, Orhan Firat, Yuan Cao, Pidong Wang, and Wolfgang Macherey.
\newblock Multilingual mix: Example interpolation improves multilingual neural machine translation.
\newblock In \emph{Proceedings of the 60th Annual Meeting of the Association for Computational Linguistics (Volume 1: Long Papers)}, pages 4092--4102, 2022.

\bibitem[Stap et~al.(2023)Stap, Niculae, and Monz]{stap2023viewing}
David Stap, Vlad Niculae, and Christof Monz.
\newblock Viewing knowledge transfer in multilingual machine translation through a representational lens.
\newblock In \emph{Findings of the Association for Computational Linguistics: EMNLP 2023}, pages 14973--14987, 2023.

\bibitem[Wang and Zhang(2022)]{wang2022parameter}
Qian Wang and Jiajun Zhang.
\newblock Parameter differentiation based multilingual neural machine translation.
\newblock In \emph{Proceedings of the AAAI Conference on Artificial Intelligence}, volume~36, pages 11440--11448, 2022.

\bibitem[Beltagy et~al.(2020)Beltagy, Peters, and Cohan]{beltagy2020longformer}
Iz~Beltagy, Matthew~E Peters, and Arman Cohan.
\newblock Longformer: The long-document transformer.
\newblock \emph{arXiv preprint arXiv:2004.05150}, 2020.

\bibitem[Su et~al.(2024)Su, Ahmed, Lu, Pan, Bo, and Liu]{su2024roformer}
Jianlin Su, Murtadha Ahmed, Yu~Lu, Shengfeng Pan, Wen Bo, and Yunfeng Liu.
\newblock Roformer: Enhanced transformer with rotary position embedding.
\newblock \emph{Neurocomputing}, 568:\penalty0 127063, 2024.

\bibitem[Peng et~al.(2023{\natexlab{b}})Peng, Wang, Dong, Hao, Huang, Ma, and Wei]{peng2023kosmos}
Zhiliang Peng, Wenhui Wang, Li~Dong, Yaru Hao, Shaohan Huang, Shuming Ma, and Furu Wei.
\newblock Kosmos-2: Grounding multimodal large language models to the world.
\newblock \emph{arXiv preprint arXiv:2306.14824}, 2023{\natexlab{b}}.

\bibitem[Liu et~al.(2024{\natexlab{b}})Liu, Li, Wu, and Lee]{liu2024visual}
Haotian Liu, Chunyuan Li, Qingyang Wu, and Yong~Jae Lee.
\newblock Visual instruction tuning.
\newblock \emph{Advances in neural information processing systems}, 36, 2024{\natexlab{b}}.

\bibitem[Gu and Dao(2023)]{gu2023mamba}
Albert Gu and Tri Dao.
\newblock Mamba: Linear-time sequence modeling with selective state spaces.
\newblock \emph{arXiv preprint arXiv:2312.00752}, 2023.

\bibitem[Qin et~al.(2023)Qin, Chen, Wei, Huang, and Che]{qin2023cross}
Libo Qin, Qiguang Chen, Fuxuan Wei, Shijue Huang, and Wanxiang Che.
\newblock Cross-lingual prompting: Improving zero-shot chain-of-thought reasoning across languages.
\newblock In \emph{Proceedings of the 2023 Conference on Empirical Methods in Natural Language Processing}, pages 2695--2709, 2023.

\bibitem[Etxaniz et~al.(2023)Etxaniz, Azkune, Soroa, de~Lacalle, and Artetxe]{etxaniz2023multilingual}
Julen Etxaniz, Gorka Azkune, Aitor Soroa, Oier~Lopez de~Lacalle, and Mikel Artetxe.
\newblock Do multilingual language models think better in english?
\newblock \emph{arXiv preprint arXiv:2308.01223}, 2023.

\bibitem[Sap et~al.(2020)Sap, Shwartz, Bosselut, Choi, and Roth]{sap2020commonsense}
Maarten Sap, Vered Shwartz, Antoine Bosselut, Yejin Choi, and Dan Roth.
\newblock Commonsense reasoning for natural language processing.
\newblock In \emph{Proceedings of the 58th Annual Meeting of the Association for Computational Linguistics: Tutorial Abstracts}, pages 27--33, 2020.

\bibitem[Yu et~al.(2023{\natexlab{a}})Yu, Zhang, and Wang]{yu2023nature}
Fei Yu, Hongbo Zhang, and Benyou Wang.
\newblock Nature language reasoning, a survey.
\newblock \emph{arXiv preprint arXiv:2303.14725}, 2023{\natexlab{a}}.

\bibitem[Liu et~al.(2023{\natexlab{d}})Liu, Liu, Cui, Teng, Duan, Zhou, and Zhang]{liu2023logiqa}
Hanmeng Liu, Jian Liu, Leyang Cui, Zhiyang Teng, Nan Duan, Ming Zhou, and Yue Zhang.
\newblock Logiqa 2.0—an improved dataset for logical reasoning in natural language understanding.
\newblock \emph{IEEE/ACM Transactions on Audio, Speech, and Language Processing}, 2023{\natexlab{d}}.

\bibitem[Wang et~al.(2022{\natexlab{b}})Wang, Wei, Schuurmans, Le, Chi, Narang, Chowdhery, and Zhou]{wang2022self}
Xuezhi Wang, Jason Wei, Dale Schuurmans, Quoc~V Le, Ed~H Chi, Sharan Narang, Aakanksha Chowdhery, and Denny Zhou.
\newblock Self-consistency improves chain of thought reasoning in language models.
\newblock In \emph{The Eleventh International Conference on Learning Representations}, 2022{\natexlab{b}}.

\bibitem[Lyu et~al.(2023)Lyu, Havaldar, Stein, Zhang, Rao, Wong, Apidianaki, and Callison-Burch]{lyu2023faithful}
Qing Lyu, Shreya Havaldar, Adam Stein, Li~Zhang, Delip Rao, Eric Wong, Marianna Apidianaki, and Chris Callison-Burch.
\newblock Faithful chain-of-thought reasoning.
\newblock In \emph{Proceedings of the 13th International Joint Conference on Natural Language Processing and the 3rd Conference of the Asia-Pacific Chapter of the Association for Computational Linguistics (Volume 1: Long Papers)}, pages 305--329, 2023.

\bibitem[Wei et~al.(2022{\natexlab{a}})Wei, Wang, Schuurmans, Bosma, Xia, Chi, Le, Zhou, et~al.]{wei2022chain}
Jason Wei, Xuezhi Wang, Dale Schuurmans, Maarten Bosma, Fei Xia, Ed~Chi, Quoc~V Le, Denny Zhou, et~al.
\newblock Chain-of-thought prompting elicits reasoning in large language models.
\newblock \emph{Advances in neural information processing systems}, 35:\penalty0 24824--24837, 2022{\natexlab{a}}.

\bibitem[Aguilar et~al.(2020)Aguilar, Kar, and Solorio]{aguilar2020lince}
Gustavo Aguilar, Sudipta Kar, and Thamar Solorio.
\newblock Lince: A centralized benchmark for linguistic code-switching evaluation.
\newblock In \emph{Proceedings of the Twelfth Language Resources and Evaluation Conference}, pages 1803--1813, 2020.

\bibitem[Srivastava and Singh(2022)]{srivastava2022overview}
Vivek Srivastava and Mayank Singh.
\newblock Overview and results of mixmt shared-task at wmt 2022.
\newblock In \emph{Proceedings of the Seventh Conference on Machine Translation (WMT)}, pages 806--811, 2022.

\bibitem[Mehnaz et~al.(2021)Mehnaz, Mahata, Gosangi, Gunturi, Jain, Gupta, Kumar, Lee, Acharya, and Shah]{mehnaz2021gupshup}
Laiba Mehnaz, Debanjan Mahata, Rakesh Gosangi, Uma~Sushmitha Gunturi, Riya Jain, Gauri Gupta, Amardeep Kumar, Isabelle~G Lee, Anish Acharya, and Rajiv Shah.
\newblock Gupshup: Summarizing open-domain code-switched conversations.
\newblock In \emph{Proceedings of the 2021 Conference on Empirical Methods in Natural Language Processing}, pages 6177--6192, 2021.

\bibitem[Zhang et~al.(2023{\natexlab{i}})Zhang, Zhou, Wang, Chen, Wu, Liu, Chen, Liu, Wang, Li, et~al.]{zhang2023speak}
Ziqiang Zhang, Long Zhou, Chengyi Wang, Sanyuan Chen, Yu~Wu, Shujie Liu, Zhuo Chen, Yanqing Liu, Huaming Wang, Jinyu Li, et~al.
\newblock Speak foreign languages with your own voice: Cross-lingual neural codec language modeling.
\newblock \emph{arXiv preprint arXiv:2303.03926}, 2023{\natexlab{i}}.

\bibitem[Huang et~al.(2024)Huang, Yang, Fu, Dunbar, and Lee]{huang2024zero}
Kuan-Po Huang, Chih-Kai Yang, Yu-Kuan Fu, Ewan Dunbar, and Hung-yi Lee.
\newblock Zero resource code-switched speech benchmark using speech utterance pairs for multiple spoken languages.
\newblock In \emph{ICASSP 2024-2024 IEEE International Conference on Acoustics, Speech and Signal Processing (ICASSP)}, pages 10006--10010. IEEE, 2024.

\bibitem[Do{\u{g}}ru{\"o}z et~al.(2021)Do{\u{g}}ru{\"o}z, Sitaram, Bullock, and Toribio]{dougruoz2021survey}
A~Seza Do{\u{g}}ru{\"o}z, Sunayana Sitaram, Barbara Bullock, and Almeida~Jacqueline Toribio.
\newblock A survey of code-switching: Linguistic and social perspectives for language technologies.
\newblock In \emph{Proceedings of the 59th Annual Meeting of the Association for Computational Linguistics and the 11th International Joint Conference on Natural Language Processing (Volume 1: Long Papers)}, pages 1654--1666, 2021.

\bibitem[Winata et~al.(2021)Winata, Cahyawijaya, Liu, Lin, Madotto, and Fung]{winata2021multilingual}
Genta~Indra Winata, Samuel Cahyawijaya, Zihan Liu, Zhaojiang Lin, Andrea Madotto, and Pascale Fung.
\newblock Are multilingual models effective in code-switching?
\newblock In \emph{Proceedings of the Fifth Workshop on Computational Approaches to Linguistic Code-Switching}, pages 142--153, 2021.

\bibitem[Das et~al.(2023)Das, Ranjan, Pathak, and Jyothi]{das-etal-2023-improving}
Richeek Das, Sahasra Ranjan, Shreya Pathak, and Preethi Jyothi.
\newblock Improving pretraining techniques for code-switched {NLP}.
\newblock In Anna Rogers, Jordan Boyd-Graber, and Naoaki Okazaki, editors, \emph{Proceedings of the 61st Annual Meeting of the Association for Computational Linguistics (Volume 1: Long Papers)}, pages 1176--1191, Toronto, Canada, July 2023. Association for Computational Linguistics.
\newblock \doi{10.18653/v1/2023.acl-long.66}.
\newblock URL \url{https://aclanthology.org/2023.acl-long.66}.

\bibitem[Gao et~al.(2023{\natexlab{a}})Gao, Xiong, Gao, Jia, Pan, Bi, Dai, Sun, and Wang]{gao2023retrieval}
Yunfan Gao, Yun Xiong, Xinyu Gao, Kangxiang Jia, Jinliu Pan, Yuxi Bi, Yi~Dai, Jiawei Sun, and Haofen Wang.
\newblock Retrieval-augmented generation for large language models: A survey.
\newblock \emph{arXiv preprint arXiv:2312.10997}, 2023{\natexlab{a}}.

\bibitem[He et~al.(2023{\natexlab{a}})He, Liang, Jiao, Zhang, Yang, Wang, Tu, Shi, and Wang]{he2023exploring}
Zhiwei He, Tian Liang, Wenxiang Jiao, Zhuosheng Zhang, Yujiu Yang, Rui Wang, Zhaopeng Tu, Shuming Shi, and Xing Wang.
\newblock Exploring human-like translation strategy with large language models, 2023{\natexlab{a}}.

\bibitem[Fan et~al.(2021)Fan, Bhosale, Schwenk, Ma, El-Kishky, Goyal, Baines, Celebi, Wenzek, Chaudhary, et~al.]{fan2021beyond}
Angela Fan, Shruti Bhosale, Holger Schwenk, Zhiyi Ma, Ahmed El-Kishky, Siddharth Goyal, Mandeep Baines, Onur Celebi, Guillaume Wenzek, Vishrav Chaudhary, et~al.
\newblock Beyond english-centric multilingual machine translation.
\newblock \emph{Journal of Machine Learning Research}, 22:\penalty0 1--48, 2021.

\bibitem[Wei et~al.(2022{\natexlab{b}})Wei, Tay, Bommasani, Raffel, Zoph, Borgeaud, Yogatama, Bosma, Zhou, Metzler, et~al.]{wei2022emergent}
Jason Wei, Yi~Tay, Rishi Bommasani, Colin Raffel, Barret Zoph, Sebastian Borgeaud, Dani Yogatama, Maarten Bosma, Denny Zhou, Donald Metzler, et~al.
\newblock Emergent abilities of large language models.
\newblock \emph{arXiv preprint arXiv:2206.07682}, 2022{\natexlab{b}}.

\bibitem[Magueresse et~al.(2020)Magueresse, Carles, and Heetderks]{magueresse2020low}
Alexandre Magueresse, Vincent Carles, and Evan Heetderks.
\newblock Low-resource languages: A review of past work and future challenges.
\newblock \emph{arXiv preprint arXiv:2006.07264}, 2020.

\bibitem[Ranathunga et~al.(2023)Ranathunga, Lee, Prifti~Skenduli, Shekhar, Alam, and Kaur]{ranathunga2023neural}
Surangika Ranathunga, En-Shiun~Annie Lee, Marjana Prifti~Skenduli, Ravi Shekhar, Mehreen Alam, and Rishemjit Kaur.
\newblock Neural machine translation for low-resource languages: A survey.
\newblock \emph{ACM Computing Surveys}, 55:\penalty0 1--37, 2023.

\bibitem[Miceli-Barone et~al.(2017)Miceli-Barone, Haddow, Germann, and Sennrich]{miceli2017regularization}
Antonio~Valerio Miceli-Barone, Barry Haddow, Ulrich Germann, and Rico Sennrich.
\newblock Regularization techniques for fine-tuning in neural machine translation.
\newblock In \emph{Proceedings of the 2017 Conference on Empirical Methods in Natural Language Processing}, pages 1489--1494, 2017.

\bibitem[Dabre et~al.(2019)Dabre, Fujita, and Chu]{dabre2019exploiting}
Raj Dabre, Atsushi Fujita, and Chenhui Chu.
\newblock Exploiting multilingualism through multistage fine-tuning for low-resource neural machine translation.
\newblock In \emph{Proceedings of the 2019 Conference on Empirical Methods in Natural Language Processing and the 9th International Joint Conference on Natural Language Processing (EMNLP-IJCNLP)}, pages 1410--1416, 2019.

\bibitem[Bapna and Firat(2019)]{bapna2019simple}
Ankur Bapna and Orhan Firat.
\newblock Simple, scalable adaptation for neural machine translation.
\newblock In \emph{Proceedings of the 2019 Conference on Empirical Methods in Natural Language Processing and the 9th International Joint Conference on Natural Language Processing (EMNLP-IJCNLP)}, pages 1538--1548, 2019.

\bibitem[Manning et~al.(2008)Manning, Raghavan, and Schütze]{manning_raghavan_schutze_2008}
Christopher~D. Manning, Prabhakar Raghavan, and Hinrich Schütze.
\newblock \emph{Introduction to Information Retrieval}.
\newblock Cambridge University Press, 2008.

\bibitem[Fluhr et~al.(1995)Fluhr, Frederking, Oard, Okumura, Ishikawa, and Satoh]{fluhr19952}
Christian Fluhr, Robert~E Frederking, Doug Oard, Akitoshi Okumura, Kai Ishikawa, and Kenji Satoh.
\newblock 2.1 multilingual information retrieval.
\newblock 1995.

\bibitem[Oard and Diekema(1998)]{oard1998cross}
Douglas~W Oard and Anne~R Diekema.
\newblock Cross-language information retrieval.
\newblock \emph{Annual Review of Information Science and Technology (ARIST)}, 33:\penalty0 223--56, 1998.

\bibitem[Nie et~al.(1999)Nie, Simard, Isabelle, and Durand]{nie1999cross}
Jian-Yun Nie, Michel Simard, Pierre Isabelle, and Richard Durand.
\newblock Cross-language information retrieval based on parallel texts and automatic mining of parallel texts from the web.
\newblock In \emph{Proceedings of the 22nd annual international ACM SIGIR conference on Research and development in information retrieval}, pages 74--81, 1999.

\bibitem[Hollink et~al.(2004)Hollink, Kamps, Monz, and De~Rijke]{hollink2004monolingual}
Vera Hollink, Jaap Kamps, Christof Monz, and Maarten De~Rijke.
\newblock Monolingual document retrieval for european languages.
\newblock \emph{Information retrieval}, 7:\penalty0 33--52, 2004.

\bibitem[Gao et~al.(2007)Gao, Niu, Nie, Zhou, Hu, Wong, and Hon]{gao2007cross}
Wei Gao, Cheng Niu, Jian-Yun Nie, Ming Zhou, Jian Hu, Kam-Fai Wong, and Hsiao-Wuen Hon.
\newblock Cross-lingual query suggestion using query logs of different languages.
\newblock In \emph{Proceedings of the 30th annual international ACM SIGIR conference on Research and development in information retrieval}, pages 463--470, 2007.

\bibitem[Grefenstette(2012)]{grefenstette2012cross}
Gregory Grefenstette.
\newblock \emph{Cross-language information retrieval}, volume~2.
\newblock Springer Science \& Business Media, 2012.

\bibitem[Peters et~al.(2012)Peters, Braschler, and Clough]{peters2012multilingual}
Carol Peters, Martin Braschler, and Paul Clough.
\newblock \emph{Multilingual information retrieval: From research to practice}.
\newblock Springer, 2012.

\bibitem[Nie(2022)]{nie2022cross}
Jian-Yun Nie.
\newblock \emph{Cross-language information retrieval}.
\newblock Springer Nature, 2022.

\bibitem[Lawrie et~al.(2022)Lawrie, Mayfield, Oard, and Yang]{lawrie2022hc4}
Dawn Lawrie, James Mayfield, Douglas~W. Oard, and Eugene Yang.
\newblock Hc4: A new suite of test collections for ad hoc clir.
\newblock In Matthias Hagen, Suzan Verberne, Craig Macdonald, Christin Seifert, Krisztian Balog, Kjetil N{\o}rv{\aa}g, and Vinay Setty, editors, \emph{Advances in Information Retrieval}, pages 351--366, Cham, 2022. Springer International Publishing.
\newblock ISBN 978-3-030-99736-6.

\bibitem[Lin et~al.(2022{\natexlab{b}})Lin, Nogueira, and Yates]{lin2022pretrained}
Jimmy Lin, Rodrigo Nogueira, and Andrew Yates.
\newblock \emph{Pretrained transformers for text ranking: Bert and beyond}.
\newblock Springer Nature, 2022{\natexlab{b}}.

\bibitem[Ai et~al.(2023)Ai, Bai, Cao, Chang, Chen, Chen, Cheng, Dong, Dou, Feng, Gao, Guo, He, Lan, Li, Liu, Lyu, Ma, Ma, Ren, Ren, Wang, Wang, Wen, Wu, Xin, Xu, Yin, Zhang, Zhang, Zhang, Zhang, and Zhu]{ai202380}
Qingyao Ai, Ting Bai, Zhao Cao, Yi~Chang, Jiawei Chen, Zhumin Chen, Zhiyong Cheng, Shoubin Dong, Zhicheng Dou, Fuli Feng, Shen Gao, Jiafeng Guo, Xiangnan He, Yanyan Lan, Chenliang Li, Yiqun Liu, Ziyu Lyu, Weizhi Ma, Jun Ma, Zhaochun Ren, Pengjie Ren, Zhiqiang Wang, Mingwen Wang, Ji-Rong Wen, Le~Wu, Xin Xin, Jun Xu, Dawei Yin, Peng Zhang, Fan Zhang, Weinan Zhang, Min Zhang, and Xiaofei Zhu.
\newblock Information retrieval meets large language models: A strategic report from chinese ir community.
\newblock \emph{AI Open}, 4:\penalty0 80--90, 2023.
\newblock ISSN 2666-6510.
\newblock \doi{https://doi.org/10.1016/j.aiopen.2023.08.001}.
\newblock URL \url{https://www.sciencedirect.com/science/article/pii/S2666651023000049}.

\bibitem[Zhu et~al.(2024)Zhu, Yuan, Wang, Liu, Liu, Deng, Chen, Liu, Dou, and Wen]{zhu2024largelanguagemodelsinformation}
Yutao Zhu, Huaying Yuan, Shuting Wang, Jiongnan Liu, Wenhan Liu, Chenlong Deng, Haonan Chen, Zheng Liu, Zhicheng Dou, and Ji-Rong Wen.
\newblock Large language models for information retrieval: A survey, 2024.
\newblock URL \url{https://arxiv.org/abs/2308.07107}.

\bibitem[Liu et~al.(2024{\natexlab{c}})Liu, Zhou, Zhu, Lian, Li, Dou, Lian, and Nie]{liu2024information}
Zheng Liu, Yujia Zhou, Yutao Zhu, Jianxun Lian, Chaozhuo Li, Zhicheng Dou, Defu Lian, and Jian-Yun Nie.
\newblock Information retrieval meets large language models.
\newblock In \emph{Companion Proceedings of the ACM Web Conference 2024}, WWW '24, page 1586–1589, New York, NY, USA, 2024{\natexlab{c}}. Association for Computing Machinery.
\newblock ISBN 9798400701726.
\newblock \doi{10.1145/3589335.3641299}.
\newblock URL \url{https://doi.org/10.1145/3589335.3641299}.

\bibitem[Haq et~al.(2024)Haq, Sharma, Khattab, Chhaya, and Bhattacharyya]{haq-etal-2024-indicirsuite}
Saiful Haq, Ashutosh Sharma, Omar Khattab, Niyati Chhaya, and Pushpak Bhattacharyya.
\newblock {I}ndic{IRS}uite: Multilingual dataset and neural information models for {I}ndian languages.
\newblock In Lun-Wei Ku, Andre Martins, and Vivek Srikumar, editors, \emph{Proceedings of the 62nd Annual Meeting of the Association for Computational Linguistics (Volume 2: Short Papers)}, pages 501--509, Bangkok, Thailand, August 2024. Association for Computational Linguistics.
\newblock \doi{10.18653/v1/2024.acl-short.46}.
\newblock URL \url{https://aclanthology.org/2024.acl-short.46}.

\bibitem[Izacard et~al.(2021)Izacard, Caron, Hosseini, Riedel, Bojanowski, Joulin, and Grave]{izacard2021contriever}
Gautier Izacard, Mathilde Caron, Lucas Hosseini, Sebastian Riedel, Piotr Bojanowski, Armand Joulin, and Edouard Grave.
\newblock Unsupervised dense information retrieval with contrastive learning, 2021.
\newblock URL \url{https://arxiv.org/abs/2112.09118}.

\bibitem[Wang et~al.(2024{\natexlab{d}})Wang, Yang, Huang, Jiao, Yang, Jiang, Majumder, and Wei]{wang2024textembeddingsweaklysupervisedcontrastive}
Liang Wang, Nan Yang, Xiaolong Huang, Binxing Jiao, Linjun Yang, Daxin Jiang, Rangan Majumder, and Furu Wei.
\newblock Text embeddings by weakly-supervised contrastive pre-training, 2024{\natexlab{d}}.
\newblock URL \url{https://arxiv.org/abs/2212.03533}.

\bibitem[Schwenk et~al.(2021)Schwenk, Wenzek, Edunov, Grave, Joulin, and Fan]{schwenk-etal-2021-ccmatrix}
Holger Schwenk, Guillaume Wenzek, Sergey Edunov, Edouard Grave, Armand Joulin, and Angela Fan.
\newblock {CCM}atrix: Mining billions of high-quality parallel sentences on the web.
\newblock In \emph{Proceedings of the 59th Annual Meeting of the Association for Computational Linguistics and the 11th International Joint Conference on Natural Language Processing (Volume 1: Long Papers)}, pages 6490--6500, Online, August 2021. Association for Computational Linguistics.
\newblock \doi{10.18653/v1/2021.acl-long.507}.
\newblock URL \url{https://aclanthology.org/2021.acl-long.507}.

\bibitem[El-Kishky et~al.(2020)El-Kishky, Chaudhary, Guzm{\'a}n, and Koehn]{el-kishky-etal-2020-ccaligned}
Ahmed El-Kishky, Vishrav Chaudhary, Francisco Guzm{\'a}n, and Philipp Koehn.
\newblock {CCA}ligned: A massive collection of cross-lingual web-document pairs.
\newblock In \emph{Proceedings of the 2020 Conference on Empirical Methods in Natural Language Processing (EMNLP)}, pages 5960--5969, Online, November 2020. Association for Computational Linguistics.
\newblock \doi{10.18653/v1/2020.emnlp-main.480}.
\newblock URL \url{https://aclanthology.org/2020.emnlp-main.480}.

\bibitem[Lin and Ma(2021)]{lin2021few}
Jimmy Lin and Xueguang Ma.
\newblock A few brief notes on deepimpact, coil, and a conceptual framework for information retrieval techniques.
\newblock \emph{arXiv preprint arXiv:2106.14807}, 2021.

\bibitem[Robertson and Zaragoza(2009)]{robertson2009probabilistic}
Stephen Robertson and Hugo Zaragoza.
\newblock The probabilistic relevance framework: {BM}25 and beyond.
\newblock \emph{Foundation and Trends in Information Retrieval}, 3\penalty0 (4):\penalty0 333–389, April 2009.
\newblock ISSN 1554-0669.

\bibitem[Zhang et~al.(2021{\natexlab{b}})Zhang, Ma, Shi, and Lin]{zhang-etal-2021-mr}
Xinyu Zhang, Xueguang Ma, Peng Shi, and Jimmy Lin.
\newblock Mr. {T}y{D}i: A multi-lingual benchmark for dense retrieval.
\newblock In \emph{Proceedings of the 1st Workshop on Multilingual Representation Learning}, pages 127--137, Punta Cana, Dominican Republic, November 2021{\natexlab{b}}. Association for Computational Linguistics.
\newblock \doi{10.18653/v1/2021.mrl-1.12}.
\newblock URL \url{https://aclanthology.org/2021.mrl-1.12}.

\bibitem[Hull and Grefenstette(1996)]{hull1996querying}
David~A Hull and Gregory Grefenstette.
\newblock Querying across languages: A dictionary-based approach to multilingual information retrieval.
\newblock In \emph{Proceedings of the 19th annual international ACM SIGIR conference on Research and development in information retrieval}, pages 49--57, 1996.

\bibitem[Zhou et~al.(2012)Zhou, Truran, Brailsford, Wade, and Ashman]{zhou2012translation}
Dong Zhou, Mark Truran, Tim Brailsford, Vincent Wade, and Helen Ashman.
\newblock Translation techniques in cross-language information retrieval.
\newblock \emph{ACM Computing Surveys (CSUR)}, 45\penalty0 (1):\penalty0 1--44, 2012.

\bibitem[Lin et~al.(2023)Lin, Alfonso-Hermelo, Jeronymo, Kamalloo, Lassance, Nogueira, Ogundepo, Rezagholizadeh, Thakur, Yang, and Zhang]{lin2023simpleeffectiveneuralranking}
Jimmy Lin, David Alfonso-Hermelo, Vitor Jeronymo, Ehsan Kamalloo, Carlos Lassance, Rodrigo Nogueira, Odunayo Ogundepo, Mehdi Rezagholizadeh, Nandan Thakur, Jheng-Hong Yang, and Xinyu Zhang.
\newblock Simple yet effective neural ranking and reranking baselines for cross-lingual information retrieval, 2023.
\newblock URL \url{https://arxiv.org/abs/2304.01019}.

\bibitem[Dai and Callan(2019)]{dai2019contextawaresentencepassagetermimportance}
Zhuyun Dai and Jamie Callan.
\newblock Context-aware sentence/passage term importance estimation for first stage retrieval, 2019.
\newblock URL \url{https://arxiv.org/abs/1910.10687}.

\bibitem[Gao et~al.(2021)Gao, Dai, and Callan]{gao-etal-2021-coil}
Luyu Gao, Zhuyun Dai, and Jamie Callan.
\newblock {COIL}: Revisit exact lexical match in information retrieval with contextualized inverted list.
\newblock In \emph{Proceedings of the 2021 Conference of the North American Chapter of the Association for Computational Linguistics: Human Language Technologies}, pages 3030--3042, Online, June 2021. Association for Computational Linguistics.
\newblock \doi{10.18653/v1/2021.naacl-main.241}.
\newblock URL \url{https://aclanthology.org/2021.naacl-main.241}.

\bibitem[Formal et~al.(2021{\natexlab{a}})Formal, Piwowarski, and Clinchant]{10.1145/3404835.3463098}
Thibault Formal, Benjamin Piwowarski, and St\'{e}phane Clinchant.
\newblock \emph{SPLADE: Sparse Lexical and Expansion Model for First Stage Ranking}, page 2288–2292.
\newblock Association for Computing Machinery, New York, NY, USA, 2021{\natexlab{a}}.
\newblock ISBN 9781450380379.
\newblock URL \url{https://doi.org/10.1145/3404835.3463098}.

\bibitem[Formal et~al.(2021{\natexlab{b}})Formal, Lassance, Piwowarski, and Clinchant]{10.48550/arxiv.2109.10086}
Thibault Formal, Carlos Lassance, Benjamin Piwowarski, and Stéphane Clinchant.
\newblock Splade v2: Sparse lexical and expansion model for information retrieval, 2021{\natexlab{b}}.
\newblock URL \url{https://arxiv.org/abs/2109.10086}.

\bibitem[Formal et~al.(2022)Formal, Lassance, Piwowarski, and Clinchant]{10.1145/3477495.3531857}
Thibault Formal, Carlos Lassance, Benjamin Piwowarski, and St\'{e}phane Clinchant.
\newblock From distillation to hard negative sampling: Making sparse neural ir models more effective.
\newblock In \emph{Proceedings of the 45th International ACM SIGIR Conference on Research and Development in Information Retrieval}, SIGIR '22, page 2353–2359, New York, NY, USA, 2022. Association for Computing Machinery.
\newblock ISBN 9781450387323.
\newblock \doi{10.1145/3477495.3531857}.
\newblock URL \url{https://doi.org/10.1145/3477495.3531857}.

\bibitem[Lassance and Clinchant(2022)]{10.1145/3477495.3531833}
Carlos Lassance and St\'{e}phane Clinchant.
\newblock An efficiency study for splade models.
\newblock In \emph{Proceedings of the 45th International ACM SIGIR Conference on Research and Development in Information Retrieval}, SIGIR '22, page 2220–2226, New York, NY, USA, 2022. Association for Computing Machinery.
\newblock ISBN 9781450387323.
\newblock \doi{10.1145/3477495.3531833}.
\newblock URL \url{https://doi.org/10.1145/3477495.3531833}.

\bibitem[Lassance(2023)]{lassance2023extendingenglishirmethods}
Carlos Lassance.
\newblock Extending english ir methods to multi-lingual ir, 2023.
\newblock URL \url{https://arxiv.org/abs/2302.14723}.

\bibitem[Conneau et~al.(2020{\natexlab{b}})Conneau, Khandelwal, Goyal, Chaudhary, Wenzek, Guzmán, Grave, Ott, Zettlemoyer, and Stoyanov]{conneau2020unsupervisedcrosslingualrepresentationlearning}
Alexis Conneau, Kartikay Khandelwal, Naman Goyal, Vishrav Chaudhary, Guillaume Wenzek, Francisco Guzmán, Edouard Grave, Myle Ott, Luke Zettlemoyer, and Veselin Stoyanov.
\newblock Unsupervised cross-lingual representation learning at scale, 2020{\natexlab{b}}.
\newblock URL \url{https://arxiv.org/abs/1911.02116}.

\bibitem[Nair et~al.(2023)Nair, Yang, Lawrie, Mayfield, and Oard]{nair2023blade}
Suraj Nair, Eugene Yang, Dawn Lawrie, James Mayfield, and Douglas~W. Oard.
\newblock Blade: Combining vocabulary pruning and intermediate pretraining for scaleable neural clir.
\newblock In \emph{Proceedings of the 46th International ACM SIGIR Conference on Research and Development in Information Retrieval}, SIGIR '23, page 1219–1229, New York, NY, USA, 2023. Association for Computing Machinery.
\newblock ISBN 9781450394086.
\newblock \doi{10.1145/3539618.3591644}.
\newblock URL \url{https://doi.org/10.1145/3539618.3591644}.

\bibitem[Asai et~al.(2021)Asai, Kasai, Clark, Lee, Choi, and Hajishirzi]{asai-etal-2021-xor}
Akari Asai, Jungo Kasai, Jonathan Clark, Kenton Lee, Eunsol Choi, and Hannaneh Hajishirzi.
\newblock {XOR} {QA}: Cross-lingual open-retrieval question answering.
\newblock In \emph{Proceedings of the 2021 Conference of the North American Chapter of the Association for Computational Linguistics: Human Language Technologies}, pages 547--564, Online, June 2021. Association for Computational Linguistics.
\newblock \doi{10.18653/v1/2021.naacl-main.46}.
\newblock URL \url{https://aclanthology.org/2021.naacl-main.46}.

\bibitem[Zhang et~al.(2023{\natexlab{j}})Zhang, Ogueji, Ma, and Lin]{zhang2023toward}
Xinyu Zhang, Kelechi Ogueji, Xueguang Ma, and Jimmy Lin.
\newblock Toward best practices for training multilingual dense retrieval models.
\newblock \emph{ACM Trans. Inf. Syst.}, 42\penalty0 (2), September 2023{\natexlab{j}}.
\newblock ISSN 1046-8188.
\newblock \doi{10.1145/3613447}.
\newblock URL \url{https://doi.org/10.1145/3613447}.

\bibitem[Li et~al.(2022{\natexlab{b}})Li, Franz, Sultan, Iyer, Lee, and Sil]{li-etal-2022-learning-cross}
Yulong Li, Martin Franz, Md~Arafat Sultan, Bhavani Iyer, Young-Suk Lee, and Avirup Sil.
\newblock Learning cross-lingual {IR} from an {E}nglish retriever.
\newblock In \emph{Proceedings of the 2022 Conference of the North American Chapter of the Association for Computational Linguistics: Human Language Technologies}, pages 4428--4436, Seattle, United States, July 2022{\natexlab{b}}. Association for Computational Linguistics.
\newblock \doi{10.18653/v1/2022.naacl-main.329}.
\newblock URL \url{https://aclanthology.org/2022.naacl-main.329}.

\bibitem[Khattab and Zaharia(2020)]{khattab2020colbert}
Omar Khattab and Matei Zaharia.
\newblock Colbert: Efficient and effective passage search via contextualized late interaction over bert.
\newblock In \emph{Proceedings of the 43rd International ACM SIGIR Conference on Research and Development in Information Retrieval}, SIGIR '20, page 39–48, New York, NY, USA, 2020. Association for Computing Machinery.
\newblock ISBN 9781450380164.
\newblock \doi{10.1145/3397271.3401075}.
\newblock URL \url{https://doi.org/10.1145/3397271.3401075}.

\bibitem[MacAvaney et~al.(2020)MacAvaney, Soldaini, and Goharian]{macavaney2020teaching}
Sean MacAvaney, Luca Soldaini, and Nazli Goharian.
\newblock Teaching a new dog old tricks: Resurrecting multilingual retrieval using zero-shot learning.
\newblock In \emph{Advances in Information Retrieval: 42nd European Conference on IR Research, ECIR 2020, Lisbon, Portugal, April 14–17, 2020, Proceedings, Part II}, page 246–254, Berlin, Heidelberg, 2020. Springer-Verlag.
\newblock ISBN 978-3-030-45441-8.
\newblock \doi{10.1007/978-3-030-45442-5_31}.
\newblock URL \url{https://doi.org/10.1007/978-3-030-45442-5_31}.

\bibitem[Shi and Lin(2019)]{shi2019crosslingualrelevancetransferdocument}
Peng Shi and Jimmy Lin.
\newblock Cross-lingual relevance transfer for document retrieval, 2019.
\newblock URL \url{https://arxiv.org/abs/1911.02989}.

\bibitem[Shi et~al.(2020)Shi, Bai, and Lin]{shi-etal-2020-cross}
Peng Shi, He~Bai, and Jimmy Lin.
\newblock Cross-lingual training of neural models for document ranking.
\newblock In \emph{Findings of the Association for Computational Linguistics: EMNLP 2020}, pages 2768--2773, Online, November 2020. Association for Computational Linguistics.
\newblock \doi{10.18653/v1/2020.findings-emnlp.249}.
\newblock URL \url{https://aclanthology.org/2020.findings-emnlp.249}.

\bibitem[Adeyemi et~al.(2024)Adeyemi, Oladipo, Zhang, Alfonso-Hermelo, Rezagholizadeh, Chen, Omotayo, Abdulmumin, Etori, Musa, Fanijo, Awoyomi, Salahudeen, Mohammed, Abolade, Lawan, Sabo~Abubakar, Nasir~Iro, Imam~Abubakar, Mohamed, Mohamed, Ajayi, and Lin]{mofetoluwa2024ciral}
Mofetoluwa Adeyemi, Akintunde Oladipo, Xinyu Zhang, David Alfonso-Hermelo, Mehdi Rezagholizadeh, Boxing Chen, Abdul-Hakeem Omotayo, Idris Abdulmumin, Naome~A. Etori, Toyib~Babatunde Musa, Samuel Fanijo, Oluwabusayo~Olufunke Awoyomi, Saheed~Abdullahi Salahudeen, Labaran~Adamu Mohammed, Daud~Olamide Abolade, Falalu~Ibrahim Lawan, Maryam Sabo~Abubakar, Ruqayya Nasir~Iro, Amina Imam~Abubakar, Shafie~Abdi Mohamed, Hanad~Mohamud Mohamed, Tunde~Oluwaseyi Ajayi, and Jimmy Lin.
\newblock Ciral: A test collection for clir evaluations in african languages.
\newblock In \emph{Proceedings of the 47th International ACM SIGIR Conference on Research and Development in Information Retrieval}, SIGIR '24, page 293–302, New York, NY, USA, 2024. Association for Computing Machinery.
\newblock ISBN 9798400704314.
\newblock \doi{10.1145/3626772.3657884}.
\newblock URL \url{https://doi.org/10.1145/3626772.3657884}.

\bibitem[Li et~al.(2024{\natexlab{d}})Li, Jin, Zhou, Zhang, Zhang, Zhu, and Dou]{li2024matchinggenerationsurveygenerative}
Xiaoxi Li, Jiajie Jin, Yujia Zhou, Yuyao Zhang, Peitian Zhang, Yutao Zhu, and Zhicheng Dou.
\newblock From matching to generation: A survey on generative information retrieval, 2024{\natexlab{d}}.
\newblock URL \url{https://arxiv.org/abs/2404.14851}.

\bibitem[Lee et~al.(2024{\natexlab{d}})Lee, Chen, Dai, Dua, Sachan, Boratko, Luan, Arnold, Perot, Dalmia, Hu, Lin, Pasupat, Amini, Cole, Riedel, Naim, Chang, and Guu]{lee2024longcontextlanguagemodelssubsume}
Jinhyuk Lee, Anthony Chen, Zhuyun Dai, Dheeru Dua, Devendra~Singh Sachan, Michael Boratko, Yi~Luan, Sébastien M.~R. Arnold, Vincent Perot, Siddharth Dalmia, Hexiang Hu, Xudong Lin, Panupong Pasupat, Aida Amini, Jeremy~R. Cole, Sebastian Riedel, Iftekhar Naim, Ming-Wei Chang, and Kelvin Guu.
\newblock Can long-context language models subsume retrieval, {RAG}, {SQL}, and more?, 2024{\natexlab{d}}.
\newblock URL \url{https://arxiv.org/abs/2406.13121}.

\bibitem[Tamber et~al.(2023)Tamber, Pradeep, and Lin]{tamber2023scalingdownlittingup}
Manveer~Singh Tamber, Ronak Pradeep, and Jimmy Lin.
\newblock {Scaling Down, LiTting Up}: Efficient zero-shot listwise reranking with seq2seq encoder-decoder models, 2023.
\newblock URL \url{https://arxiv.org/abs/2312.16098}.

\bibitem[Grattafiori et~al.(2024)Grattafiori, Dubey, Jauhri, Pandey, Kadian, Al-Dahle, Letman, Mathur, Schelten, Vaughan, Yang, Fan, Goyal, Hartshorn, Yang, Mitra, Sravankumar, Korenev, Hinsvark, Rao, Zhang, Rodriguez, Gregerson, Spataru, Roziere, Biron, Tang, Chern, Caucheteux, Nayak, Bi, Marra, McConnell, Keller, Touret, Wu, Wong, Ferrer, Nikolaidis, Allonsius, Song, Pintz, Livshits, Wyatt, Esiobu, Choudhary, Mahajan, Garcia-Olano, Perino, Hupkes, Lakomkin, AlBadawy, Lobanova, Dinan, Smith, Radenovic, Guzmán, Zhang, Synnaeve, Lee, Anderson, Thattai, Nail, Mialon, Pang, Cucurell, Nguyen, Korevaar, Xu, Touvron, Zarov, Ibarra, Kloumann, Misra, Evtimov, Zhang, Copet, Lee, Geffert, Vranes, Park, Mahadeokar, Shah, van~der Linde, Billock, Hong, Lee, Fu, Chi, Huang, Liu, Wang, Yu, Bitton, Spisak, Park, Rocca, Johnstun, Saxe, Jia, Alwala, Prasad, Upasani, Plawiak, Li, Heafield, Stone, El-Arini, Iyer, Malik, Chiu, Bhalla, Lakhotia, Rantala-Yeary, van~der Maaten, Chen, Tan, Jenkins, Martin, Madaan, Malo, Blecher,
  Landzaat, de~Oliveira, Muzzi, Pasupuleti, Singh, Paluri, Kardas, Tsimpoukelli, Oldham, Rita, Pavlova, Kambadur, Lewis, Si, Singh, Hassan, Goyal, Torabi, Bashlykov, Bogoychev, Chatterji, Zhang, Duchenne, Çelebi, Alrassy, Zhang, Li, Vasic, Weng, Bhargava, Dubal, Krishnan, Koura, Xu, He, Dong, Srinivasan, Ganapathy, Calderer, Cabral, Stojnic, Raileanu, Maheswari, Girdhar, Patel, Sauvestre, Polidoro, Sumbaly, Taylor, Silva, Hou, Wang, Hosseini, Chennabasappa, Singh, Bell, Kim, Edunov, Nie, Narang, Raparthy, Shen, Wan, Bhosale, Zhang, Vandenhende, Batra, Whitman, Sootla, Collot, Gururangan, Borodinsky, Herman, Fowler, Sheasha, Georgiou, Scialom, Speckbacher, Mihaylov, Xiao, Karn, Goswami, Gupta, Ramanathan, Kerkez, Gonguet, Do, Vogeti, Albiero, Petrovic, Chu, Xiong, Fu, Meers, Martinet, Wang, Wang, Tan, Xia, Xie, Jia, Wang, Goldschlag, Gaur, Babaei, Wen, Song, Zhang, Li, Mao, Coudert, Yan, Chen, Papakipos, Singh, Srivastava, Jain, Kelsey, Shajnfeld, Gangidi, Victoria, Goldstand, Menon, Sharma, Boesenberg,
  Baevski, Feinstein, Kallet, Sangani, Teo, Yunus, Lupu, Alvarado, Caples, Gu, Ho, Poulton, Ryan, Ramchandani, Dong, Franco, Goyal, Saraf, Chowdhury, Gabriel, Bharambe, Eisenman, Yazdan, James, Maurer, Leonhardi, Huang, Loyd, Paola, Paranjape, Liu, Wu, Ni, Hancock, Wasti, Spence, Stojkovic, Gamido, Montalvo, Parker, Burton, Mejia, Liu, Wang, Kim, Zhou, Hu, Chu, Cai, Tindal, Feichtenhofer, Gao, Civin, Beaty, Kreymer, Li, Adkins, Xu, Testuggine, David, Parikh, Liskovich, Foss, Wang, Le, Holland, Dowling, Jamil, Montgomery, Presani, Hahn, Wood, Le, Brinkman, Arcaute, Dunbar, Smothers, Sun, Kreuk, Tian, Kokkinos, Ozgenel, Caggioni, Kanayet, Seide, Florez, Schwarz, Badeer, Swee, Halpern, Herman, Sizov, Guangyi, Zhang, Lakshminarayanan, Inan, Shojanazeri, Zou, Wang, Zha, Habeeb, Rudolph, Suk, Aspegren, Goldman, Zhan, Damlaj, Molybog, Tufanov, Leontiadis, Veliche, Gat, Weissman, Geboski, Kohli, Lam, Asher, Gaya, Marcus, Tang, Chan, Zhen, Reizenstein, Teboul, Zhong, Jin, Yang, Cummings, Carvill, Shepard, McPhie,
  Torres, Ginsburg, Wang, Wu, U, Saxena, Khandelwal, Zand, Matosich, Veeraraghavan, Michelena, Li, Jagadeesh, Huang, Chawla, Huang, Chen, Garg, A, Silva, Bell, Zhang, Guo, Yu, Moshkovich, Wehrstedt, Khabsa, Avalani, Bhatt, Mankus, Hasson, Lennie, Reso, Groshev, Naumov, Lathi, Keneally, Liu, Seltzer, Valko, Restrepo, Patel, Vyatskov, Samvelyan, Clark, Macey, Wang, Hermoso, Metanat, Rastegari, Bansal, Santhanam, Parks, White, Bawa, Singhal, Egebo, Usunier, Mehta, Laptev, Dong, Cheng, Chernoguz, Hart, Salpekar, Kalinli, Kent, Parekh, Saab, Balaji, Rittner, Bontrager, Roux, Dollar, Zvyagina, Ratanchandani, Yuvraj, Liang, Alao, Rodriguez, Ayub, Murthy, Nayani, Mitra, Parthasarathy, Li, Hogan, Battey, Wang, Howes, Rinott, Mehta, Siby, Bondu, Datta, Chugh, Hunt, Dhillon, Sidorov, Pan, Mahajan, Verma, Yamamoto, Ramaswamy, Lindsay, Lindsay, Feng, Lin, Zha, Patil, Shankar, Zhang, Zhang, Wang, Agarwal, Sajuyigbe, Chintala, Max, Chen, Kehoe, Satterfield, Govindaprasad, Gupta, Deng, Cho, Virk, Subramanian, Choudhury,
  Goldman, Remez, Glaser, Best, Koehler, Robinson, Li, Zhang, Matthews, Chou, Shaked, Vontimitta, Ajayi, Montanez, Mohan, Kumar, Mangla, Ionescu, Poenaru, Mihailescu, Ivanov, Li, Wang, Jiang, Bouaziz, Constable, Tang, Wu, Wang, Wu, Gao, Kleinman, Chen, Hu, Jia, Qi, Li, Zhang, Zhang, Adi, Nam, Yu, Wang, Zhao, Hao, Qian, Li, He, Rait, DeVito, Rosnbrick, Wen, Yang, Zhao, and Ma]{grattafiori2024llama3herdmodels}
Aaron Grattafiori, Abhimanyu Dubey, Abhinav Jauhri, Abhinav Pandey, Abhishek Kadian, Ahmad Al-Dahle, Aiesha Letman, Akhil Mathur, Alan Schelten, Alex Vaughan, Amy Yang, Angela Fan, Anirudh Goyal, Anthony Hartshorn, Aobo Yang, Archi Mitra, Archie Sravankumar, Artem Korenev, Arthur Hinsvark, Arun Rao, Aston Zhang, Aurelien Rodriguez, Austen Gregerson, Ava Spataru, Baptiste Roziere, Bethany Biron, Binh Tang, Bobbie Chern, Charlotte Caucheteux, Chaya Nayak, Chloe Bi, Chris Marra, Chris McConnell, Christian Keller, Christophe Touret, Chunyang Wu, Corinne Wong, Cristian~Canton Ferrer, Cyrus Nikolaidis, Damien Allonsius, Daniel Song, Danielle Pintz, Danny Livshits, Danny Wyatt, David Esiobu, Dhruv Choudhary, Dhruv Mahajan, Diego Garcia-Olano, Diego Perino, Dieuwke Hupkes, Egor Lakomkin, Ehab AlBadawy, Elina Lobanova, Emily Dinan, Eric~Michael Smith, Filip Radenovic, Francisco Guzmán, Frank Zhang, Gabriel Synnaeve, Gabrielle Lee, Georgia~Lewis Anderson, Govind Thattai, Graeme Nail, Gregoire Mialon, Guan Pang,
  Guillem Cucurell, Hailey Nguyen, Hannah Korevaar, Hu~Xu, Hugo Touvron, Iliyan Zarov, Imanol~Arrieta Ibarra, Isabel Kloumann, Ishan Misra, Ivan Evtimov, Jack Zhang, Jade Copet, Jaewon Lee, Jan Geffert, Jana Vranes, Jason Park, Jay Mahadeokar, Jeet Shah, Jelmer van~der Linde, Jennifer Billock, Jenny Hong, Jenya Lee, Jeremy Fu, Jianfeng Chi, Jianyu Huang, Jiawen Liu, Jie Wang, Jiecao Yu, Joanna Bitton, Joe Spisak, Jongsoo Park, Joseph Rocca, Joshua Johnstun, Joshua Saxe, Junteng Jia, Kalyan~Vasuden Alwala, Karthik Prasad, Kartikeya Upasani, Kate Plawiak, Ke~Li, Kenneth Heafield, Kevin Stone, Khalid El-Arini, Krithika Iyer, Kshitiz Malik, Kuenley Chiu, Kunal Bhalla, Kushal Lakhotia, Lauren Rantala-Yeary, Laurens van~der Maaten, Lawrence Chen, Liang Tan, Liz Jenkins, Louis Martin, Lovish Madaan, Lubo Malo, Lukas Blecher, Lukas Landzaat, Luke de~Oliveira, Madeline Muzzi, Mahesh Pasupuleti, Mannat Singh, Manohar Paluri, Marcin Kardas, Maria Tsimpoukelli, Mathew Oldham, Mathieu Rita, Maya Pavlova, Melanie Kambadur,
  Mike Lewis, Min Si, Mitesh~Kumar Singh, Mona Hassan, Naman Goyal, Narjes Torabi, Nikolay Bashlykov, Nikolay Bogoychev, Niladri Chatterji, Ning Zhang, Olivier Duchenne, Onur Çelebi, Patrick Alrassy, Pengchuan Zhang, Pengwei Li, Petar Vasic, Peter Weng, Prajjwal Bhargava, Pratik Dubal, Praveen Krishnan, Punit~Singh Koura, Puxin Xu, Qing He, Qingxiao Dong, Ragavan Srinivasan, Raj Ganapathy, Ramon Calderer, Ricardo~Silveira Cabral, Robert Stojnic, Roberta Raileanu, Rohan Maheswari, Rohit Girdhar, Rohit Patel, Romain Sauvestre, Ronnie Polidoro, Roshan Sumbaly, Ross Taylor, Ruan Silva, Rui Hou, Rui Wang, Saghar Hosseini, Sahana Chennabasappa, Sanjay Singh, Sean Bell, Seohyun~Sonia Kim, Sergey Edunov, Shaoliang Nie, Sharan Narang, Sharath Raparthy, Sheng Shen, Shengye Wan, Shruti Bhosale, Shun Zhang, Simon Vandenhende, Soumya Batra, Spencer Whitman, Sten Sootla, Stephane Collot, Suchin Gururangan, Sydney Borodinsky, Tamar Herman, Tara Fowler, Tarek Sheasha, Thomas Georgiou, Thomas Scialom, Tobias Speckbacher,
  Todor Mihaylov, Tong Xiao, Ujjwal Karn, Vedanuj Goswami, Vibhor Gupta, Vignesh Ramanathan, Viktor Kerkez, Vincent Gonguet, Virginie Do, Vish Vogeti, Vítor Albiero, Vladan Petrovic, Weiwei Chu, Wenhan Xiong, Wenyin Fu, Whitney Meers, Xavier Martinet, Xiaodong Wang, Xiaofang Wang, Xiaoqing~Ellen Tan, Xide Xia, Xinfeng Xie, Xuchao Jia, Xuewei Wang, Yaelle Goldschlag, Yashesh Gaur, Yasmine Babaei, Yi~Wen, Yiwen Song, Yuchen Zhang, Yue Li, Yuning Mao, Zacharie~Delpierre Coudert, Zheng Yan, Zhengxing Chen, Zoe Papakipos, Aaditya Singh, Aayushi Srivastava, Abha Jain, Adam Kelsey, Adam Shajnfeld, Adithya Gangidi, Adolfo Victoria, Ahuva Goldstand, Ajay Menon, Ajay Sharma, Alex Boesenberg, Alexei Baevski, Allie Feinstein, Amanda Kallet, Amit Sangani, Amos Teo, Anam Yunus, Andrei Lupu, Andres Alvarado, Andrew Caples, Andrew Gu, Andrew Ho, Andrew Poulton, Andrew Ryan, Ankit Ramchandani, Annie Dong, Annie Franco, Anuj Goyal, Aparajita Saraf, Arkabandhu Chowdhury, Ashley Gabriel, Ashwin Bharambe, Assaf Eisenman, Azadeh
  Yazdan, Beau James, Ben Maurer, Benjamin Leonhardi, Bernie Huang, Beth Loyd, Beto~De Paola, Bhargavi Paranjape, Bing Liu, Bo~Wu, Boyu Ni, Braden Hancock, Bram Wasti, Brandon Spence, Brani Stojkovic, Brian Gamido, Britt Montalvo, Carl Parker, Carly Burton, Catalina Mejia, Ce~Liu, Changhan Wang, Changkyu Kim, Chao Zhou, Chester Hu, Ching-Hsiang Chu, Chris Cai, Chris Tindal, Christoph Feichtenhofer, Cynthia Gao, Damon Civin, Dana Beaty, Daniel Kreymer, Daniel Li, David Adkins, David Xu, Davide Testuggine, Delia David, Devi Parikh, Diana Liskovich, Didem Foss, Dingkang Wang, Duc Le, Dustin Holland, Edward Dowling, Eissa Jamil, Elaine Montgomery, Eleonora Presani, Emily Hahn, Emily Wood, Eric-Tuan Le, Erik Brinkman, Esteban Arcaute, Evan Dunbar, Evan Smothers, Fei Sun, Felix Kreuk, Feng Tian, Filippos Kokkinos, Firat Ozgenel, Francesco Caggioni, Frank Kanayet, Frank Seide, Gabriela~Medina Florez, Gabriella Schwarz, Gada Badeer, Georgia Swee, Gil Halpern, Grant Herman, Grigory Sizov, Guangyi, Zhang, Guna
  Lakshminarayanan, Hakan Inan, Hamid Shojanazeri, Han Zou, Hannah Wang, Hanwen Zha, Haroun Habeeb, Harrison Rudolph, Helen Suk, Henry Aspegren, Hunter Goldman, Hongyuan Zhan, Ibrahim Damlaj, Igor Molybog, Igor Tufanov, Ilias Leontiadis, Irina-Elena Veliche, Itai Gat, Jake Weissman, James Geboski, James Kohli, Janice Lam, Japhet Asher, Jean-Baptiste Gaya, Jeff Marcus, Jeff Tang, Jennifer Chan, Jenny Zhen, Jeremy Reizenstein, Jeremy Teboul, Jessica Zhong, Jian Jin, Jingyi Yang, Joe Cummings, Jon Carvill, Jon Shepard, Jonathan McPhie, Jonathan Torres, Josh Ginsburg, Junjie Wang, Kai Wu, Kam~Hou U, Karan Saxena, Kartikay Khandelwal, Katayoun Zand, Kathy Matosich, Kaushik Veeraraghavan, Kelly Michelena, Keqian Li, Kiran Jagadeesh, Kun Huang, Kunal Chawla, Kyle Huang, Lailin Chen, Lakshya Garg, Lavender A, Leandro Silva, Lee Bell, Lei Zhang, Liangpeng Guo, Licheng Yu, Liron Moshkovich, Luca Wehrstedt, Madian Khabsa, Manav Avalani, Manish Bhatt, Martynas Mankus, Matan Hasson, Matthew Lennie, Matthias Reso, Maxim
  Groshev, Maxim Naumov, Maya Lathi, Meghan Keneally, Miao Liu, Michael~L. Seltzer, Michal Valko, Michelle Restrepo, Mihir Patel, Mik Vyatskov, Mikayel Samvelyan, Mike Clark, Mike Macey, Mike Wang, Miquel~Jubert Hermoso, Mo~Metanat, Mohammad Rastegari, Munish Bansal, Nandhini Santhanam, Natascha Parks, Natasha White, Navyata Bawa, Nayan Singhal, Nick Egebo, Nicolas Usunier, Nikhil Mehta, Nikolay~Pavlovich Laptev, Ning Dong, Norman Cheng, Oleg Chernoguz, Olivia Hart, Omkar Salpekar, Ozlem Kalinli, Parkin Kent, Parth Parekh, Paul Saab, Pavan Balaji, Pedro Rittner, Philip Bontrager, Pierre Roux, Piotr Dollar, Polina Zvyagina, Prashant Ratanchandani, Pritish Yuvraj, Qian Liang, Rachad Alao, Rachel Rodriguez, Rafi Ayub, Raghotham Murthy, Raghu Nayani, Rahul Mitra, Rangaprabhu Parthasarathy, Raymond Li, Rebekkah Hogan, Robin Battey, Rocky Wang, Russ Howes, Ruty Rinott, Sachin Mehta, Sachin Siby, Sai~Jayesh Bondu, Samyak Datta, Sara Chugh, Sara Hunt, Sargun Dhillon, Sasha Sidorov, Satadru Pan, Saurabh Mahajan,
  Saurabh Verma, Seiji Yamamoto, Sharadh Ramaswamy, Shaun Lindsay, Shaun Lindsay, Sheng Feng, Shenghao Lin, Shengxin~Cindy Zha, Shishir Patil, Shiva Shankar, Shuqiang Zhang, Shuqiang Zhang, Sinong Wang, Sneha Agarwal, Soji Sajuyigbe, Soumith Chintala, Stephanie Max, Stephen Chen, Steve Kehoe, Steve Satterfield, Sudarshan Govindaprasad, Sumit Gupta, Summer Deng, Sungmin Cho, Sunny Virk, Suraj Subramanian, Sy~Choudhury, Sydney Goldman, Tal Remez, Tamar Glaser, Tamara Best, Thilo Koehler, Thomas Robinson, Tianhe Li, Tianjun Zhang, Tim Matthews, Timothy Chou, Tzook Shaked, Varun Vontimitta, Victoria Ajayi, Victoria Montanez, Vijai Mohan, Vinay~Satish Kumar, Vishal Mangla, Vlad Ionescu, Vlad Poenaru, Vlad~Tiberiu Mihailescu, Vladimir Ivanov, Wei Li, Wenchen Wang, Wenwen Jiang, Wes Bouaziz, Will Constable, Xiaocheng Tang, Xiaojian Wu, Xiaolan Wang, Xilun Wu, Xinbo Gao, Yaniv Kleinman, Yanjun Chen, Ye~Hu, Ye~Jia, Ye~Qi, Yenda Li, Yilin Zhang, Ying Zhang, Yossi Adi, Youngjin Nam, Yu, Wang, Yu~Zhao, Yuchen Hao, Yundi
  Qian, Yunlu Li, Yuzi He, Zach Rait, Zachary DeVito, Zef Rosnbrick, Zhaoduo Wen, Zhenyu Yang, Zhiwei Zhao, and Zhiyu Ma.
\newblock The llama 3 herd of models, 2024.
\newblock URL \url{https://arxiv.org/abs/2407.21783}.

\bibitem[Yu et~al.(2023{\natexlab{b}})Yu, Lin, Yu, and Xing]{yu2023gptfuzzer}
Jiahao Yu, Xingwei Lin, Zheng Yu, and Xinyu Xing.
\newblock Gptfuzzer: Red teaming large language models with auto-generated jailbreak prompts, 2023{\natexlab{b}}.

\bibitem[Li et~al.(2023{\natexlab{e}})Li, Zhou, Zhu, Yao, Liu, and Han]{li2023deepinception}
Xuan Li, Zhanke Zhou, Jianing Zhu, Jiangchao Yao, Tongliang Liu, and Bo~Han.
\newblock Deepinception: Hypnotize large language model to be jailbreaker.
\newblock \emph{arXiv preprint arXiv:2311.03191}, 2023{\natexlab{e}}.

\bibitem[Wei et~al.(2023{\natexlab{b}})Wei, Wang, and Wang]{wei2023jailbreak}
Zeming Wei, Yifei Wang, and Yisen Wang.
\newblock Jailbreak and guard aligned language models with only few in-context demonstrations.
\newblock \emph{arXiv preprint arXiv:2310.06387}, 2023{\natexlab{b}}.

\bibitem[Chao et~al.(2023)Chao, Robey, Dobriban, Hassani, Pappas, and Wong]{chao2023jailbreaking}
Patrick Chao, Alexander Robey, Edgar Dobriban, Hamed Hassani, George~J Pappas, and Eric Wong.
\newblock Jailbreaking black box large language models in twenty queries.
\newblock In \emph{R0-FoMo: Robustness of Few-shot and Zero-shot Learning in Large Foundation Models}, 2023.

\bibitem[Ding et~al.(2023)Ding, Kuang, Ma, Cao, Xian, Chen, and Huang]{ding2023wolf}
Peng Ding, Jun Kuang, Dan Ma, Xuezhi Cao, Yunsen Xian, Jiajun Chen, and Shujian Huang.
\newblock A wolf in sheep's clothing: Generalized nested jailbreak prompts can fool large language models easily.
\newblock \emph{arXiv preprint arXiv:2311.08268}, 2023.

\bibitem[Lv et~al.(2024)Lv, Wang, Zhang, Huang, Dou, Ye, Gui, Zhang, and Huang]{lv2024codechameleon}
Huijie Lv, Xiao Wang, Yuansen Zhang, Caishuang Huang, Shihan Dou, Junjie Ye, Tao Gui, Qi~Zhang, and Xuanjing Huang.
\newblock Codechameleon: Personalized encryption framework for jailbreaking large language models.
\newblock \emph{arXiv preprint arXiv:2402.16717}, 2024.

\bibitem[Zhou et~al.(2024{\natexlab{b}})Zhou, Wang, Xiong, Xia, Gu, Chai, Zhu, Huang, Dou, Xi, et~al.]{zhou2024easyjailbreak}
Weikang Zhou, Xiao Wang, Limao Xiong, Han Xia, Yingshuang Gu, Mingxu Chai, Fukang Zhu, Caishuang Huang, Shihan Dou, Zhiheng Xi, et~al.
\newblock Easyjailbreak: A unified framework for jailbreaking large language models.
\newblock \emph{arXiv preprint arXiv:2403.12171}, 2024{\natexlab{b}}.

\bibitem[Mirkovic et~al.(2008)Mirkovic, Reiher, Papadopoulos, Hussain, Shepard, Berg, and Jung]{mirkovic2008testing}
Jelena Mirkovic, Peter Reiher, Christos Papadopoulos, Alefiya Hussain, Marla Shepard, Michael Berg, and Robert Jung.
\newblock Testing a collaborative ddos defense in a red team/blue team exercise.
\newblock \emph{IEEE Transactions on Computers}, 57:\penalty0 1098--1112, 2008.

\bibitem[Mehrotra et~al.(2023)Mehrotra, Zampetakis, Kassianik, Nelson, Anderson, Singer, and Karbasi]{mehrotra2023tree}
Anay Mehrotra, Manolis Zampetakis, Paul Kassianik, Blaine Nelson, Hyrum Anderson, Yaron Singer, and Amin Karbasi.
\newblock Tree of attacks: Jailbreaking black-box llms automatically.
\newblock \emph{arXiv preprint arXiv:2312.02119}, 2023.

\bibitem[Bhardwaj and Poria(2023)]{bhardwaj2023red}
Rishabh Bhardwaj and Soujanya Poria.
\newblock Red-teaming large language models using chain of utterances for safety-alignment.
\newblock \emph{arXiv preprint arXiv:2308.09662}, 2023.

\bibitem[Bai et~al.(2022)Bai, Kadavath, Kundu, Askell, Kernion, Jones, Chen, Goldie, Mirhoseini, McKinnon, et~al.]{bai2022constitutional}
Yuntao Bai, Saurav Kadavath, Sandipan Kundu, Amanda Askell, Jackson Kernion, Andy Jones, Anna Chen, Anna Goldie, Azalia Mirhoseini, Cameron McKinnon, et~al.
\newblock Constitutional ai: Harmlessness from ai feedback.
\newblock \emph{arXiv preprint arXiv:2212.08073}, 2022.

\bibitem[Sweller(1988)]{sweller1988cognitive}
John Sweller.
\newblock Cognitive load during problem solving: Effects on learning.
\newblock \emph{Cognitive science}, 12:\penalty0 257--285, 1988.

\bibitem[Sweller(2011)]{sweller2011cognitive}
John Sweller.
\newblock Cognitive load theory.
\newblock In \emph{Psychology of learning and motivation}, volume~55, pages 37--76. Elsevier, 2011.

\bibitem[Szulewski et~al.(2021)Szulewski, Howes, van Merri{\"e}nboer, and Sweller]{szulewski2021theory}
Adam Szulewski, Daniel Howes, Jeroen~JG van Merri{\"e}nboer, and John Sweller.
\newblock From theory to practice: the application of cognitive load theory to the practice of medicine.
\newblock \emph{Academic Medicine}, 96:\penalty0 24--30, 2021.

\bibitem[Shayegani et~al.(2023)Shayegani, Dong, and Abu-Ghazaleh]{shayegani2023jailbreak}
Erfan Shayegani, Yue Dong, and Nael Abu-Ghazaleh.
\newblock Jailbreak in pieces: Compositional adversarial attacks on multi-modal language models.
\newblock In \emph{The Twelfth International Conference on Learning Representations}, 2023.

\bibitem[Gao et~al.(2023{\natexlab{b}})Gao, Han, Zhang, Lin, Geng, Zhou, Zhang, Lu, He, Yue, et~al.]{gao2023llama}
Peng Gao, Jiaming Han, Renrui Zhang, Ziyi Lin, Shijie Geng, Aojun Zhou, Wei Zhang, Pan Lu, Conghui He, Xiangyu Yue, et~al.
\newblock Llama-adapter v2: Parameter-efficient visual instruction model.
\newblock \emph{arXiv preprint arXiv:2304.15010}, 2023{\natexlab{b}}.

\bibitem[Rando and Tram{\`e}r(2023)]{rando2023universal}
Javier Rando and Florian Tram{\`e}r.
\newblock Universal jailbreak backdoors from poisoned human feedback.
\newblock In \emph{The Twelfth International Conference on Learning Representations}, 2023.

\bibitem[Wolf et~al.(2023)Wolf, Wies, Levine, and Shashua]{wolf2023fundamental}
Yotam Wolf, Noam Wies, Yoav Levine, and Amnon Shashua.
\newblock Fundamental limitations of alignment in large language models.
\newblock \emph{arXiv preprint arXiv:2304.11082}, 2023.

\bibitem[Albert(2023)]{jailbreakchat}
Alex Albert.
\newblock Jailbreak chat.
\newblock \url{https://www.jailbreakchat.com}, 2023.
\newblock Accessed: 2024-02-20.

\bibitem[Chao et~al.(2024)Chao, Debenedetti, Robey, Andriushchenko, Croce, Sehwag, Dobriban, Flammarion, Pappas, Tramer, Hassani, and Wong]{chao2024jailbreakbench}
Patrick Chao, Edoardo Debenedetti, Alexander Robey, Maksym Andriushchenko, Francesco Croce, Vikash Sehwag, Edgar Dobriban, Nicolas Flammarion, George~J. Pappas, Florian Tramer, Hamed Hassani, and Eric Wong.
\newblock Jailbreakbench: An open robustness benchmark for jailbreaking large language models, 2024.

\bibitem[Xu et~al.(2024{\natexlab{b}})Xu, Dong, Guo, Wu, and Xiong]{xu2024exploring}
Shaoyang Xu, Weilong Dong, Zishan Guo, Xinwei Wu, and Deyi Xiong.
\newblock Exploring multilingual human value concepts in large language models: Is value alignment consistent, transferable and controllable across languages?, 2024{\natexlab{b}}.

\bibitem[Mazeika et~al.(2024)Mazeika, Phan, Yin, Zou, Wang, Mu, Sakhaee, Li, Basart, Li, et~al.]{mazeika2024harmbench}
Mantas Mazeika, Long Phan, Xuwang Yin, Andy Zou, Zifan Wang, Norman Mu, Elham Sakhaee, Nathaniel Li, Steven Basart, Bo~Li, et~al.
\newblock Harmbench: A standardized evaluation framework for automated red teaming and robust refusal.
\newblock \emph{arXiv preprint arXiv:2402.04249}, 2024.

\bibitem[Yang et~al.(2023{\natexlab{d}})Yang, Liu, and Wang]{yang2023fingpt}
Hongyang Yang, Xiao-Yang Liu, and Christina~Dan Wang.
\newblock Fingpt: Open-source financial large language models.
\newblock \emph{arXiv preprint arXiv:2306.06031}, 2023{\natexlab{d}}.

\bibitem[Liu et~al.(2023{\natexlab{e}})Liu, Wang, and Zha]{liu2023fingpt}
Xiao-Yang Liu, Guoxuan Wang, and Daochen Zha.
\newblock Fingpt: Democratizing internet-scale data for financial large language models.
\newblock \emph{arXiv preprint arXiv:2307.10485}, 2023{\natexlab{e}}.

\bibitem[Chen et~al.(2023{\natexlab{b}})Chen, Wang, Long, Zhang, Lu, Li, Wang, Xu, Bai, Huang, and Wei]{chen2023disc}
Wei Chen, Qiushi Wang, Zefei Long, Xianyin Zhang, Zhongtian Lu, Bingxuan Li, Siyuan Wang, Jiarong Xu, Xiang Bai, Xuanjing Huang, and Zhongyu Wei.
\newblock Disc-finllm: A chinese financial large language model based on multiple experts fine-tuning.
\newblock \emph{arXiv preprint arXiv:2310.15205}, 2023{\natexlab{b}}.

\bibitem[Wu et~al.(2023{\natexlab{a}})Wu, Irsoy, Lu, Dabravolski, Dredze, Gehrmann, Kambadur, Rosenberg, and Mann]{wu2023bloomberggpt}
Shijie Wu, Ozan Irsoy, Steven Lu, Vadim Dabravolski, Mark Dredze, Sebastian Gehrmann, Prabhanjan Kambadur, David Rosenberg, and Gideon Mann.
\newblock Bloomberggpt: A large language model for finance.
\newblock \emph{arXiv preprint arXiv:2303.17564}, 2023{\natexlab{a}}.

\bibitem[Yang et~al.(2023{\natexlab{e}})Yang, Tang, and Tam]{yang2023investlm}
Yi~Yang, Yixuan Tang, and Kar~Yan Tam.
\newblock Investlm: A large language model for investment using financial domain instruction tuning.
\newblock \emph{arXiv preprint arXiv:2309.13064}, 2023{\natexlab{e}}.

\bibitem[Tian et~al.(2023)Tian, Gan, Song, Zhang, and Zhang]{tian2023chimed}
Yuanhe Tian, Ruyi Gan, Yan Song, Jiaxing Zhang, and Yongdong Zhang.
\newblock Chimed-gpt: A chinese medical large language model with full training regime and better alignment to human preferences.
\newblock \emph{arXiv preprint arXiv:2311.06025}, 2023.

\bibitem[Zhang et~al.(2023{\natexlab{k}})Zhang, Tian, Yang, Chen, Li, and Petzold]{zhang2023alpacareinstructiontuned}
Xinlu Zhang, Chenxin Tian, Xianjun Yang, Lichang Chen, Zekun Li, and Linda~Ruth Petzold.
\newblock Alpacare:instruction-tuned large language models for medical application, 2023{\natexlab{k}}.

\bibitem[Luo et~al.(2024)Luo, Ning, Zhao, Wang, Ding, Chen, Fu, Han, Xu, Qiu, et~al.]{luo2024taiyi}
Ling Luo, Jinzhong Ning, Yingwen Zhao, Zhijun Wang, Zeyuan Ding, Peng Chen, Weiru Fu, Qinyu Han, Guangtao Xu, Yunzhi Qiu, et~al.
\newblock Taiyi: a bilingual fine-tuned large language model for diverse biomedical tasks.
\newblock \emph{Journal of the American Medical Informatics Association}, page ocae037, 2024.

\bibitem[Yang et~al.(2023{\natexlab{f}})Yang, Zhang, Kuang, Xie, and Ananiadou]{yang2023mentalllama}
Kailai Yang, Tianlin Zhang, Ziyan Kuang, Qianqian Xie, and Sophia Ananiadou.
\newblock Mentalllama: Interpretable mental health analysis on social media with large language models.
\newblock \emph{arXiv preprint arXiv:2309.13567}, 2023{\natexlab{f}}.

\bibitem[Liu et~al.(2023{\natexlab{f}})Liu, Li, Cao, Ren, Liao, and Wu]{liu2023chatcounselor}
June~M. Liu, Donghao Li, He~Cao, Tianhe Ren, Zeyi Liao, and Jiamin Wu.
\newblock Chatcounselor: A large language models for mental health support, 2023{\natexlab{f}}.

\bibitem[Zhang et~al.(2023{\natexlab{l}})Zhang, Chen, Jiang, Yu, Chen, Li, Chen, Wu, Zhang, Xiao, Wan, Wang, and Li]{huatuogpt-2023}
Hongbo Zhang, Junying Chen, Feng Jiang, Fei Yu, Zhihong Chen, Jianquan Li, Guiming Chen, Xiangbo Wu, Zhiyi Zhang, Qingying Xiao, Xiang Wan, Benyou Wang, and Haizhou Li.
\newblock Huatuogpt, towards taming language models to be a doctor.
\newblock \emph{arXiv preprint arXiv:2305.15075}, 2023{\natexlab{l}}.

\bibitem[Xiong et~al.(2023)Xiong, Wang, Zhu, Zhao, Liu, Wang, and Shen]{xiong2023doctorglm}
Honglin Xiong, Sheng Wang, Yitao Zhu, Zihao Zhao, Yuxiao Liu, Qian Wang, and Dinggang Shen.
\newblock Doctorglm: Fine-tuning your chinese doctor is not a herculean task.
\newblock \emph{arXiv preprint arXiv:2304.01097}, 2023.

\bibitem[Yue et~al.(2023)Yue, Chen, Wang, Li, Shen, Liu, Zhou, Xiao, Yun, Huang, and Wei]{yue2023disclawllm}
Shengbin Yue, Wei Chen, Siyuan Wang, Bingxuan Li, Chenchen Shen, Shujun Liu, Yuxuan Zhou, Yao Xiao, Song Yun, Xuanjing Huang, and Zhongyu Wei.
\newblock Disc-lawllm: Fine-tuning large language models for intelligent legal services, 2023.

\bibitem[Li et~al.(2023{\natexlab{f}})Li, Ai, Chen, Dong, Wu, Liu, Chen, and Tian]{SAILER}
Haitao Li, Qingyao Ai, Jia Chen, Qian Dong, Yueyue Wu, Yiqun Liu, Chong Chen, and Qi~Tian.
\newblock Sailer: Structure-aware pre-trained language model for legal case retrieval, 2023{\natexlab{f}}.

\bibitem[Huang et~al.(2023{\natexlab{f}})Huang, Tao, Zhang, An, Jiang, Chen, Wu, and Feng]{lawyer-llama-report}
Quzhe Huang, Mingxu Tao, Chen Zhang, Zhenwei An, Cong Jiang, Zhibin Chen, Zirui Wu, and Yansong Feng.
\newblock Lawyer llama technical report.
\newblock \emph{ArXiv}, abs/2305.15062, 2023{\natexlab{f}}.

\bibitem[He et~al.(2023{\natexlab{b}})He, Wen, Zhang, Cheng, Qin, Li, Jiang, Chen, Wang, and Yang]{HanFei}
Wanwei He, Jiabao Wen, Lei Zhang, Hao Cheng, Bowen Qin, Yunshui Li, Feng Jiang, Junying Chen, Benyou Wang, and Min Yang.
\newblock Hanfei-1.0.
\newblock \url{https://github.com/siat-nlp/HanFei}, 2023{\natexlab{b}}.

\bibitem[Cui et~al.(2023{\natexlab{b}})Cui, Li, Yan, Chen, and Yuan]{cui2023chatlaw}
Jiaxi Cui, Zongjian Li, Yang Yan, Bohua Chen, and Li~Yuan.
\newblock Chatlaw: Open-source legal large language model with integrated external knowledge bases.
\newblock \emph{arXiv preprint arXiv:2306.16092}, 2023{\natexlab{b}}.

\bibitem[Dan et~al.(2023)Dan, Lei, Gu, Li, Yin, Lin, Ye, Tie, Zhou, Wang, et~al.]{dan2023educhat}
Yuhao Dan, Zhikai Lei, Yiyang Gu, Yong Li, Jianghao Yin, Jiaju Lin, Linhao Ye, Zhiyan Tie, Yougen Zhou, Yilei Wang, et~al.
\newblock Educhat: A large-scale language model-based chatbot system for intelligent education.
\newblock \emph{arXiv preprint arXiv:2308.02773}, 2023.

\bibitem[Yu et~al.(2023{\natexlab{c}})Yu, Zhu, Wang, Liu, Chang, Nie, Kong, Chong, XinLiu, An, Lu, Fang, and Zhu]{Taoli-LLama}
Jingsi Yu, Junhui Zhu, Yujie Wang, Yang Liu, Hongxiang Chang, Jinran Nie, Cunliang Kong, Ruining Chong, XinLiu, Jiyuan An, Luming Lu, Mingwei Fang, and Lin Zhu.
\newblock Taoli llama.
\newblock \url{https://github.com/blcuicall/taoli}, 2023{\natexlab{c}}.

\bibitem[Wang et~al.(2024{\natexlab{e}})Wang, Wei, Hu, and Han]{wang2024transgpt}
Peng Wang, Xiang Wei, Fangxu Hu, and Wenjuan Han.
\newblock Transgpt: Multi-modal generative pre-trained transformer for transportation.
\newblock \emph{arXiv preprint arXiv:2402.07233}, 2024{\natexlab{e}}.

\bibitem[Chen et~al.(2023{\natexlab{c}})Chen, Cano, Romanou, Bonnet, Matoba, Salvi, Pagliardini, Fan, K{\"o}pf, Mohtashami, et~al.]{chen2023meditron}
Zeming Chen, Alejandro~Hern{\'a}ndez Cano, Angelika Romanou, Antoine Bonnet, Kyle Matoba, Francesco Salvi, Matteo Pagliardini, Simin Fan, Andreas K{\"o}pf, Amirkeivan Mohtashami, et~al.
\newblock Meditron-70b: Scaling medical pretraining for large language models.
\newblock \emph{arXiv preprint arXiv:2311.16079}, 2023{\natexlab{c}}.

\bibitem[Singhal et~al.(2023)Singhal, Azizi, Tu, Mahdavi, Wei, Chung, Scales, Tanwani, Cole-Lewis, Pfohl, et~al.]{singhal2023large}
Karan Singhal, Shekoofeh Azizi, Tao Tu, S~Sara Mahdavi, Jason Wei, Hyung~Won Chung, Nathan Scales, Ajay Tanwani, Heather Cole-Lewis, Stephen Pfohl, et~al.
\newblock Large language models encode clinical knowledge.
\newblock \emph{Nature}, 620\penalty0 (7972):\penalty0 172--180, 2023.

\bibitem[Wang et~al.(2023{\natexlab{e}})Wang, Yang, Du, Fan, and Li]{wang2023clinicalgpt}
Guangyu Wang, Guoxing Yang, Zongxin Du, Longjun Fan, and Xiaohu Li.
\newblock Clinicalgpt: large language models finetuned with diverse medical data and comprehensive evaluation.
\newblock \emph{arXiv preprint arXiv:2306.09968}, 2023{\natexlab{e}}.

\bibitem[Wu et~al.(2023{\natexlab{b}})Wu, Zhang, Zhang, Wang, and Xie]{wu2023pmc}
Chaoyi Wu, Xiaoman Zhang, Ya~Zhang, Yanfeng Wang, and Weidi Xie.
\newblock Pmc-llama: Further finetuning llama on medical papers.
\newblock \emph{arXiv preprint arXiv:2304.14454}, 2023{\natexlab{b}}.

\bibitem[Han et~al.(2023)Han, Adams, Papaioannou, Grundmann, Oberhauser, L{\"o}ser, Truhn, and Bressem]{han2023medalpaca}
Tianyu Han, Lisa~C Adams, Jens-Michalis Papaioannou, Paul Grundmann, Tom Oberhauser, Alexander L{\"o}ser, Daniel Truhn, and Keno~K Bressem.
\newblock Medalpaca--an open-source collection of medical conversational ai models and training data.
\newblock \emph{arXiv preprint arXiv:2304.08247}, 2023.

\bibitem[Yunxiang et~al.(2023)Yunxiang, Zihan, Kai, Ruilong, and You]{yunxiang2023chatdoctor}
Li~Yunxiang, Li~Zihan, Zhang Kai, Dan Ruilong, and Zhang You.
\newblock Chatdoctor: A medical chat model fine-tuned on llama model using medical domain knowledge.
\newblock \emph{arXiv preprint arXiv:2303.14070}, 2023.

\bibitem[Chen et~al.(2023{\natexlab{d}})Chen, Wang, Gao, Jiang, Chen, Zhang, Song, Xie, Kong, Li, et~al.]{chen2023huatuogpt}
Junying Chen, Xidong Wang, Anningzhe Gao, Feng Jiang, Shunian Chen, Hongbo Zhang, Dingjie Song, Wenya Xie, Chuyi Kong, Jianquan Li, et~al.
\newblock Huatuogpt-ii, one-stage training for medical adaption of llms.
\newblock \emph{arXiv preprint arXiv:2311.09774}, 2023{\natexlab{d}}.

\bibitem[Conneau et~al.(2020{\natexlab{c}})Conneau, Khandelwal, Goyal, Chaudhary, Wenzek, Guzm{\'a}n, Grave, Ott, Zettlemoyer, and Stoyanov]{conneau2020unsupervised}
Alexis Conneau, Kartikay Khandelwal, Naman Goyal, Vishrav Chaudhary, Guillaume Wenzek, Francisco Guzm{\'a}n, Edouard Grave, Myle Ott, Luke Zettlemoyer, and Veselin Stoyanov.
\newblock Unsupervised cross-lingual representation learning at scale.
\newblock In \emph{Proceedings of the 58th Annual Meeting of the Association for Computational Linguistics}. Association for Computational Linguistics, 2020{\natexlab{c}}.

\bibitem[Huang et~al.(2023{\natexlab{g}})Huang, Tao, An, Zhang, Jiang, Chen, Wu, and Feng]{huang2023lawyer}
Quzhe Huang, Mingxu Tao, Zhenwei An, Chen Zhang, Cong Jiang, Zhibin Chen, Zirui Wu, and Yansong Feng.
\newblock Lawyer llama technical report.
\newblock \emph{arXiv preprint arXiv:2305.15062}, 2023{\natexlab{g}}.

\bibitem[Colombo et~al.(2024)Colombo, Pires, Boudiaf, Culver, Melo, Corro, Martins, Esposito, Raposo, Morgado, et~al.]{colombo2024saullm}
Pierre Colombo, Telmo~Pessoa Pires, Malik Boudiaf, Dominic Culver, Rui Melo, Caio Corro, Andre~FT Martins, Fabrizio Esposito, Vera~L{\'u}cia Raposo, Sofia Morgado, et~al.
\newblock Saullm-7b: A pioneering large language model for law.
\newblock \emph{arXiv preprint arXiv:2403.03883}, 2024.

\bibitem[Chalkidis et~al.(2020)Chalkidis, Fergadiotis, Malakasiotis, Aletras, and Androutsopoulos]{chalkidis2020legal}
Ilias Chalkidis, Manos Fergadiotis, Prodromos Malakasiotis, Nikolaos Aletras, and Ion Androutsopoulos.
\newblock Legal-bert: The muppets straight out of law school.
\newblock In \emph{Findings of the Association for Computational Linguistics: EMNLP 2020}, pages 2898--2904, 2020.

\bibitem[Bansal et~al.(2019)Bansal, Sharma, and Singh]{bansal2019review}
Neha Bansal, Arun Sharma, and RK~Singh.
\newblock A review on the application of deep learning in legal domain.
\newblock In \emph{Artificial Intelligence Applications and Innovations: 15th IFIP WG 12.5 International Conference, AIAI 2019, Hersonissos, Crete, Greece, May 24--26, 2019, Proceedings 15}, pages 374--381. Springer, 2019.

\bibitem[Gordon et~al.(2009)Gordon, Governatori, and Rotolo]{gordon2009rules}
Thomas~F Gordon, Guido Governatori, and Antonino Rotolo.
\newblock Rules and norms: Requirements for rule interchange languages in the legal domain.
\newblock In \emph{International Workshop on Rules and Rule Markup Languages for the Semantic Web}, pages 282--296. Springer, 2009.

\bibitem[Nguyen et~al.(2024)Nguyen, Yamada, and Satoh]{nguyen2024gpts}
Ha-Thanh Nguyen, Hiroaki Yamada, and Ken Satoh.
\newblock Gpts and language barrier: A cross-lingual legal qa examination.
\newblock \emph{arXiv preprint arXiv:2403.18098}, 2024.

\bibitem[Rabelo et~al.(2020)Rabelo, Kim, Goebel, Yoshioka, Kano, and Satoh]{rabelo2020summary}
Juliano Rabelo, Mi-Young Kim, Randy Goebel, Masaharu Yoshioka, Yoshinobu Kano, and Ken Satoh.
\newblock A summary of the coliee 2019 competition.
\newblock In \emph{New Frontiers in Artificial Intelligence: JSAI-isAI International Workshops, JURISIN, AI-Biz, LENLS, Kansei-AI, Yokohama, Japan, November 10--12, 2019, Revised Selected Papers 10}, pages 34--49. Springer, 2020.

\bibitem[Rabelo et~al.(2022)Rabelo, Goebel, Kim, Kano, Yoshioka, and Satoh]{rabelo2022overview}
Juliano Rabelo, Randy Goebel, Mi-Young Kim, Yoshinobu Kano, Masaharu Yoshioka, and Ken Satoh.
\newblock Overview and discussion of the competition on legal information extraction/entailment (coliee) 2021.
\newblock \emph{The Review of Socionetwork Strategies}, 16\penalty0 (1):\penalty0 111--133, 2022.

\bibitem[Chalkidis et~al.(2022{\natexlab{b}})Chalkidis, Jana, Hartung, Bommarito, Androutsopoulos, Katz, and Aletras]{chalkidis2022lexglue}
Ilias Chalkidis, Abhik Jana, Dirk Hartung, Michael Bommarito, Ion Androutsopoulos, Daniel Katz, and Nikolaos Aletras.
\newblock Lexglue: A benchmark dataset for legal language understanding in english.
\newblock In \emph{Proceedings of the 60th Annual Meeting of the Association for Computational Linguistics (Volume 1: Long Papers)}, pages 4310--4330, 2022{\natexlab{b}}.

\bibitem[Nicholas and Bhatia(2023)]{nicholas2023lost}
Gabriel Nicholas and Aliya Bhatia.
\newblock Lost in translation: Large language models in non-english content analysis.
\newblock \emph{arXiv preprint arXiv:2306.07377}, 2023.

\bibitem[Pereltsvaig(2020)]{pereltsvaig2020languages}
Asya Pereltsvaig.
\newblock \emph{Languages of the World}.
\newblock Cambridge University Press, 2020.

\bibitem[Schneider(2018)]{schneider2018english}
Edgar~W Schneider.
\newblock English and colonialism.
\newblock In \emph{The Routledge handbook of English language studies}, pages 42--58. Routledge, 2018.

\bibitem[Pennycook(2002)]{pennycook2002english}
Alastair Pennycook.
\newblock \emph{English and the discourses of colonialism}.
\newblock Routledge, 2002.

\bibitem[Joshi et~al.(2020)Joshi, Santy, Budhiraja, Bali, and Choudhury]{joshi-etal-2020-state}
Pratik Joshi, Sebastin Santy, Amar Budhiraja, Kalika Bali, and Monojit Choudhury.
\newblock The state and fate of linguistic diversity and inclusion in the {NLP} world.
\newblock In Dan Jurafsky, Joyce Chai, Natalie Schluter, and Joel Tetreault, editors, \emph{Proceedings of the 58th Annual Meeting of the Association for Computational Linguistics}, pages 6282--6293, Online, July 2020. Association for Computational Linguistics.
\newblock \doi{10.18653/v1/2020.acl-main.560}.
\newblock URL \url{https://aclanthology.org/2020.acl-main.560}.

\bibitem[Dodge et~al.(2021)Dodge, Sap, Marasovi{\'c}, Agnew, Ilharco, Groeneveld, Mitchell, and Gardner]{dodge2021documenting}
Jesse Dodge, Maarten Sap, Ana Marasovi{\'c}, William Agnew, Gabriel Ilharco, Dirk Groeneveld, Margaret Mitchell, and Matt Gardner.
\newblock Documenting large webtext corpora: A case study on the colossal clean crawled corpus.
\newblock In \emph{Proceedings of the 2021 Conference on Empirical Methods in Natural Language Processing}, pages 1286--1305, 2021.

\bibitem[Caswell et~al.(2020)Caswell, Breiner, van Esch, and Bapna]{caswell-etal-2020-language}
Isaac Caswell, Theresa Breiner, Daan van Esch, and Ankur Bapna.
\newblock Language {ID} in the wild: Unexpected challenges on the path to a thousand-language web text corpus.
\newblock In Donia Scott, Nuria Bel, and Chengqing Zong, editors, \emph{Proceedings of the 28th International Conference on Computational Linguistics}, pages 6588--6608, Barcelona, Spain (Online), December 2020. International Committee on Computational Linguistics.
\newblock \doi{10.18653/v1/2020.coling-main.579}.
\newblock URL \url{https://aclanthology.org/2020.coling-main.579}.

\bibitem[Kreutzer et~al.(2022)Kreutzer, Caswell, Wang, Wahab, van Esch, Ulzii-Orshikh, Tapo, Subramani, Sokolov, Sikasote, et~al.]{kreutzer2022quality}
Julia Kreutzer, Isaac Caswell, Lisa Wang, Ahsan Wahab, Daan van Esch, Nasanbayar Ulzii-Orshikh, Allahsera Tapo, Nishant Subramani, Artem Sokolov, Claytone Sikasote, et~al.
\newblock Quality at a glance: An audit of web-crawled multilingual datasets.
\newblock \emph{Transactions of the Association for Computational Linguistics}, 10:\penalty0 50--72, 2022.

\bibitem[Nekoto et~al.(2020)Nekoto, Marivate, Matsila, Fasubaa, Fagbohungbe, Akinola, Muhammad, Kabongo~Kabenamualu, Osei, Sackey, Niyongabo, Macharm, Ogayo, Ahia, Berhe, Adeyemi, Mokgesi-Selinga, Okegbemi, Martinus, Tajudeen, Degila, Ogueji, Siminyu, Kreutzer, Webster, Ali, Abbott, Orife, Ezeani, Dangana, Kamper, Elsahar, Duru, Kioko, Espoir, van Biljon, Whitenack, Onyefuluchi, Emezue, Dossou, Sibanda, Bassey, Olabiyi, Ramkilowan, {\"O}ktem, Akinfaderin, and Bashir]{nekoto-etal-2020-participatory}
Wilhelmina Nekoto, Vukosi Marivate, Tshinondiwa Matsila, Timi Fasubaa, Taiwo Fagbohungbe, Solomon~Oluwole Akinola, Shamsuddeen Muhammad, Salomon Kabongo~Kabenamualu, Salomey Osei, Freshia Sackey, Rubungo~Andre Niyongabo, Ricky Macharm, Perez Ogayo, Orevaoghene Ahia, Musie~Meressa Berhe, Mofetoluwa Adeyemi, Masabata Mokgesi-Selinga, Lawrence Okegbemi, Laura Martinus, Kolawole Tajudeen, Kevin Degila, Kelechi Ogueji, Kathleen Siminyu, Julia Kreutzer, Jason Webster, Jamiil~Toure Ali, Jade Abbott, Iroro Orife, Ignatius Ezeani, Idris~Abdulkadir Dangana, Herman Kamper, Hady Elsahar, Goodness Duru, Ghollah Kioko, Murhabazi Espoir, Elan van Biljon, Daniel Whitenack, Christopher Onyefuluchi, Chris~Chinenye Emezue, Bonaventure F.~P. Dossou, Blessing Sibanda, Blessing Bassey, Ayodele Olabiyi, Arshath Ramkilowan, Alp {\"O}ktem, Adewale Akinfaderin, and Abdallah Bashir.
\newblock Participatory research for low-resourced machine translation: A case study in {A}frican languages.
\newblock In Trevor Cohn, Yulan He, and Yang Liu, editors, \emph{Findings of the Association for Computational Linguistics: EMNLP 2020}, pages 2144--2160, Online, November 2020. Association for Computational Linguistics.
\newblock \doi{10.18653/v1/2020.findings-emnlp.195}.
\newblock URL \url{https://aclanthology.org/2020.findings-emnlp.195}.

\bibitem[Callahan and Herring(2011)]{callahan2011cultural}
Ewa~S Callahan and Susan~C Herring.
\newblock Cultural bias in wikipedia content on famous persons.
\newblock \emph{Journal of the American society for information science and technology}, 62\penalty0 (10):\penalty0 1899--1915, 2011.

\bibitem[Peris et~al.(2023)Peris, Dupuy, Majmudar, Parikh, Smaili, Zemel, and Gupta]{peris2023privacy}
Charith Peris, Christophe Dupuy, Jimit Majmudar, Rahil Parikh, Sami Smaili, Richard Zemel, and Rahul Gupta.
\newblock Privacy in the time of language models.
\newblock In \emph{Proceedings of the Sixteenth ACM International Conference on Web Search and Data Mining}, pages 1291--1292, 2023.

\bibitem[Yao et~al.(2024)Yao, Duan, Xu, Cai, Sun, and Zhang]{yao2024survey}
Yifan Yao, Jinhao Duan, Kaidi Xu, Yuanfang Cai, Zhibo Sun, and Yue Zhang.
\newblock A survey on large language model (llm) security and privacy: The good, the bad, and the ugly.
\newblock \emph{High-Confidence Computing}, page 100211, 2024.

\bibitem[Yan et~al.(2024)Yan, Li, Xu, Dong, Zhang, Ren, and Cheng]{yan2024protecting}
Biwei Yan, Kun Li, Minghui Xu, Yueyan Dong, Yue Zhang, Zhaochun Ren, and Xiuzheng Cheng.
\newblock On protecting the data privacy of large language models (llms): A survey.
\newblock \emph{arXiv preprint arXiv:2403.05156}, 2024.

\bibitem[Kandpal et~al.(2022)Kandpal, Wallace, and Raffel]{kandpal2022deduplicating}
Nikhil Kandpal, Eric Wallace, and Colin Raffel.
\newblock Deduplicating training data mitigates privacy risks in language models.
\newblock In \emph{International Conference on Machine Learning}, pages 10697--10707. PMLR, 2022.

\bibitem[Thakur et~al.(2024{\natexlab{c}})Thakur, Kazi, Luo, Lin, and Ahmad]{thakur2024miragebenchautomaticmultilingualbenchmark}
Nandan Thakur, Suleman Kazi, Ge~Luo, Jimmy Lin, and Amin Ahmad.
\newblock Mirage-bench: Automatic multilingual benchmark arena for retrieval-augmented generation systems, 2024{\natexlab{c}}.
\newblock URL \url{https://arxiv.org/abs/2410.13716}.

\bibitem[Roy et~al.(2020)Roy, Constant, Al-Rfou, Barua, Phillips, and Yang]{roy-etal-2020-lareqa}
Uma Roy, Noah Constant, Rami Al-Rfou, Aditya Barua, Aaron Phillips, and Yinfei Yang.
\newblock {LAR}e{QA}: Language-agnostic answer retrieval from a multilingual pool.
\newblock In Bonnie Webber, Trevor Cohn, Yulan He, and Yang Liu, editors, \emph{Proceedings of the 2020 Conference on Empirical Methods in Natural Language Processing (EMNLP)}, pages 5919--5930, Online, November 2020. Association for Computational Linguistics.
\newblock \doi{10.18653/v1/2020.emnlp-main.477}.
\newblock URL \url{https://aclanthology.org/2020.emnlp-main.477}.

\bibitem[Chang et~al.(2024)Chang, Wang, Wang, Wu, Yang, Zhu, Chen, Yi, Wang, Wang, et~al.]{chang2024survey}
Yupeng Chang, Xu~Wang, Jindong Wang, Yuan Wu, Linyi Yang, Kaijie Zhu, Hao Chen, Xiaoyuan Yi, Cunxiang Wang, Yidong Wang, et~al.
\newblock A survey on evaluation of large language models.
\newblock \emph{ACM Transactions on Intelligent Systems and Technology}, 15\penalty0 (3):\penalty0 1--45, 2024.

\bibitem[Penedo et~al.(2024)Penedo, Kydlíček, von Werra, and Wolf]{penedo2024fineweb}
Guilherme Penedo, Hynek Kydlíček, Leandro von Werra, and Thomas Wolf.
\newblock Fineweb, 2024.
\newblock URL \url{https://huggingface.co/datasets/HuggingFaceFW/fineweb}.

\bibitem[Gao et~al.(2020)Gao, Biderman, Black, Golding, Hoppe, Foster, Phang, He, Thite, Nabeshima, et~al.]{gao2020pile}
Leo Gao, Stella Biderman, Sid Black, Laurence Golding, Travis Hoppe, Charles Foster, Jason Phang, Horace He, Anish Thite, Noa Nabeshima, et~al.
\newblock The pile: An 800gb dataset of diverse text for language modeling.
\newblock \emph{arXiv preprint arXiv:2101.00027}, 2020.

\bibitem[Soldaini et~al.(2024)Soldaini, Kinney, Bhagia, Schwenk, Atkinson, Authur, Bogin, Chandu, Dumas, Elazar, et~al.]{soldaini2024dolma}
Luca Soldaini, Rodney Kinney, Akshita Bhagia, Dustin Schwenk, David Atkinson, Russell Authur, Ben Bogin, Khyathi Chandu, Jennifer Dumas, Yanai Elazar, et~al.
\newblock Dolma: An open corpus of three trillion tokens for language model pretraining research.
\newblock \emph{arXiv preprint arXiv:2402.00159}, 2024.

\bibitem[Zhang et~al.(2023{\natexlab{m}})Zhang, Li, Hauer, Shi, and Kondrak]{zhang2023don}
Xiang Zhang, Senyu Li, Bradley Hauer, Ning Shi, and Grzegorz Kondrak.
\newblock Don’t trust chatgpt when your question is not in english: A study of multilingual abilities and types of llms.
\newblock In \emph{Proceedings of the 2023 Conference on Empirical Methods in Natural Language Processing}, pages 7915--7927, 2023{\natexlab{m}}.

\bibitem[Zhang et~al.(2024{\natexlab{d}})Zhang, Aljunied, Gao, Chia, and Bing]{zhang2024m3exam}
Wenxuan Zhang, Mahani Aljunied, Chang Gao, Yew~Ken Chia, and Lidong Bing.
\newblock M3exam: A multilingual, multimodal, multilevel benchmark for examining large language models.
\newblock \emph{Advances in Neural Information Processing Systems}, 36, 2024{\natexlab{d}}.

\bibitem[Ferrara(2023)]{ferrara2023should}
Emilio Ferrara.
\newblock Should chatgpt be biased? challenges and risks of bias in large language models.
\newblock \emph{arXiv preprint arXiv:2304.03738}, 2023.

\bibitem[Wang et~al.(2022{\natexlab{c}})Wang, Liu, and Wang]{wang2022assessing}
Jialu Wang, Yang Liu, and Xin~Eric Wang.
\newblock Assessing multilingual fairness in pre-trained multimodal representations.
\newblock In \emph{Proceedings of Annual Meeting of Association for Computational Linguistics}, 2022{\natexlab{c}}.

\bibitem[Levy et~al.(2023)Levy, John, Liu, Vyas, Ma, Fujinuma, Ballesteros, Castelli, and Roth]{levy2023comparing}
Sharon Levy, Neha John, Ling Liu, Yogarshi Vyas, Jie Ma, Yoshinari Fujinuma, Miguel Ballesteros, Vittorio Castelli, and Dan Roth.
\newblock Comparing biases and the impact of multilingual training across multiple languages.
\newblock In \emph{Proceedings of the 2023 Conference on Empirical Methods in Natural Language Processing}, pages 10260--10280, 2023.

\bibitem[Yeh et~al.(2023)Yeh, Chi, Lian, and Hsieh]{yeh2023evaluating}
Kai-Ching Yeh, Jou-An Chi, Da-Chen Lian, and Shu-Kai Hsieh.
\newblock Evaluating interfaced llm bias.
\newblock In \emph{Proceedings of the 35th Conference on Computational Linguistics and Speech Processing (ROCLING 2023)}, pages 292--299, 2023.

\bibitem[Hutchinson et~al.(2020)Hutchinson, Prabhakaran, Denton, Webster, Zhong, and Denuyl]{hutchinson2020social}
Ben Hutchinson, Vinodkumar Prabhakaran, Emily Denton, Kellie Webster, Yu~Zhong, and Stephen Denuyl.
\newblock Social biases in nlp models as barriers for persons with disabilities.
\newblock In \emph{Proceedings of the 58th Annual Meeting of the Association for Computational Linguistics}, pages 5491--5501, 2020.

\bibitem[Nadeem et~al.(2021)Nadeem, Bethke, and Reddy]{nadeem2021stereoset}
Moin Nadeem, Anna Bethke, and Siva Reddy.
\newblock Stereoset: Measuring stereotypical bias in pretrained language models.
\newblock In \emph{Proceedings of the 59th Annual Meeting of the Association for Computational Linguistics and the 11th International Joint Conference on Natural Language Processing (Volume 1: Long Papers)}, pages 5356--5371, 2021.

\bibitem[Wan et~al.(2023)Wan, Pu, Sun, Garimella, Chang, and Peng]{wan2023kelly}
Yixin Wan, George Pu, Jiao Sun, Aparna Garimella, Kai-Wei Chang, and Nanyun Peng.
\newblock “kelly is a warm person, joseph is a role model”: Gender biases in llm-generated reference letters.
\newblock In \emph{Findings of the Association for Computational Linguistics: EMNLP 2023}, pages 3730--3748, 2023.

\bibitem[de~Wynter et~al.(2024)de~Wynter, Watts, Alt{\i}ntoprak, Wongsangaroonsri, Zhang, Farra, Baur, Claudet, Gajdusek, G{\"o}ren, et~al.]{de2024rtp}
Adrian de~Wynter, Ishaan Watts, Nektar~Ege Alt{\i}ntoprak, Tua Wongsangaroonsri, Minghui Zhang, Noura Farra, Lena Baur, Samantha Claudet, Pavel Gajdusek, Can G{\"o}ren, et~al.
\newblock Rtp-lx: Can llms evaluate toxicity in multilingual scenarios?
\newblock \emph{arXiv preprint arXiv:2404.14397}, 2024.

\bibitem[Yu et~al.(2024)Yu, Kim, Choi, and Choi]{yu2024your}
Jeongrok Yu, Seong~Ug Kim, Jacob Choi, and Jinho~D Choi.
\newblock What is your favorite gender, mlm? gender bias evaluation in multilingual masked language models.
\newblock \emph{arXiv preprint arXiv:2404.06621}, 2024.

\bibitem[Zhao et~al.(2020)Zhao, Mukherjee, Hosseini, Chang, and Awadallah]{zhao2020gender}
Jieyu Zhao, Subhabrata Mukherjee, Saghar Hosseini, Kai-Wei Chang, and Ahmed~Hassan Awadallah.
\newblock Gender bias in multilingual embeddings and cross-lingual transfer.
\newblock In \emph{Proceedings of the 58th Annual Meeting of the Association for Computational Linguistics}, pages 2896--2907, 2020.

\bibitem[Piqueras and S{\o}gaard(2022)]{piqueras2022pretrained}
Laura~Cabello Piqueras and Anders S{\o}gaard.
\newblock Are pretrained multilingual models equally fair across languages?
\newblock In \emph{Proceedings of the 29th International Conference on Computational Linguistics}, pages 3597--3605, 2022.

\bibitem[Vashishtha et~al.(2023)Vashishtha, Ahuja, and Sitaram]{vashishtha2023evaluating}
Aniket Vashishtha, Kabir Ahuja, and Sunayana Sitaram.
\newblock On evaluating and mitigating gender biases in multilingual settings.
\newblock \emph{arXiv preprint arXiv:2307.01503}, 2023.

\bibitem[Team et~al.(2024)Team, Mesnard, Hardin, Dadashi, Bhupatiraju, Pathak, Sifre, Rivi{\`e}re, Kale, Love, et~al.]{team2024gemma}
Gemma Team, Thomas Mesnard, Cassidy Hardin, Robert Dadashi, Surya Bhupatiraju, Shreya Pathak, Laurent Sifre, Morgane Rivi{\`e}re, Mihir~Sanjay Kale, Juliette Love, et~al.
\newblock Gemma: Open models based on gemini research and technology.
\newblock \emph{arXiv preprint arXiv:2403.08295}, 2024.

\bibitem[Reusens et~al.(2023)Reusens, Borchert, Mieskes, De~Weerdt, and Baesens]{reusens2023investigating}
Manon Reusens, Philipp Borchert, Margot Mieskes, Jochen De~Weerdt, and Bart Baesens.
\newblock Investigating bias in multilingual language models: Cross-lingual transfer of debiasing techniques.
\newblock In \emph{Proceedings of the 2023 Conference on Empirical Methods in Natural Language Processing}, pages 2887--2896, 2023.

\bibitem[Prabhumoye et~al.(2018)Prabhumoye, Tsvetkov, Salakhutdinov, and Black]{prabhumoye2018style}
Shrimai Prabhumoye, Yulia Tsvetkov, Ruslan Salakhutdinov, and Alan~W Black.
\newblock Style transfer through back-translation.
\newblock In \emph{Proceedings of the 56th Annual Meeting of the Association for Computational Linguistics (Volume 1: Long Papers)}, pages 866--876, 2018.

\bibitem[Hoang et~al.(2018)Hoang, Koehn, Haffari, and Cohn]{hoang2018iterative}
Vu~Cong~Duy Hoang, Philipp Koehn, Gholamreza Haffari, and Trevor Cohn.
\newblock Iterative back-translation for neural machine translation.
\newblock In \emph{Proceedings of the 2nd Workshop on Neural Machine Translation and Generation}, pages 18--24, 2018.

\bibitem[Brislin(1970)]{brislin1970back}
Richard~W Brislin.
\newblock Back-translation for cross-cultural research.
\newblock \emph{Journal of cross-cultural psychology}, 1:\penalty0 185--216, 1970.

\end{thebibliography}
\bibliographystyle{unsrtnat}

\end{document}